\documentclass{article}
\usepackage{natbib}
\usepackage[normalem]{ulem}

\usepackage[preprint]{neurips_2026}

\usepackage[utf8]{inputenc} %
\usepackage[T1]{fontenc}    %
\usepackage{hyperref}
\hypersetup{
  hidelinks,
  pdftitle={Finite-Sample Convergence in Networked Average Reward MARL: Decentralization Pitfalls and Entropy Remedies},
  pdfauthor={Yizhou Zhang, Yashaswini Murthy, Laixi Shi, Adam Wierman}
}       %
\usepackage{url}            %
\usepackage{booktabs}       %
\usepackage{longtable}      %
\usepackage{amsfonts}       %
\usepackage{nicefrac}       %
\usepackage{microtype}      %
\usepackage{xcolor}
\definecolor{networkblue}{RGB}{0,90,170}         %
\colorlet{revisionblue}{black}                 %
\usepackage{amsmath}
\usepackage{algorithm}
\usepackage{algorithmic}
\usepackage{amsthm}
\usepackage{amssymb}
\usepackage{tikz}
\usetikzlibrary{automata, positioning, arrows.meta, calc}
\usepackage{mathtools}

\newtheorem{theorem}{Theorem}
\newtheorem{lemma}{Lemma}

\newtheorem{definition}{Definition}
\newtheorem{example}[theorem]{Example}
\newtheorem{assumption}{Assumption}
\newtheorem{proposition}{Proposition}
\newtheorem{corollary}{Corollary}
\theoremstyle{remark}        %
\newtheorem*{remark}{Remark} %

\title{Finite-Sample Convergence in\\Networked Average Reward MARL:\\Decentralization Pitfalls and Entropy Remedies}

\author{%
  Yizhou Zhang\thanks{Equal contribution.} \\
  California Institute of Technology \\
  \texttt{yzhang8@caltech.edu} \\
  \And
  Yashaswini Murthy\footnotemark[1] \\
  The University of Texas at Austin \\
  \texttt{yashaswini.murthy@austin.utexas.edu} \\
  \AND
  Laixi Shi \\
  Johns Hopkins University \\
  \texttt{laixis@jhu.edu} \\
  \And
  Adam Wierman \\
  California Institute of Technology \\
  \texttt{adamw@caltech.edu} \\
}

\begin{document}

\maketitle

\begin{abstract}
  We study the problem of average reward Multi-Agent Reinforcement Learning (MARL) where agents interact within a network. Each agent in the network conducts local updates with information within its $k$-hop neighborhood and collaboratively maximizes the overall reward of the entire network. We first provide impossibility results in achieving convergence to the global optimal joint policies in our setting.
  These impossibility results highlight the challenges of decentralized learning with local information in multi-agent systems, namely the issues of \emph{multiagency}, where each agent independently optimizes its policy, and \emph{partial observability}, where agents have limited visibility into the global state. Given these challenges which indicate that decentralized policy optimization can generally be certiﬁed only up to stationarity, we provide finite-sample convergence guarantees for a decentralized actor-critic algorithm with linear function approximation that converges to an approximate stationary point with small gradient norm. To overcome such impossibility results, we further show that adding fixed-state entropy regularization reshapes the objective landscape which leads to even stronger global convergence guarantees for our proposed actor-critic algorithm.
\end{abstract}

\section{Introduction}
Multi-Agent Reinforcement Learning (MARL) has been a practically important regime for training real-world cooperative multi-agent systems, with applications ranging from autonomous driving \citep{dinneweth2022multi}, energy systems \citep{keren2024multi} and robotics \citep{orr2023multi}. When scaling up the number of agents for better cooperation in more complex multi-agent systems, we are faced with a problem known as \emph{the curse of dimensionality}: the joint state-action space scales exponentially in the number of agents, making centralized value estimation, policy optimization, and exploration statistically and computationally intractable \citep{kearns1999efficient, zhang2021multi, shi2024breaking}.

To make multi-agent systems more scalable, one promising approach is to leverage the problem structure. Among various different structures, a line of work considering \emph{networked} MARL, where agents are associated within a network and only interact with neighboring agents, has gained much attention \citep{qu2022scalable, qu2020scalable, zhang2023global, zhou2023convergence}.
Such networked structure can be used to model various multi-agent systems including social networks \citep{llas2003nonequilibrium}, communication networks \citep{zocca2019temporal} and smart transportation \citep{zhang2016control}. Leveraging the local interaction structure, \citet{qu2022scalable} formulated an \emph{exponential decay} property, which formulates the system identity that the impact of an agent decays exponentially in the distance within the network. Using this property, agents can infer the global $Q$ function and update its policy based only on the information available nearby \citep{qu2022scalable, qu2020scalable}, thereby making network MARL more scalable.

Despite this breadth, prior analyses leave two gaps. First, the optimality of stationary points of policy-gradient objective has been extensively characterized in single-agent settings across a variety of MDP formulations~\citep{kumar2024globalconvergencepolicygradient, mei2020global, even2009online, murthy2023convergence}, whereas analogous insights for networked average-reward MARL---where agents are constrained to localized policies and update with nearby information---have not been developed. Second, existing convergence guarantees in this setting are largely tabular convergence to stationary points---the policy gradient norm vanishes in the limit \citep{qu2022scalable, qu2020scalable, liu2025scalable}. Finite-time bounds that characterize the suboptimality of the policy as a function of the number of samples used for training, and that expose how the localization error from $k$-hop truncation, the function-approximation error from the feature class, and the statistical TD error from finite samples interact, have not been established for networked average-reward MARL.
This motivates the two central questions of this paper:
\begin{center}
\begin{minipage}{\linewidth}
\centering
\emph{Are global convergence guarantees achievable under decentralized localized learning?}

\emph{How do localization, function approximation, and the number of samples used for learning interact in the policy-based analyses of average-reward networked MARL?}
\end{minipage}
\end{center}

\paragraph{Our contributions.}
We address these questions through three main contributions. 
\begin{enumerate}
    \item We identify \emph{Multiagency} and \emph{Partial Observability}---two intrinsic structural obstructions to global convergence under unregularized decentralized learning. For multiagency, we show through Example~\ref{example:multiagency} that while ergodicity of the induced Markov chain guarantees global optimality of every stationary point of policy gradient in single-agent RL~\citep{kumar2024globalconvergencepolicygradient}, this guarantee fails once the joint policy is constrained to be a product of local policies, even in a fully cooperative, fully observable, ergodic MDP. For partial observability, Example~\ref{example:partial_observability} shows that when the individual policies cannot depend on the global state, a vanishing local policy gradient need not even certify a notion of unilateral optimality: the stationary point can have strictly positive Nash gap, i.e., at least one agent has a profitable deviation given the others' policies fixed. These two examples reflect
    broader coordination and local-information challenges in cooperative MARL, and
    motivate why our unregularized guarantees are stated in terms of stationarity.
    \item We connect network localization with linear function approximation and derive finite-sample guarantees for the resulting actor-critic method. We observe that a $k$-hop tabular critic is simply a linear approximation to the global $Q$-function, with features that group together global state-action pairs indistinguishable from agent i’s local view. This perspective allows each agent to use a compact linear critic on its k-hop neighborhood, rather than maintaining an exponentially large local table. It also reveals a clear tradeoff: larger neighborhoods reduce localization bias, but require richer features to avoid approximation error. Using this localized-linear critic, we prove finite-sample stationarity of unregularized decentralized actor-critic. The resulting bound groups the error into a finite-sample learning term, which captures critic tracking, Markovian sampling, and actor-critic coupling, plus two structural approximation terms: feature approximation and network localization.
    \item We show that adding strong fixed-state entropy regularization can restore global convergence. The regularized algorithm adds an analytic local entropy-gradient correction while leaving the reward critic unchanged. Theorem~\ref{thm:main-regularized-convergence-summary} gives finite-time global convergence for the fixed-state entropy objective under the entropy-dominance and uniform-exploration conditions. In contrast to the unregularized case, the regularized result admits convergence to the globally optimal regularized objective function, and improves the best-iterate guarantee to a stronger last-iterate. Additionally, the regularized suboptimality contracts geometrically up to the statistical, approximation, and localization errors.
\end{enumerate}

\section{Preliminaries}
In this section we present our problem setup and introduce basic problem properties.

\subsection{Networked Average-Reward MDP}
We work under the same problem setting as in \citet{qu2020scalable}. Consider a network of $n$ agents specified by an undirected graph $\mathcal{G}=(\mathcal{N},\mathcal{E})$ where $\mathcal{N}=\{1,2,\dots,n\}$ is the set of nodes with each node representing an agent, and $\mathcal{E}\subseteq \mathcal{N}\times\mathcal{N}$ denotes the set of edges. Each agent $i$ has a local state $s_i\in \mathcal{S}_i$ and local action $a_i\in\mathcal{A}_i$. The global state and action are the joint tuples of local states and actions respectively, namely $s_g=(s_1,\dots,s_n)\in \mathcal{S}_g:=\mathcal{S}_1\times \dots \times \mathcal{S}_n$ and $a_g=(a_1,\dots,a_n)\in \mathcal{A}_g:=\mathcal{A}_1\times \dots \times \mathcal{A}_n$. For each agent $i$, we let $N_i$ denote the set of its neighbors (including itself) in the graph, and $N_i^k$ denote the $k$-hop neighborhood consisting of all agents with graph distance at most $k$ to $i$. We use $s_i^k\in \mathcal{S}_i^k$ and $a_i^k\in \mathcal{A}_i^k$ to denote the joint state and action in the $k$-hop neighborhood of $i$ respectively.

At each time step $t$, the next local state $s_i(t+1)$ is independently generated according to the transition kernel, which only depends on the local action and the ($1$-hop) neighboring states, such that $s_i(t+1)\sim P_i(\cdot|s_i^1(t), a_i(t))$.
Accordingly, the global transition kernel is the following product:
\begin{equation}
P(s_g(t+1)|s_g(t),a_g(t))=\prod_{i=1}^n P_i(s_i(t+1)|s_{i}^1(t),a_i(t))
\end{equation}
Each agent has a policy $\pi_{\theta_i}: \mathcal{S}_i\rightarrow \Delta(\mathcal{A}_i)$ parameterized by $\theta_i$ that depends only on the local state. We assume that each agent adopts a \emph{softmax parameterization}, such that $\pi_{\theta_i}(a_i|s_i)=\frac{\exp(\theta_i(s_i,a_i))}{\sum_{a_i'}\exp(\theta_i(s_i,a_i'))}$.
The global policy $\pi_\theta(a_g|s_g)=\prod_{i=1}^n \pi_{\theta_i}(a_i|s_i)$ where $\theta=(\theta_1,\dots,\theta_n)$ denotes the joint global parameters. Let $\nu^\theta$ denote the corresponding stationary distribution over $\mathcal S_g \times \mathcal A_g$.

When acting $a_i$ at local state $s_i$, agent $i$ receives a local reward $r_i(s_i,a_i)$. Throughout the paper, we assume local rewards are uniformly bounded: there exists $\bar r>0$ such that $|r_i(s_i,a_i)|\leq \bar r$ for all $i\in\mathcal{N}$ and $(s_i,a_i)\in\mathcal{S}_i\times\mathcal{A}_i$. We define the global reward as the sum of local rewards: $r(s_g,a_g):=\sum_{i=1}^n r_i(s_i,a_i)$. The objective of all agents is to collaboratively maximize the average global reward $J(\theta):=\mathbb{E}_{(s_g,a_g)\sim \nu^\theta} [r(s_g,a_g)]$, while each agent only uses information within its $k$-hop neighborhood $N_i^k$. When the policy induces an ergodic Markov chain, the average reward is independent of initial state distribution~\cite{bertsekas2011dynamic}. We can decompose the global average reward $J(\theta)$ into local average rewards as: $J(\theta)=\sum_{i=1}^n J_i(\theta)$, where $J_i(\theta):=\mathbb{E}_{(s_i,a_i)\sim \nu^\theta} [r_i(s_i,a_i)]$ are the local average rewards of agent $i$. Define the global and local $Q$ functions as:
\begin{align}
Q^\theta(s_g,a_g):=&\mathbb{E}_{\pi_\theta}\left[\sum_{t=0}^\infty \left(r(s_g(t),a_g(t))-J(\theta)\right) \middle|s_g(0)=s_g,a_g(0)=a_g\right];\\
Q_i^\theta(s_g,a_g):=&\mathbb{E}_{\pi_\theta}\left[\sum_{t=0}^\infty \left(r_i(s_i(t),a_i(t))-J_i(\theta)\right) \middle|s_g(0)=s_g,a_g(0)=a_g\right]. 
\end{align}
Notice that both global and local $Q$ functions depend on the global state-action pair $(s_g,a_g)$ and the global policy parameter $\theta$, as the joint global policies and the starting state-action pair elsewhere will affect later local rewards of $i$ as well. We can easily verify that $Q^\theta(s_g,a_g)=\sum_{i=1}^n Q_i^\theta(s_g,a_g)$.

\paragraph{Regularized objective.} In order to obtain finite-time global convergence results, in addition to the unregularized average-reward objective $J(\theta)$, we also consider a fixed-state entropy-regularized objective. For each agent $i$, fix some local state weights $\omega_i\in\Delta(\mathcal{S}_i)$, let
\begin{equation}
    \mathcal H_\omega(\theta)
    :=
    \sum_{i=1}^n
    \sum_{s_i\in\mathcal S_i}
    \omega_i(s_i)
    h_i\!\left(\pi_{\theta_i}(\cdot\mid s_i)\right),
    \qquad
    h_i(p):=-\sum_{a_i\in\mathcal A_i}p(a_i)\log p(a_i).
    \label{eq:state-wise-entropy-regularized-objective}
\end{equation}
be the entropy weighted by $\omega_i$, the regularized objective is defined as $J_{\iota,\omega}(\theta):=J(\theta)+\iota\mathcal H_\omega(\theta)$. Notice that although fixed-state entropy regularization is different from the usual entropy-regularization used in theoretical RL settings, where the entropy is weighted by the true state-distribution under $\pi_\theta$, it is an empirically more popular choice \citep{schulman2017proximal, yu2022surprising, forkel1high}.

\subsection{The Exponential Decay Property and Truncated $Q$ Functions}
Since our goal is to let agents learn only with information within the $k$-hop neighborhood, it is required that the states and actions of agents too far away in the network won't change the local interaction structure much. Let $N_{-i}^k=\mathcal{N}\setminus N_i^k$ denote the set of agents outside of the $k$-hop neighborhood of $i$. For any global state-action pair, write $s_{-i}^k\in \mathcal{S}_{-i}^k:=\prod_{j\in N_{-i}^k}\mathcal{S}_j$ and $a_{-i}^k\in \mathcal{A}_{-i}^k:=\prod_{j\in N_{-i}^k}\mathcal{A}_j$ for the joint state and action outside $N_i^k$. We adopt the notion of \emph{exponential decay} in \citet{qu2020scalable}:
\begin{definition}
    Given $c>0$ and $\rho\in(0,1)$, we say the $(c,\rho)$ exponential decay property holds if for all $\theta$ and $i\in \mathcal{N}$, $s_i^k\in \mathcal{S}_i^k, s_{-i}^k, (s_{-i}^k)'\in \mathcal{S}_{-i}^k, a_i^k\in \mathcal{A}_i^k, a_{-i}^k, (a_{-i}^k)'\in \mathcal{A}_{-i}^k$, it holds that:
    \begin{equation}
        \left|Q_i^\theta(s_i^k,s_{-i}^k,a_i^k,a_{-i}^k)-Q_i^\theta(s_i^k,(s_{-i}^k)',a_i^k,(a_{-i}^k)')\right|\leq c\rho^{k+1}.
    \end{equation}
\end{definition}

The following theorem from \citet{qu2020scalable} gives a sufficient condition for the exponential decay property to hold in the average-reward setting. For each pair of agents $i,j\in\mathcal{N}$, define the local interaction coefficient
\begin{equation}
    C_{ij}:=
    \begin{cases}
        0, & j\notin N_i,\\
        \displaystyle \sup_{s_{N_i\setminus\{j\}},a_i}\sup_{s_j,s_j'} \mathrm{TV}\left(P_i(\cdot|s_j,s_{N_i\setminus\{j\}},a_i), P_i(\cdot|s_j',s_{N_i\setminus\{j\}},a_i)\right), & j\in N_i,\ j\neq i,\\
        \displaystyle \sup_{s_{N_i\setminus\{i\}}}\sup_{s_i,s_i',a_i,a_i'} \mathrm{TV}\left(P_i(\cdot|s_i,s_{N_i\setminus\{i\}},a_i), P_i(\cdot|s_i',s_{N_i\setminus\{i\}},a_i')\right), & j=i,
    \end{cases}
\end{equation}
where $\mathrm{TV}(\cdot,\cdot)$ is the total variation distance and $s_{N_i\setminus\{j\}}$ denotes the joint local states in $N_i\setminus\{j\}$.

\begin{theorem}[\citet{qu2020scalable}, Theorem 1]
    \label{thm:spatial-decay-sufficient-condition}
    Suppose there exist $\bar r>0$ and $\rho\in(0,1)$ such that $|r_i(s_i,a_i)|\leq \bar r$ for all $i\in\mathcal{N}$ and $(s_i,a_i)\in\mathcal{S}_i\times\mathcal{A}_i$, and $\sum_{j=1}^n C_{ij}\leq \rho, \forall i\in\mathcal{N}$.
    Then the $\left(\frac{\bar r}{1-\rho},\rho\right)$ exponential decay property holds.
\end{theorem}
The exponential decay property guarantees $Q_i^\theta$ doesn't change too much as the states and actions outside the $k$-hop neighborhood changes. Once the exponential decay property holds, the class of truncated $Q$ functions $\widetilde{Q}_i^\theta(s_i^k,a_i^k)$ that only depends on the state-action pairs in the $k$-hop neighborhood can approximate the true underlying $Q$ functions well.

\subsection{Linear Function Approximation}
The truncated critic $\widetilde{Q}_i^\theta(s_i^k,a_i^k)$ is still tabular: each agent $i$ keeps a separate parameter for every $(s_i^k, a_i^k) \in \mathcal{S}_i^k \times \mathcal{A}_i^k$, with dimension exponential in $k$. To make the role of $k$ precise, introduce the local-to-global lifting matrix $\phi_i \in \{0,1\}^{|\mathcal{S}_g||\mathcal{A}_g| \times |\mathcal{S}_i^k||\mathcal{A}_i^k|}$, $\phi_i\bigl((s_g,a_g),(s_i^k,a_i^k)\bigr) := \mathbf{1}\{\mathrm{Proj}_{\mathcal{N}_i^k}(s_g,a_g) = (s_i^k,a_i^k)\}$, which lifts any truncated critic $\widehat{Q}_i$ to a global function $\phi_i \widehat{Q}_i$ constant on each $k$-hop equivalence class (see illustration in Appendix~\ref{sss:phi_visual}). The tabular localized critic is then the projected Bellman fixed point~\citep{qu2020scalable}:
\begin{equation}
\phi_i \widehat{Q}_i^\star \;=\; \Pi^{\phi_i}_{D_{\pi_\theta}}\,\mathcal{T}^i_{\pi_\theta}\bigl(\phi_i \widehat{Q}_i^\star\bigr),
\label{eq:tabular-fp}
\end{equation}
where $\mathcal{T}^i_{\pi_\theta}$ is the local average-reward Bellman operator and $\Pi^{\phi_i}_{D_{\pi_\theta}}$ is the $D_{\pi_\theta}$-weighted projection onto $\mathrm{span}(\phi_i)$. Here $D_{\pi_\theta}:=\text{diag}(\nu^\theta).$ Read this way, the truncated tabular critic is itself a linear-function-approximation fixed point for the global $Q_i^\theta$ with feature $\phi_i$. In generic linear approximation, the resulting projection error is hard to characterize since it depends on how well the feature spans the true value function; here, the very specific structure of $\phi_i$ (it groups exactly those global state-action pairs differing only outside $\mathcal{N}_i^k$ ) ties this error directly to the exponential decay constant, yielding $\|\phi_i \widehat{Q}_i^\star - Q_i^\theta\|_\infty = O(c\rho^{k+1})$. Larger $k$ enriches $\mathrm{span}(\phi_i)$ and shrinks this error geometrically.

To handle larger neighborhoods, we further restrict the critic to a low-dimensional linear class. For each agent $i$, let $\psi_i : \mathcal{S}_i^k \times \mathcal{A}_i^k \to \mathbb{R}^{d_i}$ be a prescribed feature map and set $\widetilde{Q}_{i,w_i}(s_i^k,a_i^k) := \psi_i(s_i^k,a_i^k)^\top w_i$. With $\Psi_i$ collecting $\psi_i(\cdot,\cdot)^\top$ row-wise and $\chi_i := \phi_i \Psi_i$, the corresponding linear fixed point $w_i^\star(\theta)$ satisfies
\begin{equation}
\chi_i w_i^\star(\theta) \;=\; \Pi^{\chi_i}_{D_{\pi_\theta}}\,\mathcal{T}^i_{\pi_\theta}\bigl(\chi_i w_i^\star(\theta)\bigr).
\label{eq:lfa-fp}
\end{equation}
The two-stage embedding $\mathbb{R}^{d_i} \xrightarrow{\Psi_i} \mathbb{R}^{|\mathcal{S}_i^k||\mathcal{A}_i^k|} \xrightarrow{\phi_i} \mathbb{R}^{|\mathcal{S}_g||\mathcal{A}_g|}$ separates the two error sources cleanly:
\begin{equation}
\|\chi_i w_i^\star (\theta) - Q_i^\theta\|_{D_{\pi_\theta}} \;\le\; \underbrace{\|\phi_i \widehat{Q}_i^\star - Q_i^\theta\|_{D_{\pi_\theta}}}_{\text{localization, } O(c\rho^{k+1})} \;+\; \underbrace{\|\chi_i w_i^\star(\theta) - \phi_i \widehat{Q}_i^\star\|_{D_{\pi_\theta}}}_{\text{function approximation}}.
\label{eq:error-decomp}
\end{equation}
For maximum expressivity, $\Psi_i$ are chosen to be full rank matrices. For a fixed feature dimension $d_i$, increasing $k$ shrinks the first term but \emph{may inflate} the second, since $\Psi_i$ must now compress a higher-dimensional truncated $Q$-function into the same parameter budget. The neighborhood size $k$ is thus a tunable lever for trading localization against expressivity, a useful knob in designing decentralization schemes. For a fixed policy $\pi_\theta$, the local TD update line 5 in Algorithm~\ref{alg:decentralized_ac_lfa} with step size $\beta = \Theta(\log T / T)$ drives $\chi_i \textcolor{revisionblue}{\Pi_i^{\perp}} w_i^T$ to within $\widetilde{O}(1/\sqrt{T})$ of the sum of truncation and function approximation floor in~\eqref{eq:error-decomp}. Here $\textcolor{revisionblue}{\Pi_i^{\perp}}$ denotes the projection removing the additive-constant direction. Detailed finite-time analyses for the fixed-policy critic and the moving-target actor-critic recursion are deferred to Appendix~\ref{subsection:critic}. \textcolor{revisionblue}{The fixed-policy bound is Theorem~\ref{thm:fixed-policy-td-learning-bounds}.}

\section{Impossibility Results}
\label{sec:impossibility_result}
If we use a centralized policy that has access to the global state and directly optimizes the joint global actions, the problem reduces to a single-agent average-reward RL problem, where policy gradient has global convergence guarantee to the optimal objective function $J^*(\theta)$ \citep{kumar2024globalconvergencepolicygradient}.
However, prior works \citep{qu2022scalable, qu2020scalable} that adopted a localized policy ($\pi_{\theta_i}$ only depends on the local state $s_i$) were only able to achieve convergence results in terms of the gradient norm $\|\nabla J(\theta)\|$, instead of the more desirable global optimality $J^*(\theta)$ itself.

To investigate whether global convergence is attainable under this localized learning setting, notice that when each agent $i$ has a local policy $\pi_{\theta_i}$ that only depends on its local state $s_i$, the average-reward policy gradient theorem (Lemma~\ref{lem:avg-reward-policy-gradient}) provides a useful lens for the decentralized policy optimization objective:
For a tabular softmax local policy and every $(\bar s_i,\bar a_i)\in\mathcal S_i\times\mathcal A_i$,
\begin{align}
    \frac{\partial J(\theta)}
    {\partial\theta_i(\bar s_i,\bar a_i)}
    &=
    \sum_{s_{-i}\in\mathcal S_{-i}}
    d^\theta(\bar s_i,s_{-i})
    \pi_{\theta_i}(\bar a_i|\bar s_i)
    \bar A_i^\theta((\bar s_i,s_{-i}),\bar a_i),
    \label{eq:challenge-partial-observability-pg}\\
    \text{where}\quad \bar A_i^\theta(s_g,a_i)
    &:=
    \sum_{a_{-i}\in\mathcal A_{-i}}
    \pi_{\theta_{-i}}(a_{-i}|s_{-i})
    Q^\theta(s_g,a_i,a_{-i})
    -
    V^\theta(s_g).
    \label{eq:challenge-one-agent-averaged-advantage}
\end{align}
Here $\bar A_i^\theta(s_g,a_i)$ is the advantage of forcing agent $i$ to take action $a_i$ while averaging over the other agents' current actions. Here $V^\theta(s_g)\coloneqq \sum_{a_g}\pi_\theta(a_g|s_g)Q^\theta(s_g,a_g)$ and $d^\theta(s_g)\coloneqq \sum_{a_g}\nu^\theta(s_g,a_g).$
Based on the averaged advantage $\bar A_i^\theta(s_g,a_i)$, the policy gradient \eqref{eq:challenge-partial-observability-pg} further averages over the state $s_{-i}$ of other agents, which the policy of agent $i$ doesn't have access to.
Averaging over other agents' actions and states provides a coarse-grained gradient signal for each local action $\bar{a}_i$ under local state $\bar{s}_i$.
In this section we show that due to the averaging in \eqref{eq:challenge-partial-observability-pg} and \eqref{eq:challenge-one-agent-averaged-advantage}, global optimality cannot be guaranteed in general for decentralized product-policy actor methods, and provide two distinct sources of suboptimality---\emph{multiagency} and \emph{partial observability}.
\paragraph{Challenge 1: Multiagency.} \emph{Multiagency} is the scenario where each agent maintains and updates its own policy in a decentralized manner. Example~\ref{example:multiagency} shows that even if agents are \emph{fully collaborative} and the environment is \emph{fully observable}, the policies may get stuck in suboptimal local optima:
\begin{example}\label{example:multiagency}
Consider the one-state MDP in Figure~\ref{fig:one_state_coordination}(a), with rewards specified in Figure~\ref{fig:one_state_coordination}(b). Two cooperative agents choose actions in $\{a,b\}$, the next state is always $s$, and therefore the induced chain is irreducible and aperiodic under every policy. Thus the average-reward objective is smooth with respect to the tabular parametrization~\citep{kumar2024globalconvergencepolicygradient}.

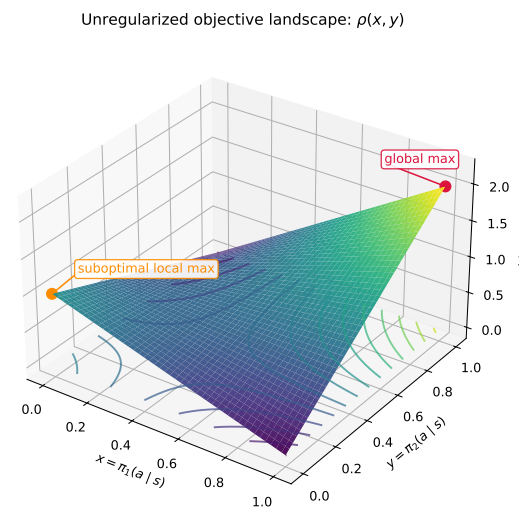
\begin{figure}[t]
\centering
\begin{minipage}[t]{0.23\textwidth}
\centering
\textbf{(a) MDP graph}\\[1mm]
\resizebox{\linewidth}{!}{%
\begin{tikzpicture}[>=Latex, every node/.style={font=\small}]
    \node[draw, circle, minimum size=1.15cm] (s) {$s$};
    \draw[->, thick]
        (s.120) to[out=55,in=125,looseness=8]
        node[above=10pt, align=center]
        {$u \in \{(a,a),(a,b),(b,a),(b,b)\}$\\
         $P(s\mid s,u)=1$}
        (s.60);
\end{tikzpicture}
}
\end{minipage}
\hfill
\begin{minipage}[t]{0.23\textwidth}
\centering
\textbf{(b) Reward table}\\[1mm]
\vspace{3mm}
\small
\resizebox{\linewidth}{!}{%
\begin{tabular}{cc}
\toprule
joint action $u$ & reward $r(u)$ \\
\midrule
$(a,a)$ & $2$ \\
$(b,b)$ & $1$ \\
$(a,b)$ & $0$ \\
$(b,a)$ & $0$ \\
\bottomrule
\end{tabular}
}
\end{minipage}
\hfill
\begin{minipage}[t]{0.25\textwidth}
\centering
\textbf{(c) Objective landscape}\\[1mm]
\includegraphics[width=\linewidth]{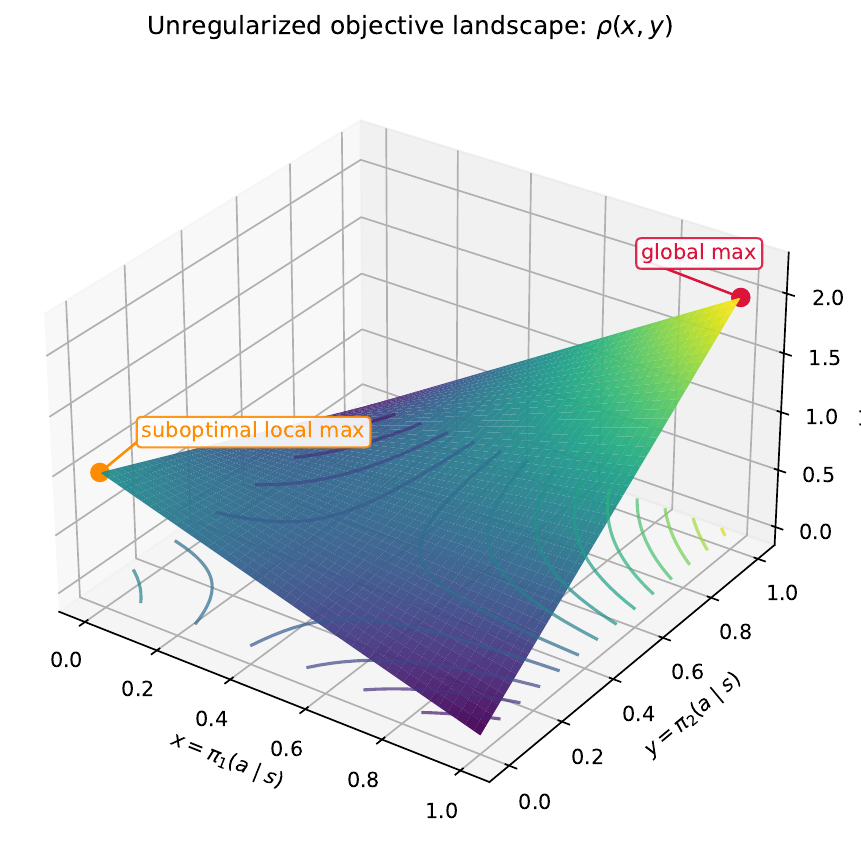}
\end{minipage}
\caption{One-state coordination example. Panel (a) shows that every joint action returns to the same state; panel (b) gives the reward table; panel (c) plots the unregularized direct product-policy objective, where the suboptimal corner $(0,0)$ is locally attractive while $(1,1)$ is globally optimal.}
\label{fig:one_state_coordination}
\end{figure}
\paragraph{Decentralized learning using product policies.}
Let $p:=\pi_1(a\mid s)$ and $q:=\pi_2(a\mid s)$. Under the product policy, we have $J(p,q)=2pq+(1-p)(1-q)=1-p-q+3pq$ and $\nabla J(p,q)=(-1+3q,\,-1+3p)$.
Hence both partial derivatives are negative on the square $p<1/3$, $q<1/3$, so projected gradient ascent pushes the policy toward $(p,q)=(0,0)$. This corner is a strict local optimum: if $0\le q<1/3$, then $\forall p>0$, we have $J(p,q)-J(0,0)=p(3q-1)-q<0$,
thus $(0,0)$ has value $J(0,0)=1$ but is suboptimal, since the global optimum is $(1,1)$ with value $J(1,1)=2$.
Figure~\ref{fig:one_state_coordination}(c) illustrates this nonconcave product-policy objective landscape with multiple local optima.

\paragraph{Centralized learning.}
For the centralized single-agent formulation with joint-action space $\mathcal U:=\{(a,a),(a,b),(b,a),(b,b)\}$ and policy $x\in\Delta(\mathcal U)$, the same objective is linear:
$J(x)=2x_{aa}+x_{bb}$.
Therefore it is simply a multi-armed bandit problem and there are no suboptimal local optima.
\end{example}
For the sake of brevity, Example~\ref{example:multiagency} is not fully consistent with the networked MDP setting. This does not invalidate the earlier observation regarding suboptimal convergence, as we show an example that is fully compatible with our framework in Appendix~\ref{app:suboptimal_convergence_example}.

Example~\ref{example:multiagency} illustrates that even in a fully cooperative and fully observable setting, decentralized learning with product policies can converge to a suboptimal local optimum.
From the viewpoint of \eqref{eq:challenge-one-agent-averaged-advantage}, each actor coordinate measures only a unilateral action change while the other agents keep their current policies. Near $(p,q)=(0,0)$, such a unilateral move mainly creates mismatched low-reward actions, while the coordinated improvement to $(1,1)$ is invisible to a single-agent gradient. This product-policy geometry turns the centralized linear objective into a non-concave landscape, and decentralized gradient dynamics can then get stuck at its suboptimal local optima.
This multiagency challenge is fundamental to decentralized learning: Once the global policy is constrained to be products of local policies in independent learning, one should not expect global optimality of policy gradient in general.
Instead, one could at best hope for achieving optimality unilaterally, which asserts that each agent is acting optimally assuming other agents are fixed. To measure this unilateral optimality, for a policy class $\overline{\Pi}_i$ available to agent $i$, define the agent-wise \emph{Nash gap}:
\begin{equation}
    \mathrm{NashGap}_i(\theta)
        :=
        \sup_{\pi_i'\in\overline{\Pi}_i}
        J(\pi_i',\theta_{-i})-J(\theta).
    \label{eq:intro-nash-gap}
\end{equation}
Here $J(\pi_i',\theta_{-i})$ is the average reward obtained when agent $i$ unilaterally switches to $\pi_i'$ while all other agents keep their current policies.
Thus $\mathrm{NashGap}_i(\theta)=0$ means that agent $i$ has no profitable unilateral deviation within its admissible local policy class, and $\sum_i\mathrm{NashGap}_i(\theta)$ measures the aggregate incentive for decentralized deviation.
Although this does not rule out local optima such as Example~\ref{example:multiagency}, it is the desirable convergence certificate in the decentralized setting: It guarantees that the product policy is locally stable against any single agent's policy improvement.

\paragraph{Challenge 2: Partial Observability.}
As a weaker optimality notion, one may hope that policy gradient based methods give convergence to near-zero Nash gap in decentralized learning.
The next example shows that this can also fail when the policy of agent $i$ can only depend on its local state $s_i$.
\begin{example}\label{example:partial_observability}
Consider two agents on a two-node network. Agent 1 has local state $s_1=\star$
and actions $\{\ell,r\}$; agent 2 has local state $s_2\in\{L,R,M\}$ and a
singleton action set. The local rewards are $r_1(\star,a_1)=0,r_2(s_2)=\mathbf{1}\{s_2\in\{L,R\}\}$.
Thus $J$ is the stationary probability of $s_2\in\{L,R\}$. The state $s_1$ is
fixed, and $s_2$ evolves according to Figure~\ref{fig:partial_observability_example}.
\begin{figure}[t]
\centering
\begin{minipage}[t]{0.23\textwidth}
\centering
\textbf{(a) Transition table}\\[1mm]
\small
\resizebox{\linewidth}{!}{%
\begin{tabular}{c|cc}
 & $a_1=\ell$ & $a_1=r$\\
\hline
$L$ & $L$ & $M$\\
$R$ & $M$ & $R$\\
$M$ & $L$ & $R$
\end{tabular}
}
\end{minipage}
\hfill
\begin{minipage}[t]{0.43\textwidth}
\centering
\textbf{(b) Transition graph}\\[1mm]
\resizebox{\linewidth}{!}{%
\begin{tikzpicture}[>=Latex, every node/.style={font=\small}]
    \node[draw, circle, minimum size=0.9cm] (L) at (-2.2,0) {$L$};
    \node[draw, circle, minimum size=0.9cm] (M) at (0,0) {$M$};
    \node[draw, circle, minimum size=0.9cm] (R) at (2.2,0) {$R$};

    \draw[->, thick] (M) to[bend left=18] node[below] {$\ell$} (L);
    \draw[->, thick] (M) to[bend left=18] node[above] {$r$} (R);
    \draw[->, thick] (L) to[bend left=18] node[above] {$r$} (M);
    \draw[->, thick] (R) to[bend left=18] node[below] {$\ell$} (M);
    \draw[->, thick]
        (L.130) to[out=145,in=215,looseness=6]
        node[left] {$\ell$}
        (L.230);
    \draw[->, thick]
        (R.50) to[out=35,in=-35,looseness=6]
        node[right] {$r$}
        (R.-50);
\end{tikzpicture}
}
\end{minipage}
\caption{Partial-observability example. Panel (a) gives the transition table for the hidden local state $s_2$; panel (b) shows the same transition graph. Agent 1 cannot condition its action on $s_2$.}
\label{fig:partial_observability_example}
\end{figure}
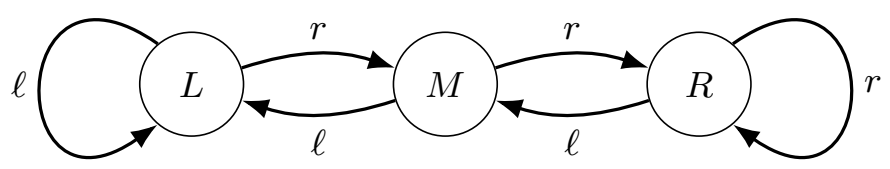
Agent 1 observes only $s_1=\star$, so write
$p:=\pi_1(\ell|\star)$ and $1-p:=\pi_1(r|\star)$. For every softmax policy
$p\in(0,1)$, the chain on $\{L,R,M\}$ is irreducible and aperiodic, and its
stationary distribution satisfies $d^p(L)=\frac{p}{1-p}d^p(M)$, $d^p(R)=\frac{1-p}{p}d^p(M)$.
Therefore $d^p(M)
    =
    \left(
        1+\frac{p}{1-p}+\frac{1-p}{p}
    \right)^{-1}$ and $J(p)=1-d^p(M)$.
At $p=1/2$, $d^{1/2}(M)=1/3$, $J(1/2)=2/3$, and by symmetry $J'(1/2)=0$.
Equivalently, under the tabular softmax parameterization, $\frac{\partial J}{\partial\theta_1(\star,\ell)}=\frac{\partial J}{\partial\theta_1(\star,r)}=0$ at $p=1/2$. Since agent 2 has no actor parameter, the joint local policy is a
stationary point of the product-policy objective.
However, its Nash gap is strictly positive: for the unilateral policy $p'=1/4$,
we have $J(1/4)=1-\left(1+\frac{1}{3}+3\right)^{-1}=\frac{10}{13}>\frac{2}{3}=J(1/2)$.
Thus $\mathrm{NashGap}_1(\theta) \geq J(1/4)-J(1/2)=\frac{4}{39}>0$ even though all local policy gradients vanish at $\theta$.
\end{example}

A fully networked-MDP-compatible version of Example~\ref{example:partial_observability} is given in Appendix~\ref{app:networked_partial_observability_example}. As the local policy $\pi_{\theta_i}$ of agent $i$ only depends on its local state $s_i$, a vanishing local-policy gradient need not rule out profitable unilateral deviations. In Example~\ref{example:partial_observability}, the state averaging in \eqref{eq:challenge-partial-observability-pg} is where the obstruction appears. At $p=1/2$, the stationary distribution is uniform on $\{L,M,R\}$, we have $\bar A_1^\theta((\star,L),\ell)=\frac{1}{3},\bar A_1^\theta((\star,M),\ell)=0$ and $\bar A_1^\theta((\star,R),\ell)=-\frac{1}{3}.$
The local averaged advantage for action $\ell$ sees only the stationary average of these three quantities, so the positive effect at $L$ is canceled by the negative effect at $R$.
Notice that the issue is not inaccurate value estimation under limited information, but the localized state dependencies of policies: even if agents have access to the exact global value/$Q$ functions that depend on the entire $s_g$, $\bar A_i^\theta(s_g,a_i)$ is averaged over hidden states sharing the same local observation.
This indicates that partial observability is a fundamental challenge: if agents are constrained to use policies that cannot see the global state, then one should not expect convergence to zero Nash gap in general.

\section{Algorithm and Analysis}
We now introduce Algorithm~\ref{alg:decentralized_ac_lfa}, our decentralized actor-critic algorithm that admits both unregularized and fixed-state entropy regularized objectives. In light of the impossibility results, we first provide a finite-time analysis for its unregularized case up to a stationary point, and then introduce fixed-state entropy regularization to remove spurious local maxima, thereby ensuring global convergence.

\begin{algorithm}[H]
\caption{Decentralized Actor-Critic for Networked Average-Reward MARL}
\label{alg:decentralized_ac_lfa}
\begin{algorithmic}[1]
\REQUIRE \! Features $\psi_i\!:\mathcal S_i^k\!\times\!\mathcal A_i^k\to\mathbb R^{d_i}$; stepsizes\! $\{\alpha_t,\beta_t\}$; $\!c_{ss}\!>\!0$; initial $(\theta_i^0,w_i^0,\mu_i^0)$ for all agents $i$.
\STATE Sample initial global state $s_g(0)$ and initial action $a_i(0).$
\FOR{$t=0,1,2,\ldots$}
    \STATE Receive reward $r_i^t:=r_i(s_i(t),a_i(t)).$ Observe $s_i^k(t\!+\!1)$. Sample $a_i(t+1)\!\sim\!\pi_{\theta_i^t}(\cdot\!\mid\! s_i(t+1))$ and observe $a_i^k(t\!+\!1)$. Define $\textcolor{revisionblue}{\psi_{i,t}}:=\psi_i(s_i^k(t),a_i^k(t))$ and $\textcolor{revisionblue}{\psi_{i,t+1}}:=\psi_i(s_i^k(t+1),a_i^k(t+1))$ for all $i$.
    \STATE Calculate TD Error: $\delta_i^t:=r_i^t-\mu_i^t+(\textcolor{revisionblue}{\psi_{i,t+1}}-\textcolor{revisionblue}{\psi_{i,t}})^\top w_i^t$ for all $i$.
    \STATE Update: $\mu_i^{t+1}:=\mu_i^t+c_{ss}\beta_t(r_i^t-\mu_i^t)$ and $w_i^{t+1}:=w_i^t+\beta_t\delta_i^t\textcolor{revisionblue}{\psi_{i,t}}$ for all $i$.
    \STATE For all $i$, define $\widehat g_i^t
        :=
        \nabla_{\theta_i}\log\pi_{\theta_i^t}(a_i(t)\mid s_i(t))
        \Big(\sum_{j\in N_i^k}(\textcolor{revisionblue}{\psi_{j,t}})^\top w_j^t\Big)$.
    \STATE Update the actor, for all $i$, denote $\Gamma_{i,\omega}(\theta_i^t)=\omega_i(s_i)\pi_{\theta_i}(a_i|s_i)
    \left[
    -\log\pi_{\theta_i}(a_i|s_i)
    -
    h_i(\pi_{\theta_i}(\cdot|s_i))
    \right]$,
    \[
        \text{Conduct policy gradient: }\theta_i^{t+1}:=
        \begin{cases}
            \theta_i^t+\alpha_t\widehat g_i^t,
            & \text{unregularized},\\[1mm]
            \theta_i^t+\alpha_t
            \left(\widehat g_i^t+\iota\Gamma_{i,\omega}(\theta_i^t)\right),
            & \text{regularized}.
        \end{cases}
    \]
\ENDFOR
\end{algorithmic}
\end{algorithm}
\begin{assumption}
\label{assump:ergodicity}
    Every global policy $\pi_\theta$ induces a Markov chain that is irreducible and aperiodic over $S_g \times A_g$. We additionally assume that the uniform restricted critic contraction gap $\Delta_{\min}^{\chi}>0$, as specified in \eqref{eq:min_constant}.
\end{assumption}
\begin{assumption}[Uniform mixing]
Let $\textcolor{revisionblue}{P_{\pi_\theta}}$ be the frozen-policy Markov kernel of the global state-action process
$(s_g(t),a_g(t))$. There exist constants $m_{\mathrm{mix}}<\infty, \rho_{\mathrm{mix}}\in(0,1),$ such that $\sup_{\theta}\sup_x
d_{\mathrm{TV}}\!\left(\textcolor{revisionblue}{P_{\pi_\theta}}^\ell(x,\cdot),\nu^\theta\right)
\le
m_{\mathrm{mix}}\rho_{\mathrm{mix}}^\ell, \ \ell\ge 0.$
\label{assump:uniform_mixing}
\end{assumption}
\begin{assumption}[Bounded critic iterates]
\label{assump:bounded_critic}
Let $\Theta_{\mathrm{ac}}$ denote the actor-parameter region considered in the
actor proofs.  There exists $W<\infty$ such that, for every $i\in\mathcal N$, $\sup_t \|w_i^t\|_2\le W.$
\end{assumption}
\begin{remark}
This is a standard assumption in the finite time analysis of actor critic algorithms. It is typically enforced through an additional projection step after the critic update. See ~\citet{wu2020finite,chen2023finite}.
\end{remark}

\begin{assumption}[Spatial decay condition for actor localization]
    \label{assump:actor-spatial-decay}
    The local interaction coefficients satisfy the hypothesis of
    Theorem~\ref{thm:spatial-decay-sufficient-condition}, and therefore satisfy $(c,\rho)$ exponential decay.
\end{assumption}

\begin{remark}
    Average-reward critics are identifiable only up to additive constants. Thus the full-space negative-definiteness assumption often used in average-reward actor-critic analyses is not natural here; in the tabular case it cannot hold because \((I-P_{\pi_\theta})\mathbf 1=0\). We project out this direction and leverage the contraction on \(S_{\Psi_i,e}^{\perp}\), the critic subspace orthogonal to constant shifts. Unlike prior works, which impose stronger conditions such as smoothness of the stationary distribution, we require only Lipschitzness of several key quantities, all derived directly from the softmax policy structure, uniform mixing, and ergodicity.
\end{remark}

\subsection{Analysis of the Critic}
The actor update is coupled with a critic whose target changes over time, so that the critic tracks $\bigl(J_i(\theta_t),w_i^\star(\theta_t)\bigr)$. Since differential value functions are unique up to an additive constant, we define the Lyapunov function $\textcolor{revisionblue}{\mathcal V_t}$ with respect to the projected critic: let $\textcolor{revisionblue}{\Pi_i^{\perp}}$ denote the Euclidean projection that removes the additive-constant direction of agent $i$. Define $\widetilde \mu_i^t := \mu_i^t-J_i(\theta_t)$ and $\widetilde w_i^t := \textcolor{revisionblue}{\Pi_i^{\perp}}\bigl(w_i^t-w_i^\star(\theta_t)\bigr),$ and finally $ \textcolor{revisionblue}{\mathcal V_t}:=\sum_{i=1}^n
    \left(q_i(\widetilde \mu_i^t)^2+\|\widetilde w_i^t\|_2^2\right),$
where the fixed weights \(q_i>0\) are chosen so that the frozen-policy
average-reward critic is uniformly contractive. Let
\begin{equation}
    \Delta_{\min}^{\chi}
    \coloneqq
    \min\nolimits_{1\le i\le n} \, 
    \inf\nolimits_\theta \, 
    \inf\nolimits_{{v\in S_{\Psi_i,e}^{\perp} , \,  \|v\|_2=1}}
    v^\top \chi_i^\top D_{\pi_\theta}(I-P_{\pi_\theta})\chi_i v
    >0.
    \label{eq:min_constant}
\end{equation}
Note that for a fixed policy $\min_i \Delta_i^{\chi_i}>0$, where $\Delta_i^{\chi_i}$ is agent $i$'s fixed-policy critic contraction coefficient. Assumption~\ref{assump:ergodicity} is to ensure it uniformly across all policies.
Furthermore, define $\textcolor{revisionblue}{\lambda_\chi}
    :=
    \frac{\Delta_{\min}^{\chi}}{4}$ 
The main source of contraction for the critic is $\Delta_{\min}^{\chi}$, which is the restricted projected Bellman gap of the global probability transition matrix. 

For deterministic critic stepsizes \(0<\beta_t\le \bar\beta\le 1\) and a window
\(\tau_t\le t\), the quantity $\Gamma_t(\alpha_t,\beta_t)
    :=
    \rho_{\rm mix}^{\tau_t-1}
    \!+\! \sum_{\ell=t-\tau_t}^{t-1}\!\beta_\ell
    \!+\! \left(1\!+\!\tau_t\right)\sum_{\ell=t-\tau_t}^{t-1}\alpha_\ell $
collects all finite-window errors: the residual mixing error, critic drift within the
window, and actor-induced movement of the kernel and stationary distribution. 
\begin{lemma}[Moving-target critic drift]\label{lem:main_moving_target_critic_drift}
Suppose Assumptions~\ref{assump:ergodicity},\ref{assump:uniform_mixing},\ref{assump:bounded_critic} and \ref{assump:actor-spatial-decay} hold. Let $c_{ss}\beta_t\le 1.$ There exist constants \(C,c_0<\infty\), independent of \(t,\alpha_t,\beta_t\), s.t.
whenever $\beta_t+\Gamma_t\le c_0,$ we have,
\begin{equation}
    \mathbb E[\textcolor{revisionblue}{\mathcal V_{t+1}}]
    \le
    (1-\textcolor{revisionblue}{\lambda_\chi}\beta_t)\mathbb E[\textcolor{revisionblue}{\mathcal V_t}]
    +
    C\left(
        \beta_t^2
        +\beta_t\Gamma_t (\alpha_t,\beta_t)
        +{\alpha_t^2}/{\beta_t}
    \right).
\end{equation}
\end{lemma}
The term \(\beta_t^2\) is the usual finite-stepsize TD noise. The term
\(\beta_t\Gamma_t\) is the Markovian and moving-policy bias accumulated over the mixing
window. The term \(\alpha_t^2/\beta_t\) is the cost of comparing the next critic iterate
to the new target \(w_i^\star(\theta_{t+1})\) rather than the old target
\(w_i^\star(\theta_t)\). Thus, critic is controlled by contraction
\(\textcolor{revisionblue}{\lambda_\chi}\), the mixing rate \(\rho_{\rm mix}\), the critic stepsize
\(\beta_t\), and the actor-to-critic speed ratio \(\alpha_t/\beta_t\).

For fixed step size $\alpha_t\equiv \alpha$, $\beta_t\equiv\beta$, we can show that:  $\frac1T\sum_{t=0}^{T-1}\mathbb E[\textcolor{revisionblue}{\mathcal V_t}]
    =
    O\!\left(
        \frac{\mathbb E[\textcolor{revisionblue}{\mathcal V_0}]}{\textcolor{revisionblue}{\lambda_\chi}\beta T}
        +\beta
        +\Gamma_t
        +\frac{\alpha^2}{\beta^2}
    \right).$ For diminishing step sizes, choose $\beta_t = \frac{\beta_0}{(t+1)^b}$, $\alpha_t = \frac{\alpha_0}{(t+1)^a}$ with $1/2<b<1$ and $\frac{1+b}{2}<a<1$ with $\beta_0$ small enough such that $c_{ss}\beta_t\le 1.$ Then, for a sufficiently large $t_0$, we can show that $\frac{1}{B_{t_0,T}}
    \sum_{t=t_0}^{T-1}\beta_t\mathbb E[\textcolor{revisionblue}{\mathcal V_t}]
    =
    O\!\left(T^{-(1-b)}\right),$ where $B_{t_0,T}:=\sum_{t=t_0}^{T-1}\beta_t.$ \textcolor{revisionblue}{See Lemma~\ref{lem:moving_actor_constant_beta} and Corollary~\ref{cor:moving_actor_time_varying_beta} in Appendix~\ref{app:moving_target_critic} for the detailed bounds.}

\subsection{Analysis of the Actor without Regularization}
Based on the critic analysis, we provide a convergence guarantee for the unregularized case  in Algorithm~\ref{alg:decentralized_ac_lfa}.
The detailed finite-time statement and proof are provided in Theorem~\ref{thm:unreg-finite-time-stationarity} in Appendix~\ref{subsection:unregularized_overall_convergence}.
\begin{theorem}[Unregularized convergence]
    \label{thm:main-unreg-convergence-summary}
    Suppose Assumptions~\ref{assump:ergodicity}--\ref{assump:actor-spatial-decay} hold. Let
    $\varepsilon_{\mathrm{FA}}$ denote the linear critic approximation error
    from \eqref{eq:error-decomp} \textcolor{revisionblue}{(uniformly defined in \eqref{eq:actor-fa-error-new})}.  The unregularized actor-critic iterates
    satisfy the following:
    \begin{enumerate}
        \item \textbf{Fixed stepsizes.}
        If $\alpha_t=\alpha$, $\beta_t=\beta$ are sufficiently small and let
        $\delta_{\alpha,\beta}:=\max\{\alpha,\beta\}$, $f(\alpha,\beta)\coloneqq \beta\log(1/\delta_{\alpha,\beta})
            +
            \alpha\log^2(1/\delta_{\alpha,\beta})
            +
            \frac{\alpha^2}{\beta^2}$ then, for
        $\overline G_T:=T^{-1}\sum_{t=0}^{T-1}
        \mathbb E\|\nabla J(\theta_t)\|_2^2$,
        \begin{equation}
            \overline G_T
            =
            O\left(
            \frac1{\alpha T}
            \! +\! 
            \frac1{\beta T}
            \! +\! 
            f(\alpha,\beta)
            \! +\! 
            \varepsilon_{\mathrm{FA}}^2
            +\! 
            \rho^{2(k+1)}
            \right).
            \label{eq:main-unreg-fixed-rate}
        \end{equation}
        \item \textbf{Diminishing stepsizes.}
        If $\beta_t=\beta_0(t+1)^{-b}$ and
        $\alpha_t=\alpha_0(t+1)^{-a}$ for constants
        $\alpha_0,\beta_0>0$,
        $\frac12<b<1$ and
        $\frac{1+b}{2}<a<1$,  for every
        $t_0$ greater than some threshold and every $T>t_0$,
        \begin{equation}
            \min_{t_0\le t<T}
            \mathbb E\|\nabla J(\theta_t)\|_2^2
            =
            O\left(
            T^{-(1-\textcolor{revisionblue}{a})}
            +
            \varepsilon_{\mathrm{FA}}^2
            +
            \rho^{2(k+1)}
            \right).
            \label{eq:main-unreg-decaying-rate}
        \end{equation}
    \end{enumerate}
\end{theorem}

Theorem~\ref{thm:main-unreg-convergence-summary} provides average/best-iterate convergence guarantee in terms of the gradient norm for both fixed and diminishing step sizes. For the actor part, we decompose the gradient approximation error into the error induced by Markovian sampling, critic bias, function approximation error, and truncation within the $k$-hop neighborhood. We then bound each individual error term to get a recursion bound for the actor. Finally, we combine the action recursion bound with Lemma~\ref{lem:main_moving_target_critic_drift} to get the overall convergence result. \textcolor{revisionblue}{$f(\alpha,\beta)$ collects critic tracking, Markovian mixing, actor drift, and target motion; hidden-constant conventions are specified in Appendix Theorems~\ref{thm:unreg-finite-time-stationarity} and~\ref{thm:strong-entropy-pl-consequence}.}

\subsection{Analysis of the Actor with Regularization}

We now present the result for the fixed-state entropy-regularized version of Algorithm~\ref{alg:decentralized_ac_lfa}. We first present two additional assumptions for the regularized case:
\begin{assumption}[Uniformly explored regularized actor region]
    \label{assump:strong-entropy-exploration-region}
    There are constants $\delta\in(0,1)$ and an actor-parameter region
    $\Theta_0$ such that the fixed-state regularized actor iterates considered
    in the global-convergence proof stay in $\Theta_0$ and $\pi_{\theta_i}(a_i|s_i)\geq\delta,\forall \theta\in\Theta_0,\ i,\ s_i,\ a_i$.
\end{assumption}

\begin{assumption}[Positive fixed-state entropy weights]
    \label{assump:strong-entropy-positive-weights}
    The fixed distribution used in the entropy regularizer has uniformly
    positive local marginals: $\omega_{\min}
        :=
        \min_{i,s_i}\omega_i(s_i)>0$.
\end{assumption}
The detailed finite-time statement and proof are provided in Theorem~\ref{thm:strong-entropy-pl-consequence} in Appendix~\ref{sec:regularized_convergence_results}.
\begin{theorem}[Regularized global convergence]
    \label{thm:main-regularized-convergence-summary}
    Suppose Assumptions~\ref{assump:ergodicity}--\ref{assump:strong-entropy-positive-weights} hold. There exists some
    $\overline L_J<\infty$ such that if $\iota>\overline L_J/\omega_{\min}$ and denote
    $\eta_{\iota,\omega}:=\delta^2(\iota\omega_{\min}-\overline L_J)$.  Let
    $\Delta_{\iota,\omega,t}:=
    J_{\iota,\omega,\Pi_0}^\star-J_{\iota,\omega}(\theta_t)$, where $\Pi_0$ is
    the uniformly explored policy region induced by $\Theta_0$.  The strongly
    entropy-regularized actor-critic iterates satisfy the following convergence:
    \begin{enumerate}
        \item \textbf{Fixed stepsizes.}
        If $\alpha_t\!=\!\alpha\!\le1/\eta_{\iota,\omega}$,
        $\beta_t\!=\!\beta$ are small enough, then
            \begin{equation}
                \mathbb E\Delta_{\iota,\omega,T} = O\Big(
            e^{-\eta_{\iota,\omega}\alpha T}
            \Delta_{\iota,\omega,0}
            +
            \frac{1}{\eta_{\iota,\omega}}
            \Big[
            e^{-\lambda_\chi\beta T}\mathbb E\mathcal V_0 + f(\alpha,\beta) + \varepsilon_{\mathrm{FA}}^2
            +
            \rho^{2(k+1)}\Big]\Big)
            \end{equation}

        \item \textbf{Sharper diminishing specialization.}
        If $\beta_t=\beta_0(t+1)^{-1/2-\mathfrak c},
            \alpha_t=\alpha_0(t+1)^{-3/4-\mathfrak c}$ for some $
            0<\mathfrak c<\frac14$,
        use $\tau_t=\lceil c_\tau\log(t+1)\rceil$ with
        $c_\tau>(1/2+\mathfrak c)/(-\log\rho_{\mathrm{mix}})$, and choose a fixed finite
        burn-in $t_0$ after which $1\le\tau_t\le t$,
        $\alpha_t\le1/\eta_{\iota,\omega}$, and the critic smallness conditions
        hold.  With $A_{t_0,T}:=\sum_{t=t_0}^{T-1}\alpha_t=\Theta(T^{1/4-\mathfrak c})$,
        \begin{equation}
            \mathbb E\Delta_{\iota,\omega,T}
            =
            O\Big(
            e^{-\eta_{\iota,\omega}A_{t_0,T}}
            \mathbb E\Delta_{\iota,\omega,t_0}
            +
            \frac{1}{\eta_{\iota,\omega}}
            \Big[
            T^{-1/2}\log T
            +
            \varepsilon_{\mathrm{FA}}^2
            +
            \rho^{2(k+1)}
            \Big]
            \Big).
            \label{eq:main-reg-decaying-sharp-gap-rate}
        \end{equation}
    \end{enumerate}
\end{theorem}

Theorem~\ref{thm:main-regularized-convergence-summary} provides a last-iterate global convergence guarantee in terms of the regularized value gap $\Delta_{\iota,\omega,t}$ for both fixed and diminishing step sizes.
Compared with the unregularized guarantee in Theorem~\ref{thm:main-unreg-convergence-summary}, Theorem~\ref{thm:main-regularized-convergence-summary} directly controls the value gap to the regularized optimum rather than only the policy-gradient norm, and therefore rules out suboptimal stationary points created by the product-policy and partial-observability obstructions discussed above.
Additionally, the regularized actor-critic admits a stretched-exponential convergence rate for its recursion term, although the overall convergence rate \textcolor{revisionblue}{above the feature-approximation and localization floor} is still $O(T^{-1/2}\log T)$ due to the error induced by mixing.

\textit{Proof sketch.}
The proof reuses the unregularized critic analysis, because fixed-state entropy only adds an exact local entropy-gradient term to the actor update. The actor proof establishes a regularized drift inequality with the same error terms as unregularized, and converts this drift to a last-iterate value-gap recursion once a product-policy PL-type inequality holds. This PL-type inequality is verified in direct policy variables: the curvature of $J$ is bounded by $\overline L_J$, while the entropy term contributes negative curvature at least $\iota\omega_{\min}$; hence $\iota\omega_{\min}>\overline L_J$ gives strong concavity, a direct-policy PL-type inequality, and finally a softmax-logit PL constant $2\delta^2(\iota\omega_{\min}-\overline L_J)$. Full proof can be found in Appendix~\ref{sec:regularized_convergence_results}.

\section{Numerical Illustration}
\label{sec:numerical-illustration}

The error decomposition~\eqref{eq:error-decomp} suggests a qualitative
tradeoff: a larger critic radius reveals neighbor information but makes a
fixed-dimensional representation compress a larger neighborhood.  We
illustrate this with a finite-state system of interacting maintenance units.
Each unit has a binary health state, chooses a binary intervention action, and
records its previous action as a one-step memory variable.  Its local reward
favors healthy operation while penalizing intervention, and its health
transition depends on its own state and action together with the health and
previous actions of adjacent units.  We test paths with
$N\in\{8,16,32,64\}$ agents and square grids with side length
$L\in\{3,4,6,8\}$, hence $N\in\{9,16,36,64\}$ agents.  We sweep
$k\in\{0,1,2,3\}$ and pairwise-hashed feature dimensions
$d_i\in\{4,8,16,32,64\}$, with a tabular critic for comparison.
Each cell uses $3{,}000$ training steps and a fresh $300$-step greedy
evaluation; each plotted point pools the four listed sizes and three seeds.
Full protocol details and additional sweeps are in
Appendix~\ref{app:numerical-localization-compression}.

\begin{figure}[!htbp]
    \centering
    \includegraphics[width=\textwidth]{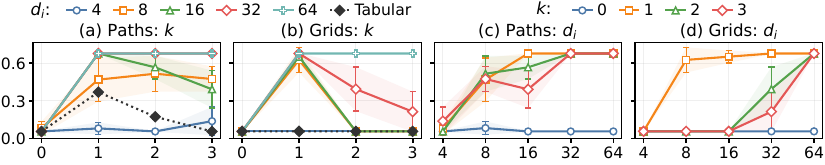}
    \caption{Greedy reward per agent versus $k$ (a,b) and $d_i$ (c,d).
    Means pool 12 runs (four sizes, three seeds each); bars show
    descriptive normal-approximation 95\% confidence intervals.}
    \label{fig:main-experiment-tradeoff}
\end{figure}

At $d_i=16$, paths score $0.676\pm0.002$ at $k=1$ but $0.392\pm0.148$ at
$k=3$, while grid $k=2,3$ stay near $0.058$; all positive radii reach about
$0.676$ by $d_i=32$ on paths or $d_i=64$ on grids.  Tabular stays near $0.058$
on grids and peaks at only $0.369\pm0.006$ on paths.  Under the same trajectory
budget, the tabular critic fails to learn effectively, showing that function
approximation is necessary in this tested growing-neighborhood regime.  Thus,
larger neighborhoods help only when encoded: grid $k=2,3$ match $k=1$ only at
$d_i=64$, and path $k=3$ matches $k=1$ by $d_i=32$.

\textbf{Limitations and Future Work.} A key limitation is that our guarantees rely on the exponential decay property in the network: they apply only to settings where each agent’s state transition is influenced only by its immediate neighbors, so relaxing this local weak-dependence structure is an important future direction. Fixed-state entropy regularization should also be viewed as a remedy rather than a universal fix, as our global convergence guarantee requires the entropy to be strong. Future work includes identifying different policy classes (in addition to localized policies) and structural assumptions on the global transition kernel of the multi-agent MDPs under which decentralized policy-optimization methods have more favorable landscapes and may admit either global convergence guarantees or finite-time convergence to zero Nash gap.

\clearpage
\begin{ack}
\textcolor{revisionblue}{Adam Wierman acknowledges support from the National Science Foundation under grant CCF-2326609.}
\end{ack}

\bibliographystyle{plainnat}   
\bibliography{references}

\newpage
\appendix

\section*{Appendix Contents}
\begingroup
\small
\renewcommand{\arraystretch}{1.14}
\noindent
\begin{tabular*}{\textwidth}{@{}p{0.12\textwidth}@{\extracolsep{\fill}}p{0.74\textwidth}r@{}}
\toprule
\textbf{App.} & \textbf{Entry} & \textbf{Page} \\
\midrule
\ref{app:table-of-notations} & \hyperref[app:table-of-notations]{\textbf{Table of Notations}} & \pageref{app:table-of-notations} \\
\ref{app:related-work} & \hyperref[app:related-work]{\textbf{Related Work}} & \pageref{app:related-work} \\
\ref{app:suboptimal_convergence_example} & \hyperref[app:suboptimal_convergence_example]{\textbf{Example of Policy Gradient Converging to a Suboptimal Stationary Point}} & \pageref{app:suboptimal_convergence_example} \\
\ref{app:networked_partial_observability_example} & \hyperref[app:networked_partial_observability_example]{\textbf{A Network-Compatible Partial-Observability Example}} & \pageref{app:networked_partial_observability_example} \\
\addlinespace
\ref{app:unregularized-convergence} & \hyperref[app:unregularized-convergence]{\textbf{Convergence Proof without Regularization}} & \pageref{app:unregularized-convergence} \\
\ref{app:unregularized-critic} & \quad\hyperref[app:unregularized-critic]{Convergence of the Critic} & \pageref{app:unregularized-critic} \\
\ref{app:unregularized-actor} & \quad\hyperref[app:unregularized-actor]{Convergence of the Actor} & \pageref{app:unregularized-actor} \\
\ref{subsection:unregularized_overall_convergence} & \quad\hyperref[subsection:unregularized_overall_convergence]{Overall Convergence} & \pageref{subsection:unregularized_overall_convergence} \\
\addlinespace
\ref{sec:regularized_convergence_results} & \hyperref[sec:regularized_convergence_results]{\textbf{Convergence Proof with Regularization}} & \pageref{sec:regularized_convergence_results} \\
\ref{app:regularized-critic} & \quad\hyperref[app:regularized-critic]{Convergence of the Critic} & \pageref{app:regularized-critic} \\
\ref{app:regularized-actor} & \quad\hyperref[app:regularized-actor]{Convergence of the Actor} & \pageref{app:regularized-actor} \\
\ref{sec:strong-entropy-pl-route} & \quad\hyperref[sec:strong-entropy-pl-route]{A Strong-Entropy Route to the Product-Policy PL Inequality} & \pageref{sec:strong-entropy-pl-route} \\
\addlinespace
\ref{sec:auxiliary-lemmas} & \hyperref[sec:auxiliary-lemmas]{\textbf{Auxiliary Lemmas}} & \pageref{sec:auxiliary-lemmas} \\
\bottomrule
\end{tabular*}
\endgroup

\begingroup
\color{revisionblue}
\paragraph{Proof map.}
\begin{enumerate}
    \item \textbf{Fixed-policy critic bound after \eqref{eq:error-decomp}.}
    Theorem~\ref{thm:fixed-policy-td-learning-bounds} combines the localization comparison in Lemma~\ref{lem:local-truncation} with fixed-policy TD contraction in Lemma~\ref{lem:fixed_actor_ss}.
    \item \textbf{Moving-target critic (Lemma~\ref{lem:main_moving_target_critic_drift}).}
    Lemma~\ref{lem:moving_actor_constant_beta} and Corollary~\ref{cor:moving_actor_time_varying_beta} use critic contraction (Lemma~\ref{lem:fixed_actor_ss}), policy Lipschitz bounds (Lemma~\ref{lem:lipschitzness}), bounded actor movement (Lemma~\ref{lem:bounded-actor-increment}), and finite-window mixing.
    \item \textbf{Unregularized convergence (Theorem~\ref{thm:main-unreg-convergence-summary}).}
    Theorem~\ref{thm:unreg-finite-time-stationarity} combines the actor estimates in Proposition~\ref{prop:actor-stationarity} and Lemmas~\ref{lem:actor-gradient-error}--\ref{lem:actor-drift} with critic tracking in Lemma~\ref{lem:moving_actor_constant_beta} and Corollary~\ref{cor:moving_actor_time_varying_beta}.
    \item \textbf{Regularized convergence (Theorem~\ref{thm:main-regularized-convergence-summary}).}
    Theorem~\ref{thm:strong-entropy-pl-consequence} combines regularized actor analysis (Theorem~\ref{thm:regularized-actor-convergence}; Lemmas~\ref{lem:fixed-state-entropy-gradient-decomposition}--\ref{lem:fixed-state-regularized-actor-drift}), curvature and PL bounds (Lemmas~\ref{lem:strong-entropy-curvature-finite}--\ref{lem:strong-entropy-softmax-pl}), and critic tracking (Lemma~\ref{lem:moving_actor_constant_beta}; Corollary~\ref{cor:moving_actor_time_varying_beta}).
\end{enumerate}
\endgroup

\section{Table of Notations}
\label{app:table-of-notations}

Symbols that are local to a single lemma or proof are defined near their first use.
Table~\ref{tab:notation} collects the notation used repeatedly across the main text and appendix.

\begingroup
\small
\renewcommand{\arraystretch}{1.14}
\begin{longtable}{@{}p{0.27\textwidth}p{0.67\textwidth}@{}}
\caption{Recurring notation used throughout the paper.}
\label{tab:notation}\\
\toprule
\textbf{Symbol} & \textbf{Meaning} \\
\midrule
\endfirsthead
\toprule
\textbf{Symbol} & \textbf{Meaning} \\
\midrule
\endhead
\midrule
\multicolumn{2}{r}{\emph{Continued on next page}}\\
\endfoot
\bottomrule
\endlastfoot

\multicolumn{2}{@{}l}{\textbf{Network and state-action spaces}}\\
\addlinespace
\(\mathcal G=(\mathcal N,\mathcal E)\) & Undirected interaction graph; \(\mathcal N=\{1,\ldots,n\}\) is the set of agents and \(\mathcal E\) is the edge set. \\
\(i,j\), \(n\) & Agent indices and total number of agents. \\
\(N_i\), \(N_i^k\) & The one-hop neighborhood of agent \(i\), including itself, and the \(k\)-hop neighborhood of \(i\). \\
\(N_{-i}^k\) & Agents outside \(N_i^k\), i.e., \(\mathcal N\setminus N_i^k\). \\
\(k\) & Localization radius. \\
\(\mathcal S_i,\mathcal A_i\) & Local state and action spaces of agent \(i\). \\
\(s_i,a_i\) & Local state and action of agent \(i\). \\
\(\mathcal S_g,\mathcal A_g\) & Global state and action spaces, \(\mathcal S_g=\prod_i\mathcal S_i\) and \(\mathcal A_g=\prod_i\mathcal A_i\). \\
\(s_g,a_g\) & Global state and action tuples. \\
\(\mathcal S_i^k,\mathcal A_i^k\) & Joint state and action spaces restricted to the agents in \(N_i^k\). \\
\(s_i^k,a_i^k\) & \(k\)-hop local state-action tuple observed by agent \(i\). \\
\(\mathcal S_{-i}^k,\mathcal A_{-i}^k\) & Joint state and action spaces outside \(N_i^k\). \\
\(s_{-i},a_{-i}\) & State and action tuples of all agents except \(i\). \\
\(\Delta(\mathcal A_i)\) & Probability simplex over the action set \(\mathcal A_i\). \\

\addlinespace
\multicolumn{2}{@{}l}{\textbf{Dynamics, policies, and distributions}}\\
\addlinespace
\(P_i(\cdot\mid s_i^1,a_i)\) & Local transition kernel for agent \(i\). \\
\(P(s_g'\mid s_g,a_g)\) & Global transition kernel, equal to the product of the local transition kernels. \\
\(\textcolor{revisionblue}{P_{\pi_\theta}^{\mathrm{s}}}\) & Global state transition kernel induced by the product policy \(\pi_\theta\). \\
\(\textcolor{revisionblue}{P_{\pi_\theta}}\) & Frozen-policy Markov kernel of the global state-action process. \\
\(\pi_{\theta_i}\) & Local softmax policy of agent \(i\), parameterized by logits \(\theta_i\). \\
\(\pi_\theta\) & Global product policy, \(\pi_\theta(a_g\mid s_g)=\prod_i\pi_{\theta_i}(a_i\mid s_i)\). \\
\(\theta_i,\theta\) & Local actor parameters and joint actor parameter \(\theta=(\theta_1,\ldots,\theta_n)\). \\
\(d^\theta\) & Stationary distribution over global states under \(\pi_\theta\). \\
\(\nu^\theta\) & Stationary distribution over global state-action pairs under \(\pi_\theta\). \\
\(D_{\pi_\theta}\) & Diagonal matrix with diagonal entries given by \(\nu^\theta\). \\
\(\bar\nu^\theta\) & Stationary one-step law used in the moving-target critic analysis. \\
\(m_{\mathrm{mix}}\) & Uniform geometric-mixing prefactor used in the main text and appendix. \\
\(\rho_{\mathrm{mix}}\) & Uniform geometric-mixing rate of the frozen kernels. \\
\(\mathrm{TV}(\cdot,\cdot)\) & Total variation distance. \\
\(S_i(\theta_i;s_i,a_i)\) & Local score \(\nabla_{\theta_i}\log\pi_{\theta_i}(a_i\mid s_i)\). \\
\(S(\theta;s_g,a_g)\) & Global product-policy score \(\nabla_\theta\log\pi_\theta(a_g\mid s_g)\). \\
\(G_i,G\) & Uniform bounds on local and global softmax scores; for tabular softmax, \(G_i=\sqrt 2\) and \(G=\sqrt{2n}\). \\
\(L_{S,i},L_K,L_\nu\) & Lipschitz constants for local scores, frozen kernels, and stationary laws. \\

\addlinespace
\multicolumn{2}{@{}l}{\textbf{Rewards, values, objectives, and gaps}}\\
\addlinespace
\(r_i(s_i,a_i)\) & Local reward of agent \(i\). \\
\(r(s_g,a_g)\) & Global reward, \(r(s_g,a_g)=\sum_i r_i(s_i,a_i)\). \\
\(\bar r,r_{\max},R\) & Reward bounds; \(R\) is often used for a global reward bound such as \(n r_{\max}\). \\
\(J(\theta)\) & Average global reward objective. \\
\(J_i(\theta)\) & Local average reward contribution of agent \(i\), with \(J(\theta)=\sum_iJ_i(\theta)\). \\
\(J_{\max},J^\star\) & Upper bound on \(J\) and optimal unregularized value over the stated policy class. \\
\(Q^\theta(s_g,a_g)\) & Global differential action-value function for the average-reward objective. \\
\(Q_i^\theta(s_g,a_g)\) & Local differential action-value function associated with reward \(r_i\). \\
\(V^\theta(s_g)\) & Global differential state-value function. \\
\(\bar A_i^\theta(s_g,a_i)\) & One-agent averaged advantage used in the partial-observability and Nash-gap arguments. \\
\(\mathrm{NashGap}_i(\theta)\) & Maximum improvement available to agent \(i\) by a unilateral policy deviation. \\
\(\omega,\omega_i\) & Fixed state distribution used for entropy regularization and its local marginals. \\
\(h_i(p)\) & Local entropy, \(h_i(p)=-\sum_{a_i}p(a_i)\log p(a_i)\). \\
\(\mathcal H_\omega(\theta)\) & Fixed-state entropy functional. \\
\(\iota\) & Entropy-regularization weight. \\
\(J_{\iota,\omega}(\theta)\) & Fixed-state entropy-regularized objective \(J(\theta)+\iota\mathcal H_\omega(\theta)\). \\
\(\Gamma_{i,\omega}(\theta_i)\) & Exact local gradient of the fixed-state entropy term for actor block \(i\). \\
\(\Theta_0,\Pi_0,\delta\) & Uniformly explored actor-parameter and product-policy regions; \(\delta\) lower bounds local action probabilities on the region. \\
\(\omega_{\min}\) & Minimum local entropy weight \(\min_{i,s_i}\omega_i(s_i)\). \\
\(\overline L_J\) & Direct-policy curvature upper bound for \(J\) on \(\Pi_0\). \\
\(\eta_{\iota,\omega}\) & Strong-entropy contraction parameter \(\delta^2(\iota\omega_{\min}-\overline L_J)\). \\
\(\Delta_{\iota,\omega,t}\) & Regularized value gap \(J_{\iota,\omega,\Pi_0}^\star-J_{\iota,\omega}(\theta_t)\). \\
\(J_{\iota,\omega,\Pi_0}^\star\) & Optimal fixed-state regularized value over \(\Pi_0\). \\
\(H_{\omega,\max}\) & Entropy upper bound \(\sum_i\log|\mathcal A_i|\). \\
\(G_\Pi(\pi)\) & Tangent-space projection of the direct-policy gradient. \\
\(\mu_{\iota,\omega}^{\Pi},\mu_{\iota,\omega,\theta}\) & Direct-policy and softmax-logit PL constants for the regularized objective. \\

\addlinespace
\multicolumn{2}{@{}l}{\textbf{Localization and critic approximation}}\\
\addlinespace
\(C_{ij}\) & Local interaction coefficient measuring the sensitivity of agent \(i\)'s transition kernel to agent \(j\). \\
\((c,\rho)\) & Exponential-decay constants; localization errors scale as \(O(c\rho^{k+1})\). \\
\(\widetilde Q_i^\theta\) & Truncated \(k\)-hop approximation of \(Q_i^\theta\). \\
\(\widehat Q_i,\widehat Q_i^\star\) & Tabular localized critic and its projected Bellman fixed point. \\
\(\phi_i\) & Local-to-global lifting matrix for \(k\)-hop tabular critics. \\
\(\psi_i,\Psi_i\) & Local feature map and corresponding feature matrix for linear function approximation. \\
\(d_i\) & Feature dimension for agent \(i\)'s linear critic. \\
\(\chi_i\) & Composite global feature matrix \(\chi_i=\phi_i\Psi_i\). \\
\(\mathcal T_{\pi_\theta}^i\) & Local average-reward Bellman operator for agent \(i\). \\
\(\Pi_{D_{\pi_\theta}}^{\phi_i},\Pi_{D_{\pi_\theta}}^{\chi_i}\) & \(D_{\pi_\theta}\)-weighted projections onto the localized tabular and linear critic spaces. \\
\(e,e_S,w_e\) & All-ones vectors on state-action and state spaces, and a critic coefficient vector satisfying \(\Psi_iw_e=e\) when it exists. \\
\(S_{\Psi_i,e},S_{\Psi_i,e}^{\perp}\) & Constant-shift subspace of critic coefficients and its Euclidean orthogonal complement. \\
\(\textcolor{revisionblue}{\Pi_i^{\perp}}\) & In critic proofs, Euclidean projection onto \(S_{\Psi_i,e}^{\perp}\); this is distinct from the local transition kernel \(P_i(\cdot\mid\cdot)\). \\
\(w_i^\star(\theta)\) & Normalized projected Bellman fixed point tracked by agent \(i\)'s linear critic. \\
\(\Delta_i^{\phi_i},\Delta_i^{\chi_i}\) & Restricted contraction coefficients for tabular localized and linear critic spaces. \\
\(\Delta_{\min}^{\chi}\) & Uniform minimum of the linear-critic contraction coefficients across agents and actor parameters. \\
\(\textcolor{revisionblue}{\lambda_\chi}\) & Critic contraction constant, typically \(\Delta_{\min}^{\chi}/4\). \\
\(\varepsilon_{\mathrm{FA}},\varepsilon_{\mathrm{FA},i}\) & Uniform and agent-wise linear function-approximation errors. \\
\(C_i^{\mathrm{tr}}\) & Agent-wise truncation-error constant in the fixed-policy TD bound. \\
\(\Lambda_i\) & Maximum eigenvalue \(\lambda_{\max}(\chi_i^\top D_{\pi_\theta}\chi_i)\). \\
\(\bar\psi\) & Uniform bound on the local feature norms. \\
\(W,W_\star,W_{\mathrm{act}}\) & Bounds on critic iterates, critic fixed points, and their maximum. \\
\(H_i,B_i^W\) & Neighborhood size \(H_i=|N_i^k|\) and actor-proof critic magnitude bound \(B_i^W=|N_i^k|W_{\mathrm{act}}\). \\

\addlinespace
\multicolumn{2}{@{}l}{\textbf{Algorithmic and finite-time analysis quantities}}\\
\addlinespace
\(\alpha_t,\beta_t\) & Actor and critic step sizes. \\
\(c_{ss}\) & Average-reward estimator speed constant in the critic update. \\
\(\mu_i^t,w_i^t\) & Agent \(i\)'s average-reward estimate and critic weights at time \(t\). \\
\(\delta_i^t\) & TD error used by agent \(i\)'s critic update. \\
\(\textcolor{revisionblue}{\psi_{i,t}},\textcolor{revisionblue}{\psi_{i,t+1}}\) & Current and next local feature vectors in Algorithm~\ref{alg:decentralized_ac_lfa}. \\
\(\widehat q_j^t\) & Current scalar critic prediction \(\psi_j(s_j^k(t),a_j^k(t))^\top w_j^t\). \\
\(\widehat g_i^t\) & Unregularized stochastic actor direction for agent \(i\). \\
\(\widehat g_{\iota,\omega,i}^t\) & Fixed-state regularized stochastic actor direction for agent \(i\). \\
\(\widetilde\mu_i^t,\widetilde w_i^t\) & Centered average-reward error and projected critic tracking error. \\
\(\textcolor{revisionblue}{\mathcal V_t}\) & Critic Lyapunov function. \\
\(q_i\) & Positive scalar weight in the critic Lyapunov function. \\
\(\tau_t,t_\tau\) & Mixing window length and delayed time \(t_\tau=t-\tau_t\). \\
\(B_t^\beta,A_t^\beta\) & Accumulated critic and actor step sizes over a mixing window. \\
\(D_{\theta,t},D_{w,t},D_{K,t}\) & Actor, critic, and kernel drift proxies over the delayed window. \\
\(\varepsilon_{\mathrm{mix}}(t)\) & Conditional Markovian mixing error for delayed actor estimates. \\
\(\Gamma_t(\alpha_t,\beta_t)\) & Main-text moving-target error scale combining residual mixing, critic drift, and actor-induced drift. \\
\(\varepsilon_{P,t}^{\mathrm{mov}}\), \(\varepsilon_{P,\beta,t}^{\mathrm{mov}}\) & Detailed moving-target critic error scales for decaying and fixed critic step sizes. \\
\(A_{t_0,T},B_{t_0,T}\) & Cumulative actor and critic step sizes over the horizon \(t_0,\ldots,T-1\). \\
\(\overline G_T,\overline G_{t_0,T}^{\alpha}\) & Fixed-step and weighted averaged policy-gradient norms. \\
\(\delta_{\alpha,\beta}\) & Maximum fixed step size, \(\max\{\alpha,\beta\}\). \\
\(L_{\alpha,\beta}\) & Stepsize-dependent logarithmic mixing window. \\
\(\mathcal R_{\iota,\omega,t}\) & One-step error aggregate in the regularized actor drift. \\
\end{longtable}
\endgroup

\section{Related Work}
\label{app:related-work}

In this section we provide a detailed discussion on related work.

\paragraph{Networked multi-agent reinforcement learning.}
Our work lies in the line of literature that designs scalable multi-agent reinforcement learning algorithms for networked systems.

Within this domain, we especially follow the framework of \citet{qu2022scalable} that leverages spatial correlation decay to achieve provable convergence guarantees. \citet{qu2020scalable} extended this local interaction structure of networks to establish an exponential spatial decay property for policy gradients in the average-reward setting. \citet{lin2021multi} further considered environments with stochastic network transitions, demonstrating that spatial decay properties can be leveraged to bound approximation errors independently of the total network size. Building upon this framework, \citet{zhang2023global} considered a wider class of policies that depends on all states in the $k$-hop neighborhood, and proved the convergence of localized policy iteration to a globally optimal policy that potentially depends on the entire global state. Furthermore, \citet{zhou2023convergence} extended this fully cooperative framework to a class of networked Markov potential games, and established finite-time global convergence rates for localized actor-critic algorithms, and more recently, \citet{ren2024scalable} leveraged spectral representations of the underlying network topology to efficiently compress high-dimensional state and action spaces.
\citet{faizal2026local} study discounted graph-structured MDPs with local rewards and transitions. Under influence-decay conditions, they establish optimal-policy approximation guarantees for cooperative policies. Their multi-agent construction selects a joint action by maximizing a sum of neighborhood-dependent $Q$-factors. The resulting policy can use neighboring states and coordinated action selection. Their multi-agent result establishes existence of a good cooperative policy, while efficient computation of its surrogate remains open; their single-active-agent setting also admits constructive algorithms.

Meanwhile, we note that there are other models for the networked MARL problem that models the interaction between agents differently. \citet{zhang2018fully} first proposed the framework of networked multi-agent MDP without considering localization of policy and $Q$ functions in the network. \citet{liu2025scalable} studied local convergence distributed policy gradient and actor-critic in the discounted setting.

Despite this progress, finite-time convergence results in average-reward networked MARL leveraging the spatial decay properties has never been established.

\paragraph{Average-reward reinforcement learning.}
Classical dynamic-programming treatments of average-cost MDPs are given in \cite{bertsekas2011dynamic}. More recently, finite-time policy-optimization guarantees have been developed for single-agent average-reward MDPs, including vanilla policy-gradient methods~\cite{kumar2024globalconvergencepolicygradient} and natural-policy-gradient or mirror-descent-type methods~\citep{murthy2023convergence,grosof2024convergence,even2009online}.

Asymptotic analysis of actor critic algorithms comprising of both policy gradient and natural policy gradient goes back to~\cite{konda1999actor, bhatnagar2009natural}. Finite-sample analyses of average-reward TD and Q-learning were developed by \cite{zhang2021finite}, and finite-time actor-critic guarantees were later obtained under two-timescale stepsizes by \cite{wu2020finite} and under single-timescale Markovian sampling by \cite{chen2023finite}. Our critic analysis builds on these standard average-reward TD and stochastic-approximation ideas, but applies them to a collection of localized network critics. These results have been used to derive finite-sample complexity guarantees for popular algorithms such as natural policy gradient ~\citep{abbasi2019politex, murthy2023performance, murthy2024performance} in the single agent setting.

Average-reward MARL has also been studied in fully decentralized communication-based settings. 
 \cite{hairi2022finite} prove finite-time convergence for a consensus-based fully decentralized
actor-critic algorithm in cooperative average-reward MARL, where agents communicate local
TD-error information to estimate a global actor direction. \cite{hairi2024sample} study the
related policy-evaluation problem and show that multiple local TD steps between communication
rounds can reduce communication complexity. These works address decentralized communication
and consensus in average-reward MARL. Our focus is different: we study networked MDPs with
spatially local state-action information, where each agent uses a \(k\)-hop localized critic rather than communicating to learn a common global critic. This leads to explicit localization and function-approximation error floors in the actor-critic convergence bound.

\paragraph{Other related work.} In addition to networked and average-reward settings, we also briefly acknowledge a wider body of literature.
\citet{mei2020global} established global convergence rates for softmax policy gradients for bandits and discounted RL.
\citet{zhang2022global} provided global convergence results for softmax (natural) policy gradients for a class of Markov potential games.
\citet{deweese2026revisiting} optimize over the full simplex of distributions on a restricted deterministic policy class in discounted MDPs. Their evaluation samples a deterministic policy, holds it fixed for several time steps, and then resamples. Under smoothness, their exact-gradient guarantees control the expected performance of a final sampled deterministic policy relative to the best deterministic member of that class, with a residual decreasing exponentially in the temporal horizon. Their multi-step horizon measures time rather than graph distance.
\citet{liu2022sample} studies sample complexity in partially observable Marko games, proposing the Optimism and Maximum Likelihood Estimation (OMLE) algorithm to establish polynomial sample complexity bounds for learning approximate Nash, correlated, and coarse correlated equilibria under weak revealability assumptions.
\citet{hong2023policy} provided finite-sample (non-asymptotic) convergence guarantees for policy gradients utilizing history-dependent policies in continuous-space POMDPs.
\citet{forkel1high} investigated high entropy in Dec-POMDPs and found that high fixed-state entropy regularization helps the convergence of policy gradient methods to symmetry equivariant policies.
While these work share somewhat similar high-level intuition as ours, the specific settings that they consider differ significantly from our networked MDP setting.

\section{Example of Policy Gradient Converging to a Suboptimal Stationary Point}\label{app:suboptimal_convergence_example}
We next construct a cooperative two-agent example showing that even under ergodicity, policy gradient with product policies can converge to a suboptimal local optimum in policy space. In contrast to Example~\ref{example:multiagency}, in this setting the reward assigned to each agent is a function solely of its own state and action, while the transition dynamics remain jointly coupled across the two agents, thereby preserving the interaction structure central to the problem formulation considered in this work. 

\paragraph{Local state and action spaces.}
For each agent \(i\in\{1,2\}\),
\[
\mathcal S_i=\{A,B,C\}, \qquad \mathcal A_i=\{a,b\}.
\]
The one-step reward of agent \(i\) depends only on \((s_i,a_i)\) and is given by
\[
r_i(A,a)=0,\qquad r_i(A,b)=1,\qquad r_i(B,\cdot)=0,\qquad r_i(C,\cdot)=10.
\]
Thus the team objective is
\[
J(\pi)=J_1(\pi)+J_2(\pi).
\]

\paragraph{Local dynamics.}
For each agent \(i\), the local dynamics are as follows.
\begin{itemize}
    \item At state \(A\):
    \[
    A \xrightarrow[a]{0.99} B,\qquad A \xrightarrow[a]{0.01} A,
    \]
    \[
    A \xrightarrow[b]{0.99} A,\qquad A \xrightarrow[b]{0.01} B.
    \]
    \item At state \(B\):
    \begin{itemize}
        \item if the other agent is also at \(B\), then
        \[
        B \to C \text{ with probability }0.9,\qquad B\to A \text{ with probability }0.1;
        \]
        \item if the other agent is not at \(B\), then
        \[
        B \to B \text{ with probability }0.9,\qquad B\to A \text{ with probability }0.1.
        \]
    \end{itemize}
    \item At state \(C\):
    \[
    C\to C \text{ with probability }0.1,\qquad C\to A \text{ with probability }0.9.
    \]
\end{itemize}

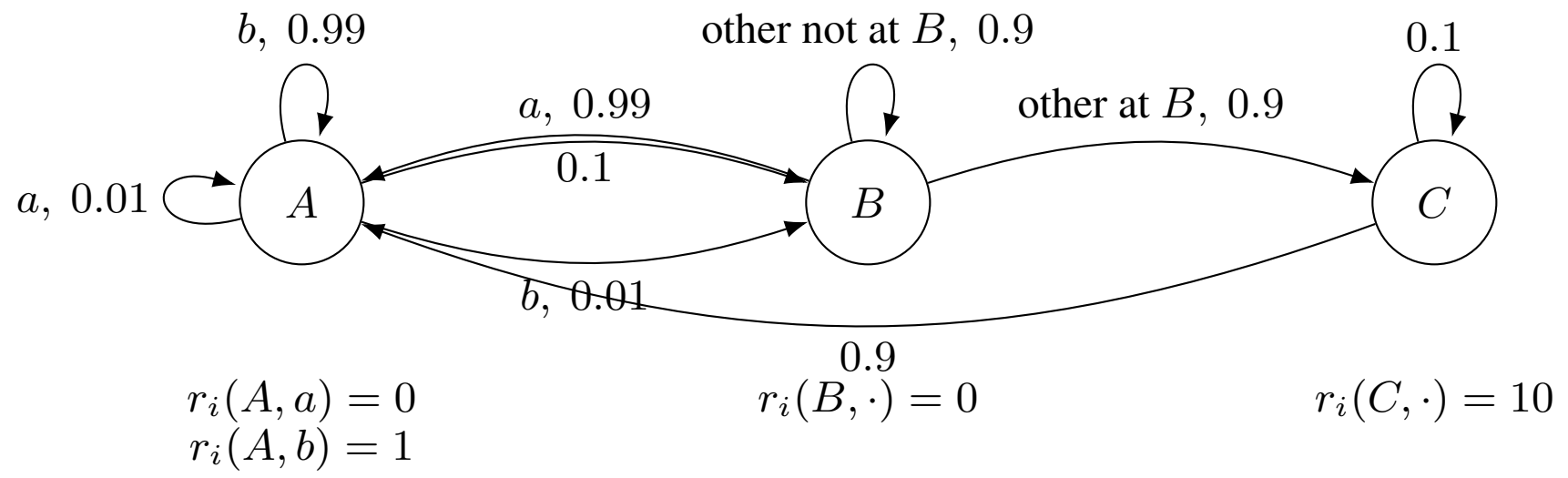
\begin{figure}[H]
\centering
\begin{tikzpicture}[
    >=Latex,
    state/.style={draw,circle,minimum size=9mm,inner sep=0pt},
    every node/.style={font=\small}
]
\node[state] (A) {$A$};
\node[state, right=3.2cm of A] (B) {$B$};
\node[state, right=3.2cm of B] (C) {$C$};

\node[below=7mm of A, align=center] {$r_i(A,a)=0$\\$r_i(A,b)=1$};
\node[below=7mm of B, align=center] {$r_i(B,\cdot)=0$};
\node[below=7mm of C, align=center] {$r_i(C,\cdot)=10$};

\draw[->]
(A) edge[loop above] node {$b,\ 0.99$} ()
(A) edge[loop left]  node {$a,\ 0.01$} ()
(A) edge[bend left=18] node[above] {$a,\ 0.99$} (B)
(A) edge[bend right=18] node[below] {$b,\ 0.01$} (B);

\draw[->]
(B) edge[loop above] node {$\text{other not at }B,\ 0.9$} ()
(B) edge[bend left=18] node[above] {$\text{other at }B,\ 0.9$} (C)
(B) edge[bend right=20] node[below] {$0.1$} (A);

\draw[->]
(C) edge[loop above] node {$0.1$} ()
(C) edge[bend left=20] node[below] {$0.9$} (A);
\end{tikzpicture}
\caption{Local dynamics for a single agent. The reward depends only on the agent's own state and action, but the transition out of \(B\) depends on whether the \emph{other} agent is also at \(B\).}
\label{fig:three-state-local-dynamics}
\end{figure}
\paragraph{Product policy with softmax parametrization.}
Only the policy at state \(A\) matters; the action at states \(B\) and \(C\) is irrelevant for both
rewards and transitions. For agent \(i\), write the two logits at state \(A\) as
\(\theta_i^a,\theta_i^b\), and define
\[
\pi_i(a\mid A)
=
\frac{e^{\theta_i^a}}{e^{\theta_i^a}+e^{\theta_i^b}},
\qquad
\pi_i(b\mid A)
=
\frac{e^{\theta_i^b}}{e^{\theta_i^a}+e^{\theta_i^b}}.
\]
Let
\[
p_i:=\pi_i(a\mid A)
=
\frac{e^{\theta_i^a}}{e^{\theta_i^a}+e^{\theta_i^b}},
\qquad
1-p_i=\pi_i(b\mid A).
\]
Equivalently, if we write the logit difference
\[
\Delta_i:=\theta_i^a-\theta_i^b,
\]
then
\[
p_i=\frac{e^{\Delta_i}}{1+e^{\Delta_i}}.
\]
From the specified transition probability kernel, it follows that, for any admissible policy, the induced Markov chain on the joint state space
\[
\{AA,AB,AC,BA,BB,BC,CA,CB,CC\}
\]
is irreducible and aperiodic for every softmax parameter vector.

\paragraph{Performance of deterministic policies.}
Let \(d^{aa}\) and \(d^{bb}\) denote the stationary distributions under the deterministic policies
\((a,a)\) and \((b,b)\), respectively. Ordering the joint states as
\[
(AA,AB,AC,BA,BB,BC,CA,CB,CC),
\]
one obtains
\[
d^{aa}\approx
(0.2099,\,0.0579,\,0.0507,\,0.0579,\,0.3090,\,0.0055,\,0.0507,\,0.0055,\,0.2528),
\]
and
\[
d^{bb}\approx
(0.8423,\,0.0772,\,0.00027,\,0.0772,\,0.00147,\,0.00000029,\,0.00027,\,0.00000029,\,0.00121).
\]
Hence
\[
J(1,1)\approx 6.1801,
\qquad
J(0,0)\approx 1.8691.
\]
Thus the policy \((a,a)\) is strictly better than \((b,b)\).
The surprising fact is that the bad corner \((p_1,p_2)=(0,0)\) is locally attractive under softmax policy gradient.
\begin{proposition}
Consider the continuous-time policy-gradient dynamics in the softmax logits,
\[
\dot{\theta}_i^a(t)=\frac{\partial J}{\partial \theta_i^a}(\theta(t)),
\qquad
\dot{\theta}_i^b(t)=\frac{\partial J}{\partial \theta_i^b}(\theta(t)),
\qquad i\in\{1,2\},
\]
where the corresponding action probabilities \(p_i(t)=\pi_i(a\mid A)\) are induced by the
softmax parametrization defined above. Then there exists \(\delta>0\) such that, if
\[
0<p_1(0)\le \delta,
\qquad
0<p_2(0)\le \delta,
\]
the resulting trajectory satisfies
\[
\theta_i^a(t)-\theta_i^b(t)\to -\infty,
\qquad i\in\{1,2\},
\]
and equivalently,
\[
p_1(t)\to 0,
\qquad
p_2(t)\to 0.
\]
Thus policy gradient converges in policy space to the suboptimal boundary point
\[
(p_1,p_2)=(0,0).
\]
\end{proposition}

\begin{proof}
The sign of \(\partial J/\partial p_i\) at \((0,0)\) is obtained by differentiating the stationary
average reward \(J(p_1,p_2)\). Since
\[
\frac{\partial J}{\partial p_1}(0,0)
=
\frac{\partial J}{\partial p_2}(0,0)
\approx -4.6881<0,
\]
continuity implies that there exists \(\delta>0\) such that
\[
\frac{\partial J}{\partial p_1}(p_1,p_2)<0,
\qquad
\frac{\partial J}{\partial p_2}(p_1,p_2)<0
\]
for all \((p_1,p_2)\in[0,\delta]^2\).

Now fix \(i\in\{1,2\}\). It is straightforward to see that,
\[
\frac{\partial p_i}{\partial \theta_i^a}=p_i(1-p_i),
\qquad
\frac{\partial p_i}{\partial \theta_i^b}=-\,p_i(1-p_i).
\]
Therefore,
\[
\frac{\partial J}{\partial \theta_i^a}
=
\frac{\partial J}{\partial p_i}\,
\frac{\partial p_i}{\partial \theta_i^a}
=
p_i(1-p_i)\frac{\partial J}{\partial p_i},
\]
and similarly
\[
\frac{\partial J}{\partial \theta_i^b}
=
\frac{\partial J}{\partial p_i}\,
\frac{\partial p_i}{\partial \theta_i^b}
=
-\,p_i(1-p_i)\frac{\partial J}{\partial p_i}.
\]
Hence, throughout the box \([0,\delta]^2\),
\[
\dot{\theta}_i^a<0,
\qquad
\dot{\theta}_i^b>0.
\]

Define the logit gap
\[
\Delta_i(t):=\theta_i^a(t)-\theta_i^b(t).
\]
Then
\[
\dot{\Delta}_i(t)
=
\dot{\theta}_i^a(t)-\dot{\theta}_i^b(t)
=
2p_i(t)(1-p_i(t))\frac{\partial J}{\partial p_i}(p_1(t),p_2(t)).
\]
Thus, whenever \((p_1(t),p_2(t))\in[0,\delta]^2\),
\[
\dot{\Delta}_i(t)<0.
\]
Since \(p_i(t)\) is a strictly increasing function of \(\Delta_i(t)\), it follows that
\[
\dot p_i(t)<0
\]
throughout \([0,\delta]^2\). Hence this box is invariant: if
\[
0<p_1(0)\le \delta,
\qquad
0<p_2(0)\le \delta,
\]
then
\[
0<p_1(t)\le \delta,
\qquad
0<p_2(t)\le \delta
\]
for all \(t\ge 0\), and both coordinates are monotone decreasing.

Since each \(p_i(t)\) is nonnegative and monotone, it has a limit. Suppose that
\[
\lim_{t\to\infty} p_i(t)=\bar p_i>0
\]
for some \(i\). Then \(\Delta_i(t)\) remains bounded below, and hence
\(p_i(t)(1-p_i(t))\) is bounded away from \(0\) for all sufficiently large \(t\).
Because \(\partial J/\partial p_i\) remains strictly negative on \([0,\delta]^2\), it follows that
\[
\dot{\Delta}_i(t)\le -c<0
\]
for all sufficiently large \(t\), a contradiction. Therefore
\[
\Delta_i(t)\to -\infty,
\qquad
p_i(t)\to 0,
\qquad i\in\{1,2\}.
\]
This proves convergence in policy space to the suboptimal boundary point \((0,0)\).
\end{proof}

Near \((p_1,p_2)=(0,0)\), each agent mostly plays \(b\) at \(A\), receives the safe reward \(1\), and tends to remain near state \(A\). If one agent slightly increases its probability of playing \(a\), then with high probability it leaves \(A\), lands in \(B\), and receives reward \(0\) while waiting for the other agent. This is a {first-order} loss. In contrast, the beneficial event in which {both} agents leave \(A\), meet at \(BB\), and then move to the high-reward state \(C\) is only a {second-order} effect when both \(p_1\) and \(p_2\) are small. Thus, near \((0,0)\), the one-agent gradient signal is negative even though the jointly coordinated policy \((1,1)\) is globally superior.

\section{A Network-Compatible Partial-Observability Example}\label{app:networked_partial_observability_example}
We now give a version of Example~\ref{example:partial_observability} that is fully compatible with the networked MDP formulation in Section~\ref{sec:impossibility_result}: each local reward depends only on the agent's own state-action pair, and each local transition depends only on neighboring states and the agent's own action.

Consider a two-node network. Agent 1 has
\[
\mathcal S_1=\{D,E_\ell,E_r\},
\qquad
\mathcal A_1=\{\ell,r\},
\]
and agent 2 has
\[
\mathcal S_2=\{L,M,R\},
\qquad
\mathcal A_2=\{\ast\}.
\]
The rewards are local:
\[
r_1(s_1,a_1)=0,
\qquad
r_2(L,\ast)=r_2(R,\ast)=1,
\qquad
r_2(M,\ast)=0.
\]
Thus \(J\) is the stationary probability that \(s_2\in\{L,R\}\).

The local transition of agent 1 stores its chosen action in its own next state. For every \(s_2\) and \(a\in\{\ell,r\}\),
\[
P_1(D\mid D,s_2,a)=\frac12,
\qquad
P_1(E_a\mid D,s_2,a)=\frac12,
\]
and from the two execution states agent 1 returns to \(D\):
\[
P_1(D\mid E_\ell,s_2,a)=P_1(D\mid E_r,s_2,a)=1.
\]
The local transition of agent 2 depends on the neighboring state \(s_1\), but not on agent 1's action. If \(s_1=D\), then \(s_2\) is unchanged. If \(s_1=E_\ell\), then
\[
L\mapsto L,\qquad M\mapsto L,\qquad R\mapsto M,
\]
and if \(s_1=E_r\), then
\[
L\mapsto M,\qquad M\mapsto R,\qquad R\mapsto R.
\]
\begin{center}
\centering
\begin{minipage}[t]{0.45\textwidth}
\centering
\textbf{(a) Agent 1 local state}\\[1mm]
\resizebox{0.82\linewidth}{!}{%
\begin{tikzpicture}[
    >=Latex,
    state/.style={draw,circle,minimum size=9mm,inner sep=0pt},
    every node/.style={font=\small}
]
\node[state] (D) at (0,0) {$D$};
\node[state] (El) at (-1.7,-1.7) {$E_\ell$};
\node[state] (Er) at (1.7,-1.7) {$E_r$};

\draw[->, thick]
(D) edge[loop above] node {$1/2$} ();
\draw[->, thick]
(D) to[bend right=10] node[left] {$\ell,\ 1/2$} (El);
\draw[->, thick]
(D) to[bend left=10] node[right] {$r,\ 1/2$} (Er);
\draw[->, thick]
(El) to[bend right=10] node[below right] {$1$} (D);
\draw[->, thick]
(Er) to[bend left=10] node[below left] {$1$} (D);
\end{tikzpicture}
}
\end{minipage}
\hfill
\begin{minipage}[t]{0.50\textwidth}
\centering
\textbf{(b) Agent 2 local state}\\[1mm]
\resizebox{\linewidth}{!}{%
\begin{tikzpicture}[
    >=Latex,
    state/.style={draw,circle,minimum size=9mm,inner sep=0pt},
    every node/.style={font=\small}
]
\node[state] (M) at (0,0) {$M$};
\node[state] (L) at (-1.7,-1.7) {$L$};
\node[state] (R) at (1.7,-1.7) {$R$};

\draw[->, thick]
(M) to[bend right=10] node[left] {$E_\ell$} (L);
\draw[->, thick]
(R) to[bend right=10] node[right] {$E_\ell$} (M);
\draw[->, thick]
(L) edge[loop left] node {$E_\ell$} ();

\draw[->, thick, dashed]
(L) to[bend right=10] node[right] {$E_r$} (M);
\draw[->, thick, dashed]
(M) to[bend right=10] node[left] {$E_r$} (R);
\draw[->, thick, dashed]
(R) edge[loop right] node {$E_r$} ();

\node[below=8mm of L, xshift=1.7cm, align=center] {$s_1=D$: \(L,M,R\) unchanged};
\end{tikzpicture}
}
\end{minipage}
\par\vspace{1mm}
\small\textbf{Illustration.} Network-compatible partial-observability example. Agent 1's action affects agent 2 only after being stored in agent 1's own local state \(E_\ell\) or \(E_r\); agent 2's transition therefore depends on the neighboring state \(s_1\), not directly on agent 1's action.
\end{center}
Consequently the global transition kernel factors as
\[
P(s_g'\mid s_g,a_g)
=
P_1(s_1'\mid s_1,s_2,a_1)
P_2(s_2'\mid s_1,s_2,\ast),
\]
which has exactly the product form required in the networked MDP setting.

Let \(p:=\pi_1(\ell\mid D)\). For every softmax policy \(p\in(0,1)\), the induced chain on \(\mathcal S_1\times\mathcal S_2\) is irreducible and aperiodic: from \(D\), both execution states \(E_\ell\) and \(E_r\) are reachable with positive probability, and \(D\) has a self-loop. Actions at \(E_\ell\) and \(E_r\) do not affect rewards or transitions.

It remains to compute the stationary objective. Let \(\mu^p\) be the stationary distribution of the embedded hidden chain on \(\{L,M,R\}\) obtained by looking at the updates to \(s_2\). The holding at \(D\) is independent of \(s_2\), so it does not change the stationary marginal over \(s_2\). Therefore \(\mu^p\) satisfies the same balance equations as in Example~\ref{example:partial_observability}:
\[
\mu^p(L)=\frac{p}{1-p}\mu^p(M),
\qquad
\mu^p(R)=\frac{1-p}{p}\mu^p(M).
\]
Hence
\begin{equation}
    \mu^p(M)
    =
    \left(
        1+\frac{p}{1-p}+\frac{1-p}{p}
    \right)^{-1},
    \qquad
    J(p)=1-\mu^p(M).
    \label{eq:networked-partial-observability-objective}
\end{equation}
At \(p=1/2\), \(J(1/2)=2/3\) and \(J'(1/2)=0\). Thus the two softmax gradients at state \(D\) vanish. The softmax gradients at \(E_\ell\) and \(E_r\) also vanish because the actions there affect neither rewards nor transitions, and agent 2 has no actor parameter. The product policy is therefore stationary for the local-policy gradient.

Nevertheless, the Nash gap is strictly positive. If agent 1 unilaterally changes only its policy at \(D\) to \(p'=1/4\), then \eqref{eq:networked-partial-observability-objective} gives
\[
J(1/4)=1-\left(1+\frac13+3\right)^{-1}=\frac{10}{13}
>
\frac23=J(1/2).
\]
Thus
\[
\mathrm{NashGap}_1(\theta)
\ge
J(1/4)-J(1/2)
=
\frac{4}{39}
>0,
\]
even though all local policy gradients vanish. This confirms that the partial-observability obstruction is not an artifact of the simplified transition notation in Example~\ref{example:partial_observability}; it persists under the networked MDP transition and reward restrictions.

\section{Convergence Proof without Regularization}
\label{app:unregularized-convergence}
In this section we present the convergence proof of the decentralized actor-critic algorithm without regularization.
We separate the proof into three steps: The convergence of the critic, the convergence of the actor, and the overall convergence.

\subsection{Convergence of the Critic}
\label{subsection:critic}
\label{app:unregularized-critic}
This section establishes finite-time convergence of the critic component of the decentralized actor-critic algorithm.

\paragraph{Proof roadmap.}
The critic proof has two layers: fixed-policy approximation and moving-target stability.  Lemma~\ref{lem:solution_Existence} first resolves the average-reward constant-shift ambiguity by selecting a unique normalized projected Bellman fixed point, and Lemma~\ref{lem:contrac_coeffs} gives the contraction coefficients that make this fixed point stable.  Proposition~\ref{prop:basic_props_constrained_proj} supplies the constrained-projection identities used by Lemma~\ref{lem:local-truncation},which bounds the error of the localized tabular projected fixed point relative to the true local \(Q\)-function in terms of restricted projected Bellman gap of the global transition matrix. Lemma~\ref{lem:fixed_actor_ss} then proves fixed-policy TD contraction, and Theorem~\ref{thm:fixed-policy-td-learning-bounds} converts that contraction plus the approximation floor into finite-time local and global value-function bounds.  Finally, Lemma~\ref{lem:lipschitzness}, Lemma~\ref{lem:bounded-actor-increment}, Lemma~\ref{lem:moving_actor_constant_beta}, and Corollary~\ref{cor:moving_actor_time_varying_beta} show that the same critic tracking remains controlled when the actor moves.

Proposition 2 supplies the constrained-projection
identities used by Lemma 4, which bounds the error of the localized tabular projected
fixed point relative to the true local \(Q\)-function, i.e., the spatial localization part of
the approximation floor.

\subsubsection{Localized projected Bellman equations}
The exact local $Q$-function $Q_i^\theta : S_g \times A_g \to \mathbb{R}$ depends on the full global state-action pair, but under decentralization, agent $i$ can only observe its $k$-hop neighborhood $(s_i^k, a_i^k) \in S_{N_i^k} \times A_{N_i^k}$. The critic must therefore approximate $Q_i^\theta$ using only local information, which introduces two distinct layers of error that the analysis must separate and quantify.

Under Assumption~\ref{assump:ergodicity}, the global chain has a unique stationary distribution $D_{\pi_\theta}\coloneqq \text{diag}(\nu^\theta)$, and the $0$-$1$ localization matrix $\phi_i \in \mathbb{R}^{|S_g||A_g| \times |S_{N_i^k}||A_{N_i^k}|}$, defined by
$$\phi_i((s_g, a_g),\, (s_i^k, a_i^k)) = \mathbf{1}\!\left[\mathrm{Proj}_{N_i^k}(s_g, a_g) = (s_i^k, a_i^k)\right],$$
where $\mathrm{Proj}_{N_i^k} : S_g \times A_g \to S_{N_i^k} \times A_{N_i^k}$ extracts the states and actions of agents in the $k$-hop neighborhood: $\mathrm{Proj}_{N_i^k}(s_g, a_g) := (s_p, a_p)_{p \in N_i^k}$. The matrix $\phi_i^\top D_{\pi_\theta} \phi_i$ is a diagonal matrix of dimension $|S_{N_i^k}||A_{N_i^k}|$. This follows directly from the $0$-$1$ structure: each row of $\phi_i$ has exactly one nonzero entry, so two distinct neighborhood pairs $(s_i^k, a_i^k) \neq (s_i^{k'}, a_i^{k'})$ cannot simultaneously have $\phi_i((s_g,a_g),\cdot) = 1$ for the same global $(s_g, a_g)$, making all off-diagonal entries zero. See \ref{sss:phi_visual} for more details on $\phi_i$. The diagonal entry at $(s_i^k, a_i^k)$ is $\sum_{(s_g,a_g):\,\mathrm{Proj}_{N_i^k}(s_g,a_g) = (s_i^k,a_i^k)} D_{\pi_\theta}(s_g,a_g)$, the marginal stationary probability of the neighborhood pair. Under Assumption 1 these marginals are strictly positive, so $\phi_i^\top D_{\pi_\theta}\phi_i$ is invertible. We denote this marginal by $\widehat{D}_{\pi_\theta} := \phi_i^\top D_{\pi_\theta} \phi_i$.

Consider first approximating $Q_i^\theta$ by a function $\widehat{Q}_i : S_{N_i^k} \times A_{N_i^k} \to \mathbb{R}$ that depends only on the $k$-hop neighborhood. The tabular updates are:
\begin{equation}
    \mu_i^{t+1} = \mu_i^t + c_{ss}\,\beta_t\,\bigl(r_i(s_i(t), a_i(t)) - \mu_i^t\bigr),
\end{equation}
\small{
\begin{equation}
    \widehat{Q}_i^{t+1}(s_i^k(t), a_i^k(t))\! =\! \widehat{Q}_i^t(s_i^k(t), a_i^k(t))\! + \!\beta_t\Bigl(r_i(s_i(t), a_i(t))\! -\! \mu_i^t \!+\! \widehat{Q}_i^t(s_i^k(t+1), a_i^k(t+1))\! - \!\widehat{Q}_i^t(s_i^k(t), a_i^k(t))\Bigr).
\end{equation}
}
\normalsize{}
\noindent The first update tracks the local average reward $J_i(\theta)$; the second is a standard average-reward TD update, operating entirely in the tabular neighborhood space. At stationarity the update for $\widehat{Q}_i$ has the fixed point
$$\widehat{Q}_i^\infty = (\phi_i^\top D_{\pi_\theta} \phi_i)^{-1} \phi_i^\top D_{\pi_\theta}\, T^i_{\pi_\theta}\, \phi_i\, \widehat{Q}_i^\infty,$$
where $T^i_{\pi_\theta} \nu := r_i - J_i(\theta)\mathbf{1} + \textcolor{revisionblue}{P_{\pi_\theta}}\nu$ is the global average-reward Bellman operator for agent $i$. Left-multiplying both sides by $\phi_i$ and recognizing the $D_{\pi_\theta}$-weighted projection
$$\Pi^{\phi_i}_{D_{\pi_\theta}} := \phi_i(\phi_i^\top D_{\pi_\theta} \phi_i)^{-1}\phi_i^\top D_{\pi_\theta},$$
the fixed point condition on the global space is exactly
\begin{equation}
    \phi_i\widehat{Q}_i^\infty = \Pi^{\phi_i}_{D_{\pi_\theta}}\, T^i_{\pi_\theta}\, \phi_i\widehat{Q}_i^\infty.
\end{equation}
This is a projected Bellman equation: the true Bellman operator $T^i_{\pi_\theta}$ is first applied, then the result is projected back onto $\mathrm{span}(\phi_i)$, the subspace of global functions that are constant on each neighborhood equivalence class. \textcolor{revisionblue}{The tabular local fixed point $\phi_i\widehat{Q}_i^\infty$ solves this projected Bellman equation; its localization error is bounded in Lemma~\ref{lem:local-truncation}.}

Working directly in the tabular space $\mathbb{R}^{|S_{N_i^k}||A_{N_i^k}|}$ becomes intractable for large neighborhoods. We therefore introduce a feature matrix $\Psi_i \in \mathbb{R}^{|S_{N_i^k}||A_{N_i^k}| \times d}$ and approximate $\widehat{Q}_i^\infty$ by $\Psi_i w$ for some $w \in \mathbb{R}^d$. This trades a function-approximation error (from compressing into $d$ dimensions) for computational tractability and the ability to use larger $k$. Let $\psi_i(s_i^k, a_i^k) \in \mathbb{R}^d$ denote the row of $\Psi_i$ corresponding to $(s_i^k, a_i^k)$. The critic runs the following updates on neighborhood samples $(s_i^k(t), a_i^k(t))$:

\begin{equation}
    \mu_i^{t+1} = \mu_i^t + c_{ss}\, \beta_t\, \bigl(r_i(s_i(t), a_i(t)) - \mu_i^t\bigr),
    \label{eq:update1}
\end{equation}
\begin{equation}
    \delta_i^t = r_i(s_i(t), a_i(t)) - \mu_i^t + \psi_i^\top(s_i^k(t+1), a_i^k(t+1))\, w_i^t - \psi_i^\top(s_i^k(t), a_i^k(t))\, w_i^t,
    \label{eq:update2}
\end{equation}
\begin{equation}
    w_i^{t+1} = w_i^t + \beta_t\, \delta_i^t\, \psi_i(s_i^k(t), a_i^k(t)),
    \label{eq:update3}
\end{equation}
where $\beta_t > 0$ is the critic step size and $c_{ss} \geq 1$ is a speed-ratio constant. Equation~\eqref{eq:update1} tracks the local average reward; Equation~\eqref{eq:update2} is the average-reward TD error computed using the linear critic $\psi_i^\top w_i^t$; Equation~\eqref{eq:update3} moves $w_i^t$ in the direction of the TD error weighted by the feature vector. In matrix form these read $\mathcal{W}_i^{t+1} = \mathcal{W}_i^t + \beta_t(A_i^t \mathcal{W}_i^t + b_i^t)$ with augmented state $\mathcal{W}_i^t := (\mu_i^t, w_i^t)^\top$ and
\begin{equation}
    A_i^t \coloneqq \begin{pmatrix}
- c_{ss} & 0 \\
-\psi_i(s^k_i(t),a^k_i(t)) &
\psi_i(s^k_i(t),a^k_i(t))(\psi_i(s^k_i(t\!+\!1),a^k_i(t\!+\!1))^\intercal-\psi_i(s^k_i(t),a^k_i(t))^\intercal)
\end{pmatrix}
\end{equation}
and
\begin{equation}
    b_i^t \coloneqq 
\begin{pmatrix}
c_{ss}\,r_i(s_i(t),a_i(t))\\[4pt]
\psi_i(s^k_i(t),a^k_i(t))\,r_i(s_i(t),a_i(t))
\end{pmatrix}
\end{equation}
The fixed point $\mathcal{W}_i^\infty := (J_i(\theta), w_i^\star)^\top$ satisfies $A_i^\infty \mathcal{W}_i^\infty + b_i^\infty = 0$, where $A_i^\infty := \mathbb{E}_{\pi_\theta}[A_i^t]$ and $b_i^\infty := \mathbb{E}_{\pi_\theta}[b_i^t]$ are the mean-field matrices at stationarity. The local Bellman operator on the truncated space is
\begin{equation}
    \widehat{T}^i_{\pi_\theta} := (\phi_i^\top D_{\pi_\theta} \phi_i)^{-1} \phi_i^\top D_{\pi_\theta}\, T^i_{\pi_\theta}\, \phi_i,
\end{equation}
and the fixed point $w_i^\star$ that Equations~\eqref{eq:update1}-\eqref{eq:update3} track is \textcolor{revisionblue}{the unique normalized solution, with the additive-constant ambiguity fixed as in Lemma~\ref{lem:solution_Existence},} satisfying
\begin{equation*}
    w_i^\star = (\Psi_i^\top \widehat{D}_{\pi_\theta} \Psi_i)^{-1} \Psi_i^\top \widehat{D}_{\pi_\theta}\, \widehat{T}^i_{\pi_\theta}\, \Psi_i w_i^\star, \quad \text{equivalently,} \quad \Psi_i w_i^\star = \Pi^{\widehat{D}_{\pi_\theta}}_{\Psi_i}\, \widehat{T}^i_{\pi_\theta}(\Psi_i w_i^\star).
\end{equation*}
This is the projected Bellman fixed point within the linear function class $\mathrm{span}(\Psi_i)$: the local Bellman operator $\widehat{T}^i_{\pi_\theta}$ is applied and then projected back onto $\mathrm{span}(\Psi_i)$ under the marginal stationary measure $\widehat{D}_{\pi_\theta}$. The condition $c_{ss} \geq \Delta_i^{\chi_i} + \sqrt{(\Delta_i^{\chi_i})^{-2} - 1}$ ensures that $\mu_i^t$~\eqref{eq:update1} adapts faster than $w_i^t$~\eqref{eq:update3}, so the average-reward estimate stays close to $J_i(\theta)$ throughout and does not slow the contraction of $w_i^t$.
Define the composite feature matrix $\chi_i := \phi_i \Psi_i \in \mathbb{R}^{|S_g||A_g| \times d}$. Expanding the local fixed point equation and substituting $\widehat{D}_{\pi_\theta} = \phi_i^\top D_{\pi_\theta}\phi_i$:
\begin{equation}
    \phi_i\Psi_i w_i^\star = \phi_i\Psi_i(\Psi_i^\top \phi_i^\top D_{\pi_\theta}\phi_i\Psi_i)^{-1}\Psi_i^\top \phi_i^\top D_{\pi_\theta}\, T^i_{\pi_\theta}\,\phi_i\Psi_i w_i^\star = \Pi^{D_{\pi_\theta}}_{\chi_i}\, T^i_{\pi_\theta}(\chi_i w_i^\star),
\end{equation}
which is precisely the global projected Bellman equation
\begin{equation}
    \Pi^{D_{\pi_\theta}}_{\chi_i}\, T^i_{\pi_\theta}\,\chi_i w_i^\star = \chi_i w_i^\star. 
    \label{eq:lfa_fp}
\end{equation}
The matrices $\phi_i$ and $\Psi_i$ define a two-stage embedding $\mathbb{R}^d \xrightarrow{\Psi_i} \mathbb{R}^{|S_{N_i^k}||A_{N_i^k}|} \xrightarrow{\phi_i} \mathbb{R}^{|S_g||A_g|}$, so $\mathrm{span}(\chi_i)$ is exactly the set of global value functions representable by agent $i$'s architecture. The equivalence between the local equation and Equation~\eqref{eq:lfa_fp} holds because the $0$-$1$ structure of $\phi_i$ means that the $D_{\pi_\theta}$-projection onto $\mathrm{span}(\chi_i)$ decomposes cleanly: localization via $\widehat{D}_{\pi_\theta} = \phi_i^\top D_{\pi_\theta}\phi_i$, then linear compression via $\Pi^{\widehat{D}_{\pi_\theta}}_{\Psi_i}$. Consequently, the fixed point $\chi_i w_i^\star$ incurs two independent error sources relative to the true $Q_i^\theta$: (i) Localization error: $O(c\rho^{k+1})$ from restricting observations to the $k$-hop neighborhood (controlled by increasing $k$) (ii) Function Approximation Error: $\|\phi_i\widehat{Q}_i^\infty - \chi_i w_i^\star\|_{D_{\pi_\theta}}$ from projecting the tabular fixed point onto $\mathrm{span}(\Psi_i)$ (depends on expressivity of $\Psi_i$). 

\subsubsection{Visualizing the localization matrix $\phi_i$}
\label{sss:phi_visual}
The localization matrix \(\phi_i\) has a very specific structure. It is a
\(\{0,1\}\)-valued matrix of size
\[
\phi_i \in \{0,1\}^{|\mathcal S_g\times \mathcal A_g|\times |\mathcal S_i^k\times \mathcal A_i^k|},
\]
whose rows are indexed by global state-action pairs and whose columns are indexed by
local state-action pairs on the \(k\)-hop neighborhood of agent \(i\).
For a global state-action pair \(x_g\in \mathcal S_g\times \mathcal A_g\) and a local
state-action pair \(x_i^k\in \mathcal S_i^k\times \mathcal A_i^k\), the entry of
\(\phi_i\) is given by
\[
[\phi_i]_{x_g,x_i^k}
=
\mathbf 1\!\left\{x_i^k = \mathrm{Proj}_{N_i^k}(x_g)\right\},
\]
where \(\mathrm{Proj}_{N_i^k}(x_g)\) denotes the restriction of the global
state-action pair \(x_g\) to the \(k\)-hop neighborhood \(N_i^k\) of agent \(i\).

Thus, each row of \(\phi_i\) has exactly one nonzero entry, and that nonzero entry is
equal to \(1\). Indeed, every global state-action pair has a unique projection onto the
local \(k\)-hop neighborhood. In this sense, \(\phi_i\) precisely encodes the mapping
from the global state-action space to the local state-action space. Also, different columns of \(\phi_i\) have disjoint supports, and each column is nonzero. Therefore, the columns are linearly independent, so \(\phi_i\) has full column rank.

Equivalently, for any local tabular vector \(q_i\in \mathbb R^{|\mathcal S_i^k\times \mathcal A_i^k|}\),
the lifted vector \(\phi_i q_i\in \mathbb R^{|\mathcal S_g\times \mathcal A_g|}\) is given by
\[
(\phi_i q_i)(x_g)
=
q_i\!\left(\mathrm{Proj}_{N_i^k}(x_g)\right).
\]
That is, \(\phi_i q_i\) is the global vector obtained by replicating the same local value
across all global state-action pairs that agree on agent \(i\)'s \(k\)-hop neighborhood.

\paragraph{Example.}
Consider a network with \(4\) agents, where each agent has \(2\) states and a single
action. Let the graph have edges $(1,2), (2,3), (2,4).$ Thus, agent \(1\) has only agent \(2\) as a neighbor, and its \(1\)-hop neighborhood is $N_1^1 = \{1,2\}.$

\begin{figure}[h]
\centering
\begin{tikzpicture}[
    every node/.style={circle, draw, minimum size=8mm, inner sep=0pt},
    scale=1
]
\node[fill=blue!15, thick] (1) at (0,0) {1};
\node[fill=blue!15, thick] (2) at (2,0) {2};
\node (3) at (4,1) {3};
\node (4) at (4,-1) {4};

\draw[thick] (1) -- (2);
\draw[thick] (2) -- (3);
\draw[thick] (2) -- (4);

\node[draw=none, rectangle] at (2,-2) {$N_1^1=\{1,2\}$};
\end{tikzpicture}
\caption{Network in the example. The shaded nodes form the \(1\)-hop neighborhood of agent \(1\).}
\label{fig:phi_example_graph}
\end{figure}
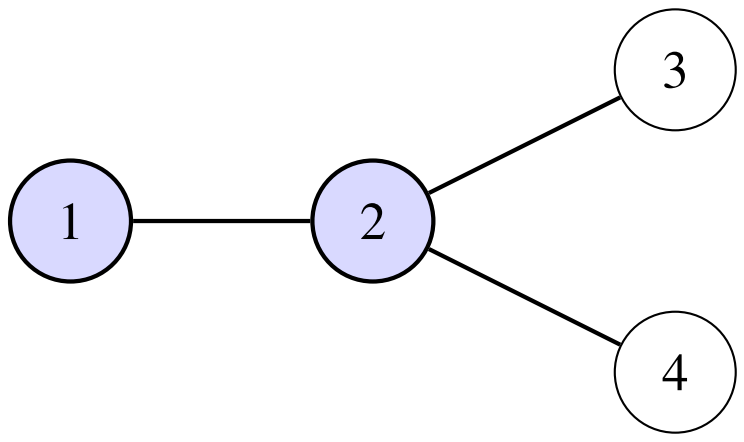

Since each agent has a single action, the action coordinate is trivial and will be
suppressed in the labels below. Hence,
\[
|\mathcal S_g\times \mathcal A_g| = 2^4 = 16,
\qquad
|\mathcal S_1^1\times \mathcal A_1^1| = 2^2 = 4.
\]
Therefore, \(\phi_1\) is a \(16\times 4\) matrix.

We order the global state-action pairs lexicographically as
\[
(0,0,0,0),\ (0,0,0,1),\ (0,0,1,0),\ (0,0,1,1),\ \ldots,\ (1,1,1,1),
\]
and the local state-action pairs for agent \(1\) as
\[
(0,0),\ (0,1),\ (1,0),\ (1,1),
\]
where \((s_1,s_2)\) denotes the local pair on \(N_1^1=\{1,2\}\).

With this ordering, the localization matrix \(\phi_1\) is
\begingroup
\setlength{\arraycolsep}{6pt}
\renewcommand{\arraystretch}{1.1}
\[
\phi_1
=
\left[
\begin{array}{c|cccc}
\text{global }(s_1,s_2,s_3,s_4)\backslash \text{ local }(s_1,s_2)
& (0,0) & (0,1) & (1,0) & (1,1)
\\ \hline
(0,0,0,0) & 1 & 0 & 0 & 0 \\
(0,0,0,1) & 1 & 0 & 0 & 0 \\
(0,0,1,0) & 1 & 0 & 0 & 0 \\
(0,0,1,1) & 1 & 0 & 0 & 0 \\
(0,1,0,0) & 0 & 1 & 0 & 0 \\
(0,1,0,1) & 0 & 1 & 0 & 0 \\
(0,1,1,0) & 0 & 1 & 0 & 0 \\
(0,1,1,1) & 0 & 1 & 0 & 0 \\
(1,0,0,0) & 0 & 0 & 1 & 0 \\
(1,0,0,1) & 0 & 0 & 1 & 0 \\
(1,0,1,0) & 0 & 0 & 1 & 0 \\
(1,0,1,1) & 0 & 0 & 1 & 0 \\
(1,1,0,0) & 0 & 0 & 0 & 1 \\
(1,1,0,1) & 0 & 0 & 0 & 1 \\
(1,1,1,0) & 0 & 0 & 0 & 1 \\
(1,1,1,1) & 0 & 0 & 0 & 1
\end{array}
\right].
\]
\endgroup

This matrix simply records the projection of a global state-action pair onto the local
coordinates \((s_1,s_2)\). For instance,
\[
\mathrm{Proj}_{N_1^1}(0,1,1,0) = (0,1),
\]
and therefore the row indexed by \((0,1,1,0)\) has a \(1\) in the column indexed by
\((0,1)\) and zeros elsewhere.

In particular, if \(q_1\in\mathbb R^4\) is indexed by the local pairs
\((0,0),(0,1),(1,0),(1,1)\), then
\[
(\phi_1 q_1)(s_1,s_2,s_3,s_4)=q_1(s_1,s_2).
\]
Thus, \(\phi_1\) lifts a local function of \((s_1,s_2)\) to a global function on
\((s_1,s_2,s_3,s_4)\) by replicating the same value over all global tuples that share the
same local projection. In this example, one also has
\[
\phi_1^\top \phi_1 = 4 I_4,
\]
which makes the full-column-rank property immediate.

\subsubsection{Constant-shift geometry for average-reward TD}
In the average-reward setting the Bellman equation $Q_i^\theta = T^i_{\pi_\theta} Q_i^\theta$ holds for any additive constant shift of $Q_i$ (since $T^i_{\pi_\theta}(Q_i + c\mathbf{1}) = T^i_{\pi_\theta} Q_i+c\mathbf{1}$ because $\textcolor{revisionblue}{P_{\pi_\theta}}\mathbf{1} = \mathbf{1}$ and $J_i(\theta)\mathbf{1}$ absorbs the shift). This means that the fixed point in Equation~\eqref{eq:lfa_fp} can have a one-parameter family of solutions. To pin down a unique fixed point, define
$$S_{\Psi_i, e} := \text{span}\{w_e \mid \Psi_i w_e = e\}, \qquad S_{\Psi_i, e}^\perp := \{\beta \mid \beta^\top w_e = 0\},$$
where $e$ is the all-ones vector and $w_e$ satisfies $\Psi_i w_e = e$. Since $\phi_i$ is a $0$-$1$ matrix with $\phi_i v = e \Leftrightarrow v = e$, it follows that $S_{\Psi_i,e} = S_{\chi_i,e}$. The existence of the fixed point solution to Equation~\eqref{eq:lfa_fp} is characterized in the following lemma.

\begin{lemma}[Fixed-point structure, \cite{zhang2021finite}]
    Under Assumption 1, the fixed points of Equation~\eqref{eq:lfa_fp} are
    \begin{equation}
        W_i^\star = \begin{cases} w_i^\star & \text{if } S_{\Psi_i,e} = \emptyset, \\ w_i^\star + c\,w_e, \; c \in \mathbb{R} & \text{if } S_{\Psi_i,e} \neq \emptyset, \end{cases}
    \end{equation}
    where $w_i^\star$ is the unique element of $S_{\Psi_i,e}^\perp$ satisfying
    \begin{equation*}
        \Pi_{D_{\pi_\theta},\, S_{\Psi_i,e}^\perp}^{\chi_i}\, T^i_{\pi_\theta}\, \chi_i w_i^\star \;=\; \chi_i w_i^\star,
    \end{equation*}
    and $\Pi_{D_{\pi_\theta}, S_{\Psi_i,e}^\perp}^{\chi_i}$ denotes the $D_{\pi_\theta}$-weighted projection onto $\{\chi_i w \mid w \in S_{\Psi_i,e}^\perp\}$.
    \label{lem:solution_Existence}
\end{lemma}

The constant-shift ambiguity is the direct consequence of the average-reward Bellman operator satisfying $T^i_{\pi_\theta}(Q + c\mathbf{1}) = c\mathbf{1} + T^i_{\pi_\theta} Q$ for any $c \in \mathbb{R}$, so Equation~\eqref{eq:lfa_fp} cannot distinguish $\chi_i w$ from $\chi_i w + c\mathbf{1} = \chi_i(w + cw_e)$. Restricting the projection to $S_{\Psi_i,e}^\perp$ breaks this symmetry: it enforces $\beta^\top w_e = 0$, which pins the constant component of $w$ to zero and makes the fixed point unique.

\begin{lemma}[Contraction Coefficients, \cite{zhang2021finite}]
\label{lem:contrac_coeffs}
    Under Assumption 1,
    \begin{equation}
        \Delta_i^{\phi_i} := \min_{\substack{\|v\|_2=1 \\ v \in S_{\phi_i,e}^\perp}} v^\top \phi_i^\top D_{\pi_\theta}(I - \textcolor{revisionblue}{P_{\pi_\theta}})\phi_i\, v > 0, \qquad \Delta_i^{\chi_i} := \min_{\substack{\|v\|_2=1 \\ v \in S_{\Psi_i,e}^\perp}} v^\top \chi_i^\top D_{\pi_\theta}(I - \textcolor{revisionblue}{P_{\pi_\theta}})\chi_i\, v > 0.
    \end{equation}
\end{lemma}
The quantity $\Delta_i^{\chi_i}$ is the minimum Rayleigh quotient of $\chi_i^\top D_{\pi_\theta}(I-\textcolor{revisionblue}{P_{\pi_\theta}})\chi_i$ over unit vectors in $S_{\Psi_i,e}^\perp$. It plays the role of a strong-monotonicity constant for the projected Bellman operator. Larger the value of $\Delta_i^{\chi_i}$, the iterate is pulled more strongly toward $w_i^\star$, hence faster contraction.

The analysis of the critic requires careful handling of the $D_{\pi_\theta}$-weighted projection onto the constrained subspace. We collect the needed properties of constrained projection in the following proposition.

\begin{proposition}[Basic properties of the constrained projection]
Let $S_{\phi_i,e}^{\perp} := \{\beta \in \mathbb{R}^{|S_{N_i^k}||A_{N_i^k}|} : \beta^\top e = 0\}$,
and let
\(
\Pi^{\phi_i}_{D_{\pi_\theta},S_{\phi_i,e}^{\perp}}
\)
denote the $D_{\pi_\theta}$-orthogonal projection onto the subspace
\(
\{\phi_i \nu : \nu \in S_{\phi_i,e}^{\perp}\}.
\)
Then the following hold:
\begin{enumerate}
    \item $\Pi^{\phi_i}_{D_{\pi_\theta},S_{\phi_i,e}^{\perp}}$ is idempotent, i.e.
    \( \left(\Pi^{\phi_i}_{D_{\pi_\theta},S_{\phi_i,e}^{\perp}}\right)^2 = \Pi^{\phi_i}_{D_{\pi_\theta},S_{\phi_i,e}^{\perp}}.
    \)
    \item $\Pi^{\phi_i}_{D_{\pi_\theta},S_{\phi_i,e}^{\perp}}$ is self-adjoint with respect to
$\langle \cdot,\cdot\rangle_{D_{\pi_\theta}}$, i.e. for any
$x,y \in \mathbb{R}^{|S_g||A_g|}$,
\(
\left\langle
x,\Pi^{\phi_i}_{D_{\pi_\theta},S_{\phi_i,e}^{\perp}}y
\right\rangle_{D_{\pi_\theta}}
=
\left\langle
\Pi^{\phi_i}_{D_{\pi_\theta},S_{\phi_i,e}^{\perp}}x,y
\right\rangle_{D_{\pi_\theta}}.
\)
\item Consequently, for any $e_i^{(\bar c)} \in \mathbb{R}^{|S_g||A_g|}$, if we define
\[
u_i^{(\bar c)}
:=
\Pi^{\phi_i}_{D_{\pi_\theta},S_{\phi_i,e}^{\perp}} e_i^{(\bar c)},
\qquad
b_i^{(\bar c)}
:=
\left(I-\Pi^{\phi_i}_{D_{\pi_\theta},S_{\phi_i,e}^{\perp}}\right)e_i^{(\bar c)},
\]
then
\[
\left\langle
u_i^{(\bar c)},b_i^{(\bar c)}
\right\rangle_{D_{\pi_\theta}}
=0.
\]
\end{enumerate}
\label{prop:basic_props_constrained_proj}
\end{proposition}
\begin{proof}
    Let $E_i$ be any full-column-rank matrix whose columns form a basis of $S_{\phi_i,e}^{\perp}$. Then
\[
\{\phi_i \nu : \nu \in S_{\phi_i,e}^{\perp}\}
=
\{\phi_i E_i \alpha : \alpha \in \mathbb{R}^{|S_{N_i^k}||A_{N_i^k}|-1}\}.
\]
Hence, for any $x \in \mathbb{R}^{|S_g||A_g|}$, the vector
$\Pi^{\phi_i}_{D_{\pi_\theta},S_{\phi_i,e}^{\perp}}x = \phi_i E_i \alpha^*$ is the unique minimizer of
\[
\min_{\alpha}
\|x-\phi_i E_i \alpha\|_{D_{\pi_\theta}}^2.
\]
where $\alpha^*$ is the optimal choice of $\alpha$. The corresponding normal equations are
\[
E_i^\top \phi_i^\top D_{\pi_\theta}\phi_i E_i \,\alpha
=
E_i^\top \phi_i^\top D_{\pi_\theta}x,
\]
and therefore
\[
\Pi^{\phi_i}_{D_{\pi_\theta},S_{\phi_i,e}^{\perp}}x
=
\phi_i E_i
\left(E_i^\top \phi_i^\top D_{\pi_\theta}\phi_i E_i\right)^{-1}
E_i^\top \phi_i^\top D_{\pi_\theta}x.
\]
Thus,
\[
\Pi^{\phi_i}_{D_{\pi_\theta},S_{\phi_i,e}^{\perp}}
=
\phi_i E_i
\left(E_i^\top \phi_i^\top D_{\pi_\theta}\phi_i E_i\right)^{-1}
E_i^\top \phi_i^\top D_{\pi_\theta}.
\]
Define
\[
A_i := \phi_i E_i,
\qquad
M_i := A_i^\top D_{\pi_\theta} A_i.
\]
Then
\[
\Pi^{\phi_i}_{D_{\pi_\theta},S_{\phi_i,e}^{\perp}}
=
A_i M_i^{-1} A_i^\top D_{\pi_\theta}.
\]
Since $A_i$ has full column rank and $D_{\pi_\theta}$ is positive definite, $M_i$ is symmetric
positive definite.
\begin{enumerate}
    \item We first prove idempotence. Using the above representation,
\begin{align*}
\left(\Pi^{\phi_i}_{D_{\pi_\theta},S_{\phi_i,e}^{\perp}}\right)^2
&=
A_i M_i^{-1} A_i^\top D_{\pi_\theta}
A_i M_i^{-1} A_i^\top D_{\pi_\theta} \\
&=
A_i M_i^{-1}
\left(A_i^\top D_{\pi_\theta} A_i\right)
M_i^{-1} A_i^\top D_{\pi_\theta} \\
&=
A_i M_i^{-1} M_i M_i^{-1} A_i^\top D_{\pi_\theta} \\
&=
A_i M_i^{-1} A_i^\top D_{\pi_\theta} \\
&=
\Pi^{\phi_i}_{D_{\pi_\theta},S_{\phi_i,e}^{\perp}}.
\end{align*}
Hence $\Pi^{\phi_i}_{D_{\pi_\theta},S_{\phi_i,e}^{\perp}}$ is idempotent.
\item Next we prove self-adjointness in the $D_{\pi_\theta}$-inner product. For any
$x,y \in \mathbb{R}^{|S_g||A_g|}$,
\begin{align*}
\left\langle
x,\Pi^{\phi_i}_{D_{\pi_\theta},S_{\phi_i,e}^{\perp}}y
\right\rangle_{D_{\pi_\theta}}
&=
x^\top D_{\pi_\theta} A_i M_i^{-1} A_i^\top D_{\pi_\theta} y \\
&=
\left(A_i M_i^{-1} A_i^\top D_{\pi_\theta} x\right)^\top D_{\pi_\theta} y \\
&=
\left\langle
\Pi^{\phi_i}_{D_{\pi_\theta},S_{\phi_i,e}^{\perp}}x,y
\right\rangle_{D_{\pi_\theta}},
\end{align*}
where we used the symmetry of $M_i^{-1}$. Hence
$\Pi^{\phi_i}_{D_{\pi_\theta},S_{\phi_i,e}^{\perp}}$ is self-adjoint with respect to
$\langle \cdot,\cdot\rangle_{D_{\pi_\theta}}$.
\item Finally, we prove $\left\langle
u_i^{(\bar c)},b_i^{(\bar c)}
\right\rangle_{D_{\pi_\theta}}
=0$. 
\begin{align*}
\left\langle
u_i^{(\bar c)},b_i^{(\bar c)}
\right\rangle_{D_{\pi_\theta}}
&=
\left\langle
\Pi^{\phi_i}_{D_{\pi_\theta},S_{\phi_i,e}^{\perp}} e_i^{(\bar c)},
\left(I-\Pi^{\phi_i}_{D_{\pi_\theta},S_{\phi_i,e}^{\perp}}\right)e_i^{(\bar c)}
\right\rangle_{D_{\pi_\theta}} \\
&=
\left\langle
\Pi^{\phi_i}_{D_{\pi_\theta},S_{\phi_i,e}^{\perp}} e_i^{(\bar c)},
e_i^{(\bar c)}
\right\rangle_{D_{\pi_\theta}}
-
\left\langle
\Pi^{\phi_i}_{D_{\pi_\theta},S_{\phi_i,e}^{\perp}} e_i^{(\bar c)},
\Pi^{\phi_i}_{D_{\pi_\theta},S_{\phi_i,e}^{\perp}} e_i^{(\bar c)}
\right\rangle_{D_{\pi_\theta}}.
\end{align*}
Using idempotence and self-adjointness,
\begin{align*}
    \left\langle
\Pi^{\phi_i}_{D_{\pi_\theta},S_{\phi_i,e}^{\perp}} e_i^{(\bar c)},
e_i^{(\bar c)}
\right\rangle_{D_{\pi_\theta}}
&=
\left\langle
\left(\Pi^{\phi_i}_{D_{\pi_\theta},S_{\phi_i,e}^{\perp}}\right)^2 e_i^{(\bar c)},
e_i^{(\bar c)}
\right\rangle_{D_{\pi_\theta}} \\&
=\left\langle
\Pi^{\phi_i}_{D_{\pi_\theta},S_{\phi_i,e}^{\perp}} e_i^{(\bar c)},
\Pi^{\phi_i}_{D_{\pi_\theta},S_{\phi_i,e}^{\perp}} e_i^{(\bar c)}
\right\rangle_{D_{\pi_\theta}}.
\end{align*}
Therefore
\[
\left\langle
u_i^{(\bar c)},b_i^{(\bar c)}
\right\rangle_{D_{\pi_\theta}}
=0.
\]
\end{enumerate}
\end{proof}

Before analyzing the TD iterates, we bound the error between the approximation subspace fixed point $\phi_i \widehat{Q}_i^\star$ and the true $Q$-function $Q_i^\theta$. This truncation error quantifies the price of restricting to $k$-hop neighborhood observations. 

\subsubsection{Approximation floor: localization and linear FA}
\begin{lemma}[Local truncation error bound]\label{lem:local-truncation}
Let $D_{\pi_\theta}$ denote the diagonal matrix of the stationary distribution $\pi_\theta$,
and let $\Pi^{\phi_i}_{D_{\pi_\theta},\,S_{\phi_i,e}^{\perp}}$ denote the $D_{\pi_\theta}$-orthogonal projection
onto the approximation subspace associated with $\phi_i$ under the constraint $S_{\phi_i,e}^{\perp}$.
Let $\widehat Q_i^\star$ be a fixed point in this  such that
\[
\phi_i \widehat Q_i^\star = \Pi^{\phi_i}_{D_{\pi_\theta},S_{\phi_i,e}^{\perp}}
T_{\pi_\theta}^i(\phi_i \widehat Q_i^\star).
\]
For any scalar $\bar c \in \mathbb{R}$, define
\(
Q_i^{(\bar c)} := Q^\theta_i + \bar c e,\) and \(
e_i^{(\bar c)} := \phi_i \widehat Q_i^\star - Q_i^{(\bar c)},
\)
where $Q^\theta_i$ is any solution to the original Bellman equation $Q^\theta_i = T^i_{\pi_\theta}Q^\theta_i$. Then, 
\small{
\[
\!\bigl\| \phi_i \widehat Q_i^\star \!-\! (Q^\theta_i \!+ \!\bar c e)\bigr\|_{D_{\pi_\theta}}
\!\!\le \!
\sqrt{
\!1\!+\!\frac{\lambda_{\max}^2(\phi_i^\top D_{\pi_\theta}\phi_i)}{(\Delta_i^{\phi_i})^2}
}\!
\sqrt{\!
\lambda_{\max}(D_{\pi_\theta})|S_g||A_g|\,c^2 \rho^{2(k+1)}
\!+\!
\frac{\bigl(e^\top q_i(\bar c)\bigr)^2}
{e^\top (\phi_i^\top D_{\pi_\theta}\phi_i)^{-1} e}
},
\]
}
\normalsize
{where \(
q_i(\bar c)
:=
(\phi_i^\top D_{\pi_\theta}\phi_i)^{-1}\phi_i^\top D_{\pi_\theta}(Q^\theta_i+\bar c e).
\)}

\end{lemma}
\begin{proof}
Recall the usual $D_{\pi_\theta}$-projection onto $\mathrm{span}(\phi_i)$:
\[
\Pi^{\phi_i}_{D_{\pi_\theta}}
\;:=\;
\phi_i\bigl(\phi_i^\intercal D_{\pi_\theta}\phi_i\bigr)^{-1}\phi_i^\intercal D_{\pi_\theta}.
\]
In particular, the (unconstrained) projected fixed point $\widehat Q_i^\infty$ is the solution to
\[
\phi_i \widehat Q_i^\infty
=
\phi_i\bigl(\phi_i^\intercal D_{\pi_\theta}\phi_i\bigr)^{-1}\phi_i^\intercal D_{\pi_\theta}
\,T_{\pi_\theta}^i\!\bigl(\phi_i \widehat Q_i^\infty\bigr)
=
\Pi^{\phi_i}_{D_{\pi_\theta}}\,T_{\pi_\theta}^i\!\bigl(\phi_i \widehat Q_i^\infty\bigr).
\]
Following Lemma~\ref{lem:solution_Existence} the fixed point need not be unique due to
the additive-constant ambiguity; we therefore work with
$\Pi^{\phi_i}_{D_{\pi_\theta},\,S_{\phi_i,e}^{\perp}}$, defined analogously as in Lemma~\ref{lem:solution_Existence} where a unique solution $\widehat Q_i^\star$ exists, satisfying
\begin{equation}
\label{eq:unique_fp}
    \phi_i \widehat Q_i^\star
=
\Pi^{\phi_i}_{D_{\pi_\theta},\,S_{\phi_i,e}^{\perp}}\,T_{\pi_\theta}^i\!\bigl(\phi_i \widehat Q_i^\star\bigr).
\end{equation}
Since $Q^\theta_i = T_{\pi_\theta}^i Q^\theta_i$ and
\[
T_{\pi_\theta}^i(Q^\theta_i+\bar c e)
=
r_i - J_i(\theta)e + P_{\pi_\theta}(Q^\theta_i+\bar c e)
=
r_i - J_i(\theta)e + P_{\pi_\theta}Q^\theta_i + \bar c P_{\pi_\theta}e
=
Q^\theta_i + \bar c e,
\]
where we used $P_{\pi_\theta}e=e$, it follows that
\[
Q_i^{(\bar c)} = T_{\pi_\theta}^i Q_i^{(\bar c)}.
\]
Define
\[
u_i^{(\bar c)}
:=
\Pi^{\phi_i}_{D_{\pi_\theta},S_{\phi_i,e}^{\perp}} e_i^{(\bar c)},
\qquad
b_i^{(\bar c)}
:=
\bigl(I-\Pi^{\phi_i}_{D_{\pi_\theta},S_{\phi_i,e}^{\perp}}\bigr)e_i^{(\bar c)},
\]
so that
\[
e_i^{(\bar c)} = u_i^{(\bar c)} + b_i^{(\bar c)}.
\]
Since, 
\begin{align*}
    \langle u_i^{(\bar c)}, b_i^{(\bar c)}\rangle_{D_{\pi_\theta}} & =
    \left\langle
    \Pi^{\phi_i}_{D_{\pi_\theta},S_{\phi_i,e}^{\perp}} e_i^{(\bar c)},
    \bigl(I-\Pi^{\phi_i}_{D_{\pi_\theta},S_{\phi_i,e}^{\perp}}\bigr)e_i^{(\bar c)}
    \right\rangle_{D_{\pi_\theta}} = 0
\end{align*}
where the last equality follows from Proposition~\ref{prop:basic_props_constrained_proj}. Therefore,
\begin{equation}
    \|e_i^{(\bar c)}\|_{D_{\pi_\theta}}^2 = \|u_i^{(\bar c)}\|_{D_{\pi_\theta}}^2 + \|b_i^{(\bar c)}\|_{D_{\pi_\theta}}^2.
\label{eq:error_decomp}
\end{equation}
\noindent Using the definition of $u_i^{(\bar c)}$, 
\begin{align*}
    u_i^{(\bar c)} = \Pi^{\phi_i}_{D_{\pi_\theta},S_{\phi_i,e}^{\perp}} \bigl(\phi_i\widehat Q_i^\star - Q_i^{(\bar c)}\bigr) &= \phi_i\widehat Q_i^\star - \Pi^{\phi_i}_{D_{\pi_\theta},S_{\phi_i,e}^{\perp}}Q_i^{(\bar c)} \\
    &= \Pi^{\phi_i}_{D_{\pi_\theta},S_{\phi_i,e}^{\perp}} T_{\pi_\theta}^i(\phi_i\widehat Q_i^\star) - \Pi^{\phi_i}_{D_{\pi_\theta},S_{\phi_i,e}^{\perp}} T_{\pi_\theta}^i(Q_i^{(\bar c)}) \\
    &= \Pi^{\phi_i}_{D_{\pi_\theta},S_{\phi_i,e}^{\perp}} P_{\pi_\theta} \bigl(\phi_i\widehat Q_i^\star - Q_i^{(\bar c)}\bigr) \\
    & = 
    \Pi^{\phi_i}_{D_{\pi_\theta},S_{\phi_i,e}^{\perp}} P_{\pi_\theta} \bigl(u_i^{(\bar c)} + b_i^{(\bar c)}\bigr).
\end{align*}
Rearranging,
\begin{equation}
    \Bigl(
I-\Pi^{\phi_i}_{D_{\pi_\theta},S_{\phi_i,e}^{\perp}}P_{\pi_\theta}
\Bigr)
u_i^{(\bar c)}
=
\Pi^{\phi_i}_{D_{\pi_\theta},S_{\phi_i,e}^{\perp}}P_{\pi_\theta}b_i^{(\bar c)}.
\label{eq:relation_between_u_and_b}
\end{equation}
Taking $D_{\pi_\theta}$-norms gives
\[
\left\|
\Bigl(
I-\Pi^{\phi_i}_{D_{\pi_\theta},S_{\phi_i,e}^{\perp}}P_{\pi_\theta}
\Bigr)
u_i^{(\bar c)}
\right\|_{D_{\pi_\theta}}
=
\left\|
\Pi^{\phi_i}_{D_{\pi_\theta},S_{\phi_i,e}^{\perp}}P_{\pi_\theta}b_i^{(\bar c)}
\right\|_{D_{\pi_\theta}}
\le
\|P_{\pi_\theta}b_i^{(\bar c)}\|_{D_{\pi_\theta}},
\]
where the last inequality uses the non-expansivity of an orthogonal projection in the induced norm. Since $(P_{\pi_\theta}b_i^{(\bar c)})(s_g,a_g) = \mathbb E_{P_{\pi_\theta}}\bigl[b_i^{(\bar c)}(s_g^+,a_g^+) \mid s_g,a_g\bigr]$, we obtain the following,
\begin{align*}
    \|P_{\pi_\theta}b_i^{(\bar c)}\|_{D_{\pi_\theta}}^2 = \mathbb E_{D_{\pi_\theta}} \left[ \mathbb E\bigl[b_i^{(\bar c)}(s_g^+,a_g^+) \mid s_g,a_g\bigr]^2 \right]
    &\le \mathbb E_{D_{\pi_\theta}} \left[\mathbb E\bigl[b_i^{(\bar c)}(s_g^+,a_g^+)^2 \mid s_g,a_g\bigr] \right] \\
    & = \mathbb E_{D_{\pi_\theta}} \bigl[b_i^{(\bar c)}(s_g,a_g)^2\bigr] = \|b_i^{(\bar c)}\|_{D_{\pi_\theta}}^2
\end{align*}
where the inequality follows from Jensen and the last equality uses stationarity of $D_{\pi_\theta}$. Hence
\begin{equation}
    \left\| \Bigl(I-\Pi^{\phi_i}_{D_{\pi_\theta},S_{\phi_i,e}^{\perp}}P_{\pi_\theta} \Bigr)u_i^{(\bar c)}\right\|_{D_{\pi_\theta}}\le \|b_i^{(\bar c)}\|_{D_{\pi_\theta}}.
    \label{eq:norms_of_u_and_b}
\end{equation}
Consider 
\[
\left\langle
u_i^{(\bar c)},
\Bigl(
I-\Pi^{\phi_i}_{D_{\pi_\theta},S_{\phi_i,e}^{\perp}}P_{\pi_\theta}
\Bigr)
u_i^{(\bar c)}
\right\rangle_{D_{\pi_\theta}}.
\]
Upon expanding, 
\[
\left\langle
u_i^{(\bar c)},
\Bigl(
I-\Pi^{\phi_i}_{D_{\pi_\theta},S_{\phi_i,e}^{\perp}}P_{\pi_\theta}
\Bigr)
u_i^{(\bar c)}
\right\rangle_{D_{\pi_\theta}}
=
\langle u_i^{(\bar c)},u_i^{(\bar c)}\rangle_{D_{\pi_\theta}}
-
\left\langle
u_i^{(\bar c)},
\Pi^{\phi_i}_{D_{\pi_\theta},S_{\phi_i,e}^{\perp}}P_{\pi_\theta}u_i^{(\bar c)}
\right\rangle_{D_{\pi_\theta}}.
\]
Using self-adjointness of the orthogonal projection and the fact that
$u_i^{(\bar c)}$ lies in the range of $\Pi^{\phi_i}_{D_{\pi_\theta},S_{\phi_i,e}^{\perp}}$,
\[
\left\langle
u_i^{(\bar c)},
\Pi^{\phi_i}_{D_{\pi_\theta},S_{\phi_i,e}^{\perp}}P_{\pi_\theta}u_i^{(\bar c)}
\right\rangle_{D_{\pi_\theta}}
=
\left\langle
\Pi^{\phi_i}_{D_{\pi_\theta},S_{\phi_i,e}^{\perp}}u_i^{(\bar c)},
P_{\pi_\theta}u_i^{(\bar c)}
\right\rangle_{D_{\pi_\theta}}
=
\langle
u_i^{(\bar c)},P_{\pi_\theta}u_i^{(\bar c)}
\rangle_{D_{\pi_\theta}}.
\]
Therefore,
\[
\left\langle
u_i^{(\bar c)},
\Bigl(
I-\Pi^{\phi_i}_{D_{\pi_\theta},S_{\phi_i,e}^{\perp}}P_{\pi_\theta}
\Bigr)
u_i^{(\bar c)}
\right\rangle_{D_{\pi_\theta}}
=
u_i^{(\bar c)\top}D_{\pi_\theta}(I-P_{\pi_\theta})u_i^{(\bar c)}.
\]
Since $u_i^{(\bar c)}$ lies in the constrained projection subspace, we may write
\[
u_i^{(\bar c)} = \phi_i v_i^{(\bar c)}
\]
for some coefficient vector $v_i^{(\bar c)} \in S_{\phi_i,e}^{\perp}$. Then
\[
u_i^{(\bar c)\top}D_{\pi_\theta}(I-P_{\pi_\theta})u_i^{(\bar c)}
=
v_i^{(\bar c)\top}\phi_i^\top D_{\pi_\theta}(I-P_{\pi_\theta})\phi_i v_i^{(\bar c)}
\ge
\Delta_i^{\phi_i}\|v_i^{(\bar c)}\|_2^2,
\]
by the definition of $\Delta_i^{\phi_i}$ in Lemma~\ref{lem:contrac_coeffs}. Moreover,
\[
\|u_i^{(\bar c)}\|_{D_{\pi_\theta}}^2
=
v_i^{(\bar c)\top}\phi_i^\top D_{\pi_\theta}\phi_i v_i^{(\bar c)}
\le
\lambda_{\max}(\phi_i^\top D_{\pi_\theta}\phi_i)\|v_i^{(\bar c)}\|_2^2.
\]
Hence
\[
\|v_i^{(\bar c)}\|_2^2
\ge
\frac{\|u_i^{(\bar c)}\|_{D_{\pi_\theta}}^2}
{\lambda_{\max}(\phi_i^\top D_{\pi_\theta}\phi_i)}.
\]
Combining, 
\begin{equation}
    \left\langle u_i^{(\bar c)}, \Bigl(I-\Pi^{\phi_i}_{D_{\pi_\theta},S_{\phi_i,e}^{\perp}}P_{\pi_\theta} \Bigr) u_i^{(\bar c)} \right\rangle_{D_{\pi_\theta}}
    \ge \frac{\Delta_i^{\phi_i}} {\lambda_{\max}(\phi_i^\top D_{\pi_\theta}\phi_i)} \|u_i^{(\bar c)}\|_{D_{\pi_\theta}}^2.
\label{eq:bound_u}
\end{equation}
Finally, by Cauchy--Schwarz,
\[ 
    \left\langle u_i^{(\bar c)}, \Bigl( I-\Pi^{\phi_i}_{D_{\pi_\theta},S_{\phi_i,e}^{\perp}}P_{\pi_\theta} \Bigr) u_i^{(\bar c)} \right\rangle_{D_{\pi_\theta}}
    \le \|u_i^{(\bar c)}\|_{D_{\pi_\theta}} \left\| \Bigl(I-\Pi^{\phi_i}_{D_{\pi_\theta},S_{\phi_i,e}^{\perp}}P_{\pi_\theta}\Bigr) u_i^{(\bar c)} \right\|_{D_{\pi_\theta}}.
\]
Together with \eqref{eq:bound_u}, this yields,
\[ \frac{\Delta_i^{\phi_i}} {\lambda_{\max}(\phi_i^\top D_{\pi_\theta}\phi_i)} \|u_i^{(\bar c)}\|_{D_{\pi_\theta}}
\le\left\| \Bigl(I-\Pi^{\phi_i}_{D_{\pi_\theta},S_{\phi_i,e}^{\perp}}P_{\pi_\theta}\Bigr)u_i^{(\bar c)} \right\|_{D_{\pi_\theta}}.
\]
Using \eqref{eq:norms_of_u_and_b}, we obtain,
\[
\|u_i^{(\bar c)}\|_{D_{\pi_\theta}}
\le
\frac{\lambda_{\max}(\phi_i^\top D_{\pi_\theta}\phi_i)}
{\Delta_i^{\phi_i}}
\|b_i^{(\bar c)}\|_{D_{\pi_\theta}}.
\]
Plugging this into \eqref{eq:error_decomp}, we obtain, 
\begin{equation}
\|e_i^{(\bar c)}\|_{D_{\pi_\theta}}^2
=
\|u_i^{(\bar c)}\|_{D_{\pi_\theta}}^2
+
\|b_i^{(\bar c)}\|_{D_{\pi_\theta}}^2
\le
\left(
1+
\frac{\lambda_{\max}^2(\phi_i^\top D_{\pi_\theta}\phi_i)}
{(\Delta_i^{\phi_i})^2}
\right)
\|b_i^{(\bar c)}\|_{D_{\pi_\theta}}^2.
\label{eq:semi_Error_bound}
\end{equation}
We now bound $b_i^{(\bar c)}$. Define the following unconstrained projection vector,
\[
q_i(\bar c)
:=
(\phi_i^\top D_{\pi_\theta}\phi_i)^{-1}\phi_i^\top D_{\pi_\theta}(Q_i+\bar c e),
\]
so that
\[
\Pi^{\phi_i}_{D_{\pi_\theta}}(Q^\theta_i+\bar c e)
=
\phi_i q_i(\bar c).
\]
By definition of the constrained projection, the projection vector of
$\Pi^{\phi_i}_{D_{\pi_\theta},S_{\phi_i,e}^{\perp}}(Q^\theta_i+\bar c e)$ is obtained by solving
\[
\min_{\nu:\, \nu^\top e = 0}
\| (Q^\theta_i+\bar c e) - \phi_i \nu \|_{D_{\pi_\theta}}^2.
\]
Since $\phi_i q_i(\bar c)$ is the unconstrained $D_{\pi_\theta}$-projection of $Q^\theta_i+\bar c e$
onto $\mathrm{span}(\phi_i)$, we may write
\[
(Q^\theta_i+\bar c e)-\phi_i\nu
=
\bigl((Q^\theta_i+\bar c e)-\phi_i q_i(\bar c)\bigr)
+
\phi_i\bigl(q_i(\bar c)-\nu\bigr).
\]
The first term is $D_{\pi_\theta}$-orthogonal to $\mathrm{span}(\phi_i)$, while the second term lies
in $\mathrm{span}(\phi_i)$. Hence the cross term vanishes, and minimizing the previous display is
equivalent to minimizing
\[
\|\phi_i(q_i(\bar c)-\nu)\|_{D_{\pi_\theta}}^2
=
(\nu-q_i(\bar c))^\top (\phi_i^\top D_{\pi_\theta}\phi_i)(\nu-q_i(\bar c))
\]
subject to $\nu^\top e=0$. Hence, consider the corresponding Lagrangian
\[
\mathcal{L}(\nu,\lambda)
=
(\nu-q_i(\bar c))^\top (\phi_i^\top D_{\pi_\theta}\phi_i)(\nu-q_i(\bar c))
+
\lambda \,\nu^\top e.
\]
Setting the gradient with respect to $\nu$ equal to zero gives
\[
2(\phi_i^\top D_{\pi_\theta}\phi_i)(\nu-q_i(\bar c)) + \lambda e = 0.
\]
This yields, 
\[
\nu
=
q_i(\bar c)
-
\frac{\lambda}{2}
(\phi_i^\top D_{\pi_\theta}\phi_i)^{-1}e.
\]
Imposing the constraint $\nu^\top e=0$ results in
\[
0
=
e^\top q_i(\bar c)
-
\frac{\lambda}{2}
e^\top (\phi_i^\top D_{\pi_\theta}\phi_i)^{-1}e,
\]
and therefore
\[
\frac{\lambda}{2}
=
\frac{e^\top q_i(\bar c)}
{e^\top (\phi_i^\top D_{\pi_\theta}\phi_i)^{-1}e}.
\]
Substituting back gives the constrained projection coefficients
\begin{equation}
    \nu_i(\bar c)
=
q_i(\bar c)
-
\frac{e^\top q_i(\bar c)}
{e^\top (\phi_i^\top D_{\pi_\theta}\phi_i)^{-1}e}
(\phi_i^\top D_{\pi_\theta}\phi_i)^{-1}e.
\label{eq:coeff_v_constrained}
\end{equation}
Therefore,
\[
\Pi^{\phi_i}_{D_{\pi_\theta},S_{\phi_i,e}^{\perp}}(Q^\theta_i+\bar c e)
=
\phi_i \nu_i(\bar c).
\]
Subtracting the unconstrained projection,
\[
\Pi^{\phi_i}_{D_{\pi_\theta}}(Q^\theta_i+\bar c e)
-
\Pi^{\phi_i}_{D_{\pi_\theta},S_{\phi_i,e}^{\perp}}(Q^\theta_i+\bar c e)
=
\phi_i\bigl(q_i(\bar c)-\nu_i(\bar c)\bigr)
\]
\[
=
\frac{e^\top q_i(\bar c)}
{e^\top (\phi_i^\top D_{\pi_\theta}\phi_i)^{-1}e}
\;
\phi_i(\phi_i^\top D_{\pi_\theta}\phi_i)^{-1}e.
\]
Taking the $D_{\pi_\theta}$ norm yields, 
\begin{equation}
    \left\|
\Pi_{\phi_i}^{D_{\pi_\theta}}(Q^\theta_i+\bar c e)
-
\Pi^{\phi_i}_{D_{\pi_\theta},S_{\phi_i,e}^{\perp}}(Q^\theta_i+\bar c e)
\right\|_{D_{\pi_\theta}}^2
=
\frac{\bigl(e^\top q_i(\bar c)\bigr)^2}
{e^\top (\phi_i^\top D_{\pi_\theta}\phi_i)^{-1}e}.
\label{eq:difference_between_projections}
\end{equation}
Note that $b_i^{(\bar c)}$ can be rewritten as:
\[
b_i^{(\bar c)} = \Bigl(
\Pi^{\phi_i}_{D_{\pi_\theta},S_{\phi_i,e}^{\perp}}(Q^\theta_i+\bar c e)
-
\Pi^{\phi_i}_{D_{\pi_\theta}}(Q^\theta_i+\bar c e)
\Bigr)
+
\Bigl(
\Pi^{\phi_i}_{D_{\pi_\theta}}(Q^\theta_i+\bar c e)
-
(Q^\theta_i+\bar c e)
\Bigr).
\]
Here, the first term lies in $\mathrm{span}(\phi_i)$, while the second term is the residual of the
unconstrained orthogonal projection onto $\mathrm{span}(\phi_i)$, hence is $D_{\pi_\theta}$-orthogonal
to $\mathrm{span}(\phi_i)$. Therefore these two terms are orthogonal, and
\begin{equation}
\begin{aligned}
    \|b_i^{(\bar c)}\|_{D_{\pi_\theta}}^2
&=
\left\|
\Pi^{\phi_i}_{D_{\pi_\theta},S_{\phi_i,e}^{\perp}}(Q^\theta_i+\bar c e)
-
\Pi_{\phi_i}^{D_{\pi_\theta}}(Q^\theta_i+\bar c e)
\right\|_{D_{\pi_\theta}}^2
\\
&\quad+
\left\|
\Pi_{\phi_i}^{D_{\pi_\theta}}(Q^\theta_i+\bar c e)
-
(Q^\theta_i+\bar c e)
\right\|_{D_{\pi_\theta}}^2.
\end{aligned}
\label{eq:decomp_of_b}
\end{equation}
We now simplify the second term in \eqref{eq:decomp_of_b}. 
Since $\phi_i e=e$, we have
\[
\Pi^{\phi_i}_{D_{\pi_\theta}} e
=
\phi_i(\phi_i^\top D_{\pi_\theta}\phi_i)^{-1}\phi_i^\top D_{\pi_\theta}e
=
\phi_i(\phi_i^\top D_{\pi_\theta}\phi_i)^{-1}\phi_i^\top D_{\pi_\theta}\phi_i e
=
\phi_i e
=
e.
\]
Hence
\[
\Pi^{\phi_i}_{D_{\pi_\theta}}(Q^\theta_i+\bar c e) - (Q^\theta_i+\bar c e)
=
\Pi^{\phi_i}_{D_{\pi_\theta}}Q^\theta_i - Q^\theta_i.
\]
Substituting the above and \eqref{eq:difference_between_projections} into \eqref{eq:decomp_of_b} gives
\begin{equation}
    \|b_i^{(\bar c)}\|_{D_{\pi_\theta}}^2
=
\left\|
\Pi_{\phi_i}^{D_{\pi_\theta}}Q^\theta_i - Q^\theta_i
\right\|_{D_{\pi_\theta}}^2
+
\frac{\bigl(e^\top q_i(\bar c)\bigr)^2}
{e^\top (\phi_i^\top D_{\pi_\theta}\phi_i)^{-1}e}.
\label{eq:bound_on_b}
\end{equation}
From \cite{qu2020scalable}, the unconstrained projection $\Pi^{\phi_i}_{D_{\pi_\theta}}Q^\theta_i$ averages over
the coordinates outside $N_i^k$. Therefore, by the exponential decay property,
\[
\left\|
\Pi^{\phi_i}_{D_{\pi_\theta}}Q^\theta_i - Q^\theta_i
\right\|_{\infty}
\le
c\rho^{k+1}.
\]
Finally, converting the $\infty$-norm to the $D_{\pi_\theta}$-norm,
\[
\left\|
\Pi_{\phi_i}^{D_{\pi_\theta}}Q^\theta_i - Q^\theta_i
\right\|_{D_{\pi_\theta}}
\le
\sqrt{\lambda_{\max}(D_{\pi_\theta})}\,
\left\|
\Pi_{\phi_i}^{D_{\pi_\theta}}Q^\theta_i - Q^\theta_i
\right\|_2
\le
\sqrt{\lambda_{\max}(D_{\pi_\theta})\,|S_g|\,|A_g|}\,
c\rho^{k+1}.
\]
Combining this with \eqref{eq:bound_on_b}, we obtain, 
\[
\|b_i^{(\bar c)}\|_{D_{\pi_\theta}}
\le
\sqrt{
\lambda_{\max}(D_{\pi_\theta})\,|S_g|\,|A_g|\,c^2 \rho^{2(k+1)}
+
\frac{\bigl(e^\top q_i(\bar c)\bigr)^2}
{e^\top (\phi_i^\top D_{\pi_\theta}\phi_i)^{-1}e}
}.
\]
Substituting this bound into \eqref{eq:semi_Error_bound}, 
\[
\|e_i^{(\bar c)}\|_{D_{\pi_\theta}}
\le
\sqrt{
1+\frac{\lambda_{\max}^2(\phi_i^\top D_{\pi_\theta}\phi_i)}{(\Delta_i^{\phi_i})^2}
}
\;
\sqrt{
\lambda_{\max}(D_{\pi_\theta})\,|S_g|\,|A_g|\,c^2 \rho^{2(k+1)}
+
\frac{\bigl(e^\top q_i(\bar c)\bigr)^2}
{e^\top (\phi_i^\top D_{\pi_\theta}\phi_i)^{-1}e}
}.
\]
\end{proof}

\begin{remark}
\label{remark2}
Since \(\phi_i e = e\), the additional term $\frac{(e^\top q_i(\bar c))^2}{e^\top(\phi_i^\top D_{\pi_\theta}\phi_i)^{-1}e}$ in Lemma~\ref{lem:local-truncation} is minimized by choosing 
\[
\bar c_i^\star
=
-\frac{e^\top (\phi_i^\top D_{\pi_\theta}\phi_i)^{-1}\phi_i^\top D_{\pi_\theta}Q^\theta_i }{|S_{N_i^k}||A_{N_i^k}|},
\]
for which \(e^\top q_i(\bar c_i^\star)=0\). In particular, this choice removes the
constraint-mismatch term entirely, and yields
\[
\inf_{\bar c\in\mathbb R}
\|\phi_i\widehat Q_i^\star-(Q^\theta_i+\bar c e)\|_{D_{\pi_\theta}}
\le
\sqrt{1+\frac{\lambda_{\max}^2(\phi_i^\top D_{\pi_\theta}\phi_i)}{(\Delta_i^{\phi_i})^2}}\,
\sqrt{\lambda_{\max}(D_{\pi_\theta})\,|S_g|\,|A_g|\,c^2\rho^{2(k+1)}}.
\]
\end{remark}

\subsubsection{Fixed-policy TD convergence}
Define $\tau(\beta)\coloneqq \min\{\tau:\|\mathbb E \left[ A_i(X_t)|X_0\right]-A_i^\infty \|\leq \beta, \|\mathbb E \left[ b_i(X_t)|X_0\right]-b_i^\infty \|\leq \beta \}$. 

\begin{lemma}[Single step contraction for fixed policy]
    Let $V_i^t := \|\textcolor{revisionblue}{\Pi_i^{\perp}}(w_i^t - w_i^\star)\|_2^2,$ where $\textcolor{revisionblue}{\Pi_i^{\perp}} := \Pi_{2, S_{\Psi_i,e}^\perp}$ is the Euclidean projection onto $S_{\Psi_i,e}^\perp$. Consider a fixed policy $\pi_\theta$, and a fixed step size $\beta$ satisfying $\beta \Delta_i^{\chi_i} < 2$, $c_{ss}\beta\le 1$, $\beta\le \frac{\Delta_i^{\chi_i}}{32\bar\psi^4},$ and $\beta\, \tau(\beta) \leq \min\!\left\{\frac{1}{4\sqrt{L_i}},\; \frac{\Delta_i^{\chi_i}}{228L_i}\right\},$ where $\bar\psi
:=
\max_{j\in\mathcal N}
\max_{s_j^k,a_j^k}
\|\psi_j(s_j^k,a_j^k)\|_2 $, $L_i:=9c_{ss}^2+\bar\psi^2(3+4\bar\psi)^2,$ and $\tau(\beta)$ is the mixing time of the Markov chain at rate $\beta$, and $c_{ss} \geq \Delta_i^{\chi_i} + \sqrt{(\Delta_i^{\chi_i})^{-2} - 1}$. Then for all $t \geq T_{\mu,i}(\beta) + \tau(\beta)$:
    \begin{equation}
        \mathbb{E}[V_i^{t+1}] \leq \left(1 - \frac{\Delta_i^{\chi_i}}{2}\,\beta\right) \mathbb{E}[V_i^t] + \beta^2\, C_{\mathrm{bias},i},
    \end{equation}
    where $C_{\mathrm{bias},i}
:=
{\color{black}
\frac{2\|\bar\psi_i\|_2^2}{\Delta_i^{\chi_i}}C_{\mu,i}
}
+
{\color{black}
32\bar\psi^2
\left(
r_{\max}^2+\bar\psi^2\|w_i^\star\|_2^2
\right)
+
4\bar\psi^2 C_{\mu,i} } +
56{\color{black}L_i}\tau(\beta)
\left(
{\color{black}
B_i^2+C_{\mu,i}
}
\right). $ Here $B_i^2:=r_{\max}^2+(J_i(\theta))^2+\|w_i^\star\|_2^2+1,$ $\widehat\sigma_i^2
:=
\sigma_i^2+8r_{\max}^2C_{\rho_\text{cov}}+\frac{4r_{\max}}{c_{ss}},$  $C_{\mu,i}:=1+c_{ss}\widehat\sigma_i^2,$ $\bar\psi_i
:=
\mathbb E_{D_{\pi_\theta}}[\textcolor{revisionblue}{\psi_{i,t}}],$ where $\sigma_i^2:=\mathbb E[(r_i^t-J_i(\theta))^2]$ and $C_{\rho_\text{cov}}$ is the geometric-mixing covariance constant.
\label{lem:fixed_actor_ss}
\end{lemma}
\begin{proof}
    Abbreviate $\textcolor{revisionblue}{\psi_{i,t}} := \psi_i(s_t^k, a_t^k)$, $\textcolor{revisionblue}{\psi_{i,t+1}} := \psi_i(s_{t+1}^k, a_{t+1}^k)$, $r_i^t := r_i(s_t, a_t)$. Define the projected error $\tilde{w}_i^t := \textcolor{revisionblue}{\Pi_i^{\perp}}(w_i^t - w_i^\star)$ and the average-reward error $\tilde\mu_i^t := \mu_i^t - J_i(\theta)$. It thus follows that $V_i^t = \|\tilde{w}_i^t\|_2^2$. Since $w_i^\star \in S_{\Psi_i,e}^\perp$ and $\textcolor{revisionblue}{\Pi_i^{\perp}}$ is the Euclidean projection onto $S_{\Psi_i,e}^\perp$, we have $\textcolor{revisionblue}{\Pi_i^{\perp}} w_i^\star = w_i^\star$.

    From update equation~\eqref{eq:update3},
    $$w_i^{t+1} - w_i^\star = (w_i^t - w_i^\star) + \beta\,\delta_i^t\,\textcolor{revisionblue}{\psi_{i,t}}.$$
    Applying $\textcolor{revisionblue}{\Pi_i^{\perp}}$ to both sides,
    $$\tilde{w}_i^{t+1} = \tilde{w}_i^t + \beta\,\textcolor{revisionblue}{\Pi_i^{\perp}}(\delta_i^t\textcolor{revisionblue}{\psi_{i,t}}).$$
    Squaring yields,
    \begin{equation}
        V_i^{t+1} = V_i^t + 2\beta\,\langle\tilde{w}_i^t,\,\textcolor{revisionblue}{\Pi_i^{\perp}}(\delta_i^t\textcolor{revisionblue}{\psi_{i,t}})\rangle + \beta^2\,\|\textcolor{revisionblue}{\Pi_i^{\perp}}(\delta_i^t\textcolor{revisionblue}{\psi_{i,t}})\|_2^2.
        \label{eq:one_step_drift}
    \end{equation}
    Since $\Psi_i w_e = e$, every feature satisfies $\psi_i(s,a)^\top w_e = 1$, so $(\textcolor{revisionblue}{\psi_{i,t+1}} - \textcolor{revisionblue}{\psi_{i,t}})^\top w_e = 0$. The TD error is therefore insensitive to the $S_{\Psi_i,e}$-component of $w_i^t$. Decomposing $w_i^t = w_i^\star + \tilde{w}_i^t + z_i^t$ with $z_i^t := (I-\textcolor{revisionblue}{\Pi_i^{\perp}})(w_i^t) \in S_{\Psi_i,e}$, the $z_i^t$ term vanishes:
    \begin{equation}
        \delta_i^t = r_i^t - Ji(\theta) - \tilde\mu_i^t + (\textcolor{revisionblue}{\psi_{i,t+1}} - \textcolor{revisionblue}{\psi_{i,t}})^\top(w_i^\star + \tilde{w}_i^t).
    \end{equation}
    Consider the following definitions:  $\bar{h}_i := \mathbb{E}_{D_{\pi_\theta}}[\delta_i^t\textcolor{revisionblue}{\psi_{i,t}} \mid \tilde{w}_i^t,\tilde\mu_i^t],$ $\bar\psi_i := \mathbb{E}_{D_{\pi_\theta}}[\textcolor{revisionblue}{\psi_{i,t}}]$, $A_i(\theta) := \chi_i^\top D_{\pi_\theta}(I-\textcolor{revisionblue}{P_{\pi_\theta}})\chi_i$ and $b_i(\theta) := \chi_i^\top D_{\pi_\theta}(r_i - J_i(\theta)\mathbf{1})$. Then,
    \begin{equation}
        \mathbb{E}_{D_{\pi_\theta}}[(r_i^t - \mu_i^t)\textcolor{revisionblue}{\psi_{i,t}}] = b_i(\theta) - \tilde\mu_i^t\bar\psi_i,
    \end{equation}
    and
    \begin{equation}                \mathbb{E}_{D_{\pi_\theta}}\bigl[(\textcolor{revisionblue}{\psi_{i,t+1}}-\textcolor{revisionblue}{\psi_{i,t}})^\top(w_i^\star+\tilde{w}_i^t)\,\textcolor{revisionblue}{\psi_{i,t}}\bigr] = -A_i(\theta)(w_i^\star+\tilde{w}_i^t).
    \end{equation}
    From Equation~\eqref{eq:lfa_fp}, it follows that, 
    \begin{equation}
        \bar{h}_i =  -A_i(\theta)\tilde{w}_i^t - \tilde\mu_i^t\bar\psi_i.
        \label{eq:h_bar}
    \end{equation}
    Hence, we can decompose update as follows:
    \begin{equation}
        \textcolor{revisionblue}{\Pi_i^{\perp}}(\delta_i^t\textcolor{revisionblue}{\psi_{i,t}}) = \textcolor{revisionblue}{\Pi_i^{\perp}}\bar{h}_i + \textcolor{revisionblue}{\Pi_i^{\perp}}\eta_i^t, \qquad \eta_i^t := \delta_i^t\textcolor{revisionblue}{\psi_{i,t}} - \bar{h}_i.
    \end{equation}
    Since $\textcolor{revisionblue}{\Pi_i^{\perp}}$ is symmetric and $\textcolor{revisionblue}{\Pi_i^{\perp}} \tilde{w}_i^t = \tilde{w}_i^t$,
    \begin{equation}
        \langle\tilde{w}_i^t, \textcolor{revisionblue}{\Pi_i^{\perp}}\bar{h}_i\rangle = -\tilde{w}_i^{t\top}A_i(\theta)\tilde{w}_i^t - \tilde\mu_i^t\,(\tilde{w}_i^t)^\top \textcolor{revisionblue}{\Pi_i^{\perp}}\bar\psi_i.
    \end{equation}
    Since $\tilde{w}_i^t \in S_{\Psi_i,e}^\perp$, we obtain 
    \begin{equation}
\tilde{w}_i^{t\top}A_i(\theta)\tilde{w}_i^t \geq \Delta_i^{\chi_i}\|\tilde{w}_i^t\|_2^2 = \Delta_i^{\chi_i}V_i^t.
\label{eq:term1}
    \end{equation}
    From Young's inequality ($2|ab| \leq \epsilon a^2 + \epsilon^{-1}b^2$) with $\epsilon = \Delta_i^{\chi_i}/2$:
    \begin{equation}
        2\beta\,|\tilde\mu_i^t\,(\tilde{w}_i^t)^\top \textcolor{revisionblue}{\Pi_i^{\perp}}\bar\psi_i| \leq 2\beta\,|\tilde\mu_i^t|\,\|\tilde{w}_i^t\|\,\|\bar\psi_i\| \leq \frac{\Delta_i^{\chi_i}\beta}{2}V_i^t + \frac{2\beta\|\bar\psi_i\|_2^2}{\Delta_i^{\chi_i}}(\tilde\mu_i^t)^2.
        \label{eq:term2_old}
    \end{equation}
    Combining Equations~\eqref{eq:term1} and \eqref{eq:term2_old} in Equation~\eqref{eq:one_step_drift}, we obtain,
    \begin{equation}
        2\beta\,\mathbb{E}[\langle\tilde{w}_i^t, \textcolor{revisionblue}{\Pi_i^{\perp}}\bar{h}_i\rangle] \leq  -\frac{3\Delta_i^{\chi_i}\beta}{2}\mathbb{E}[V_i^t] + \frac{2\beta\|\bar\psi_i\|_2^2}{\Delta_i^{\chi_i}}\mathbb{E}[(\tilde\mu_i^t)^2].
        \label{eq:final_bound_ss}
    \end{equation}
    We now bound the term $\mathbb{E}[(\tilde\mu_i^t)^2]$. 
    Recall that, 
    \begin{equation*}
        (\tilde\mu_i^{t+1})^2 = (1-c_{ss}\beta)^2(\tilde\mu_i^t)^2 + 2(1-c_{ss}\beta)c_{ss}\beta\,\tilde\mu_i^t(r_i^t - J_i(\theta)) + c_{ss}^2\beta^2(r_i^t - J_i(\theta))^2.
    \end{equation*}
    Decompose $r_i^t - J_i(\theta) = (r_i^t - \bar{r}_i(s_t)) + (\bar{r}_i(s_t) - J_i(\theta))$, where $\bar{r}_i(s) := \sum_a \pi_\theta(a|s)r_i(s,a)$. Then,
    \begin{equation*}
        \mathbb{E}\bigl[\tilde\mu_i^t(r_i^t - \bar{r}_i(s_t))\bigr] = \mathbb{E}\Bigl[\tilde\mu_i^t\cdot\mathbb{E}[r_i^t - \bar{r}_i(s_t)\mid\mathcal{G}_{t},s_t]\Bigr] = 0,
    \end{equation*}
    where $\mathcal G_t$ is the pre-action sigma-field containing the history up to time $t$, the current state $s_t$, and the current estimate $\mu_i^t$. Conditional on $\mathcal G_t$, the only remaining randomness in $r_i^t-\bar r_i(s_t)$ is the action drawn from the fixed policy $\pi_\theta$.
    To handle the second term, unroll $\tilde\mu_i^t$ to obtain,
    \begin{equation*}
        \tilde\mu_i^t = c_{ss}\beta\sum_{k=0}^{t-1}(1-c_{ss}\beta)^{t-1-k}(r_i^k - J_i(\theta)) + (1-c_{ss}\beta)^t\tilde\mu_i^0.
    \end{equation*}
     Let $M_{\mu,i}(\beta)
:=
\max\left\{
1,\,
\frac{
\mathbb E[(\widetilde\mu_i^0)^2]
+
\sqrt{\mathbb E[(\widetilde\mu_i^0)^2]}
}{\beta}
\right\}.$ Define the average reward burn in period $T_{\mu,i}(\beta)
:=
\left\lceil
\frac{\log M_{\mu,i}(\beta)}
{-\log(1-c_{ss}\beta)}
\right\rceil.$ By the definition of $T_{\mu,i}(\beta)$, whenever $t\ge T_{\mu,i}(\beta)$,
    \[
    (1-c_{ss}\beta)^t \mathbb E[(\tilde\mu_i^0)^2]\le \beta,
    \qquad
    (1-c_{ss}\beta)^t \sqrt{\mathbb E[(\tilde\mu_i^0)^2]}\le \beta.
    \]
    Hence the initialization transient contributes only an $O(\beta)$ term to
    $\mathbb E[\tilde\mu_i^t(r_i^t-J_i(\theta))]$, and therefore only an $O(\beta^2)$ term to the
    recursion for $\mathbb E[(\tilde\mu_i^{t+1})^2]$. This contribution is absorbed into the
    additional $\frac{4r_{\max}}{c_{ss}}$ term in the definition of $\widehat\sigma_i^2$. It then follows that, 
    \begin{equation}
        \left|\mathbb{E}\bigl[\tilde\mu_i^t\,(r_i^t - J_i(\theta))\bigr]\right| \le c_{ss}\beta\sum_{k=0}^{t-1}(1-c_{ss}\beta)^{t-1-k}\left|\operatorname{Cov}(r_i^k,\,r_i^t)\right| + 2 r_{\max}\beta
    \end{equation}
    By geometric mixing of the chain, $|\operatorname{Cov}(r_i^k, r_i^t)|\leq 4r_{\max}^2\rho^{t-k}_{\text{cov}}$ for some $\rho\in(0,1)$ set by the Markov chain's spectral gap. Since $(1-c_{ss}\beta){\rho_\text{cov}} < 1$, the geometric sum converges:
$$\sum_{k=0}^{t-1}(1-c_{ss}\beta)^{t-1-k}\rho^{t-k}_\text{cov} \leq \frac{{\rho_\text{cov}}}{1-{\rho_\text{cov}}(1-c_{ss}\beta)} \leq \frac{1}{1-{\rho_\text{cov}}} =: C_{\rho_\text{cov}}.$$
    Thus the cross term can be bounded as:
    \begin{equation}
        \left|2(1-c_{ss}\beta)c_{ss}\beta\,\mathbb{E}\bigl[\tilde\mu_i^t(\bar{r}_i(s_t)-J_i(\theta))\bigr]\right| \leq 8c_{ss}^2\beta^2 r_{\max}^2 C_{\rho_\text{cov}} +
        4c_{ss}\beta^2 r_{\max}.
    \end{equation}
    Let the variance of instantaneous reward be defined as $\sigma_i^2\coloneqq \mathbb{E}[(r_i^t - J_i(\theta))^2]$. Define $\hat\sigma_i^2 := \sigma_i^2 + 8r_{\max}^2C_{\rho_\text{cov}}+\frac{4r_{\max}}{c_{ss}}$. Since $(1-c_{ss}\beta)^2 \leq (1-c_{ss}\beta)$, we obtain, 
    \begin{equation}
        \mathbb{E}[(\tilde\mu_i^{t+1})^2] \leq (1-c_{ss}\beta)\mathbb{E}[(\tilde\mu_i^t)^2] + c_{ss}^2\beta^2\hat\sigma_i^2.
        \label{eq:mu_recursion_postburnin}
    \end{equation}

    Iterating \eqref{eq:mu_recursion_postburnin} gives, for every $s\ge0$,
    \[
    \mathbb E[(\widetilde\mu_i^s)^2]
    \le
    (1-c_{ss}\beta)^s\mathbb E[(\widetilde\mu_i^0)^2]
    +
    c_{ss}\beta\widehat\sigma_i^2.
    \]
    Hence, for every $s\ge T_{\mu,i}(\beta)$,
    \[
    \mathbb E[(\widetilde\mu_i^s)^2]
    \le
    \beta(1+c_{ss}\widehat\sigma_i^2)
    =
    \beta C_{\mu,i}.
    \]
    Since the lemma is invoked only for $t\ge T_{\mu,i}(\beta)+\tau(\beta)$, the bound
    \begin{equation}
        \mathbb E[(\widetilde\mu_i^s)^2]\le \beta C_{\mu,i}
        \qquad
        \text{holds for every }s\in\{t-\tau(\beta),\ldots,t\}.
        \label{eq:mu_post_burnin_bound}
    \end{equation}
    The role of $c_{ss} \geq \Delta_i^{\chi_i} + \sqrt{(\Delta_i^{\chi_i})^{-2} - 1}$ is to ensure that this average-reward burn-in is no slower
than the critic burn-in scale. Indeed, since \(c_{ss}\ge\Delta_i^{\chi_i}\) and
\(c_{ss}\beta\le1\),
\[
(1-c_{ss}\beta)^s
\le
(1-\Delta_i^{\chi_i}\beta)^s
\le
\left(1-\frac{\Delta_i^{\chi_i}}{2}\beta\right)^s,
\]
for all \(s\ge0\). Thus, once the critic transient is in the post-burn-in regime, the
average-reward transient is also controlled.
    
We now bound the markovian noise cross term $2\beta\mathbb E\left[\left\langle \widetilde w_i^t, \textcolor{revisionblue}{\Pi_i^{\perp}}\eta_i^t \right\rangle\right]$. Since the instantaneous rewards and features are bounded, with $\|\textcolor{revisionblue}{\Pi_i^{\perp}}\|\le 1$, it follows that:
    \begin{equation}
        \|\textcolor{revisionblue}{\Pi_i^{\perp}}(\delta_i^t\textcolor{revisionblue}{\psi_{i,t}})\|_2 \le \sqrt{L_i}R_i^t.
        \label{eq:bound_Pi}
    \end{equation}
    where $L_i\coloneqq 9c_{ss}^2+\bar\psi^2(3+4\bar\psi)^2 $, and $R_i^t \coloneqq
\left(
B_i^2+(\widetilde\mu_i^t)^2+V_i^t
\right)^{1/2}$ with $B_i^2 \coloneqq r_{\max}^2+(J_i(\theta))^2+\|w_i^\star\|_2^2+1.$ Thus, 
\begin{equation}
    \|\widetilde w_i^{\ell+1}-\widetilde w_i^\ell\|_2
\le
\beta\sqrt{L_i}R_i^\ell.
\label{eq:bound_w_diff}
\end{equation}
Moreover, the average-reward recursion gives
\[
|\widetilde\mu_i^{\ell+1}-\widetilde\mu_i^\ell|
\le
c_{ss}\beta\bigl(|r_i^\ell-J_i(\theta)|+|\widetilde\mu_i^\ell|\bigr)
\le
\beta\sqrt{L_i}\,R_i^\ell.
\]
Combining this with \eqref{eq:bound_w_diff} yields the corresponding one-step bound for the pair
$(\widetilde\mu_i^\ell,\widetilde w_i^\ell)$.
If $\beta\tau(\beta)\le \frac{1}{4\sqrt{L_i}},$ from reverse triangle inequality, we obtain,
\begin{equation}
    R_i^\ell\le 2R_i^t,
\qquad
\forall \ell\in[t-\tau(\beta),t].
\label{eq:R_bound}
\end{equation}
Telescoping Equation~\eqref{eq:bound_w_diff}, yields
\begin{equation}
    \|\widetilde w_i^t-\widetilde w_i^{t-\tau(\beta)}\|_2
\le
2\sqrt{L_i}\beta\tau(\beta)R_i^t.
\label{eq:bound_w_tau}
\end{equation}

Define $\textcolor{revisionblue}{\Pi_i^{\perp}}\eta_i^{t,\tau}
:=
\textcolor{revisionblue}{\Pi_i^{\perp}}\left(
\delta_i^{t,\tau}\textcolor{revisionblue}{\psi_{i,t}}-\bar h_i^{t,\tau}
\right),$ where $ \delta_i^{t,\tau}
:=
r_i^t - J_i(\theta) -\widetilde\mu_i^{t-\tau}
+
(\textcolor{revisionblue}{\psi_{i,t+1}}-\textcolor{revisionblue}{\psi_{i,t}})^\top
\left(w_i^\star+\widetilde w_i^{t-\tau}\right),$ and $\bar h_i^{t,\tau}
:=
\mathbb E_{D_{\pi_\theta}}
\left[
\delta_i^{t,\tau}\textcolor{revisionblue}{\psi_{i,t}}
\;\middle|\;
\widetilde\mu_i^{t-\tau},\widetilde w_i^{t-\tau}
\right].$ Then, from markov chain mixing, we obtain,
\begin{equation}
    \left\|
\mathbb E[\textcolor{revisionblue}{\Pi_i^{\perp}}\eta_i^{t,\tau}\mid \mathcal F_{t-\tau}]
\right\|_2
\le
\sqrt{L_i}\beta R_i^{t-\tau}.
\label{eq:conseq_of_mixing}
\end{equation}
Decompose
\begin{equation}
    \left\langle
\widetilde w_i^t,
\textcolor{revisionblue}{\Pi_i^{\perp}}\eta_i^t
\right\rangle
= \left\langle
\widetilde w_i^{t-\tau},
\textcolor{revisionblue}{\Pi_i^{\perp}}\eta_i^{t,\tau}
\right\rangle + \left\langle
\widetilde w_i^t-\widetilde w_i^{t-\tau},
\textcolor{revisionblue}{\Pi_i^{\perp}}\eta_i^t
\right\rangle + \left\langle
\widetilde w_i^{t-\tau},
\textcolor{revisionblue}{\Pi_i^{\perp}}\eta_i^t-\textcolor{revisionblue}{\Pi_i^{\perp}}\eta_i^{t,\tau}
\right\rangle .
\label{eq:decomp}
\end{equation}
The first term of the decomposition can be bounded using Equations~\eqref{eq:conseq_of_mixing} and~\eqref{eq:R_bound}, and the fact that $\tau(\beta)> 1$ as follows: 
\begin{equation}
   2\beta
\left|
\mathbb E
\left[
\left\langle
\widetilde w_i^{t-\tau},
\textcolor{revisionblue}{\Pi_i^{\perp}}\eta_i^{t,\tau}
\right\rangle
\right]
\right|
\le
8L_i\beta^2\tau(\beta)\mathbb E[(R_i^t)^2]. 
\label{eq:eq_61}
\end{equation}
Using Equation~\eqref{eq:bound_w_tau} and the observation that $\|\textcolor{revisionblue}{\Pi_i^{\perp}}\eta_i^t\|_2\le \sqrt{L_i}R_i^t,$, the second term is bounded in the following manner,
\begin{equation}
    2\beta
\left|
\mathbb E
\left[
\left\langle
\widetilde w_i^t-\widetilde w_i^{t-\tau},
\textcolor{revisionblue}{\Pi_i^{\perp}}\eta_i^t
\right\rangle
\right]
\right|
\le
16L_i\beta^2\tau(\beta)\mathbb E[(R_i^t)^2].
\label{eq:term2}
\end{equation}
Since $\textcolor{revisionblue}{\Pi_i^{\perp}}\eta_i^t$ is Lipschitz in
$(\widetilde\mu_i^t,\widetilde w_i^t)$ with Lipschitz constant at most $\sqrt{L_i}$, it follows that $\|\textcolor{revisionblue}{\Pi_i^{\perp}}\eta_i^t-\textcolor{revisionblue}{\Pi_i^{\perp}}\eta_i^{t,\tau}\|_2
\le
2L_i\beta\tau(\beta)R_i^t.$ Using $R_i^{t-\tau}\le 2R_i^t$ the third term can thus be bounded as,
\begin{equation}
    2\beta
\left|
\mathbb E
\left[
\left\langle
\widetilde w_i^{t-\tau},
\textcolor{revisionblue}{\Pi_i^{\perp}}\eta_i^t-\textcolor{revisionblue}{\Pi_i^{\perp}}\eta_i^{t,\tau}
\right\rangle
\right]
\right|
\le
32L_i\beta^2\tau(\beta)\mathbb E[(R_i^t)^2].
\label{eq:term3}
\end{equation}
Combining Equations~\eqref{eq:eq_61},\eqref{eq:term2} and ~\eqref{eq:term3}, we obtain, 
\begin{equation}
    \left|
2\beta\mathbb E
\left[
\left\langle
\widetilde w_i^t,
\textcolor{revisionblue}{\Pi_i^{\perp}}\eta_i^t
\right\rangle
\right]
\right|
\le
56L_i\beta^2\tau(\beta)\mathbb E[(R_i^t)^2].
\label{eq:semi_final_bd}
\end{equation}
Recall that $\mathbb E[(\widetilde\mu_i^t)^2]
\le
\beta C_{\mu,i}
\le
C_{\mu,i}.$ Hence, we obtain
\begin{equation}
    \left|
2\beta\mathbb E
\left[
\left\langle
\widetilde w_i^t,
\textcolor{revisionblue}{\Pi_i^{\perp}}\eta_i^t
\right\rangle
\right]
\right|
\le
56L_i\beta^2\tau(\beta)\mathbb E[V_i^t]
+
56L_i\beta^2\tau(\beta)
\left(
B_i^2+C_{\mu,i}
\right).
\end{equation}
When $\beta\tau(\beta)
\le
\frac{\Delta_i^{\chi_i}}{228L_i},$, we obtain that $56L_i\beta^2\tau(\beta)\mathbb E[V_i^t]
\le
\frac{\Delta_i^{\chi_i}\beta}{4}\mathbb E[V_i^t].$
Thus we obtain
\begin{equation}
    \left|
2\beta\mathbb E
\left[
\left\langle
\widetilde w_i^t,
\textcolor{revisionblue}{\Pi_i^{\perp}}\eta_i^t
\right\rangle
\right]
\right|
\le
\frac{\Delta_i^{\chi_i}\beta}{4}\mathbb E[V_i^t]
+
56L_i\beta^2\tau(\beta)
\left(
B_i^2+C_{\mu,i}
\right),
\label{eq:final_bound_cross_term}
\end{equation}
Since $\|\textcolor{revisionblue}{\Pi_i^{\perp}}\|_2\le 1$ and $|\delta_i^t|^2\|\textcolor{revisionblue}{\psi_{i,t}}\|_2^2
\le
32\bar\psi^2
\left(
r_{\max}^2+
\bar\psi^2\|w_i^\star\|_2^2
\right)
+
4\bar\psi^2(\widetilde\mu_i^t)^2
+
8\bar\psi^4 V_i^t .$ Therefore,using \eqref{eq:mu_post_burnin_bound} and $c_{ss}\beta\le1$,
\begin{equation}
    \beta^2\mathbb E\left[
\|\textcolor{revisionblue}{\Pi_i^{\perp}}(\delta_i^t\textcolor{revisionblue}{\psi_{i,t}})\|_2^2
\right]
\le
8\bar\psi^4\beta^2\mathbb E[V_i^t]
+
\beta^2 C_{\mathrm{sq},i},
\end{equation}
where
\[
{\color{black}
C_{\mathrm{sq},i}
:=
32\bar\psi^2
\left(
r_{\max}^2+
\bar\psi^2\|w_i^\star\|_2^2
\right)
+
4\bar\psi^2C_{\mu,i}.
}
\]
If $\beta
\le
\frac{\Delta_i^{\chi_i}}{32\bar\psi^4},$ then, 
\begin{equation}
    \beta^2\mathbb E\left[
\|\textcolor{revisionblue}{\Pi_i^{\perp}}(\delta_i^t\textcolor{revisionblue}{\psi_{i,t}})\|_2^2
\right]
\le
\frac{\Delta_i^{\chi_i}\beta}{4}\mathbb E[V_i^t]
+
\beta^2 C_{\mathrm{sq},i}.
\label{eq:sq_term}
\end{equation}
Combining Equations~\eqref{eq:sq_term}, \eqref{eq:final_bound_cross_term} and \eqref{eq:final_bound_ss} yields the final result.
\end{proof}

\begin{theorem}[Finite Time TD Learning Bounds for Local and Global Value Functions]
\label{thm:fixed-policy-td-learning-bounds}
Fix a policy $\pi_\theta$. Let $D_{\pi_\theta}$ be the stationary distribution over
global state-action pairs $(s_g,a_g)$, and let $d_{\pi_\theta}$ be the stationary
distribution over global states. Define $\Lambda_i\coloneqq
\lambda_{\max}(\chi_i^\top D_{\pi_\theta}\chi_i)$,  $C_i^{\mathrm{tr}}:=\sqrt{1+ \frac{ \lambda_{\max}^2(\phi_i^\top D_{\pi_\theta}\phi_i)
}{(\Delta_i^{\phi_i})^2}}\sqrt{
\lambda_{\max}(D_{\pi_\theta})|S_g||A_g|}$,  $\varepsilon_{\mathrm{FA},i}
:=
\|\phi_i\widehat Q_i^\star-\chi_iw_i^\star\|_{D_{\pi_\theta}}$ and $\zeta_{1,i}
:=
\mathbb E\left[
\|\textcolor{revisionblue}{\Pi_i^{\perp}}(w_i^{t_{0,i}}-w_i^\star)\|_2^2
\right].$
For $t\ge t_{0,i}:=T_{\mu,i}(\beta)+\tau(\beta)$, define the local TD error as $\mathcal E_{\mathrm{TD},i}(t,\beta)
:=
\sqrt{\Lambda_i}
\left[
\zeta_{1,i}
\left(
1-\frac{\Delta_i^{\chi_i}}{2}\beta
\right)^{t-t_{0,i}}
+
\frac{2\beta}{\Delta_i^{\chi_i}}
C_{\mathrm{bias},i}
\right]^{1/2}$, where $C_{\mathrm{bias},i}:= \frac{2\|\bar\psi_i\|_2^2}{\Delta_i^{\chi_i}}C_{\mu,i}+ 32\bar\psi^2
\left(
r_{\max}^2+\bar\psi^2\|w_i^\star\|_2^2
\right)
+
4\bar\psi^2C_{\mu,i}+ 56L_i\tau(\beta)(B_i^2+C_{\mu,i}).$ Under Assumption~\ref{assump:ergodicity}, the following hold
\begin{itemize}
    \item Single agent TD Learning bound: For every agent $i$ and every $t\ge t_{0,i}$,
    \begin{equation}
        \begin{aligned}
\inf_{\bar c\in\mathbb R}
\mathbb E
\left[
\left\|
(Q_i^\theta+\bar c e)-\chi_i\textcolor{revisionblue}{\Pi_i^{\perp}}w_i^t
\right\|_{D_{\pi_\theta}}
\right]
\le{}&
\underbrace{C_i^{\mathrm{tr}}c\rho^{k+1}}_{\textnormal{truncation error}}
+
\underbrace{\varepsilon_{\mathrm{FA},i}}_{\textnormal{function approximation error}}
\\
&\quad
+
\underbrace{\mathcal E_{\mathrm{TD},i}(t,\beta)}_{\textnormal{TD learning error}}.
\end{aligned}
\label{eq:local_td_learning_bound}
    \end{equation}
    \item Global value-function TD learning bound: Define the learned global
action-value approximation
\(
\widehat Q_{\mathrm{glob}}^t
:=
\sum_{i=1}^n \chi_i\textcolor{revisionblue}{\Pi_i^{\perp}}w_i^t,
\) and the corresponding learned global state-value approximation
\(
\widehat V_{\mathrm{glob}}^t(s_g)
:=
\sum_{a_g\in A_g}
\pi_\theta(a_g\mid s_g)
\widehat Q_{\mathrm{glob}}^t(s_g,a_g).
\)
Let
\(
t_0:=\max_{1\le i\le n}t_{0,i}.
\)
For every $t\ge t_0$,
\begin{equation}
    \inf_{\bar c\in\mathbb R}
\mathbb E
\left[
\left\|
(V^\theta+\bar c e_S)-\widehat V_{\mathrm{glob}}^t
\right\|_{d_{\pi_\theta}}
\right]
\le \underbrace{
\sum_{i=1}^n C_i^{\mathrm{tr}}c\rho^{k+1}
}_{\textnormal{global truncation error}} +
\underbrace{
\sum_{i=1}^n \varepsilon_{\mathrm{FA},i}
}_{\textnormal{global FA error}}+
\underbrace{
\sum_{i=1}^n \mathcal E_{\mathrm{TD},i}(t,\beta)
}_{\textnormal{global TD learning error}}.
\label{eq:global_value_td_learning_bound}
\end{equation}
\end{itemize}
\end{theorem}
\begin{proof}
    We first prove the single-agent bound. Fix an agent $i$. Since
$w_i^\star\in S_{\Psi_i,e}^{\perp}$, we have
\(
\textcolor{revisionblue}{\Pi_i^{\perp}}w_i^\star=w_i^\star.
\)
For any scalar $\bar c\in\mathbb R$,
\[
(Q_i^\theta+\bar c e)-\chi_i\textcolor{revisionblue}{\Pi_i^{\perp}}w_i^t
=
\bigl((Q_i^\theta+\bar c e)-\phi_i\widehat Q_i^\star\bigr)
+
\bigl(\phi_i\widehat Q_i^\star-\chi_iw_i^\star\bigr)
+
\chi_i(\textcolor{revisionblue}{\Pi_i^{\perp}}w_i^\star-\textcolor{revisionblue}{\Pi_i^{\perp}}w_i^t).
\]
Taking the $D_{\pi_\theta}$-norm, expectation, and then the infimum over $\bar c$ gives
\begin{align}
\nonumber
    \inf_{\bar c\in\mathbb R}
\mathbb E
\left[
\left\|
(Q_i^\theta+\bar c e)-\chi_i\textcolor{revisionblue}{\Pi_i^{\perp}}w_i^t
\right\|_{D_{\pi_\theta}}
\right]
\le & \inf_{\bar c\in\mathbb R}
\left\|
(Q_i^\theta+\bar c e)-\phi_i\widehat Q_i^\star
\right\|_{D_{\pi_\theta}}+
\left\|
\phi_i\widehat Q_i^\star-\chi_iw_i^\star
\right\|_{D_{\pi_\theta}}\\ &+
\mathbb E
\left[
\left\|
\chi_i(\textcolor{revisionblue}{\Pi_i^{\perp}}w_i^\star-\textcolor{revisionblue}{\Pi_i^{\perp}}w_i^t)
\right\|_{D_{\pi_\theta}}
\right].
\label{eq:thm4_Eq1}
\end{align}
By the local truncation bound,
\begin{equation}
    \inf_{\bar c\in\mathbb R}
\left\|
(Q_i^\theta+\bar c e)-\phi_i\widehat Q_i^\star
\right\|_{D_{\pi_\theta}}
\le
C_i^{\mathrm{tr}}c\rho^{k+1}.
\label{eq:thm4_eq2}
\end{equation}
For the third term, since $\chi_i=\phi_i\Psi_i$, 
\begin{equation}
    \left\|
\chi_i(\textcolor{revisionblue}{\Pi_i^{\perp}}w_i^\star-\textcolor{revisionblue}{\Pi_i^{\perp}}w_i^t)
\right\|_{D_{\pi_\theta}}^2
=
(\textcolor{revisionblue}{\Pi_i^{\perp}}(w_i^\star-w_i^t))^\top
\chi_i^\top D_{\pi_\theta}\chi_i
\textcolor{revisionblue}{\Pi_i^{\perp}}(w_i^\star-w_i^t)
 \le
\Lambda_i
\|\textcolor{revisionblue}{\Pi_i^{\perp}}(w_i^t-w_i^\star)\|_2^2.
\label{eq:thm4_eq3}
\end{equation}
Therefore, by Jensen's inequality,
\begin{equation}
    \mathbb E
    \left[
    \left\|
    \chi_i(\textcolor{revisionblue}{\Pi_i^{\perp}}w_i^\star-\textcolor{revisionblue}{\Pi_i^{\perp}}w_i^t)
    \right\|_{D_{\pi_\theta}}
    \right]
    \le
    \sqrt{
    \Lambda_i
    \mathbb E
    \left[
    \|\textcolor{revisionblue}{\Pi_i^{\perp}}(w_i^t-w_i^\star)\|_2^2
    \right]}
    \label{eq:thm4_eq4}
\end{equation}
Iterating Lemma~\ref{lem:fixed_actor_ss} from $t_{0,i}$ to $t$ gives
\begin{equation}
\mathbb E
\left[
\|\textcolor{revisionblue}{\Pi_i^{\perp}}(w_i^t-w_i^\star)\|_2^2
\right]
\le
\zeta_{1,i}
\left(
1-\frac{\Delta_i^{\chi_i}}{2}\beta
\right)^{t-t_{0,i}}
+
\frac{2\beta}{\Delta_i^{\chi_i}}C_{\mathrm{bias},i}.
\label{eq:thm4_eq5}
\end{equation}
Combining Equations~\eqref{eq:thm4_Eq1}-\eqref{eq:thm4_eq5} proves \eqref{eq:local_td_learning_bound}.

We next prove the global value-function bound. For arbitrary constants $\bar c_1,\ldots,\bar c_n$, let $\bar c:=\sum_{i=1}^n\bar c_i.$ Then $$(Q^\theta+\bar c e)-\widehat Q_{\mathrm{glob}}^t
=
\sum_{i=1}^n
\left[
(Q_i^\theta+\bar c_i e)-\chi_i\textcolor{revisionblue}{\Pi_i^{\perp}}w_i^t
\right].$$
By the triangle inequality,
\[
\left\|
(Q^\theta+\bar c e)-\widehat Q_{\mathrm{glob}}^t
\right\|_{D_{\pi_\theta}}
\le
\sum_{i=1}^n
\left\|
(Q_i^\theta+\bar c_i e)-\chi_i\textcolor{revisionblue}{\Pi_i^{\perp}}w_i^t
\right\|_{D_{\pi_\theta}}.
\]
Taking expectation and optimizing over each $\bar c_i$ gives
\begin{equation}
\inf_{\bar c\in\mathbb R}
\mathbb E
\left[
\left\|
(Q^\theta+\bar c e)-\widehat Q_{\mathrm{glob}}^t
\right\|_{D_{\pi_\theta}}
\right]
\le
\sum_{i=1}^n
\inf_{\bar c_i\in\mathbb R}
\mathbb E
\left[
\left\|
(Q_i^\theta+\bar c_i e)-\chi_i\textcolor{revisionblue}{\Pi_i^{\perp}}w_i^t
\right\|_{D_{\pi_\theta}}
\right].
\label{eq:thm4_eq6}   
\end{equation}
Applying the single-agent bound to each term yields
\begin{equation}
    \inf_{\bar c\in\mathbb R}
\mathbb E
\left[
\left\|
(Q^\theta+\bar c e)-\widehat Q_{\mathrm{glob}}^t
\right\|_{D_{\pi_\theta}}
\right]
\le
\sum_{i=1}^n
\left[
C_i^{\mathrm{tr}}c\rho^{k+1}
+
\varepsilon_{\mathrm{FA},i}
+
\mathcal E_{\mathrm{TD},i}(t,\beta)
\right].
\label{eq:thm4_eq7}
\end{equation}
Finally, relate the global state-value error to the global action-value error. For any global
state-action error
\[
F(s_g,a_g)
:=
(Q^\theta(s_g,a_g)+\bar c)-\widehat Q_{\mathrm{glob}}^t(s_g,a_g),
\]
we have
\[
(V^\theta(s_g)+\bar c)-\widehat V_{\mathrm{glob}}^t(s_g)
=
\sum_{a_g\in A_g}\pi_\theta(a_g\mid s_g)F(s_g,a_g).
\]
By Jensen's inequality,
\[
\left|
(V^\theta(s_g)+\bar c)-\widehat V_{\mathrm{glob}}^t(s_g)
\right|^2
\le
\sum_{a_g\in A_g}
\pi_\theta(a_g\mid s_g)F(s_g,a_g)^2.
\]
Multiplying by $d_{\pi_\theta}(s_g)$ and summing over $s_g$ gives
\[
\left\|
(V^\theta+\bar c e_S)-\widehat V_{\mathrm{glob}}^t
\right\|_{d_{\pi_\theta}}
\le
\left\|
(Q^\theta+\bar c e)-\widehat Q_{\mathrm{glob}}^t
\right\|_{D_{\pi_\theta}}.
\]
Taking expectation, the infimum over $\bar c$, and then using Equation~\ref{eq:thm4_eq7} proves
\eqref{eq:global_value_td_learning_bound}.
\end{proof}

The fixed-policy Lyapunov drift in Lemma~\ref{lem:fixed_actor_ss} is adapted from the average-reward TD analysis of~\cite{zhang2021finite}, specialized to $\lambda=0$ and applied to the localized composite feature matrix $\chi_i=\phi_i \psi_i$. The contribution lies in the fixed point characterization of the local networked structure and in explicitly distinguishing between localization errors and feature-approximation errors.

\subsubsection{Moving-target critic stability.}
\label{app:moving_target_critic}
\begin{lemma}[Lipschitzness with respect to softmax parameter $\theta$]
\label{lem:lipschitzness}
Let $\mathcal X := \mathcal S_g \times \mathcal A_g$. Define $m\coloneqq |\mathcal X|$. Let \(\textcolor{revisionblue}{P_{\pi_\theta}}\) be the frozen state-action Markov kernel $\textcolor{revisionblue}{P_{\pi_\theta}}\bigl((s_g,a_g),(s_g^+,a_g^+)\bigr)
    :=
    P(s_g^+\mid s_g,a_g)\pi_\theta(a_g^+\mid s_g^+).$ Let \(\nu^\theta\)
denote its unique stationary distribution on
\(\mathcal S_g \times \mathcal A_g\).  For
\(\theta\in\Theta_{\mathrm{ac}}\), define
\begin{equation}
    \Delta_i^{\chi_i}(\theta)
    :=
    \inf_{\substack{\|v\|_2=1\\ v\in S_{\Psi_i,e}^{\perp}}}
    v^\top
    \chi_i^\top D_{\pi_\theta}(I-\textcolor{revisionblue}{P_{\pi_\theta}})\chi_i
    v,
    \qquad
    \underline\Delta_i
    :=
    \inf_{\theta\in\Theta_{\mathrm{ac}}}
    \Delta_i^{\chi_i}(\theta).
    \label{eq:uniform-restricted-critic-gap-def}
\end{equation}
Then under Assumptions~\ref{assump:ergodicity} and~\ref{assump:uniform_mixing},
the following hold:
    \begin{enumerate}
        \item The kernel is globally Lipschitz in the actor parameter: 
        \begin{equation}
            \sup_{x\in\mathcal X}
        d_{\rm TV}\bigl(\textcolor{revisionblue}{P_{\pi_\theta}}(x,\cdot),\textcolor{revisionblue}{P_{\pi_{\theta'}}}(x,\cdot)\bigr)
        \le
        L_K\|\theta-\theta'\|_2,
        \qquad
        L_K:=\sqrt{2n}.
        \end{equation}
        \item The stationary state-action distribution is globally Lipschitz:
        \begin{equation}
            d_{\rm TV}(\nu^\theta,\nu^{\theta'})
        \le
        L_\nu^{\mathcal X}\|\theta-\theta'\|_2,
        \qquad
        L_\nu^{\mathcal X}:=
    \frac{2m_{\rm mix}L_K}{1-\rho_{\rm mix}}.
        \end{equation}
        \item The stationary one-step law is globally Lipschitz:
        \begin{equation}
            d_{\rm TV}(\bar\nu^\theta,\bar\nu^{\theta'})
        \le
        L_\nu\|\theta-\theta'\|_2,
        \qquad
        L_\nu:=L_\nu^{\mathcal X}+L_K.
        \end{equation}
        \item Each local average reward is globally Lipschitz:
        \begin{equation}
            |J_i(\theta)-J_i(\theta')|
        \le
        L_{J,i}\|\theta-\theta'\|_2,
        \qquad
        L_{J,i}:=2r_{\max,i}L_\nu^{\mathcal X}.
        \end{equation}
        \item The unique restricted critic target \(w_i^\star(\theta)\in W_i^\star\) satisfies
\[
    \|w_i^\star(\theta)\|_2
    \le
    W_{\star,i}, \qquad W_{\star,i}
    :=
    \frac{2r_{\max,i}\bar\psi}{\underline\Delta_i},
\]
and is Lipschitz
    \[\|w_i^\star(\theta)-w_i^\star(\theta')\|_2
    \le
    L_{w,i}\|\theta-\theta'\|_2, \qquad L_{w,i}
    :=
    \frac{L_{b,i}+L_{A,i}W_{\star,i}}{\underline\Delta_i}.\]
    Consequently,
    \[
        W_\star:=\max_{1\le i\le n}W_{\star,i}
        <\infty,
        \qquad
        \sup_{\theta\in\Theta_{\mathrm{ac}}}
        \|w_i^\star(\theta)\|_2\le W_\star .
    \]
    \end{enumerate}  
\end{lemma}
\begin{proof}

    For \(x=(s_g,a_g)\), define the global softmax score $S_\theta(x)
    :=
    \nabla_\theta \log \pi_\theta(a_g\mid s_g).$ By the product form of the policy, $S_\theta(s_g,a_g)
    =
    \bigl(
        \nabla_{\theta_1}\log \pi_{\theta_1}(a_1\mid s_1),
        \ldots,
        \nabla_{\theta_n}\log \pi_{\theta_n}(a_n\mid s_n)
    \bigr).$
    For one local softmax block, 
    \begin{equation*}
        \frac{\partial}{\partial \theta_i(s_i,b_i)}
    \log \pi_{\theta_i}(a_i\mid s_i)
    =
    \mathbf 1\{a_i=b_i\}-\pi_{\theta_i}(b_i\mid s_i).
    \end{equation*}
    Hence $\bigl\|\nabla_{\theta_i}\log \pi_{\theta_i}(a_i\mid s_i)\bigr\|_2
    \le \sqrt 2,$ and therefore 
    \begin{equation}
        \|S_\theta(s_g,a_g)\|_2\le \sqrt{2n}=:G.
        \label{eq:grad_bound_early}
    \end{equation} 

    We first prove the kernel Lipschitz bound. For any fixed \(s_g^+\), the mean-value theorem and
the score bound give
    \begin{align}
    \sum_{a_g^+\in\mathcal A_g}
    \left|
        \pi_\theta(a_g^+\mid s_g^+)
        -
        \pi_{\theta'}(a_g^+\mid s_g^+)
    \right|
    &\le
    \sup_{\vartheta}
    \sum_{a_g^+\in\mathcal A_g}
    \bigl\|
        \nabla_\vartheta \pi_\vartheta(a_g^+\mid s_g^+)
    \bigr\|_2
    \|\theta-\theta'\|_2                                      \\
    &=
    \sup_{\vartheta}
    \sum_{a_g^+\in\mathcal A_g}
    \pi_\vartheta(a_g^+\mid s_g^+)
    \bigl\|
        \nabla_\vartheta \log \pi_\vartheta(a_g^+\mid s_g^+)
    \bigr\|_2
    \|\theta-\theta'\|_2                                      \\
    &\le
    G\|\theta-\theta'\|_2 .
\end{align}
Since the environment transition \(P(s_g^+\mid s_g,a_g)\) does not depend on \(\theta\), 
\begin{align}
    d_{\rm TV}\bigl(\textcolor{revisionblue}{P_{\pi_\theta}}(x,\cdot),\textcolor{revisionblue}{P_{\pi_{\theta'}}}(x,\cdot)\bigr)
    &\le
    \sum_{s_g^+}
    P(s_g^+\mid s_g,a_g)
    d_{\rm TV}\bigl(
        \pi_\theta(\cdot\mid s_g^+),
        \pi_{\theta'}(\cdot\mid s_g^+)
    \bigr)                                                     \\
    &\le
    G\|\theta-\theta'\|_2 .
\end{align}
Thus \(L_K=G=\sqrt{2n}\). 

We next prove Lipschitzness of the stationary distribution. Define $d_{\rm TV}(p,q) = \frac12\sum_{x\in\mathcal X}|p(x)-q(x)|.$ For signed measures \(\mu\) with total mass zero, define $\|\mu\|_{\rm TV}\coloneqq \frac12\sum_{x\in\mathcal X}|\mu(x)|.$ Fix \(\theta,\theta'\). For brevity, write $K:=\textcolor{revisionblue}{P_{\pi_\theta}}, K':=\textcolor{revisionblue}{P_{\pi_{\theta'}}}, \nu:=\nu^\theta,$ and $ \nu':=\nu^{\theta'}.$  Define $\Delta K:=K'-K,$ and $\delta:=\nu'\Delta K.$ From its definition, it follows that \(\delta\) has total mass zero. Moreover,
\begin{equation*}
    \|\delta\|_{\rm TV} = d_{\rm TV}(\nu'K',\nu'K) \le
    \sum_{x\in\mathcal X}
    \nu'(x)
    d_{\rm TV}\bigl(K'(x,\cdot),K(x,\cdot)\bigr) \le
    L_K\|\theta-\theta'\|_2 .
\end{equation*}
Since \(\nu'\) is stationary for \(K'\), we have \(\nu'K'=\nu'\). Therefore, for every integer
\(T\ge 1\),
\begin{equation}
    \nu'-\nu'K^T = 
    \sum_{\ell=0}^{T-1}
    \left(
        \nu'K^\ell-\nu'K^{\ell+1}
    \right) =
    \sum_{\ell=0}^{T-1}
    \nu'(I-K)K^\ell  =
    \sum_{\ell=0}^{T-1}
    \nu'(K'-K)K^\ell   =
    \sum_{\ell=0}^{T-1}
    \delta K^\ell .
\end{equation}
By the uniform mixing assumption,
\begin{equation}
    d_{\rm TV}(\nu'K^T,\nu)
    \le
    \sum_{x\in\mathcal X}
    \nu'(x)d_{\rm TV}\bigl(K^T(x,\cdot),\nu\bigr)
    \le
    m_{\rm mix}\rho_{\rm mix}^T
    \longrightarrow 0.
\end{equation}
Thus, letting \(T\to\infty\), 
\begin{equation}
    \nu'-\nu
    =
    \sum_{\ell=0}^{\infty}\delta K^\ell .
\end{equation}
It remains to bound each term \(\delta K^\ell\). For any set \(A\subseteq\mathcal X\), since
\(\delta(\mathcal X)=0\),
\begin{equation*}
    \delta K^\ell(A) =
    \sum_{x\in\mathcal X}\delta(x)K^\ell(x,A) =
    \sum_{x\in\mathcal X}
    \delta(x)
    \left[
        K^\ell(x,A)-\nu(A)
    \right].
\end{equation*}
Therefore,
\begin{align*}
    |\delta K^\ell(A)|
    &\le
    \sum_{x\in\mathcal X}
    |\delta(x)|
    \left|
        K^\ell(x,A)-\nu(A)
    \right|                                                               \\
    &\le
    \sum_{x\in\mathcal X}
    |\delta(x)|
    d_{\rm TV}\bigl(K^\ell(x,\cdot),\nu\bigr)                              \\
    &\le
    m_{\rm mix}\rho_{\rm mix}^{\ell}
    \sum_{x\in\mathcal X}
    |\delta(x)|                                                           \\
    &=
    2m_{\rm mix}\rho_{\rm mix}^{\ell}\|\delta\|_{\rm TV}.
\end{align*}
Taking the supremum over \(A\subseteq\mathcal X\) gives $\|\delta K^\ell\|_{\rm TV}
    \le
    2m_{\rm mix}\rho_{\rm mix}^{\ell}\|\delta\|_{\rm TV}.$
Hence,
\begin{align*}
    d_{\rm TV}(\nu^\theta,\nu^{\theta'})
    =
    d_{\rm TV}(\nu,\nu') &\le
    \sum_{\ell=0}^{\infty}
    \|\delta K^\ell\|_{\rm TV} \le
    2m_{\rm mix}\|\delta\|_{\rm TV}
    \sum_{\ell=0}^{\infty}
    \rho_{\rm mix}^{\ell}   \\ & 
    =
    \frac{2m_{\rm mix}}{1-\rho_{\rm mix}}
    \|\delta\|_{\rm TV} \le
    \frac{2m_{\rm mix}L_K}{1-\rho_{\rm mix}}
    \|\theta-\theta'\|_2.
\end{align*}
Finally, for the stationary one-step law, $\bar\nu^\theta=\nu^\theta \textcolor{revisionblue}{P_{\pi_\theta}},$ and $\bar\nu^{\theta'}=\nu^{\theta'}\textcolor{revisionblue}{P_{\pi_{\theta'}}},$ using the triangle inequality and contraction of total variation under a Markov kernel,
\begin{align*}
    d_{\rm TV}\bigl(\bar\nu^\theta,\bar\nu^{\theta'}\bigr)
    &=
    d_{\rm TV}\bigl(\nu^\theta \textcolor{revisionblue}{P_{\pi_\theta}},\nu^{\theta'}\textcolor{revisionblue}{P_{\pi_{\theta'}}}\bigr)       \\
    &\le
    d_{\rm TV}\bigl(\nu^\theta \textcolor{revisionblue}{P_{\pi_\theta}},\nu^{\theta'}\textcolor{revisionblue}{P_{\pi_\theta}}\bigr)
    +
    d_{\rm TV}\bigl(\nu^{\theta'}\textcolor{revisionblue}{P_{\pi_\theta}},\nu^{\theta'}\textcolor{revisionblue}{P_{\pi_{\theta'}}}\bigr)    \\
    &\le
    d_{\rm TV}\bigl(\nu^\theta,\nu^{\theta'}\bigr)
    +
    \sup_{x\in\mathcal X}
    d_{\rm TV}\bigl(\textcolor{revisionblue}{P_{\pi_\theta}}(x,\cdot),\textcolor{revisionblue}{P_{\pi_{\theta'}}}(x,\cdot)\bigr)            \\
    &\le
    \left(
        L_{\nu}^{\mathcal X}+L_K
    \right)
    \|\theta-\theta'\|_2.
\end{align*}

Next, we prove Lipschitzness of $J_i(\theta)$. Recall $J_i(\theta)
    =
    \sum_{(s_g,a_g)\in\mathcal X}
    \nu^\theta(s_g,a_g)r_i(s_i,a_i).$ 
    Since \(|r_i(s_i,a_i)|\le r_{\max,i}\),
    \begin{align}
        |J_i(\theta)-J_i(\theta')|
    &\le
    \sum_{x\in\mathcal X}
    |\nu^\theta(x)-\nu^{\theta'}(x)|\,|r_i(x)|       \\
    &\le
    2r_{\max,i}
    d_{\rm TV}(\nu^\theta,\nu^{\theta'})             \\
    &\le
    2r_{\max,i}L_\nu^{\mathcal X}\|\theta-\theta'\|_2.
    \end{align}
We next prove boundedness of $\|w_i^\star(\theta)\|.$ By Lemma~\ref{lem:solution_Existence}, \(w_i^\star(\theta)\) is the unique fixed point in the restricted subspace
\(W_i^\star\). Recall that $\textcolor{revisionblue}{\Pi_i^{\perp}}$ is the euclidean projection onto \(S_{\psi_i,e}^\perp\). Equivalently, it solves the restricted linear system $\textcolor{revisionblue}{\Pi_i^{\perp}}A_i(\theta)\textcolor{revisionblue}{\Pi_i^{\perp}}w_i^\star(\theta)=\textcolor{revisionblue}{\Pi_i^{\perp}}b_i(\theta).$ Since \(w_i^\star(\theta)\in W_i^\star\), we have \(\textcolor{revisionblue}{\Pi_i^{\perp}}w_i^\star(\theta)=w_i^\star(\theta)\). Define $B_i(\theta):=\textcolor{revisionblue}{\Pi_i^{\perp}}A_i(\theta)\textcolor{revisionblue}{\Pi_i^{\perp}}.$ Then
\begin{equation*}
    B_i(\theta)w_i^\star(\theta)=\textcolor{revisionblue}{\Pi_i^{\perp}}b_i(\theta).
\end{equation*}
We first show that \(B_i(\theta)\) is uniformly invertible on \(W_i^\star\). For any \(w\in W_i^\star\),
\begin{equation}
    w^\top B_i(\theta)w
    =
    w^\top \textcolor{revisionblue}{\Pi_i^{\perp}}A_i(\theta)\textcolor{revisionblue}{\Pi_i^{\perp}}w
    =
    w^\top A_i(\theta)w
    \ge
    \underline\Delta_i\|w\|_2^2.
\end{equation}
By Cauchy's inequality, $w^\top B_i(\theta)w
    \le
    \|w\|_2\|B_i(\theta)w\|_2.$
    Therefore, 
    \begin{equation}
        \|B_i(\theta)w\|_2
    \ge
    \underline\Delta_i\|w\|_2,
    \qquad
    w\in W_i^\star.
    \end{equation}
Hence \(B_i(\theta)\) is invertible on \(W_i^\star\), and $\|B_i(\theta)^{-1}\|_{\rm op}
    \le
    \frac{1}{\underline\Delta_i}.$
    Since \(|r_i(X)|\le r_{\max,i}\), we also have $|J_i(\theta)|\le r_{\max,i}.$ Thus,
    \begin{align*}
         \|b_i(\theta)\|_2
    &=
    \left\|
        \mathbb E_{\nu^\theta}
        \left[
            \psi_i(X)\bigl(r_i(X)-J_i(\theta)\bigr)
        \right]
    \right\|_2                                      \\
    &\le
    \mathbb E_{\nu^\theta}
    \left[
        \|\psi_i(X)\|_2
        |r_i(X)-J_i(\theta)|
    \right]                                         \\
    &\le
    2r_{\max,i}\bar\psi.
    \end{align*}
    Using \(w_i^\star(\theta)=B_i(\theta)^{-1}\textcolor{revisionblue}{\Pi_i^{\perp}}b_i(\theta)\) and \(\|\textcolor{revisionblue}{\Pi_i^{\perp}}\|_{\rm op}\le1\),
\[
\begin{aligned}
    \|w_i^\star(\theta)\|_2
    &\le
    \|B_i(\theta)^{-1}\|_{\rm op}\|\textcolor{revisionblue}{\Pi_i^{\perp}}b_i(\theta)\|_2        \\
    &\le
    \frac{1}{\underline\Delta_i}\|b_i(\theta)\|_2            \\
    &\le
    \frac{2r_{\max,i}\bar\psi}{\underline\Delta_i}.
\end{aligned}
\]
Therefore $M_{w,i}:=\frac{2r_{\max,i}\bar\psi}{\underline\Delta_i}$.

It remains to prove Lipschitzness. Recall that $A_i(\theta)
    :=
    \mathbb E_{\bar\nu^\theta}
    \left[
        \psi_i(X)
        \bigl(\psi_i(X)-\psi_i(X^+)\bigr)^\top
    \right]
    =
    \chi_i^\top D_{\pi_\theta}(I-P_{\pi_\theta})\chi_i,$. First, we bound \(A_i(\theta)-A_i(\theta')\). Let 
    \begin{equation*}
        F_i(x,x^+)
    :=
    \psi_i(x)\bigl(\psi_i(x)-\psi_i(x^+)\bigr)^\top.
    \end{equation*}
    Then $A_i(\theta)
    =
    \mathbb E_{\bar\nu^\theta}[F_i(X,X^+)].$ Since
    \begin{equation*}
        \|F_i(x,x^+)\|_{\rm op}
    \le
    \|\psi_i(x)\|_2
    \|\psi_i(x)-\psi_i(x^+)\|_2
    \le
    2\bar\psi^2,
    \end{equation*}
    the total-variation bound gives
    \begin{align*}
        \|A_i(\theta)-A_i(\theta')\|_{\rm op}
    &\le
    2\sup_{x,x^+}\|F_i(x,x^+)\|_{\rm op}
    d_{\rm TV}(\bar\nu^\theta,\bar\nu^{\theta'})        \\
    &\le
    4\bar\psi_i^2L_{\nu}\|\theta-\theta'\|_2.
    \end{align*}
    Thus $\|A_i(\theta)-A_i(\theta')\|_{\rm op}
    \le
    L_{A,i}\|\theta-\theta'\|_2,
    $ with $L_{A,i}:=4\bar\psi^2L_{\nu}.$ 
    Recall $ b_i(\theta)
    :=
    \mathbb E_{\nu^\theta}
    \left[
        \psi_i(X)\bigl(r_i(X)-J_i(\theta)\bigr)
    \right]
    =
    \chi_i^\top D_{\pi_\theta}\bigl(r_i-J_i(\theta)\mathbf 1\bigr).$ Next, we bound \(b_i(\theta)-b_i(\theta')\).
    \begin{equation*}
        b_i(\theta)
    =
    \mathbb E_{\nu^\theta}[\psi_i(X)r_i(X)]
    -
    J_i(\theta)\mathbb E_{\nu^\theta}[\psi_i(X)].
    \end{equation*}
    Therefore,
    \begin{align*}
        \|b_i(\theta)-b_i(\theta')\|_2
    &\le
    \left\|
        \mathbb E_{\nu^\theta}[\psi_i(X)r_i(X)]
        -
        \mathbb E_{\nu^{\theta'}}[\psi_i(X)r_i(X)]
    \right\|_2                                           \\
    &\quad
    +
    |J_i(\theta)|
    \left\|
        \mathbb E_{\nu^\theta}[\psi_i(X)]
        -
        \mathbb E_{\nu^{\theta'}}[\psi_i(X)]
    \right\|_2                                           \\
    &\quad
    +
    |J_i(\theta)-J_i(\theta')|
    \left\|
        \mathbb E_{\nu^{\theta'}}[\psi_i(X)]
    \right\|_2.
    \end{align*}
    Using \(|r_i|\le r_{\max,i}\), \(|J_i(\theta)|\le r_{\max,i}\), and
\(\|\psi_i(x)\|_2\le\bar\psi\),
\begin{align*}
    \|b_i(\theta)-b_i(\theta')\|_2
    &\le
    2r_{\max,i}\bar\psi
    d_{\rm TV}(\nu^\theta,\nu^{\theta'})
    +
    2r_{\max,i}\bar\psi
    d_{\rm TV}(\nu^\theta,\nu^{\theta'})                  
    +
    \bar\psi|J_i(\theta)-J_i(\theta')|                  \\
    &\le
    \bigl(
        4r_{\max,i}\bar\psi L_{\nu}^\chi
        +
        \bar\psi L_{J,i}
    \bigr)
    \|\theta-\theta'\|_2.
\end{align*}
Thus $\|b_i(\theta)-b_i(\theta')\|_2
    \le
    L_{b,i}\|\theta-\theta'\|_2,$ where $L_{b,i}
    :=
    4r_{\max,i}\bar\psi L_{\nu}^\chi
    +
    \bar\psi L_{J,i}.$
Finally, compare the two restricted linear systems:
\[
    B_i(\theta)w_i^\star(\theta)=\textcolor{revisionblue}{\Pi_i^{\perp}}b_i(\theta),
    \qquad
    B_i(\theta')w_i^\star(\theta')=\textcolor{revisionblue}{\Pi_i^{\perp}}b_i(\theta').
\]
\begin{equation*}
     B_i(\theta)
    \bigl(
        w_i^\star(\theta)-w_i^\star(\theta')
    \bigr)
    =
    \textcolor{revisionblue}{\Pi_i^{\perp}}\bigl(b_i(\theta)-b_i(\theta')\bigr)
    +
    \bigl(B_i(\theta')-B_i(\theta)\bigr)
    w_i^\star(\theta').
\end{equation*}
Applying \(B_i(\theta)^{-1}\) and using \(\|\textcolor{revisionblue}{\Pi_i^{\perp}}\|_{\rm op}\le1\),
\begin{align*}
    \|w_i^\star(\theta)-w_i^\star(\theta')\|_2
    &\le
    \frac{1}{\underline\Delta_i}
    \left(
        \|b_i(\theta)-b_i(\theta')\|_2
        +
        \|B_i(\theta)-B_i(\theta')\|_{\rm op}
        \|w_i^\star(\theta')\|_2
    \right)                                      \\
    &\le
    \frac{1}{\underline\Delta_i}
    \left(
        L_{b,i}
        +
        L_{A,i}W_{\star,i}
    \right)
    \|\theta-\theta'\|_2.
\end{align*}
In the last step we used
\[
    \|B_i(\theta)-B_i(\theta')\|_{\rm op}
    =
    \|\textcolor{revisionblue}{\Pi_i^{\perp}}(A_i(\theta)-A_i(\theta'))\textcolor{revisionblue}{\Pi_i^{\perp}}\|_{\rm op}
    \le
    \|A_i(\theta)-A_i(\theta')\|_{\rm op}.
\]
Therefore the Lipschitz bound holds with
\[
    L_{w,i}
    :=
    \frac{L_{b,i}+L_{A,i}W_{\star,i}}{\underline\Delta_i}.
\]
This completes the proof.
\end{proof}

\begin{lemma}[Bounded Actor Increment]
    \label{lem:bounded-actor-increment}
From Assumption~\ref{assump:bounded_critic} and from uniform bound on the gradient in Equation~\eqref{eq:grad_bound_early}, it follows that there exists a constant $C_\theta<\infty$ such that 
\begin{equation}
    \|\theta^{t+1}-\theta^t\|_2\le C_\theta\alpha_t .
\end{equation}
\end{lemma}
\begin{proof}
    By the unregularized branch of Algorithm~\ref{alg:decentralized_ac_lfa},
\[
\theta_i^{t+1}-\theta_i^t
=
\alpha_t S_i^t
\sum_{j\in N_i^k}(\textcolor{revisionblue}{\psi_{j,t}})^\top w_j^t .
\]
Using \(\|\textcolor{revisionblue}{\psi_{j,t}}\|_2\le \bar\psi\), \(\|S_i^t\|_2\le G_i\), and
\(\|w_j^t\|_2\le W\),
\[
\|\theta_i^{t+1}-\theta_i^t\|_2
\le
\alpha_t G_i\bar\psi
\sum_{j\in N_i^k}W.
\]
Therefore,
\[
\|\theta^{t+1}-\theta^t\|_2^2
=
\sum_{i=1}^n\|\theta_i^{t+1}-\theta_i^t\|_2^2
\le
\alpha_t^2\bar\psi^2
\sum_{i=1}^n
G_i^2
\left(\sum_{j\in N_i^k}W\right)^2.
\]
Taking square roots gives the claim with
\[
C_\theta
:=
\bar\psi
\left[
\sum_{i=1}^n
G_i^2
\left(\sum_{j\in N_i^k}W\right)^2
\right]^{1/2}.
\]
\end{proof}

\begin{lemma}[Moving target contraction]
Suppose the fixed-policy critic assumptions of Lemma~\ref{lem:fixed_actor_ss}
hold uniformly over the actor
parameters, and suppose Assumptions~\ref{assump:ergodicity},
\ref{assump:uniform_mixing}, and \ref{assump:bounded_critic} hold.  Use
Lemma~\ref{lem:lipschitzness} to define
the derived uniform critic contraction constants
\(\underline\Delta_i\) and \(\Delta_{\min}^{\chi}\).  Let
$\mathcal V_t:=\sum_{i=1}^n\mathcal V_i^t,$ where $\mathcal V_i^t
:=
q_i(\widetilde\mu_i^t)^2+\|\widetilde w_i^t\|_2^2.$
In addition to conditions over critic step size in Lemma~\ref{lem:fixed_actor_ss}, let the following hold: 
\begin{equation}
    C_{\mathrm{sq}}\beta
+
C_{\mathrm{mov}}\varepsilon_{P,\beta,t}^{\mathrm{mov}}
\le
\frac{\Delta_{\min}^{\chi}}{8} \qquad \max\!\left\{C_z\beta\tau(\beta),\,L_\star C_\theta A_t^\beta\right\}
\le \frac14 .
\label{eq:moving-beta-rules}
\end{equation}
Here $A_t^\beta\coloneqq \sum_{l=t-\tau(\beta)}^{t-1}\alpha_l.$ The constants \(C_{\mathrm{sq}},C_{\mathrm{mov}},C_z, C_{\mathrm{tar}}\in(0,\infty)\) are finite
constants specified in the proof; they depend only on the problem constants and
the constants in the stated assumptions, and are independent of \(t\), \(\alpha_t\),
and \(\beta\). 
Then, for every $t\ge \tau(\beta)$, 
\begin{equation}
\mathbb E[\mathcal V_{t+1}]
\le{}
\left(1-\frac{\Delta_{\min}^\chi}{4}\beta\right)
\mathbb E[\mathcal V_t]
+
C_{\mathrm{sq}}\beta^2
+
C_{\mathrm{mov}}\beta\varepsilon_{P,\beta,t}^{\mathrm{mov}}
+
C_{\mathrm{tar}}C_\theta^2\frac{\alpha_t^2}{\beta}.
\end{equation}
The auxiliary quantities $L_\star, \tau(\beta),
\varepsilon_{P,\beta,t}^{\mathrm{mov}}$ are specified in the proof below.
\label{lem:moving_actor_constant_beta}
\end{lemma}
\begin{proof}
Define:
\begin{equation}
    \widetilde \mu_i^t
:=
\mu_i^t-J_i(\theta_t),
\qquad
\widetilde w_i^t
:=
\textcolor{revisionblue}{\Pi_i^{\perp}}\bigl(w_i^t-w_i^\star(\theta_t)\bigr).
\end{equation}
Let $q_i \geq 1$. Define $\mathcal V_i^t \coloneqq q_i(\widetilde\mu_i^t)^2+\|\widetilde w_i^t\|_2^2,$ and $\mathcal V_t\coloneqq \sum_{i=1}^n\mathcal V_i^t.$ Define $\|(u,v)\|_{q_i}^2\coloneqq q_i \|u\|^2+\|v\|^2.$
From definition of $\textcolor{revisionblue}{\Pi_i^{\perp}}$, the TD learning error can alternatively be represented as $\delta_i^t = r_i^t-\mu_i^t+(\textcolor{revisionblue}{\psi_{i,t+1}}-\textcolor{revisionblue}{\psi_{i,t}})^\top \textcolor{revisionblue}{\Pi_i^{\perp}}w_i^t.$
Applying $\textcolor{revisionblue}{\Pi_i^{\perp}}$ to the raw critic update gives $\textcolor{revisionblue}{\Pi_i^{\perp}}w_i^{t+1} = \textcolor{revisionblue}{\Pi_i^{\perp}}w_i^t+\beta\delta_i^t\textcolor{revisionblue}{\Pi_i^{\perp}}\textcolor{revisionblue}{\psi_{i,t}}.$
Thus the projected recursion is closed in $\textcolor{revisionblue}{\Pi_i^{\perp}}w_i^t$.
For each agent $i$, define the error vector
\[
e_i^t:=(\widetilde\mu_i^t,\widetilde w_i^t).
\]
For a generic error $e_i=(\widetilde\mu_i,\widetilde w_i)$ and one-step sample $Y$, define
\[
H_i^P(e_i,Y;\theta)
:=
\begin{pmatrix}
c_{ss}(r_i-\mu_i)\\[1mm]
\delta_i \textcolor{revisionblue}{\Pi_i^{\perp}}\psi_i
\end{pmatrix},
\]
where
\[
\mu_i=J_i(\theta)+\widetilde\mu_i,
\qquad
\textcolor{revisionblue}{\Pi_i^{\perp}}w_i=w_i^\star(\theta)+\widetilde w_i,
\]
and
\[
\delta_i
=
r_i-\mu_i+(\psi_i^+-\psi_i)^\top
\left(w_i^\star(\theta)+\widetilde w_i\right).
\]
Let
\[
\overline H_i^P(e_i,\theta)
:=
\mathbb E_{\bar\nu^\theta}\!\left[
H_i^P(e_i,Y;\theta)
\right].
\]
Using the projected Bellman equation at $\theta$, we have
\[
\overline H_i^P(e_i,\theta)
=
\begin{pmatrix}
-c_{ss}\widetilde\mu_i\\[1mm]
-\textcolor{revisionblue}{\Pi_i^{\perp}}\mathbb E_{\nu^\theta}[\psi_i]\widetilde\mu_i
-
\textcolor{revisionblue}{\Pi_i^{\perp}}A_i(\theta)\widetilde w_i
\end{pmatrix}.
\]
Therefore,
\begin{equation}
    2\langle e_i^t,\overline H_i^P(e_i^t,\theta_t)\rangle_{q_i} = -2q_ic_{ss}(\widetilde\mu_i^t)^2
-2(\widetilde w_i^t)^\top \textcolor{revisionblue}{\Pi_i^{\perp}}A_i(\theta_t)\widetilde w_i^t -
2\widetilde\mu_i^t
(\widetilde w_i^t)^\top
\textcolor{revisionblue}{\Pi_i^{\perp}}\mathbb E_{\nu^{\theta_t}}[\psi_i].
\end{equation}
Since $\textcolor{revisionblue}{\Pi_i^{\perp}}$ is idempotent,
\[
(\widetilde w_i^t)^\top \textcolor{revisionblue}{\Pi_i^{\perp}}A_i(\theta_t)\widetilde w_i^t
\ge
\underline\Delta_i\|\widetilde w_i^t\|_2^2.
\]
Define $\bar m_i
:=
\sup_{\theta}
\left\|
\textcolor{revisionblue}{\Pi_i^{\perp}}\mathbb E_{\nu^\theta}[\textcolor{revisionblue}{\psi_{i,t}}]
\right\|_2
<\infty.$ The cross term satisfies,
\[
2\left|
\widetilde\mu_i^t
(\widetilde w_i^t)^\top
\textcolor{revisionblue}{\Pi_i^{\perp}}\mathbb E_{\nu^{\theta_t}}[\psi_i]
\right|
\le
\underline\Delta_i\|\widetilde w_i^t\|_2^2
+
\frac{\bar m_i^2}{\underline\Delta_i}(\widetilde\mu_i^t)^2.
\]
Let $q_i\ge \frac{\bar m_i^2}{c_{ss}\underline\Delta_i},$ we get, 
\begin{equation}
    2\langle e_i^t,\overline H_i^P(e_i^t,\theta_t)\rangle_{q_i} \le
-q_ic_{ss}(\widetilde\mu_i^t)^2
-
\underline\Delta_i\|\widetilde w_i^t\|_2^2.
\end{equation}
Hence, 
\begin{equation}
    2\langle e_i^t,\overline H_i^P(e_i^t,\theta_t)\rangle_{q_i}
\le
-\Delta_{\min}^\chi\mathcal V_i^t.
\label{eq:frozen-drift-projected}
\end{equation}
Define $z_i^\star(\theta_t)\coloneqq(J_i(\theta_t),w_i^\star(\theta_t)).$ Define the old-target next error by measuring the next iterate against the current target
$z_i^\star(\theta_t)$:
\[
e_i^{t+1,\mathrm{old}}
:=
\left(
\mu_i^{t+1}-J_i(\theta_t),
\,
\textcolor{revisionblue}{\Pi_i^{\perp}}w_i^{t+1}-w_i^\star(\theta_t)
\right).
\]
Then
\[
e_i^{t+1,\mathrm{old}}
=
e_i^t+\beta H_i^P(e_i^t,Y_t;\theta_t).
\]
Therefore,
\begin{equation}
    \|e_i^{t+1,\mathrm{old}}\|_{q_i}^2 = \mathcal V_i^t
+
2\beta
\langle e_i^t,H_i^P(e_i^t,Y_t;\theta_t)\rangle_{q_i} +
\beta^2
\|H_i^P(e_i^t,Y_t;\theta_t)\|_{q_i}^2. 
\label{eq:old-target-expansion}
\end{equation}
Define $b_i^t
:=
\mathbb E\!\left[
H_i^P(e_i^t,Y_t;\theta_t)\mid\mathcal F_t
\right]
-
\overline H_i^P(e_i^t,\theta_t),$ and $\zeta_i^{t+1}
:=
H_i^P(e_i^t,Y_t;\theta_t)
-
\mathbb E\!\left[
H_i^P(e_i^t,Y_t;\theta_t)\mid\mathcal F_t
\right].$ Since $e_i^t$ and $\theta_t$ are $\mathcal F_t$-measurable, $b_i^t$ is $\mathcal F_t$-measurable and $\mathbb E[\zeta_i^{t+1}\mid\mathcal F_t]=0.$ Taking conditional expectation in \eqref{eq:old-target-expansion}, using
\eqref{eq:frozen-drift-projected}, and then taking total expectation gives 
\begin{equation}
    \mathbb E\!\left[
\sum_i\|e_i^{t+1,\mathrm{old}}\|_{q_i}^2
\right]
\le (1-\Delta_{\min}^\chi\beta)\mathbb E[\mathcal V_t] +
2\beta\,
\mathbb E\!\left[
\sum_i\langle e_i^t,b_i^t\rangle_{q_i}
\right] +
\beta^2
\mathbb E\!\left[
\sum_i\|H_i^P(e_i^t,Y_t;\theta_t)\|_{q_i}^2
\right].
\label{eq:old-target-before-bias}
\end{equation}
Since rewards, features, and $w_i^\star(\theta)$ are bounded, and $H_i^P$ is affine in
$e_i$, there exists $C_{\mathrm{sq}}<\infty$ such that
\begin{equation}
    \sum_i\|H_i^P(e_i^t,Y_t;\theta_t)\|_{q_i}^2
\le
C_{\mathrm{sq}}(1+\mathcal V_t).
\label{eq:H-square-bound}
\end{equation}
Therefore,
\begin{equation}
    \beta^2
\mathbb E\!\left[
\sum_i\|H_i^P(e_i^t,Y_t;\theta_t)\|_{q_i}^2
\right]
\le
C_{\mathrm{sq}}\beta^2
+
C_{\mathrm{sq}}\beta^2\mathbb E[\mathcal V_t].
\label{eq:H-square-bound-2}
\end{equation}
Define
\begin{equation}
    \Lambda^P(e,\theta,Y)
:=
2\sum_{i=1}^n
\left\langle
e_i,\,
H_i^P(e_i,Y;\theta)-\overline H_i^P(e_i,\theta)
\right\rangle_{q_i}.
\end{equation}
Recall that $b_i^t
:=
\mathbb E\!\left[
H_i^P(e_i^t,Y_t;\theta_t)\mid\mathcal F_t
\right]
-
\overline H_i^P(e_i^t,\theta_t).$ Since \(e_i^t\), \(\theta_t\), and
\(\overline H_i^P(e_i^t,\theta_t)\) are \(\mathcal F_t\)-measurable, we have 
\begin{equation}
    \mathbb E\!\left[
\Lambda^P(e^t,\theta_t,Y_t)\mid\mathcal F_t
\right] = 2\sum_{i=1}^n
\left\langle
e_i^t,\,
\mathbb E\!\left[
H_i^P(e_i^t,Y_t;\theta_t)\mid\mathcal F_t
\right]
-
\overline H_i^P(e_i^t,\theta_t)
\right\rangle_{q_i} =
2\sum_{i=1}^n
\left\langle
e_i^t,b_i^t
\right\rangle_{q_i}.
\end{equation}
Taking total expectation gives
\begin{equation}
    2\mathbb E\!\left[
\sum_i\langle e_i^t,b_i^t\rangle_{q_i}
\right]
=
\mathbb E[\Lambda^P(e^t,\theta_t,Y_t)].
\label{eq:Lambda-cross-term}
\end{equation}
We now bound the right-hand side. Let $\tau:=\tau(\beta).$ Decompose 
\begin{equation}
    \Lambda^P(e^t,\theta_t,Y_t)
= \Lambda^P(e^{t-\tau},\theta_{t-\tau},Y_t) +
\left[
\Lambda^P(e^t,\theta_t,Y_t)
-
\Lambda^P(e^{t-\tau},\theta_{t-\tau},Y_t)
\right].
\label{eq:Lambda-lag-decomp}
\end{equation}
Because \(H_i^P(e_i,Y;\theta)\) and
\(\overline H_i^P(e_i,\theta)\) are affine in \(e_i\), and because rewards,
features, \(J_i(\theta)\), \(w_i^\star(\theta)\),
\(\textcolor{revisionblue}{\Pi_i^{\perp}}\mathbb E_{\nu^\theta}[\psi_i]\), and \(\textcolor{revisionblue}{\Pi_i^{\perp}}A_i(\theta)\) are uniformly
bounded, there exists a finite constant \(C_\Lambda\) such that 
\begin{equation}
    |\Lambda^P(e,\theta,Y)|
\le
C_\Lambda(1+\|e\|^2).
\label{eq:Lambda-growth}
\end{equation}
Moreover, by the Lipschitz assumptions on \(J_i(\theta)\), \(w_i^\star(\theta)\),
and the stationary moments \(\textcolor{revisionblue}{\Pi_i^{\perp}}\mathbb E_{\nu^\theta}[\psi_i]\) and
\(\textcolor{revisionblue}{\Pi_i^{\perp}}A_i(\theta)\), there exists a finite constant \(C_\Lambda\) such that
\begin{equation}
    |\Lambda^P(e,\theta,Y)-\Lambda^P(e',\theta',Y)| \le
C_\Lambda(1+\|e\|+\|e'\|)\|e-e'\| +
C_\Lambda(1+L_\nu)(1+\|e\|^2+\|e'\|^2)
\|\theta-\theta'\|_2 .
\label{eq:Lambda-growth-lip}
\end{equation}
The first term in \eqref{eq:Lambda-growth-lip} reflects that
\(\Lambda^P\) is quadratic in \(e\), while the second term reflects the
Lipschitz dependence of the moving target and the stationary one-step law on
\(\theta\).

By the bounded actor increment assumption,
\begin{equation}
    \|\theta_t-\theta_{t-\tau}\|_2
\le
\sum_{\ell=t-\tau}^{t-1}\|\theta_{\ell+1}-\theta_\ell\|_2
\le
C_\theta
\sum_{\ell=t-\tau}^{t-1}\alpha_\ell
=
C_\theta A_t^\beta
\label{eq:theta-window-bound}
\end{equation}
where $A_t^\beta
:=
\sum_{\ell=t-\tau(\beta)}^{t-1}\alpha_\ell.$ We next control the change of the critic error over the same window. Since the
projected critic recursion is affine in \(e^s\), and since the moving target
\(z_i^\star(\theta)\) is Lipschitz, there exists a finite constant \(C_z\) such
that, for all \(s\), 
\begin{equation}
    \|e^{s+1}-e^s\|
\le
C_z\beta(1+\|e^s\|)
+
L_\star C_\theta\alpha_s
\label{eq:e-window-one-step}
\end{equation}
where $L_\star^2
:=
\sum_{i=1}^n(q_iL_{J,i}^2+L_{w,i}^2).$ Set $D_s:=1+\|e^s\|.$ Summing \eqref{eq:e-window-one-step} over \(s\in[t-\tau,t-1]\) gives 
\begin{equation}
    \|e^t-e^{t-\tau}\|
\le
C_z\beta\sum_{\ell=t-\tau}^{t-1}D_\ell
+
L_\star C_\theta A_t^\beta.
\label{eq:e-window-sum}
\end{equation}
Moreover, for every \(s\in[t-\tau,t]\),
\begin{equation}
    D_s
\le
D_t+\sum_{\ell=s}^{t-1}\|e^{\ell+1}-e^\ell\|
\le
D_t+
C_z\beta\sum_{\ell=t-\tau}^{t-1}D_\ell
+
L_\star C_\theta A_t^\beta.
\end{equation}

Set
\[
M_t:=\max_{t-\tau\le s\le t} D_s.
\]
Then the preceding inequality implies
\[
M_t
\le
D_t + C_z\beta\tau(\beta) M_t + L_\star C_\theta A_t^\beta.
\] 
The condition
\(C_z\beta\tau(\beta)\le \frac14\) allows the term involving \(M_t\) to be
absorbed into the left-hand side, while
\(L_\star C_\theta A_t^\beta\le \frac14\) ensures that the target-motion term
is \(O(D_t)\), since \(D_t=1+\|e^t\|\ge 1\). Hence
\[
M_t
\le
\frac{D_t+L_\star C_\theta A_t^\beta}{1-C_z\beta\tau(\beta)}
\le
\frac{D_t+\frac14 D_t}{1-\frac14}
\le
2D_t.
\]
Therefore,
\begin{equation}
    \max_{t-\tau\le s\le t}D_s
\le
C D_t.
\label{eq:window-stability}
\end{equation}
Substituting \eqref{eq:window-stability} into \eqref{eq:e-window-sum} yields the
finite-window comparison bound 
\begin{equation}
    \|e^t-e^{t-\tau}\|
\le
C\left(\beta\tau(\beta)+C_\theta A_t^\beta\right)(1+\|e^t\|).
\label{eq:e-window-pathwise}
\end{equation}
Consequently,
\begin{equation}
    \mathbb E\!\left[
(1+\|e^t\|+\|e^{t-\tau}\|)
\|e^t-e^{t-\tau}\|
\right]
\le
C
\left(
\beta\tau(\beta)+C_\theta A_t^\beta
\right)
\left(1+\mathbb E[\mathcal V_t]\right),
\label{eq:e-window-bound}
\end{equation}
where we used \(q_i\ge1\), so that \(\|e^t\|^2\le \mathcal V_t\). Combining \eqref{eq:Lambda-growth-lip}, \eqref{eq:theta-window-bound}, and
\eqref{eq:e-window-bound}, and using
\begin{equation}
    \mathbb E\!\left[
(1+\|e^t\|^2+\|e^{t-\tau}\|^2)
\|\theta_t-\theta_{t-\tau}\|_2
\right]
\le
C C_\theta A_t^\beta(1+\mathbb E[\mathcal V_t]),
\end{equation}
we get,
\begin{equation}
    \left|
\mathbb E\!\left[
\Lambda^P(e^t,\theta_t,Y_t)
-
\Lambda^P(e^{t-\tau},\theta_{t-\tau},Y_t)
\right]
\right| \le
C(1+\mathbb E[\mathcal V_t])
\left(
\beta\tau(\beta)
+
C_\theta A_t^\beta
+
L_\nu C_\theta A_t^\beta
\right).
\label{eq:Lambda-parameter-motion}
\end{equation}
The term with \(L_\nu\) records the dependence of the stationary one-step mean
\(\overline H_i^P(e,\theta)\) on the stationary law \(\bar\nu^\theta\). Next, consider the lagged term
$\Lambda^P(e^{t-\tau},\theta_{t-\tau},Y_t).$ The pair \((e^{t-\tau},\theta_{t-\tau})\) is
\(\mathcal F_{t-\tau}\)-measurable, which is why the lagging step is valid.
Construct an auxiliary sample \(\widetilde Y_t\) by freezing the actor at
\(\theta_{t-\tau}\) during the interval \([t-\tau,t]\). The
nonhomogeneous-kernel perturbation bound gives 
\begin{equation}
    \left|
\mathbb E\!\left[
\Lambda^P(e^{t-\tau},\theta_{t-\tau},Y_t)
-
\Lambda^P(e^{t-\tau},\theta_{t-\tau},\widetilde Y_t)
\right]
\right|
\le
C(1+\mathbb E[\mathcal V_t])
\tau(\beta)L_KC_\theta A_t^\beta.
\label{eq:nonhomogeneous-kernel-bound}
\end{equation}
Indeed, the total variation distance between the true nonhomogeneous one-step
law and the auxiliary frozen one-step law is bounded by a constant multiple of 
\begin{equation}
    \tau(\beta)L_K
\sum_{\ell=t-\tau}^{t-1}\|\theta_{\ell+1}-\theta_\ell\|_2
\le
\tau(\beta)L_KC_\theta A_t^\beta,
\end{equation}
and \eqref{eq:Lambda-growth} together with \eqref{eq:window-stability} supplies
the factor \(1+\mathbb E[\mathcal V_t]\). Under the auxiliary frozen chain, uniform mixing gives 
\begin{equation}
    \left|
\mathbb E\!\left[
\Lambda^P(e^{t-\tau},\theta_{t-\tau},\widetilde Y_t)
\mid
\mathcal F_{t-\tau}
\right]
-
\mathbb E_{\bar\nu^{\theta_{t-\tau}}}
\!\left[
\Lambda^P(e^{t-\tau},\theta_{t-\tau},Y)
\right]
\right|\le
C(1+\|e^{t-\tau}\|^2)
m_{\mathrm{mix}}\rho_{\mathrm{mix}}^{\tau-1}.
\label{eq:frozen-mixing-bound}
\end{equation}
By the definition of \(\overline H_i^P\),
\begin{equation}
    \mathbb E_{\bar\nu^{\theta_{t-\tau}}}
\!\left[
\Lambda^P(e^{t-\tau},\theta_{t-\tau},Y)
\right]
=0.
\label{eq:stationary-centering-zero}
\end{equation}
Taking expectation in \eqref{eq:frozen-mixing-bound} and using
\eqref{eq:window-stability} gives
\begin{equation}
    \left|
\mathbb E\!\left[
\Lambda^P(e^{t-\tau},\theta_{t-\tau},\widetilde Y_t)
\right]
\right|
\le
C(1+\mathbb E[\mathcal V_t])
m_{\mathrm{mix}}\rho_{\mathrm{mix}}^{\tau-1}.
\label{eq:frozen-mixing-unconditional}
\end{equation}
Combining
\eqref{eq:Lambda-parameter-motion},
\eqref{eq:nonhomogeneous-kernel-bound},
\eqref{eq:frozen-mixing-unconditional},
and \eqref{eq:stationary-centering-zero}, we obtain
\begin{equation}
    \left|
\mathbb E[\Lambda^P(e^t,\theta_t,Y_t)]
\right|
\le
C_{\mathrm{mov}}
\varepsilon_{P,\beta,t}^{\mathrm{mov}}
(1+\mathbb E[\mathcal V_t]).
\label{eq:moving-markovian-bias-final}
\end{equation}
Here $\varepsilon_{P,\beta,t}^{\mathrm{mov}}
:=
m_{\mathrm{mix}}\rho_{\mathrm{mix}}^{\tau(\beta)-1}
+
\beta\tau(\beta)
+
C_\theta A_t^\beta
+
L_\nu C_\theta A_t^\beta
+
\tau(\beta)L_KC_\theta A_t^\beta .$ Using \eqref{eq:Lambda-cross-term} and
\eqref{eq:moving-markovian-bias-final} in
\eqref{eq:old-target-before-bias}, together with
\eqref{eq:H-square-bound-2}, gives
\begin{equation}
    \mathbb E\!\left[
\sum_i\|e_i^{t+1,\mathrm{old}}\|_{q_i}^2
\right]
\le{} \left(
1-\Delta_{\min}^\chi\beta
+
C_{\mathrm{sq}}\beta^2
+
C_{\mathrm{mov}}\beta\varepsilon_{P,\beta,t}^{\mathrm{mov}}
\right)
\mathbb E[\mathcal V_t] +
C_{\mathrm{sq}}\beta^2
+
C_{\mathrm{mov}}\beta\varepsilon_{P,\beta,t}^{\mathrm{mov}}.
\end{equation}
By the smallness condition $C_{\mathrm{sq}}\beta
+
C_{\mathrm{mov}}\varepsilon_{P,\beta,t}^{\mathrm{mov}}
\le
\frac{\Delta_{\min}^\chi}{8},$ we thus obtain 
\begin{equation}
    \mathbb E\!\left[
\sum_i\|e_i^{t+1,\mathrm{old}}\|_{q_i}^2
\right]
\le
\left(1-\frac{7\Delta_{\min}^\chi}{8}\beta\right)
\mathbb E[\mathcal V_t]
+
C_{\mathrm{sq}}\beta^2
+
C_{\mathrm{mov}}\beta\varepsilon_{P,\beta,t}^{\mathrm{mov}}.
\label{eq:old-target-contracted}
\end{equation}
The actual next Lyapunov function is measured against
$
z_i^\star(\theta_{t+1})
=
(J_i(\theta_{t+1}),w_i^\star(\theta_{t+1})).
$ Define $d_i^t
:=
\left(
J_i(\theta_{t+1})-J_i(\theta_t),
\,
w_i^\star(\theta_{t+1})-w_i^\star(\theta_t)
\right).$ By Lipschitzness of the target and the bounded actor increment assumption,
\begin{equation}
    \sum_i\|d_i^t\|_{q_i}^2
\le
L_\star^2\|\theta_{t+1}-\theta_t\|_2^2
\le
L_\star^2C_\theta^2\alpha_t^2.
\label{eq:target-motion-bound}
\end{equation}
Since $e_i^{t+1}=e_i^{t+1,\mathrm{old}}-d_i^t,$ Young's inequality gives, for any \(\eta>0\), 
\begin{equation}
    \|e_i^{t+1}\|_{q_i}^2
\le
(1+\eta)\|e_i^{t+1,\mathrm{old}}\|_{q_i}^2
+
(1+\eta^{-1})\|d_i^t\|_{q_i}^2.
\end{equation}
Choose $\eta:=\frac{\Delta_{\min}^{\chi}}{8}\beta.$ For \(\beta\le1\), $1+\eta^{-1}\le \frac{16}{\Delta_{\min}^{\chi}\beta}.$ Therefore, using \eqref{eq:target-motion-bound}, 
\begin{equation}
    \mathbb E[\mathcal V_{t+1}]
\le
\left(1+\frac{\Delta_{\min}^{\chi}}{8}\beta\right)
\mathbb E\!\left[
\sum_i\|e_i^{t+1,\mathrm{old}}\|_{q_i}^2
\right]
+
\frac{16L_\star^2}{\Delta_{\min}^{\chi}}
C_\theta^2
\frac{\alpha_t^2}{\beta}.
\label{eq:target-young}
\end{equation}
Let $C_{\mathrm{tar}}:=\frac{16L_\star^2}{\Delta_{\min}^{\chi}}.$ Substituting \eqref{eq:old-target-contracted} into \eqref{eq:target-young}, and using $\left(1+\frac{\Delta_{\min}^{\chi}}{8}\beta\right)
\left(1-\frac{7\Delta_{\min}^{\chi}}{8}\beta\right)
\le
1-\frac{\Delta_{\min}^{\chi}}{4}\beta$ for sufficiently small \(\beta\), yields, after increasing
\(C_{\mathrm{sq}}\) and \(C_{\mathrm{mov}}\) by universal constants if necessary, 
\begin{equation}
    \mathbb E[\mathcal V_{t+1}]
\le
\left(1-\frac{\Delta_{\min}^{\chi}}{4}\beta\right)
\mathbb E[\mathcal V_t]
+
C_{\mathrm{sq}}\beta^2
+
C_{\mathrm{mov}}\beta\varepsilon_{P,\beta,t}^{\mathrm{mov}}
+
C_{\mathrm{tar}}C_\theta^2\frac{\alpha_t^2}{\beta}.
\end{equation}
This proves the result.
\end{proof}

\begin{corollary}[Moving target contraction with decaying step size]
    Suppose the assumptions of Lemma~\ref{lem:moving_actor_constant_beta} hold, but
replace the constant critic stepsize \(\beta\) by a deterministic sequence
\(\{\beta_t\}_{t\ge0}\), with \(0<\beta_t\le\bar\beta\le1\). Let
\(\tau_t\in\{1,\ldots,t\}\), and define $B_t^\beta
:=
\sum_{\ell=t-\tau_t}^{t-1}\beta_\ell,$ $A_t^\beta
:=
\sum_{\ell=t-\tau_t}^{t-1}\alpha_\ell,$ and $\varepsilon_{P,t}^{\mathrm{mov}}
:=
m_{\mathrm{mix}}\rho_{\mathrm{mix}}^{\tau_t-1}
+
B_t^\beta
+
C_\theta A_t^\beta
+
L_\nu C_\theta A_t^\beta
+
\tau_t L_KC_\theta A_t^\beta.$ 
Assume that, for all relevant \(t\), 
\begin{equation}
    C_{\mathrm{sq}}\beta_t
+
C_{\mathrm{mov}}\varepsilon_{P,t}^{\mathrm{mov}}
\le
\frac{\Delta_{\min}^{\chi}}{8},
\qquad
\max\!\left\{
C_zB_t^\beta,\,
L_\star C_\theta A_t^\beta
\right\}
\le
\frac14 .
\end{equation}
Then, for every \(t\ge\tau_t\),
\begin{equation}
    \mathbb E[\mathcal V_{t+1}]
\le
\left(1-\frac{\Delta_{\min}^\chi}{4}\beta_t\right)
\mathbb E[\mathcal V_t]
+
C_{\mathrm{sq}}\beta_t^2
+
C_{\mathrm{mov}}\beta_t\varepsilon_{P,t}^{\mathrm{mov}}
+
C_{\mathrm{tar}}C_\theta^2\frac{\alpha_t^2}{\beta_t}.
\end{equation}
\label{cor:moving_actor_time_varying_beta}
\end{corollary}
\begin{proof}
    The proof is identical to that of
Lemma~\ref{lem:moving_actor_constant_beta}, with fixed step size $\beta$ being replaced with $\beta_t$. 
 In the
finite-window comparison, the constant-window term \(\beta\tau(\beta)\) is
replaced by
\(
B_t^\beta=\sum_{\ell=t-\tau_t}^{t-1}\beta_\ell,
\)
and the actor-window term is
\(
A_t^\beta=\sum_{\ell=t-\tau_t}^{t-1}\alpha_\ell.
\)
The target-motion Young parameter is chosen as
\(
\eta=\frac{\Delta_{\min}^{\chi}}{8}\beta_t,
\)
which gives the term
\(
C_{\mathrm{tar}}C_\theta^2\frac{\alpha_t^2}{\beta_t}.
\)
All remaining steps are unchanged.
\end{proof}

The moving-target critic analysis uses standard delayed-window and freezing arguments from finite-time stochastic-approximation analyses with Markovian samples; see, e.g., finite-time actor-critic analyses such as~\citet{chen2023finite} for a related single-timescale coupling argument.

\begin{remark}[Choice of decaying stepsizes and order of the error terms]
The step-size conditions in the preceding corollary are local finite-time
conditions. A standard way to satisfy them eventually is to take polynomial
critic and actor stepsizes with a logarithmic mixing window:
\[
\beta_t=\frac{\beta_0}{(t+1)^b},
\qquad
\alpha_t=\frac{\alpha_0}{(t+1)^a},
\qquad
b\in\left(\frac12,1\right),
\qquad
a\in\left(\frac{1+b}{2},1\right),
\]
and
\[
\tau_t=\left\lceil c_\tau \log(t+1)\right\rceil,
\qquad
c_\tau>\frac{b}{-\log \rho_{\mathrm{mix}}},
\]
with \(t_0\) chosen as a finite burn-in time large enough so that
\(1\le \tau_t\le t\) and the local smallness conditions of the preceding
corollary hold for all \(t\ge t_0\). The constants \(\alpha_0,\beta_0>0\) are chosen so that
\(0<\beta_t\le \bar\beta\le1\).

Under this choice,
\[
\beta_t\to 0,
\qquad
\sum_{t=0}^{\infty}\beta_t=\infty,
\qquad
\sum_{t=0}^{\infty}\beta_t^2<\infty.
\]
Moreover, since \(b<1\), the condition
\(a>(1+b)/2\) implies \(a>b\). Hence
\[
\frac{\alpha_t}{\beta_t}
=
\frac{\alpha_0}{\beta_0}(t+1)^{-(a-b)}
\longrightarrow 0.
\]

We now identify the size of the finite-window quantities appearing in the
preceding corollary. Since \(\tau_t=O(\log(t+1))=o(t)\), for
\(\ell\in\{t-\tau_t,\ldots,t-1\}\) we have
\[
\beta_\ell=O(\beta_t),
\qquad
\alpha_\ell=O(\alpha_t).
\]
Therefore,
\[
B_t^\beta
=
\sum_{\ell=t-\tau_t}^{t-1}\beta_\ell
=
O(\beta_t\tau_t),
\qquad
A_t^\beta
=
\sum_{\ell=t-\tau_t}^{t-1}\alpha_\ell
=
O(\alpha_t\tau_t),
\]
and
\[
\tau_t A_t^\beta
=
O(\alpha_t\tau_t^2).
\]
For the mixing term, let
\[
r_\rho:=c_\tau(-\log\rho_{\mathrm{mix}}).
\]
Since \(c_\tau>b/(-\log\rho_{\mathrm{mix}})\), we have \(r_\rho>b\). Hence
\[
m_{\mathrm{mix}}\rho_{\mathrm{mix}}^{\tau_t-1}
=
O((t+1)^{-r_\rho})
=
O(\beta_t).
\]
Consequently,
\[
\varepsilon_{P,t}^{\mathrm{mov}}
=
m_{\mathrm{mix}}\rho_{\mathrm{mix}}^{\tau_t-1}
+
B_t^\beta
+
C_\theta A_t^\beta
+
L_\nu C_\theta A_t^\beta
+
\tau_t L_KC_\theta A_t^\beta
\]
satisfies
\[
\varepsilon_{P,t}^{\mathrm{mov}}
=
O(\beta_t)
+
O(\beta_t\tau_t)
+
O(\alpha_t\tau_t)
+
O(\alpha_t\tau_t^2).
\]
Equivalently,
\[
\varepsilon_{P,t}^{\mathrm{mov}}
=
O\!\left((t+1)^{-b}\log(t+1)
+
(t+1)^{-a}\log^2(t+1)\right).
\]
Since \(a>b\), the actor-window contribution is of smaller polynomial order
than the critic-window contribution, and hence
\[
\varepsilon_{P,t}^{\mathrm{mov}}
=
O(\beta_t\tau_t)
=
O\!\left((t+1)^{-b}\log(t+1)\right)
\longrightarrow 0.
\]

Thus the local smallness conditions in the preceding corollary hold for all
sufficiently large \(t\). Indeed,
\[
C_{\mathrm{sq}}\beta_t
+
C_{\mathrm{mov}}\varepsilon_{P,t}^{\mathrm{mov}}
=
O(\beta_t\tau_t)
\longrightarrow0,
\]
and
\[
C_zB_t^\beta
=
O(\beta_t\tau_t)
\longrightarrow0,
\qquad
L_\star C_\theta A_t^\beta
=
O(\alpha_t\tau_t)
\longrightarrow0.
\]

Finally, the three additive perturbation terms in the one-step recursion have
orders
\[
C_{\mathrm{sq}}\beta_t^2
=
O((t+1)^{-2b}),
\]
\[
C_{\mathrm{mov}}\beta_t\varepsilon_{P,t}^{\mathrm{mov}}
=
O(\beta_t^2\tau_t)
=
O\!\left((t+1)^{-2b}\log(t+1)\right),
\]
and
\[
C_{\mathrm{tar}}C_\theta^2\frac{\alpha_t^2}{\beta_t}
=
O\!\left((t+1)^{-(2a-b)}\right).
\]
Because \(b>1/2\), the first two terms are summable:
\[
\sum_{t=0}^{\infty}\beta_t^2<\infty,
\qquad
\sum_{t=0}^{\infty}
\beta_t\varepsilon_{P,t}^{\mathrm{mov}}<\infty.
\]
Because \(a>(1+b)/2\), we have \(2a-b>1\), and therefore
\[
\sum_{t=0}^{\infty}\frac{\alpha_t^2}{\beta_t}<\infty.
\]
Thus this choice of stepsizes makes the moving-target and target-drift error
terms asymptotically negligible, while the condition
\(\sum_t\beta_t=\infty\) ensures that the critic contraction accumulates over
time.
\end{remark}

\subsection{Convergence of the Actor}
\label{app:unregularized-actor}

We now prove the actor-side stationarity estimate for the unregularized
objective.  The proof is written so that it can be coupled with the
single-timescale moving-target critic recursion from the preceding critic
subsection.
The conclusion is a bound on the policy-gradient norm; no unregularized global
optimality or Nash-gap conclusion is used in this subsection.

\paragraph{Proof roadmap.}
The actor proof first identifies the exact gradient target and then bounds each source of bias around it.
Lemma~\ref{lem:product-policy-score-identities} and Lemma~\ref{lem:softmax-score-kernel-lipschitz} give the score identities, score bounds, and kernel Lipschitz constants needed for all actor estimates, serving as the regularity conditions for the actor proof throughout.  
Lemma~\ref{lem:conditional-markovian-mixing-error} turns delayed Markovian samples into current-stationary expectations through leveraging the mixing properties of the induced Markov chain.
Lemma~\ref{lem:localized-critic-approximation} and Lemma~\ref{lem:localized-actor-direction-bias-softmax} quantify the localization bias caused by using only nearby critics.  
Lemma~\ref{lem:smoothness-lipschitz-policy}, Lemma~\ref{lem:actor-gradient-error}, and Lemma~\ref{lem:actor-second-moment} control smoothness, critic-induced gradient error, and stochastic variance respectively; Lemma~\ref{lem:actor-drift} combines these estimates into a one-step ascent recursion.  
Proposition~\ref{prop:actor-stationarity} telescopes this drift into a stationarity bound.
The auxiliary comparison facts, Lemma~\ref{lem:avg-reward-unilateral-performance-difference} and Lemma~\ref{lem:nash-gap-gradient-domination}, are the formal comparison counterparts of the partial-observability challenge: the first rewrites unilateral improvement through the one-agent averaged advantage, and the second explains when a softmax-gradient bound can or cannot control the Nash gap.
Theorem~\ref{thm:unreg-finite-time-stationarity} then combines the actor and critic estimates into the final unregularized finite-time result.

For the linear-function-approximation actor update, define
\begin{equation}
    \widehat q_j^t
    :=
    \psi_j(s_j^k(t),a_j^k(t))^\top w_j^t,
    \qquad
    S_i^t
    :=
    \nabla_{\theta_i}
    \log\pi_{\theta_i^t}(a_i(t)|s_i(t)),
\end{equation}
and write
\begin{equation}
    \theta_i^{t+1}
    =
    \theta_i^t+\alpha_t\widehat g_i^t,
    \qquad
    \widehat g_i^t
    :=
    S_i^t\sum_{j\in N_i^k}\widehat q_j^t.
    \label{eq:unreg-actor-increment-new}
\end{equation}
Let $\widehat g^t:=(\widehat g_1^t,\ldots,\widehat g_n^t)$ and write
\[
    X_t:=(s_g(t),a_g(t)).
\]
Let $\mathcal H_t$ denote the history through the sample $X_t$ and through all
updates completed at time $t$.  The actor proof uses a delayed mixing window
$\tau_t\le t$ and the shorthand $t_\tau:=t-\tau_t$.  For a fixed actor
parameter $\theta$, let $\nu^\theta$ denote the stationary state-action
distribution of $X=(s_g,a_g)$ under $\pi_\theta$.

We use the following notation throughout this subsection.  Set
\begin{equation}
    \bar\psi
    :=
    \max_{1\le j\le n}\max_{s_j^k,a_j^k}
    \|\psi_j(s_j^k,a_j^k)\|_2,
    \qquad
    H_i:=|N_i^k|.
    \label{eq:actor-proof-basic-constants}
\end{equation}
Define the window-drift quantities
\begin{equation}
    D_{\theta,t}
    :=
    \sum_{\ell=t_\tau}^{t-1}
    \|\theta_{\ell+1}-\theta_\ell\|_2,
    \qquad
    D_{w,t}
    :=
    \left(
    \sum_{i=1}^n
    \left(
    \sum_{\ell=t_\tau}^{t-1}
    \|w_i^{\ell+1}-w_i^\ell\|_2
    \right)^2
    \right)^{1/2}.
    \label{eq:actor-window-drift-defs}
\end{equation}
We also use the accumulated kernel-drift proxy
\begin{equation}
    D_{K,t}
    :=
    \sum_{\ell=t_\tau}^{t-1}
    \|\theta_\ell-\theta_{t_\tau}\|_2 .
    \label{eq:actor-kernel-window-drift-def}
\end{equation}
Since
$\|\theta_\ell-\theta_{t_\tau}\|_2
\leq
\sum_{r=t_\tau}^{\ell-1}\|\theta_{r+1}-\theta_r\|_2$, one always has the crude
bound $D_{K,t}\leq \tau_t D_{\theta,t}$.
Let $\textcolor{revisionblue}{\Pi_i^{\perp}}:=\Pi_{2,S_{\Psi_i,e}^\perp}$ be the Euclidean projection onto $S_{\Psi_i,e}^\perp$ and define the
critic tracking error used by the actor proof as
\begin{equation}
    \widetilde w_i^t
    :=
    \textcolor{revisionblue}{\Pi_i^{\perp}}\bigl(w_i^t-w_i^\star(\theta_t)\bigr),
    \qquad
    \mathcal V_t
    :=
    \sum_{i=1}^n
    \left(q_i(\widetilde\mu_i^t)^2+\|\widetilde w_i^t\|_2^2\right),
    \label{eq:actor-proof-critic-tracking-error}
\end{equation}
where $q_i>0$ is the scalar weight used in the moving-target critic Lyapunov
function and $\widetilde\mu_i^t:=\mu_i^t-J_i(\theta_t)$.  The actor update is
insensitive to additive constants in the critic; hence we work with the
normalized representatives $w_i^t,w_i^\star(\theta_t)\in S_{\Psi_i,e}^\perp$, so
$w_i^t-w_i^\star(\theta_t)=\widetilde w_i^t$.

The actor-side primitive inputs used below are stated in the main body:
ergodicity in Assumption~\ref{assump:ergodicity}, uniform mixing in
Assumption~\ref{assump:uniform_mixing}, bounded critic iterates in
Assumption~\ref{assump:bounded_critic}, the derived critic-target
bound in Lemma~\ref{lem:lipschitzness}, and spatial decay in
Assumption~\ref{assump:actor-spatial-decay}.  Let
$\widetilde Q_i^\theta$ denote the localized tabular fixed point
$\widehat Q_i^\star$ in \eqref{eq:tabular-fp}, written as a function of the
local state-action pair.  Define
\begin{equation}
    \varepsilon_{\mathrm{FA}}
    :=
    \max_{1\le i\le n}
    \sup_{\theta\in\Theta_{\mathrm{ac}}}
    \sup_{s_i^k,a_i^k}
    \left|
    \psi_i(s_i^k,a_i^k)^\top w_i^\star(\theta)
    -
    \widetilde Q_i^\theta(s_i^k,a_i^k)
    \right|.
    \label{eq:actor-fa-error-new}
\end{equation}
Define
\begin{equation}
    W_{\mathrm{act}}:=\max\{W,W_\star\},
    \qquad
    B_i^W:=|N_i^k|W_{\mathrm{act}}.
    \label{eq:actor-BW-def-new}
\end{equation}

\subsubsection{Provable regularity consequences}

\begin{lemma}[Product-policy score identities]
    \label{lem:product-policy-score-identities}
    Under the product policy
    $\pi_\theta(a_g|s_g)=\prod_{i=1}^n\pi_{\theta_i}(a_i|s_i)$, for every
    agent $i\in\mathcal{N}$,
    \begin{equation}
        \nabla_{\theta_i}\log\pi_\theta(a_g|s_g)
        =
        \nabla_{\theta_i}\log\pi_{\theta_i}(a_i|s_i).
    \end{equation}
    Moreover, under the softmax parameterization
    \[
        \pi_{\theta_i}(a_i|s_i)
        =
        \frac{\exp(\theta_i(s_i,a_i))}
        {\sum_{a_i'\in\mathcal{A}_i}\exp(\theta_i(s_i,a_i'))},
    \]
    for every $a\in\mathcal{A}_i$,
    \begin{equation}
        \frac{\partial}{\partial \theta_i(s_i,a)}
        \log\pi_{\theta_i}(a_i|s_i)
        =
        \mathbb{I}\{a_i=a\}-\pi_{\theta_i}(a|s_i).
    \end{equation}
    In particular, for finite action spaces, the local softmax score is
    uniformly bounded as
    \begin{equation}
        \left\|
        \nabla_{\theta_i}\log\pi_{\theta_i}(a_i|s_i)
        \right\|_2
        \leq
        \sqrt{2}
        \qquad
        \forall \theta_i,s_i,a_i.
        \label{eq:softmax-score-sqrt-two-bound}
    \end{equation}
    Moreover, it has zero conditional mean:
    \begin{equation}
        \sum_{a_i\in\mathcal A_i}
        \pi_{\theta_i}(a_i|s_i)
        \nabla_{\theta_i}\log\pi_{\theta_i}(a_i|s_i)
        =
        0.
        \label{eq:softmax-score-zero-mean}
    \end{equation}
\end{lemma}

\begin{proof}
    See Lemma~\ref{auxproof:lem:product-policy-score-identities} for the
    restated statement and proof.
\end{proof}

\begin{lemma}[Softmax score and kernel Lipschitz constants]
    \label{lem:softmax-score-kernel-lipschitz}
    Suppose each local policy is the finite-action tabular softmax policy
    \begin{equation}
        \pi_{\theta_i}(a_i|s_i)
        =
        \frac{\exp(\theta_i(s_i,a_i))}
        {\sum_{u\in\mathcal A_i}\exp(\theta_i(s_i,u))},
        \qquad
        \pi_\theta(a_g|s_g)=\prod_{i=1}^n\pi_{\theta_i}(a_i|s_i).
    \end{equation}
    Define the local score
    $S_i(\theta_i;s_i,a_i):=\nabla_{\theta_i}\log\pi_{\theta_i}(a_i|s_i)$
    and the global score
    $S(\theta;s_g,a_g):=\nabla_\theta\log\pi_\theta(a_g|s_g)$.
    Then, for every $i,\theta_i,s_i,a_i$,
    \begin{equation}
        \|S_i(\theta_i;s_i,a_i)\|_2\le G_i,
        \qquad
        G_i:=\sqrt2,
        \label{eq:actor-score-bound-new}
    \end{equation}
    and, for every $\theta,s_g,a_g$,
    \begin{equation}
        \|S(\theta;s_g,a_g)\|_2\le G,
        \qquad
        G:=\sqrt{\sum_{i=1}^nG_i^2}=\sqrt{2n}.
    \end{equation}
    Moreover, for every $i,s_i,a_i,\theta_i,\theta_i'$,
    \begin{equation}
        \left\|
        S_i(\theta_i;s_i,a_i)
        -
        S_i(\theta_i';s_i,a_i)
        \right\|_2
        \le
        L_{S,i}\|\theta_i-\theta_i'\|_2,
        \qquad
        L_{S,i}:=1 .
        \label{eq:actor-score-lipschitz-new}
    \end{equation}
    Finally, define the frozen global state-action kernel
    \begin{equation}
        \textcolor{revisionblue}{P_{\pi_\theta}}((s_g,a_g),(s_g',a_g'))
        :=
        P(s_g'|s_g,a_g)\pi_\theta(a_g'|s_g').
    \end{equation}
    For every $\theta,\theta'$,
    \begin{equation}
        \sup_x
        \|\textcolor{revisionblue}{P_{\pi_\theta}}(x,\cdot)-\textcolor{revisionblue}{P_{\pi_{\theta'}}}(x,\cdot)\|_{\mathrm{TV}}
        \leq
        L_K\|\theta-\theta'\|_2,
        \qquad
        L_K:=G=\sqrt{2n}.
        \label{eq:actor-kernel-lipschitz-new}
    \end{equation}
\end{lemma}

\begin{proof}
    See Lemma~\ref{auxproof:lem:softmax-score-kernel-lipschitz} for the
    restated statement and proof.
\end{proof}

\begin{lemma}[Conditional Markovian mixing error]
    \label{lem:conditional-markovian-mixing-error}
    Let $\Theta$ be the actor-parameter set considered in the actor proof.  For
    each $\theta\in\Theta$, define the frozen global state-action kernel
    \begin{equation}
        \textcolor{revisionblue}{P_{\pi_\theta}}((s_g,a_g),(s_g',a_g'))
        =
        P(s_g'|s_g,a_g)\pi_\theta(a_g'|s_g').
    \end{equation}
    Suppose there exist constants
    $m_{\mathrm{mix}}\geq1$, $\rho_{\mathrm{mix}}\in(0,1)$, and
    $L_K<\infty$ such that, for all $\theta,\theta'\in\Theta$ and all
    $\ell\ge0$,
    \begin{align}
        \sup_x\|\textcolor{revisionblue}{P_{\pi_\theta}}^\ell(x,\cdot)-\nu^\theta\|_{\mathrm{TV}}
        &\leq
        m_{\mathrm{mix}}\rho_{\mathrm{mix}}^\ell,
        \label{eq:aux-uniform-geometric-mixing}\\
        \sup_x\|\textcolor{revisionblue}{P_{\pi_\theta}}(x,\cdot)-\textcolor{revisionblue}{P_{\pi_{\theta'}}}(x,\cdot)\|_{\mathrm{TV}}
        &\leq
        L_K\|\theta-\theta'\|,
        \label{eq:aux-kernel-lipschitz-for-mixing}
    \end{align}
    Define
    \begin{equation}
        L_\nu
        :=
        \frac{2m_{\mathrm{mix}}L_K}{1-\rho_{\mathrm{mix}}}.
        \label{eq:actor-stationary-lipschitz-constant}
    \end{equation}
    Then
    \begin{equation}
        \|\nu^\theta-\nu^{\theta'}\|_{\mathrm{TV}}
        \leq
        L_\nu\|\theta-\theta'\|.
        \label{eq:stationary-lipschitz-for-mixing}
    \end{equation}

    Let $X_t:=(s_g(t),a_g(t))$ and let $\mathcal H_t$ be the history through
    sample $X_t$ and all updates up to time $t$.  Fix a time $t$, a window
    $\tau_t\le t$, and write $s:=t-\tau_t$.  Suppose that, conditional on
    $\mathcal H_s$, the adaptive trajectory over the window is generated by
    \begin{equation}
        \mathbb P(X_{\ell+1}\in A\mid\mathcal H_\ell)
        =
        \textcolor{revisionblue}{P_{\pi_{\vartheta_\ell}}}(X_\ell,A),
        \qquad
        \ell=s,\ldots,t-1,
        \label{eq:adaptive-window-kernel-condition}
    \end{equation}
    for $\mathcal H_\ell$-measurable parameters
    $\vartheta_\ell\in\Theta$.  In the linear-FA algorithm with
    $\mathcal H_s$ interpreted after the time-$s$ updates, the effective choice
    is $\vartheta_\ell=\theta_{\ell+1}$; under a pre-update convention it is
    $\vartheta_\ell=\theta_\ell$.

    Then, for every $\mathcal H_s$-measurable test function
    $\phi_t:\mathcal S_g\times\mathcal A_g\to\mathbb R^d$ and every
    nonnegative $\mathcal H_s$-measurable random variable $B_t$ satisfying
    $\sup_x\|\phi_t(x)\|\le B_t$, the delayed conditional bias satisfies
    \begin{equation}
        \left\|
        \mathbb{E}[\phi_t(X_t)\mid\mathcal{H}_{t-\tau_t}]
        -
        \mathbb{E}_{\nu^{\theta_t}}[\phi_t(X)]
        \right\|
        \leq
        B_t\varepsilon_{\mathrm{mix}}(t),
    \end{equation}
    where
    \begin{equation}
        \bar\varepsilon_{\mathrm{mix}}(t)
        :=
        m_{\mathrm{mix}}\rho_{\mathrm{mix}}^{\tau_t}
        +
        L_K
        \sum_{\ell=t-\tau_t}^{t-1}
        \|\vartheta_\ell-\theta_{t-\tau_t}\|
        +
        L_\nu
        \sum_{\ell=t-\tau_t}^{t-1}
        \|\theta_{\ell+1}-\theta_\ell\|.
        \label{eq:actor-epsilon-mix-raw-definition}
    \end{equation}
    With the paper's normalization of total variation, distances between
    probability laws are at most one; hence the same bound holds with the
    clipped quantity
    \begin{equation}
        \varepsilon_{\mathrm{mix}}(t)
        :=
        \min\{1,\bar\varepsilon_{\mathrm{mix}}(t)\}.
        \label{eq:epsilon-mix-definition}
    \end{equation}
    If, in addition,
    $\|\theta_{\ell+1}-\theta_\ell\|\le C_\theta\alpha_\ell$,
    $\alpha_\ell\le C_\alpha(\ell+1)^{-a}$ for some $a\in(0,1]$, and
    $\vartheta_\ell\in\{\theta_\ell,\theta_{\ell+1}\}$ over the window, then
    the choice $\tau_t=\lceil c_\tau\log(t+2)\rceil$ gives
    \begin{equation}
        \varepsilon_{\mathrm{mix}}(t)
        \leq
        m_{\mathrm{mix}}(t+2)^{-c_\tau|\log\rho_{\mathrm{mix}}|}
        +
        L_K C_\theta C_\alpha
        \tau_t^2(t-\tau_t+1)^{-a}
        +
        L_\nu C_\theta C_\alpha
        \tau_t(t-\tau_t+1)^{-a},
    \end{equation}
    and hence
    \begin{equation}
        \varepsilon_{\mathrm{mix}}(t)
        =
        O\left(t^{-c_\tau|\log\rho_{\mathrm{mix}}|}\right)
        +
        O\left((\log t)^2t^{-a}\right).
    \end{equation}
\end{lemma}

\begin{proof}
    See Lemma~\ref{auxproof:lem:conditional-markovian-mixing-error} for the
    restated statement and proof.
\end{proof}

\begin{lemma}[Localized critic approximation; \citet{qu2020scalable}]
    \label{lem:localized-critic-approximation}
    Suppose the $(c,\rho)$ exponential decay property holds for the
    unregularized local $Q$ functions. For each agent $i$, define a truncated
    critic by
    \begin{align}
        \widetilde Q_i^\theta(s_i^k,a_i^k)
        &:=
        \sum_{s_{-i}^k\in\mathcal{S}_{-i}^k}
        \sum_{a_{-i}^k\in\mathcal{A}_{-i}^k}
        \omega_i(s_{-i}^k,a_{-i}^k;s_i^k,a_i^k)
        Q_i^\theta(s_i^k,s_{-i}^k,a_i^k,a_{-i}^k),
    \end{align}
    where the weights are nonnegative and satisfy
    \begin{equation}
        \sum_{s_{-i}^k,a_{-i}^k}
        \omega_i(s_{-i}^k,a_{-i}^k;s_i^k,a_i^k)=1
        \qquad
        \forall (s_i^k,a_i^k).
    \end{equation}
    Then:
    \begin{enumerate}
        \item The truncated critic uniformly approximates the full local critic:
    \begin{equation}
        \sup_{s_g,a_g}
        \left|
        Q_i^\theta(s_g,a_g)
        -
        \widetilde Q_i^\theta(s_i^k,a_i^k)
        \right|
        \leq
        c\rho^{k+1}.
    \end{equation}
        \item If
        $\|\nabla_{\theta_i}\log\pi_{\theta_i}(a_i|s_i)\|\leq G_i$, and
        \begin{equation}
            \widetilde g_i(\theta)
            :=
            \mathbb{E}_{\nu^\theta}
            \left[
            \nabla_{\theta_i}\log\pi_{\theta_i}(a_i|s_i)
            \sum_{j\in N_i^k}
            \widetilde Q_j^\theta(s_j^k,a_j^k)
            \right],
        \end{equation}
        then, under the sum-reward convention
        $J(\theta)=\sum_{j=1}^nJ_j(\theta)$ used in this paper,
        \begin{equation}
            \left\|
            \widetilde g_i(\theta)-\nabla_{\theta_i}J(\theta)
            \right\|
            \leq
            nG_i c\rho^{k+1}.
        \end{equation}
    \end{enumerate}
\end{lemma}

\begin{proof}
    See Lemma~\ref{auxproof:lem:localized-critic-approximation} for the
    restated statement and proof.
\end{proof}

\begin{lemma}[Localized actor-direction bias under softmax]
    \label{lem:localized-actor-direction-bias-softmax}
    Suppose the $(c,\rho)$ exponential decay property holds for the
    unregularized local $Q$ functions, and define the truncated critics
    $\widetilde Q_j^\theta$ by the convex averaging construction in
    Lemma~\ref{lem:localized-critic-approximation}.  For each agent $i$, define
    the localized stationary actor direction
    \begin{equation}
        \widetilde g_i(\theta)
        :=
        \mathbb E_{\nu^\theta}
        \left[
        \nabla_{\theta_i}\log\pi_{\theta_i}(a_i|s_i)
        \sum_{j\in N_i^k}
        \widetilde Q_j^\theta(s_j^k,a_j^k)
        \right].
        \label{eq:aux-localized-actor-direction}
    \end{equation}
    Under the tabular softmax parameterization,
    \begin{equation}
        \left\|
        \widetilde g_i(\theta)-\nabla_{\theta_i}J(\theta)
        \right\|_2
        \leq
        n\sqrt{2}\,c\rho^{k+1}.
        \label{eq:aux-localized-actor-direction-bias-softmax}
    \end{equation}
    More generally, if
    $\|\nabla_{\theta_i}\log\pi_{\theta_i}(a_i|s_i)\|_2\leq G_i$, then the
    right-hand side is $nG_i c\rho^{k+1}$.
\end{lemma}

\begin{proof}
    See Lemma~\ref{auxproof:lem:localized-actor-direction-bias-softmax} for the
    restated statement and proof.
\end{proof}

\begin{lemma}[Smoothness of the unregularized objective under softmax]
    \label{lem:smoothness-lipschitz-policy}
    Suppose the policy is the finite-action product softmax policy and that
    $|r_i(s_i,a_i)|\le r_{\max}$ for every $i,s_i,a_i$.  Let
    \[
        R:=nr_{\max},
        \qquad
        G:=\sqrt{2n},
        \qquad
        H:=1 .
    \]
    Assume that the policy-induced global kernels are uniformly ergodic in the
    sense that
    \begin{equation}
        K_{\mathrm{mix}}
        :=
        \sup_\theta
        \sum_{t=0}^{\infty}
        \sup_{s_g\in\mathcal{S}_g}
        \|(\textcolor{revisionblue}{P_{\pi_\theta}^{\mathrm{s}}})^t(s_g,\cdot)-d^\theta\|_1
        <\infty.
    \end{equation}
    Define
    \begin{align}
        L_d&:=K_{\mathrm{mix}}G,&
        L_J&:=R(L_d+G),&
        L_{\nu}^{\mathrm{sm}}&:=L_d+G,\\
        B_V&:=2RK_{\mathrm{mix}},&
        B_Q&:=2R+B_V,\\
        L_V&:=K_{\mathrm{mix}}\left(RG(K_{\mathrm{mix}}+2)+GB_V\right)
        +L_dB_V,&
        L_Q&:=L_J+L_V,
    \end{align}
    and
    \begin{equation}
        L_{\nabla J}
        :=
        L_dGB_Q+(G^2+H)B_Q+GL_Q .
    \end{equation}
    Then the objective-smoothness and objective-upper-bound conclusions used in
    the actor proof hold with this $L_{\nabla J}$ and with
    \begin{equation}
        J_{\max}:=\sup_\theta J(\theta)\le nr_{\max}.
        \label{eq:actor-Jmax-new}
    \end{equation}
    That is, for all $\theta,\theta'$,
    \begin{equation}
        \|\nabla J(\theta)-\nabla J(\theta')\|
        \leq
        L_{\nabla J}\|\theta-\theta'\|,
        \label{eq:actor-J-smooth-new}
    \end{equation}
    and $J$ is bounded above by $J_{\max}$.
\end{lemma}

\begin{proof}
    See Lemma~\ref{auxproof:lem:smoothness-lipschitz-policy} for the restated
    statement and proof.
\end{proof}

\begin{lemma}[Expected actor ascent-direction error bound]
    \label{lem:actor-gradient-error}
    Suppose Assumptions~\ref{assump:ergodicity}-\ref{assump:actor-spatial-decay} hold, and use
    Lemma~\ref{lem:lipschitzness}, \eqref{eq:actor-BW-def-new}, and
    \eqref{eq:actor-fa-error-new}.
    Define, for each
    actor block $i$,
    \begin{equation}\begin{aligned}
        &A_i^{\mathrm{tr}}
        :=
        G_i\bar\psi\sqrt{H_i},
        \qquad
        A_i^{\mathrm{FA}}
        :=
        G_iH_i,
        \qquad
        A_i^{\mathrm{loc}}
        :=
        nG_i c,
        \\
        &A_i^{\mathrm{mix},0}
        :=
        G_i\bar\psi B_i^W,
        \qquad
        A_i^{\mathrm{mix},V}
        :=
        G_i\bar\psi\sqrt{H_i},
        \\
        &A_i^{\mathrm{win},0,\theta}
        :=
        2L_{S,i}\bar\psi B_i^W,
        \qquad
        A_i^{\mathrm{win},V,\theta}
        :=
        2L_{S,i}\bar\psi\sqrt{H_i},
        \qquad
        A_i^{\mathrm{win},w}
        :=
        2G_i\bar\psi\sqrt{H_i}.
        \label{eq:block-actor-error-constants-new}
    \end{aligned}\end{equation}
    Define the corresponding vector constants
    \begin{equation}\begin{aligned}
        &C_{\mathrm{tr}}
        :=
        \left(\sum_{i=1}^n (A_i^{\mathrm{tr}})^2\right)^{1/2},
        \quad
        C_{\mathrm{FA}}
        :=
        \left(\sum_{i=1}^n (A_i^{\mathrm{FA}})^2\right)^{1/2},
        \\
        &C_{\mathrm{loc}}
        :=
        \left(\sum_{i=1}^n (A_i^{\mathrm{loc}})^2\right)^{1/2},
        \quad
        C_{\mathrm{mix},0}
        :=
        \left(\sum_{i=1}^n (A_i^{\mathrm{mix},0})^2\right)^{1/2},
        \\
        &C_{\mathrm{mix},V}
        :=
        \left(\sum_{i=1}^n (A_i^{\mathrm{mix},V})^2\right)^{1/2},
        \quad
        C_{\mathrm{win},0}^\theta
        :=
        \left(\sum_{i=1}^n (A_i^{\mathrm{win},0,\theta})^2\right)^{1/2},
        \\
        &C_{\mathrm{win},V}^\theta
        :=
        \left(\sum_{i=1}^n (A_i^{\mathrm{win},V,\theta})^2\right)^{1/2},
        \quad
        C_{\mathrm{win}}^w
        :=
        \left(\sum_{i=1}^n (A_i^{\mathrm{win},w})^2\right)^{1/2}.
        \label{eq:vector-actor-error-constants-new}
    \end{aligned}\end{equation}
    Define the frozen-martingale correlation constants
    \begin{equation}
        B_{\tau,0}:=2C_{\mathrm{mix},0},
        \qquad
        B_{\tau,V}:=2C_{\mathrm{mix},V},
        \qquad
        B_{\tau,w}:=2C_{\mathrm{mix},V},
        \label{eq:frozen-martingale-correlation-constants}
    \end{equation}
    and
    \begin{equation}
        R_{\mathrm{corr},t}
        :=
        L_{\nabla J}D_{\theta,t}
        \left(
        B_{\tau,0}
        +
        B_{\tau,V}\sqrt{\mathcal V_t}
        +
        B_{\tau,w}D_{w,t}
        \right).
        \label{eq:actor-correlation-remainder}
    \end{equation}
    Let $g_t:=\nabla J(\theta_t)$.  Then, for every $t$,
    \begin{align}
        \mathbb E\langle g_t,\widehat g^t\rangle
        &\geq
        \frac12\mathbb E\|g_t\|_2^2
        -
        4C_{\mathrm{tr}}^2\mathbb E\mathcal V_t
        -
        4C_{\mathrm{FA}}^2\varepsilon_{\mathrm{FA}}^2
        -
        4C_{\mathrm{loc}}^2\rho^{2(k+1)}
        \notag\\
        &\quad
        -
        4C_{\mathrm{mix},0}^2
        \mathbb E[\varepsilon_{\mathrm{mix}}(t)]
        -
        4C_{\mathrm{mix},V}^2
        \mathbb E[
        \varepsilon_{\mathrm{mix}}(t)^2\mathcal V_{t_\tau}]
        \notag\\
        &\quad
        -
        4\max\{(C_{\mathrm{win},0}^\theta)^2,(C_{\mathrm{win}}^w)^2\}
        \mathbb E[D_{\theta,t}^2+D_{w,t}^2]
        \notag\\
        &\quad
        -
        4(C_{\mathrm{win},V}^\theta)^2
        \mathbb E[D_{\theta,t}^2\mathcal V_t]
        -
        \mathbb E[R_{\mathrm{corr},t}].
        \label{eq:actor-gradient-error-new}
    \end{align}
\end{lemma}

\begin{proof}
    Fix $t$, write $s:=t_\tau$, and let
    $g_t:=\nabla J(\theta_t)$ and $g_s:=\nabla J(\theta_s)$.  Define the
    delayed frozen estimator
    \begin{equation}
        \widehat g_i^{t,\tau}
        :=
        \nabla_{\theta_i^s}\log\pi_{\theta_i^s}(a_i(t)|s_i(t))
        \sum_{j\in N_i^k}
        \psi_j(s_j^k(t),a_j^k(t))^\top w_j^s,
        \qquad
        \widehat g^{t,\tau}
        :=
        (\widehat g_1^{t,\tau},\ldots,\widehat g_n^{t,\tau}),
        \label{eq:delayed-frozen-actor-estimator}
    \end{equation}
    and set
    \begin{equation}
        m_t^\tau
        :=
        \mathbb E[\widehat g^{t,\tau}\mid\mathcal H_s],
        \qquad
        M_t^\tau
        :=
        \widehat g^{t,\tau}-m_t^\tau .
        \label{eq:frozen-actor-mean-and-martingale}
    \end{equation}
    By construction $\mathbb E[M_t^\tau\mid\mathcal H_s]=0$.

    We first bound the deterministic bias of the delayed frozen mean.  For a
    stationary sample $(s_g,a_g)\sim\nu^{\theta_t}$, define
    \begin{equation}
        S_i
        :=
        \nabla_{\theta_i^t}\log\pi_{\theta_i^t}(a_i|s_i),
        \qquad
        S_i^\tau
        :=
        \nabla_{\theta_i^s}\log\pi_{\theta_i^s}(a_i|s_i),
        \qquad
        \psi_j
        :=
        \psi_j(s_j^k,a_j^k).
    \end{equation}
    Let
    \begin{align}
        \bar h_{i,0}^\tau
        &:=
        \mathbb E[\widehat g_i^{t,\tau}\mid\mathcal H_s],
        \notag\\
        \bar h_{i,1}^\tau
        &:=
        \mathbb E_{\nu^{\theta_t}}
        \left[
        S_i^\tau\sum_{j\in N_i^k}\psi_j^\top w_j^s
        \right],
        \notag\\
        h_{i,1}
        &:=
        \mathbb E_{\nu^{\theta_t}}
        \left[
        S_i\sum_{j\in N_i^k}\psi_j^\top w_j^t
        \right],
        \notag\\
        h_{i,2}
        &:=
        \mathbb E_{\nu^{\theta_t}}
        \left[
        S_i\sum_{j\in N_i^k}\psi_j^\top w_j^\star(\theta_t)
        \right],
        \notag\\
        h_{i,3}
        &:=
        \mathbb E_{\nu^{\theta_t}}
        \left[
        S_i\sum_{j\in N_i^k}\widetilde Q_j^{\theta_t}(s_j^k,a_j^k)
        \right],
        \notag\\
        h_{i,4}
        &:=
        \mathbb E_{\nu^{\theta_t}}
        \left[
        S_i\sum_{j=1}^n Q_j^{\theta_t}(s_g,a_g)
        \right].
        \label{eq:actor-error-intermediate-directions-new}
    \end{align}
    By the average-reward policy-gradient theorem,
    $h_{i,4}=\nabla_{\theta_i}J(\theta_t)$.

    The frozen test function
    \begin{equation}
        f_i^\tau(s_g,a_g)
        :=
        S_i^\tau(s_g,a_g)
        \sum_{j\in N_i^k}
        \psi_j(s_j^k,a_j^k)^\top w_j^s
        \label{eq:delayed-frozen-actor-test-function}
    \end{equation}
    is $\mathcal H_s$-measurable.  Since
    \begin{equation}
        \|f_i^\tau\|_\infty
        \leq
        G_i\bar\psi
        \left(B_i^W+\sqrt{H_i}\sqrt{\mathcal V_s}\right),
        \label{eq:frozen-test-function-dynamic-bound}
    \end{equation}
    Lemma~\ref{lem:conditional-markovian-mixing-error}, applied only to the
    $\mathcal H_s$-measurable function $f_i^\tau$, gives
    \begin{equation}
        \|\bar h_{i,0}^\tau-\bar h_{i,1}^\tau\|_2
        \leq
        A_i^{\mathrm{mix},0}\varepsilon_{\mathrm{mix}}(t)
        +
        A_i^{\mathrm{mix},V}\varepsilon_{\mathrm{mix}}(t)
        \sqrt{\mathcal V_s}.
        \label{eq:h01-frozen-mixing-bound}
    \end{equation}
    The current-to-frozen stationary comparison satisfies
    \begin{align}
        \|\bar h_{i,1}^\tau-h_{i,1}\|_2
        &\leq
        L_{S,i}D_{\theta,t}\bar\psi
        \left(B_i^W+\sqrt{H_i}\sqrt{\mathcal V_t}\right)
        +
        G_i\bar\psi
        \sum_{j\in N_i^k}\|w_j^t-w_j^s\|_2
        \notag\\
        &\leq
        \frac12A_i^{\mathrm{win},0,\theta}D_{\theta,t}
        +
        \frac12A_i^{\mathrm{win},V,\theta}
        D_{\theta,t}\sqrt{\mathcal V_t}
        +
        \frac12A_i^{\mathrm{win},w}D_{w,t}.
        \label{eq:stationary-current-frozen-window-bound}
    \end{align}
    The remaining deterministic terms are unchanged:
    \begin{align}
        \|h_{i,1}-h_{i,2}\|_2
        &\leq
        A_i^{\mathrm{tr}}\sqrt{\mathcal V_t},
        \label{eq:h12-bound-new}\\
        \|h_{i,2}-h_{i,3}\|_2
        &\leq
        A_i^{\mathrm{FA}}\varepsilon_{\mathrm{FA}},
        \label{eq:h23-bound-new}\\
        \|h_{i,3}-h_{i,4}\|_2
        &\leq
        A_i^{\mathrm{loc}}\rho^{k+1}.
        \label{eq:h34-bound-new}
    \end{align}
    Combining
    \eqref{eq:h01-frozen-mixing-bound}--\eqref{eq:h34-bound-new} and then
    applying Minkowski's inequality over actor blocks yields
    \begin{align}
        \|m_t^\tau-g_t\|_2
        &\leq
        C_{\mathrm{tr}}\sqrt{\mathcal V_t}
        +C_{\mathrm{FA}}\varepsilon_{\mathrm{FA}}
        +C_{\mathrm{loc}}\rho^{k+1}
        +C_{\mathrm{mix},0}\varepsilon_{\mathrm{mix}}(t)
        +C_{\mathrm{mix},V}\varepsilon_{\mathrm{mix}}(t)
        \sqrt{\mathcal V_s}
        \notag\\
        &\quad
        +\frac12C_{\mathrm{win},0}^\theta D_{\theta,t}
        +\frac12C_{\mathrm{win},V}^\theta
        D_{\theta,t}\sqrt{\mathcal V_t}
        +\frac12C_{\mathrm{win}}^wD_{w,t}.
        \label{eq:frozen-mean-gradient-bound-new}
    \end{align}
    The same score-Lipschitz and critic-window comparison gives the pathwise
    current-to-frozen estimator bound
    \begin{equation}
        \|\widehat g^t-\widehat g^{t,\tau}\|_2
        \leq
        \frac12C_{\mathrm{win},0}^\theta D_{\theta,t}
        +\frac12C_{\mathrm{win},V}^\theta
        D_{\theta,t}\sqrt{\mathcal V_t}
        +\frac12C_{\mathrm{win}}^wD_{w,t}.
        \label{eq:current-frozen-estimator-vector-bound}
    \end{equation}
    Define
    \begin{align}
        \mathcal A_t
        &:=
        C_{\mathrm{tr}}\sqrt{\mathcal V_t}
        +C_{\mathrm{FA}}\varepsilon_{\mathrm{FA}}
        +C_{\mathrm{loc}}\rho^{k+1}
        +C_{\mathrm{mix},0}\varepsilon_{\mathrm{mix}}(t)
        +C_{\mathrm{mix},V}\varepsilon_{\mathrm{mix}}(t)
        \sqrt{\mathcal V_s}
        \notag\\
        &\quad
        +C_{\mathrm{win},0}^\theta D_{\theta,t}
        +C_{\mathrm{win},V}^\theta D_{\theta,t}\sqrt{\mathcal V_t}
        +C_{\mathrm{win}}^wD_{w,t}.
        \label{eq:actor-total-bias-envelope}
    \end{align}
    Then
    \begin{equation}
        \|m_t^\tau-g_t\|_2
        +
        \|\widehat g^t-\widehat g^{t,\tau}\|_2
        \leq
        \mathcal A_t .
        \label{eq:actor-total-bias-envelope-bound}
    \end{equation}

    It remains to control the frozen martingale correlation with the current
    gradient.  Since $g_s$ is $\mathcal H_s$-measurable and
    $\mathbb E[M_t^\tau\mid\mathcal H_s]=0$,
    \begin{equation}
        \mathbb E\langle g_t,M_t^\tau\rangle
        =
        \mathbb E\langle g_t-g_s,M_t^\tau\rangle .
        \label{eq:frozen-martingale-correlation-identity}
    \end{equation}
    Smoothness of $J$ gives
    $\|g_t-g_s\|_2\leq L_{\nabla J}D_{\theta,t}$.  Moreover, for every sample
    $x$,
    \begin{equation}
        \|\widehat g^{t,\tau}(x)\|_2
        \leq
        C_{\mathrm{mix},0}
        +
        C_{\mathrm{mix},V}\sqrt{\mathcal V_t}
        +
        C_{\mathrm{mix},V}D_{w,t}.
        \label{eq:frozen-estimator-envelope-current-time}
    \end{equation}
    Therefore
    $\|M_t^\tau\|_2
    \leq
    B_{\tau,0}
    +
    B_{\tau,V}\sqrt{\mathcal V_t}
    +
    B_{\tau,w}D_{w,t}$, and
    \begin{equation}
        \left|
        \mathbb E\langle g_t,M_t^\tau\rangle
        \right|
        \leq
        \mathbb E[R_{\mathrm{corr},t}].
        \label{eq:frozen-martingale-correlation-bound}
    \end{equation}

    Finally,
    \begin{align}
        \mathbb E\langle g_t,\widehat g^t\rangle
        &=
        \mathbb E\|g_t\|_2^2
        +
        \mathbb E\langle g_t,m_t^\tau-g_t\rangle
        +
        \mathbb E\langle g_t,\widehat g^t-\widehat g^{t,\tau}\rangle
        +
        \mathbb E\langle g_t,M_t^\tau\rangle
        \notag\\
        &\geq
        \mathbb E\|g_t\|_2^2
        -
        \frac12\mathbb E\|g_t\|_2^2
        -
        \frac12\mathbb E[\mathcal A_t^2]
        -
        \mathbb E[R_{\mathrm{corr},t}].
        \label{eq:expected-ascent-before-square}
    \end{align}
    Using $(x_1+\cdots+x_8)^2\leq8\sum_{\ell=1}^8x_\ell^2$,
    $0\leq\varepsilon_{\mathrm{mix}}(t)\leq1$, and
    \eqref{eq:actor-total-bias-envelope} gives exactly
    \eqref{eq:actor-gradient-error-new}.
\end{proof}

\begin{lemma}[Actor estimator second moment]
    \label{lem:actor-second-moment}
    Suppose Assumption~\ref{assump:bounded_critic} and
    Lemma~\ref{lem:lipschitzness} hold, with $B_i^W$ defined in
    \eqref{eq:actor-BW-def-new}.
    Define
    \begin{equation}
        \sigma_{a,0}^2
        :=
        2\bar\psi^2
        \sum_{i=1}^n G_i^2(B_i^W)^2,
        \qquad
        \sigma_{a,V}^2
        :=
        2\bar\psi^2
        \sum_{i=1}^n G_i^2H_i.
        \label{eq:actor-second-moment-constants-new}
    \end{equation}
    Then, for every $t$,
    \begin{equation}
        \mathbb E\|\widehat g^t\|_2^2
        \leq
        \sigma_{a,0}^2+\sigma_{a,V}^2\mathbb E\mathcal V_t.
        \label{eq:actor-second-moment-new}
    \end{equation}
\end{lemma}

\begin{proof}
    From \eqref{eq:unreg-actor-increment-new}, the score bound, and the
    critic target-plus-tracking decomposition,
    \begin{equation}
        \|\widehat g_i^t\|_2
        \leq
        G_i\bar\psi
        \left(B_i^W+\sqrt{H_i}\sqrt{\mathcal V_t}\right).
    \end{equation}
    Using $(a+b)^2\leq2a^2+2b^2$ gives
    \begin{equation}
        \|\widehat g_i^t\|_2^2
        \leq
        2G_i^2\bar\psi^2(B_i^W)^2
        +
        2G_i^2\bar\psi^2H_i\mathcal V_t.
    \end{equation}
    Summing over $i$ proves \eqref{eq:actor-second-moment-new} with the
    constants in \eqref{eq:actor-second-moment-constants-new}.
\end{proof}

\begin{lemma}[Actor drift inequality]
    \label{lem:actor-drift}
    Suppose the hypotheses of Lemmas~\ref{lem:actor-gradient-error} and
    \ref{lem:actor-second-moment} hold and $\alpha_t\leq1$.  Let
    \begin{align}
        C_V^{\mathrm{act}}
        &:=
        4C_{\mathrm{tr}}^2
        +
        \frac{L_{\nabla J}}{2}\sigma_{a,V}^2,
        \notag\\
        C_{\mathrm{FA}}^{\mathrm{act}}
        &:=
        4C_{\mathrm{FA}}^2,
        \qquad
        C_{\mathrm{loc}}^{\mathrm{act}}
        :=
        4C_{\mathrm{loc}}^2,
        \notag\\
        C_{\mathrm{mix},0}^{\mathrm{act}}
        &:=
        4C_{\mathrm{mix},0}^2,
        \qquad
        C_{\mathrm{mix},V}^{\mathrm{act}}
        :=
        4C_{\mathrm{mix},V}^2,
        \notag\\
        C_{\mathrm{win},0}^{\mathrm{act}}
        &:=
        4\max\{(C_{\mathrm{win},0}^\theta)^2,(C_{\mathrm{win}}^w)^2\},
        \qquad
        C_{\mathrm{win},V}^{\mathrm{act}}
        :=
        4(C_{\mathrm{win},V}^\theta)^2,
        \notag\\
        C_{\mathrm{var}}^{\mathrm{act}}
        &:=
        \frac{L_{\nabla J}}{2}\sigma_{a,0}^2.
        \label{eq:actor-drift-constants-new}
    \end{align}
    Then
    \begin{align}
        \mathbb E[J(\theta_{t+1})]
        &\geq
        \mathbb E[J(\theta_t)]
        +
        \frac{\alpha_t}{2}\mathbb E\|\nabla J(\theta_t)\|_2^2
        -
        C_V^{\mathrm{act}}\alpha_t\mathbb E\mathcal V_t
        \notag\\
        &\quad
        -
        C_{\mathrm{FA}}^{\mathrm{act}}\alpha_t\varepsilon_{\mathrm{FA}}^2
        -
        C_{\mathrm{loc}}^{\mathrm{act}}\alpha_t\rho^{2(k+1)}
        -
        C_{\mathrm{mix},0}^{\mathrm{act}}\alpha_t
        \mathbb E[\varepsilon_{\mathrm{mix}}(t)]
        -
        C_{\mathrm{mix},V}^{\mathrm{act}}\alpha_t
        \mathbb E[
        \varepsilon_{\mathrm{mix}}(t)^2\mathcal V_{t_\tau}]
        \notag\\
        &\quad
        -
        C_{\mathrm{win},0}^{\mathrm{act}}\alpha_t
        \mathbb E\left[D_{\theta,t}^2+D_{w,t}^2\right]
        -
        C_{\mathrm{win},V}^{\mathrm{act}}\alpha_t
        \mathbb E[D_{\theta,t}^2\mathcal V_t]
        -
        \alpha_t\mathbb E[R_{\mathrm{corr},t}]
        -
        C_{\mathrm{var}}^{\mathrm{act}}\alpha_t^2 .
        \label{eq:actor-drift-new}
    \end{align}
\end{lemma}

\begin{proof}
    Write $g_t:=\nabla J(\theta_t)$.  By $L_{\nabla J}$-smoothness,
    \begin{equation}
        J(\theta_t+u)
        \geq
        J(\theta_t)+\langle g_t,u\rangle
        -\frac{L_{\nabla J}}{2}\|u\|_2^2
        \qquad
        \forall u .
    \end{equation}
    Applying this with $u=\alpha_t\widehat g^t$ and taking total expectation
    gives
    \begin{align}
        \mathbb E[J(\theta_{t+1})-J(\theta_t)]
        &\geq
        \alpha_t\mathbb E\langle g_t,\widehat g^t\rangle
        -
        \frac{L_{\nabla J}}{2}\alpha_t^2
        \mathbb E\|\widehat g^t\|_2^2.
        \label{eq:actor-drift-proof-start-new}
    \end{align}
    Using Lemmas~\ref{lem:actor-gradient-error} and
    \ref{lem:actor-second-moment} in
    \eqref{eq:actor-drift-proof-start-new}, and using $\alpha_t\leq1$ to
    absorb $\alpha_t^2\mathbb E\mathcal V_t$ into
    $\alpha_t\mathbb E\mathcal V_t$, gives exactly
    \eqref{eq:actor-drift-new} with the constants in
    \eqref{eq:actor-drift-constants-new}.
\end{proof}

\begin{proposition}[Actor stationarity with an explicit critic-tracking term]
    \label{prop:actor-stationarity}
    Suppose the hypotheses of Lemma~\ref{lem:actor-drift} hold for all
    $t$ in the horizon.  For integers
    $0\leq t_0<T$, define
    \begin{equation}
        A_{t_0,T}:=\sum_{t=t_0}^{T-1}\alpha_t,
        \qquad
        \overline G_{t_0,T}
        :=
        \frac{1}{A_{t_0,T}}
        \sum_{t=t_0}^{T-1}\alpha_t
        \mathbb E\|\nabla J(\theta_t)\|_2^2,
        \label{eq:actor-average-gradient-def-new}
    \end{equation}
    and
    \begin{align}
        \overline V_{t_0,T}
        &:=
        \frac{1}{A_{t_0,T}}
        \sum_{t=t_0}^{T-1}\alpha_t\mathbb E\mathcal V_t,
        \notag\\
        \overline M_{t_0,T}
        &:=
        \frac{1}{A_{t_0,T}}
        \sum_{t=t_0}^{T-1}
        \alpha_t\mathbb E[\varepsilon_{\mathrm{mix}}(t)],
        \notag\\
        \overline M_{V,t_0,T}
        &:=
        \frac{1}{A_{t_0,T}}
        \sum_{t=t_0}^{T-1}
        \alpha_t
        \mathbb E[
        \varepsilon_{\mathrm{mix}}(t)^2\mathcal V_{t_\tau}],
        \notag\\
        \overline S_{t_0,T}
        &:=
        \frac{1}{A_{t_0,T}}
        \sum_{t=t_0}^{T-1}\alpha_t^2 .
        \label{eq:actor-average-error-defs-new}
    \end{align}
    Define also the averaged window-drift terms
    \begin{equation}\begin{aligned}
        \overline D_{\theta,t_0,T}^2
        &:=
        \frac{1}{A_{t_0,T}}
        \sum_{t=t_0}^{T-1}\alpha_t\mathbb E[D_{\theta,t}^2],
        \qquad
        \overline D_{w,t_0,T}^2
        :=
        \frac{1}{A_{t_0,T}}
        \sum_{t=t_0}^{T-1}\alpha_t\mathbb E[D_{w,t}^2],
        \\
        \overline D_{\theta,V,t_0,T}^2
        &:=
        \frac{1}{A_{t_0,T}}
        \sum_{t=t_0}^{T-1}
        \alpha_t\mathbb E[D_{\theta,t}^2\mathcal V_t].
        \label{eq:actor-average-window-drift-defs-new}
    \end{aligned}\end{equation}
    Define the averaged delayed-correlation correction
    \begin{equation}
        \overline R_{\mathrm{corr},t_0,T}
        :=
        \frac{1}{A_{t_0,T}}
        \sum_{t=t_0}^{T-1}
        \alpha_t\mathbb E[R_{\mathrm{corr},t}].
        \label{eq:actor-average-correlation-correction-def}
    \end{equation}
    Then
    \begin{align}
        \overline G_{t_0,T}
        &\leq
        \frac{2\bigl(J_{\max}-\mathbb E J(\theta_{t_0})\bigr)}
        {A_{t_0,T}}
        +
        2C_V^{\mathrm{act}}\overline V_{t_0,T}
        +
        2C_{\mathrm{FA}}^{\mathrm{act}}\varepsilon_{\mathrm{FA}}^2
        \notag\\
        &\quad
        +
        2C_{\mathrm{loc}}^{\mathrm{act}}\rho^{2(k+1)}
        +
        2C_{\mathrm{mix},0}^{\mathrm{act}}\overline M_{t_0,T}
        +
        2C_{\mathrm{mix},V}^{\mathrm{act}}\overline M_{V,t_0,T}
        +
        2C_{\mathrm{win},0}^{\mathrm{act}}
        \left(
        \overline D_{\theta,t_0,T}^2
        +
        \overline D_{w,t_0,T}^2
        \right)
        \notag\\
        &\quad
        +
        2C_{\mathrm{win},V}^{\mathrm{act}}
        \overline D_{\theta,V,t_0,T}^2
        +
        2\overline R_{\mathrm{corr},t_0,T}
        +
        2C_{\mathrm{var}}^{\mathrm{act}}\overline S_{t_0,T} .
        \label{eq:actor-stationarity-bound-new}
    \end{align}
\end{proposition}

\begin{proof}
    Take total expectation in Lemma~\ref{lem:actor-drift}, sum from $t=t_0$ to
    $T-1$, and rearrange.  Since $J(\theta_T)\leq J_{\max}$,
    \begin{align}
        \frac12
        \sum_{t=t_0}^{T-1}\alpha_t
        \mathbb E\|\nabla J(\theta_t)\|_2^2
        &\leq
        J_{\max}-\mathbb E J(\theta_{t_0})
        +
        C_V^{\mathrm{act}}
        \sum_{t=t_0}^{T-1}\alpha_t\mathbb E\mathcal V_t
        \notag\\
        &\quad
        +
        C_{\mathrm{FA}}^{\mathrm{act}}
        \varepsilon_{\mathrm{FA}}^2
        \sum_{t=t_0}^{T-1}\alpha_t
        +
        C_{\mathrm{loc}}^{\mathrm{act}}
        \rho^{2(k+1)}
        \sum_{t=t_0}^{T-1}\alpha_t
        \notag\\
        &\quad
        +
        C_{\mathrm{mix},0}^{\mathrm{act}}
        \sum_{t=t_0}^{T-1}
        \alpha_t\mathbb E[\varepsilon_{\mathrm{mix}}(t)]
        +
        C_{\mathrm{mix},V}^{\mathrm{act}}
        \sum_{t=t_0}^{T-1}
        \alpha_t
        \mathbb E[
        \varepsilon_{\mathrm{mix}}(t)^2\mathcal V_{t_\tau}]
        \notag\\
        &\quad
        +
        C_{\mathrm{win},0}^{\mathrm{act}}
        \sum_{t=t_0}^{T-1}\alpha_t
        \mathbb E\left[D_{\theta,t}^2+D_{w,t}^2\right]
        +
        C_{\mathrm{win},V}^{\mathrm{act}}
        \sum_{t=t_0}^{T-1}
        \alpha_t
        \mathbb E[D_{\theta,t}^2\mathcal V_t]
        \notag\\
        &\quad
        +
        \sum_{t=t_0}^{T-1}
        \alpha_t\mathbb E[R_{\mathrm{corr},t}]
        +
        C_{\mathrm{var}}^{\mathrm{act}}
        \sum_{t=t_0}^{T-1}\alpha_t^2 .
    \end{align}
    Dividing by $A_{t_0,T}/2$ gives \eqref{eq:actor-stationarity-bound-new}.
\end{proof}

\paragraph{Auxiliary comparison facts.}
The following two facts are not used to prove the unregularized stationarity
result above.  They are kept here only as comparison tools for later
regularized arguments.

\begin{lemma}[Average-reward unilateral performance difference]
    \label{lem:avg-reward-unilateral-performance-difference}
    Fix a product softmax policy $\pi_\theta$ and an agent $i$.  Let $\pi_i'$
    be any local policy in the closure of the local simplex and set
    $\widetilde\pi=(\pi_i',\pi_{\theta_{-i}})$.  Suppose both $\pi_\theta$ and
    $\widetilde\pi$ induce unique stationary distributions, denoted
    $d^\theta$ and $d^{(\pi_i',\theta_{-i})}$.  Define
    \begin{equation}
        \bar A_i^\theta(s_g,a_i)
        :=
        \sum_{a_{-i}}\pi_{\theta_{-i}}(a_{-i}|s_{-i})
        Q^\theta(s_g,a_i,a_{-i})
        -
        V^\theta(s_g),
        \label{eq:one-agent-averaged-advantage}
    \end{equation}
    where $Q^\theta=\sum_{j=1}^n Q_j^\theta$ is the global differential
    action-value function for $J$.  Then
    \begin{equation}
        J(\pi_i',\theta_{-i})-J(\theta)
        =
        \sum_{s_g}
        d^{(\pi_i',\theta_{-i})}(s_g)
        \sum_{a_i}\pi_i'(a_i|s_i)
        \bar A_i^\theta(s_g,a_i).
        \label{eq:avg-reward-unilateral-pdl}
    \end{equation}
\end{lemma}

\begin{proof}
    Write $\widetilde d=d^{(\pi_i',\theta_{-i})}$ and
    $\widetilde\pi(a_g|s_g)=\pi_i'(a_i|s_i)\pi_{\theta_{-i}}(a_{-i}|s_{-i})$.
    Since $\widetilde d$ is stationary under $\widetilde\pi$,
    \begin{align}
        J(\pi_i',\theta_{-i})-J(\theta)
        &=
        \sum_{s_g}\widetilde d(s_g)
        \sum_{a_g}\widetilde\pi(a_g|s_g)
        \left(r(s_g,a_g)-J(\theta)\right).
    \end{align}
    The Bellman relation
    \[
        Q^\theta(s_g,a_g)
        =
        r(s_g,a_g)-J(\theta)
        +
        \sum_{s_g'}P(s_g'|s_g,a_g)V^\theta(s_g')
    \]
    gives
    \[
        r(s_g,a_g)-J(\theta)
        =
        Q^\theta(s_g,a_g)
        -
        \sum_{s_g'}P(s_g'|s_g,a_g)V^\theta(s_g').
    \]
    Substitution and stationarity of $\widetilde d$ make the transition term
    equal to $\sum_{s_g}\widetilde d(s_g)V^\theta(s_g)$.  Thus
    \begin{align}
        J(\pi_i',\theta_{-i})-J(\theta)
        &=
        \sum_{s_g}\widetilde d(s_g)
        \left[
        \sum_{a_g}\widetilde\pi(a_g|s_g)Q^\theta(s_g,a_g)
        -
        V^\theta(s_g)
        \right]  \\
        &=
        \sum_{s_g}\widetilde d(s_g)
        \sum_{a_i}\pi_i'(a_i|s_i)\bar A_i^\theta(s_g,a_i),
    \end{align}
    which proves the claim.
\end{proof}

\begin{lemma}[Nonuniform Nash-gap domination by the softmax gradient]
    \label{lem:nash-gap-gradient-domination}
    Define
    \begin{equation}
        \mathrm{NashGap}_i(\theta)
        :=
        \sup_{\pi_i'\in\overline\Pi_i}J(\pi_i',\theta_{-i})-J(\theta).
    \end{equation}
    Let $\pi_i^*(\theta_{-i})$ be a best response in $\overline\Pi_i$ and
    write $d_i^*(s_g):=d^{(\pi_i^*(\theta_{-i}),\theta_{-i})}(s_g)$.  Suppose
    there are a local function
    $\bar A_{i,\mathrm{loc}}^\theta:\mathcal S_i\times\mathcal A_i\to\mathbb R$
    and a constant $\delta_i^{\mathrm{adv}}(\theta)\ge0$ such that
    \begin{equation}
        \sup_{s_g,a_i}
        \left|
        \bar A_i^\theta(s_g,a_i)
        -
        \bar A_{i,\mathrm{loc}}^\theta(s_i,a_i)
        \right|
        \leq
        \delta_i^{\mathrm{adv}}(\theta).
        \label{eq:advantage-localization-error-for-domination}
    \end{equation}
    Define the coverage constants
    \begin{align}
        \zeta_i(\theta)
        &:=
        \min_{s_g:d_i^*(s_g)>0}
        \frac{d^\theta(s_g)}{d_i^*(s_g)}, \\
        k_i(\theta)
        &:=
        \min_{\substack{s_g,a_i:\ d_i^*(s_g)>0,\\
        \pi_i^*(a_i|s_i)>0}}
        \frac{\pi_{\theta_i}(a_i|s_i)}
        {\pi_i^*(a_i|s_i)} ,
    \end{align}
    and let
    $D_\theta:=\sum_i|\mathcal S_i||\mathcal A_i|$.  Then
    \begin{equation}
        \sum_{i=1}^n
        \zeta_i(\theta)k_i(\theta)\mathrm{NashGap}_i(\theta)
        \leq
        \sqrt{D_\theta}\|\nabla J(\theta)\|_2
        +
        2\sum_{i=1}^n\delta_i^{\mathrm{adv}}(\theta).
        \label{eq:weighted-nash-gap-domination}
    \end{equation}
\end{lemma}

\begin{proof}
    Fix $i$.  Lemma~\ref{lem:avg-reward-unilateral-performance-difference}
    applied to the best response gives
    \[
        \mathrm{NashGap}_i(\theta)
        =
        \sum_{s_g,a_i}
        d_i^*(s_g)\pi_i^*(a_i|s_i)
        \bar A_i^\theta(s_g,a_i).
    \]
    Using \eqref{eq:advantage-localization-error-for-domination} and
    $|\sum p_x e_x|\le\sup_x|e_x|$,
    \[
        \mathrm{NashGap}_i(\theta)
        \leq
        \sum_{s_g,a_i}
        d_i^*(s_g)\pi_i^*(a_i|s_i)
        |\bar A_{i,\mathrm{loc}}^\theta(s_i,a_i)|
        +
        \delta_i^{\mathrm{adv}}(\theta).
    \]
    Multiplying by $\zeta_i(\theta)k_i(\theta)$ and using the definitions
    of the coverage ratios yields
    \[
        \zeta_i(\theta)k_i(\theta)\mathrm{NashGap}_i(\theta)
        \leq
        \sum_{s_g,a_i}
        d^\theta(s_g)\pi_{\theta_i}(a_i|s_i)
        |\bar A_{i,\mathrm{loc}}^\theta(s_i,a_i)|
        +
        \delta_i^{\mathrm{adv}}(\theta).
    \]
    The softmax policy-gradient identity gives, coordinatewise,
    \[
        \frac{\partial J(\theta)}{\partial\theta_i(s_i,a_i)}
        =
        \sum_{s_{-i}}
        d^\theta(s_g)\pi_{\theta_i}(a_i|s_i)
        \bar A_i^\theta(s_g,a_i).
    \]
    Replacing $\bar A_i^\theta$ by
    $\bar A_{i,\mathrm{loc}}^\theta$ changes the $\ell_1$ norm of the vector of
    coordinates by at most $\delta_i^{\mathrm{adv}}(\theta)$.  Hence
    \[
        \zeta_i(\theta)k_i(\theta)\mathrm{NashGap}_i(\theta)
        \leq
        \|\nabla_{\theta_i}J(\theta)\|_1
        +
        2\delta_i^{\mathrm{adv}}(\theta).
    \]
    Summing over $i$ and using
    $\|\nabla J(\theta)\|_1\le\sqrt{D_\theta}\|\nabla J(\theta)\|_2$ proves
    \eqref{eq:weighted-nash-gap-domination}.
\end{proof}

\subsection{Overall Convergence}\label{subsection:unregularized_overall_convergence}

We now combine the moving-target critic recursion with the actor stationarity
bound.  The result is finite-time and is stated for the unregularized objective
only in terms of policy-gradient stationarity.  It does not assert global
optimality.

\begin{theorem}[Finite-time unregularized policy-gradient stationarity]
    \label{thm:unreg-finite-time-stationarity}
    Suppose Assumptions~\ref{assump:ergodicity}-\ref{assump:actor-spatial-decay} hold, and use Lemma~\ref{lem:lipschitzness}, \eqref{eq:actor-BW-def-new}, and
    \eqref{eq:actor-fa-error-new}.
    Define
    \begin{equation}
        \lambda_\chi:=\frac{\Delta_{\min}^{\chi}}{4}.
        \label{eq:overall-lambda-chi-def}
    \end{equation}
    Then the following two stepsize regimes give finite-time stationarity
    bounds.
    \begin{enumerate}
        \item \textbf{Fixed stepsizes.}
        Let $\alpha_t=\alpha\le1$ and $\beta_t=\beta>0$ on
        $t=0,\ldots,T-1$.  Use one fixed actor/critic proof window
        $L\ge1$, assume the corresponding bounded-increment and
        fixed-window moving-target critic premises from
        Lemma~\ref{lem:moving_actor_constant_beta}, and define
        \begin{equation}
            \overline G_T
            :=
            \frac1T\sum_{t=0}^{T-1}
            \mathbb E\|\nabla J(\theta_t)\|_2^2 .
            \label{eq:overall-gradient-average-def}
        \end{equation}
        Then
        $\min_{0\le t<T}\mathbb E\|\nabla J(\theta_t)\|_2^2\le \overline G_T$.
        Moreover, with
        \begin{equation}
            \delta_{\alpha,\beta}:=\max\{\alpha,\beta\},
            \qquad
            L_{\alpha,\beta}
            :=
            \left\lceil
            \frac{\log(1/\delta_{\alpha,\beta})}
            {|\log\rho_{\mathrm{mix}}|}
            \right\rceil ,
            \label{eq:stepsize-dependent-window-def}
        \end{equation}
        if $0<\delta_{\alpha,\beta}\le e^{-1}$,
        $L=L_{\alpha,\beta}$, and
        $\alpha L_{\alpha,\beta}\le1$, $\beta L_{\alpha,\beta}\le1$, then
        \begin{align}
            \overline G_T
            =
            O\left(
            \frac1{\alpha T}
            +
            \frac1{\beta T}
            +
            \beta\log(1/\delta_{\alpha,\beta})
            +
            \alpha\log^2(1/\delta_{\alpha,\beta})
            +
            \frac{\alpha^2}{\beta^2}
            +
            \varepsilon_{\mathrm{FA}}^2
            +
            \rho^{2(k+1)}
            \right).
            \label{eq:finite-time-unreg-gradient-bound-stepsize-window}
        \end{align}

        \item \textbf{Diminishing stepsizes.}
        For later use in the regularized analysis, fix exponents
        $\mathfrak b,\mathfrak a$ satisfying
        \begin{equation}
            \frac12<\mathfrak b<1,
            \qquad
            \frac{1+\mathfrak b}{2}<\mathfrak a<1,
            \label{eq:overall-decaying-exponent-region}
        \end{equation}
        and set
        \begin{equation}
            \beta_t:=\frac{\beta_0}{(t+1)^{\mathfrak b}},
            \qquad
            \alpha_t:=\frac{\alpha_0}{(t+1)^{\mathfrak a}}.
            \label{eq:overall-decaying-stepsizes}
        \end{equation}
        Use the logarithmic window
        \begin{equation}
            \tau_t:=\left\lceil c_\tau\log(t+1)\right\rceil,
            \qquad
            c_\tau>\frac{\mathfrak b}{-\log\rho_{\mathrm{mix}}},
            \label{eq:overall-decaying-window}
        \end{equation}
        Let $t_0$ be any finite burn-in time after which the hypotheses of
        Corollary~\ref{cor:moving_actor_time_varying_beta} and the actor-side
        bounded-increment premises hold for
        \eqref{eq:overall-decaying-stepsizes}--\eqref{eq:overall-decaying-window}.
        For $T>t_0$, define
        \begin{equation}
            A_{t_0,T}:=\sum_{t=t_0}^{T-1}\alpha_t,
            \qquad
            \overline G_{t_0,T}^{\alpha}
            :=
            \frac{1}{A_{t_0,T}}
            \sum_{t=t_0}^{T-1}
            \alpha_t\mathbb E\|\nabla J(\theta_t)\|_2^2 .
            \label{eq:overall-decaying-gradient-average-def}
        \end{equation}
        Then
        \begin{equation}
            \min_{t_0\le t<T}
            \mathbb E\|\nabla J(\theta_t)\|_2^2
            \le
            \overline G_{t_0,T}^{\alpha}
            =
            O\left(
            T^{-(1-\mathfrak a)}
            +
            \varepsilon_{\mathrm{FA}}^2
            +
            \rho^{2(k+1)}
            \right),
            \label{eq:finite-time-unreg-gradient-bound-decaying}
        \end{equation}
        In particular, taking
        $\mathfrak b=1/2+\mathfrak c$ and
        $\mathfrak a=3/4+\mathfrak c$ for any
        $0<\mathfrak c<1/4$ gives
        $O(T^{-1/4+\mathfrak c}
        +\varepsilon_{\mathrm{FA}}^2+\rho^{2(k+1)})$.
    \end{enumerate}
    The hidden constants in both big-$O$ bounds are independent of
    $T,k,\varepsilon_{\mathrm{FA}}$, and in the fixed-stepsize case also of
    $\alpha,\beta$.
\end{theorem}

\begin{proof}
    \leavevmode\paragraph{Fixed stepsizes.}
    Set
    \begin{equation}
        \tau_t:=\min\{t,L\},
        \label{eq:overall-fixed-window-def}
    \end{equation}
    and interpret the moving-target critic recursion with the full fixed
    window, $\tau(\beta)=L$, once $t\ge L$.  The actor increment bound below follows from
    Lemma~\ref{lem:bounded-actor-increment}, which uses
    Assumption~\ref{assump:bounded_critic} and
    \eqref{eq:actor-score-bound-new}; the critic increment bound is the
    fixed-window premise used in the overall combination step:
    \begin{equation}
        \|\theta_{t+1}-\theta_t\|_2\le C_\theta\alpha,
        \qquad
        \|w_i^{t+1}-w_i^t\|_2\le C_{w,i}\beta,
        \qquad
        0\le t<T,\quad i=1,\ldots,n,
        \label{eq:overall-increment-premise}
    \end{equation}
    and we write
    \begin{equation}
        C_w:=\left(\sum_{i=1}^n C_{w,i}^2\right)^{1/2}.
        \label{eq:overall-Cw-def}
    \end{equation}
    By Lemmas~\ref{lem:moving_actor_constant_beta} and \ref{lem:lipschitzness} under
    Assumptions~\ref{assump:ergodicity}, \ref{assump:uniform_mixing},
    \ref{assump:bounded_critic}, and the smallness rules
    \eqref{eq:moving-beta-rules}, the fixed-window moving-target critic
    estimate used below is
    \begin{equation}
        \mathbb E[\mathcal V_{t+1}]
        \le
        \left(1-\lambda_\chi\beta\right)
        \mathbb E[\mathcal V_t]
        +
        C_{\mathrm{sq}}\beta^2
        +
        C_{\mathrm{mov}}\beta\varepsilon_{P,\beta,t}^{\mathrm{mov}}
        +
        C_{\mathrm{tar}}C_\theta^2\frac{\alpha^2}{\beta},
        \qquad L\le t<T.
        \label{eq:overall-critic-recursion-premise}
    \end{equation}
    Define the fixed-window actor- and critic-mixing envelopes
    \begin{align}
        \mathfrak e_{\mathrm{act}}(L,T)
        &:=
        \frac{L}{T}
        +
        m_{\mathrm{mix}}\rho_{\mathrm{mix}}^L
        +
        \frac12L_KC_\theta\alpha L(L+1)
        +
        L_\nu C_\theta\alpha L,
        \label{eq:overall-actor-mixing-envelope}
        \\
        \mathfrak e_{\mathrm{crit}}(L)
        &:=
        m_{\mathrm{mix}}\rho_{\mathrm{mix}}^{L-1}
        +
        \beta L
        +
        C_\theta(1+L_\nu)\alpha L
        +
        L_KC_\theta\alpha L^2 .
        \label{eq:overall-critic-mixing-envelope}
    \end{align}
    Let $v_t:=\mathbb E\mathcal V_t$.  By
    Assumption~\ref{assump:bounded_critic}, Lemma~\ref{lem:lipschitzness},
    bounded rewards, and the finite initialization of the average-reward
    estimates, there is a constant $V_{\max}<\infty$ such that
    \begin{equation}
        v_t\le V_{\max},
        \qquad t\ge0.
        \label{eq:overall-fixed-window-Vmax}
    \end{equation}
    Define also
    \begin{equation}
        \mathfrak V_{L,T}
        :=
        \frac{LV_{\max}}{T}
        +
        \frac{V_{\max}}{\lambda_\chi\beta T}
        +
        \frac{C_{\mathrm{sq}}\beta}{\lambda_\chi}
        +
        \frac{C_{\mathrm{mov}}\mathfrak e_{\mathrm{crit}}(L)}
        {\lambda_\chi}
        +
        \frac{C_{\mathrm{tar}}C_\theta^2}{\lambda_\chi}
        \frac{\alpha^2}{\beta^2}.
        \label{eq:overall-critic-average-envelope}
    \end{equation}

    Define
    \begin{equation}
        d_t
        :=
        C_{\mathrm{sq}}\beta^2
        +
        C_{\mathrm{mov}}\beta\varepsilon_{P,\beta,t}^{\mathrm{mov}}
        +
        C_{\mathrm{tar}}C_\theta^2\frac{\alpha^2}{\beta}.
        \label{eq:overall-critic-recursion-noise}
    \end{equation}
    The critic recursion premise \eqref{eq:overall-critic-recursion-premise}
    gives, for $L\le t<T$,
    \begin{equation}
        v_{t+1}\le(1-\lambda_\chi\beta)v_t+d_t.
        \label{eq:overall-critic-recursion-short}
    \end{equation}
    Rearranging,
    $\lambda_\chi\beta v_t\le v_t-v_{t+1}+d_t$ for $L\le t<T$.
    Splitting off the initial window and summing the recursion only from
    $t=L$ to $T-1$ yields
    \[
        \frac1T\sum_{t=0}^{T-1}v_t
        \le
        \frac{LV_{\max}}{T}
        +
        \frac{V_{\max}}{\lambda_\chi\beta T}
        +
        \frac{1}{\lambda_\chi\beta T}
        \sum_{t=L}^{T-1}d_t ,
    \]
    with the last sum interpreted as zero if $T\le L$.  By the definition of
    $d_t$, this is
    \begin{equation}
        \frac1T\sum_{t=0}^{T-1}\mathbb E\mathcal V_t
        \le
        \frac{LV_{\max}}{T}
        +
        \frac{V_{\max}}{\lambda_\chi\beta T}
        +
        \frac{C_{\mathrm{sq}}\beta}{\lambda_\chi}
        +
        \frac{C_{\mathrm{mov}}}{\lambda_\chi}
        \frac1T\sum_{t=L}^{T-1}
        \varepsilon_{P,\beta,t}^{\mathrm{mov}}
        +
        \frac{C_{\mathrm{tar}}C_\theta^2}{\lambda_\chi}
        \frac{\alpha^2}{\beta^2}.
        \label{eq:finite-time-critic-average-bound}
    \end{equation}
    From the definition of the moving-target error in
    Lemma~\ref{lem:moving_actor_constant_beta}, with $\tau(\beta)=L$ and
    $A_t^\beta=\sum_{\ell=t-L}^{t-1}\alpha\le\alpha L$,
    \begin{equation}
        \varepsilon_{P,\beta,t}^{\mathrm{mov}}
        \le
        m_{\mathrm{mix}}\rho_{\mathrm{mix}}^{L-1}
        +
        \beta L
        +
        C_\theta(1+L_\nu)\alpha L
        +
        L_KC_\theta\alpha L^2
        =
        \mathfrak e_{\mathrm{crit}}(L).
        \label{eq:overall-critic-mixing-envelope-proof}
    \end{equation}
    Therefore \eqref{eq:finite-time-critic-average-bound} gives
    \begin{equation}
        \frac1T\sum_{t=0}^{T-1}\mathbb E\mathcal V_t
        \le
        \mathfrak V_{L,T}.
        \label{eq:overall-Vbar-envelope-proof}
    \end{equation}

    We next bound the actor-side averages in
    Proposition~\ref{prop:actor-stationarity}.  If $t<L$, then
    $\varepsilon_{\mathrm{mix}}(t)\le1$ by the clipped definition
    \eqref{eq:epsilon-mix-definition}.  If $t\ge L$, then
    Lemma~\ref{lem:conditional-markovian-mixing-error}, specifically
    \eqref{eq:actor-epsilon-mix-raw-definition}, the increment bound
    \eqref{eq:overall-increment-premise}, and
    the fixed-window mixing bound imply
    \begin{align}
        \varepsilon_{\mathrm{mix}}(t)
        &\le
        m_{\mathrm{mix}}\rho_{\mathrm{mix}}^L
        +
        L_K
        \sum_{\ell=t-L}^{t-1}
        \|\vartheta_\ell-\theta_{t-L}\|_2
        +
        L_\nu C_\theta\alpha L
        \notag\\
        &\le
        m_{\mathrm{mix}}\rho_{\mathrm{mix}}^L
        +
        \frac12L_KC_\theta\alpha L(L+1)
        +
        L_\nu C_\theta\alpha L .
        \label{eq:overall-actor-late-mixing-bound}
    \end{align}
    Hence
    \begin{equation}
        \frac1T\sum_{t=0}^{T-1}
        \mathbb E[\varepsilon_{\mathrm{mix}}(t)]
        \le
        \mathfrak e_{\mathrm{act}}(L,T).
        \label{eq:overall-actor-mixing-average-proof}
    \end{equation}
    Since $t_\tau=0$ for $t<L$, and since
    \eqref{eq:overall-actor-late-mixing-bound} is at most
    $\mathfrak e_{\mathrm{act}}(L,T)$, we also have
    \begin{equation}
        \frac1T\sum_{t=0}^{T-1}
        \mathbb E[
        \varepsilon_{\mathrm{mix}}(t)^2\mathcal V_{t_\tau}]
        \le
        \frac{L}{T}\mathbb E\mathcal V_0
        +
        \mathfrak e_{\mathrm{act}}(L,T)^2\mathfrak V_{L,T}.
        \label{eq:overall-mixed-tracking-envelope-proof}
    \end{equation}
    By the window-drift definitions in \eqref{eq:actor-window-drift-defs}, the
    increment bounds give, pathwise,
    \begin{equation}
        D_{\theta,t}\le C_\theta\alpha L,
        \qquad
        D_{w,t}\le C_w\beta L .
        \label{eq:overall-window-pathwise-proof}
    \end{equation}
    Combining \eqref{eq:overall-window-pathwise-proof} with
    \eqref{eq:overall-Vbar-envelope-proof} yields
    \begin{align}
        \frac1T\sum_{t=0}^{T-1}
        \mathbb E[D_{\theta,t}^2+D_{w,t}^2]
        &\le
        L^2(C_\theta^2\alpha^2+C_w^2\beta^2),
        \label{eq:overall-window-average-proof}\\
        \frac1T\sum_{t=0}^{T-1}
        \mathbb E[D_{\theta,t}^2\mathcal V_t]
        &\le
        C_\theta^2\alpha^2L^2\mathfrak V_{L,T}.
        \label{eq:overall-window-tracking-proof}
    \end{align}
    Finally, by the delayed-correlation remainder from
    Lemma~\ref{lem:actor-gradient-error}, namely
    \eqref{eq:actor-correlation-remainder},
    \eqref{eq:overall-window-pathwise-proof}, and Jensen's inequality,
    \begin{align}
        \frac1T\sum_{t=0}^{T-1}\mathbb E[R_{\mathrm{corr},t}]
        &\le
        L_{\nabla J}C_\theta\alpha L
        \left(
        B_{\tau,0}
        +
        B_{\tau,w}C_w\beta L
        +
        B_{\tau,V}
        \frac1T\sum_{t=0}^{T-1}\mathbb E\sqrt{\mathcal V_t}
        \right)
        \notag\\
        &\le
        L_{\nabla J}C_\theta\alpha L
        \left(
        B_{\tau,0}
        +
        B_{\tau,w}C_w\beta L
        +
        B_{\tau,V}\sqrt{\mathfrak V_{L,T}}
        \right).
        \label{eq:overall-correlation-envelope-proof}
    \end{align}

    Proposition~\ref{prop:actor-stationarity} with $t_0=0$ and
    $\alpha_t=\alpha$ gives $A_{0,T}=\alpha T$ and
    $\overline S_{0,T}=\alpha$.  Therefore,
    \begin{align}
        \overline G_T
        &\le
        \frac{2\bigl(J_{\max}-\mathbb E J(\theta_0)\bigr)}{\alpha T}
        +
        2C_V^{\mathrm{act}}
        \frac1T\sum_{t=0}^{T-1}\mathbb E\mathcal V_t
        \notag\\
        &\quad
        +
        2C_{\mathrm{FA}}^{\mathrm{act}}\varepsilon_{\mathrm{FA}}^2
        +
        2C_{\mathrm{loc}}^{\mathrm{act}}\rho^{2(k+1)}
        +
        2C_{\mathrm{mix},0}^{\mathrm{act}}
        \frac1T\sum_{t=0}^{T-1}
        \mathbb E[\varepsilon_{\mathrm{mix}}(t)]
        +
        2C_{\mathrm{mix},V}^{\mathrm{act}}
        \frac1T\sum_{t=0}^{T-1}
        \mathbb E[
        \varepsilon_{\mathrm{mix}}(t)^2\mathcal V_{t_\tau}]
        \notag\\
        &\quad
        +
        2C_{\mathrm{win},0}^{\mathrm{act}}
        \frac1T\sum_{t=0}^{T-1}
        \mathbb E[D_{\theta,t}^2+D_{w,t}^2]
        +
        2C_{\mathrm{win},V}^{\mathrm{act}}
        \frac1T\sum_{t=0}^{T-1}
        \mathbb E[D_{\theta,t}^2\mathcal V_t]
        \notag\\
        &\quad
        +
        \frac2T\sum_{t=0}^{T-1}
        \mathbb E[R_{\mathrm{corr},t}]
        +
        2C_{\mathrm{var}}^{\mathrm{act}}\alpha .
        \label{eq:finite-time-actor-before-substitution}
    \end{align}
    Substituting \eqref{eq:overall-Vbar-envelope-proof},
    \eqref{eq:overall-actor-mixing-average-proof},
    \eqref{eq:overall-mixed-tracking-envelope-proof},
    \eqref{eq:overall-window-average-proof},
    \eqref{eq:overall-window-tracking-proof}, and
    \eqref{eq:overall-correlation-envelope-proof} into
    \eqref{eq:finite-time-actor-before-substitution} gives the detailed
    fixed-window bound
    \begin{align}
        \overline G_T
        &\le
        \frac{2\bigl(J_{\max}-\mathbb E J(\theta_0)\bigr)}{\alpha T}
        +
        2C_V^{\mathrm{act}}\mathfrak V_{L,T}
        \notag\\
        &\quad
        +
        2C_{\mathrm{FA}}^{\mathrm{act}}\varepsilon_{\mathrm{FA}}^2
        +
        2C_{\mathrm{loc}}^{\mathrm{act}}\rho^{2(k+1)}
        +
        2C_{\mathrm{mix},0}^{\mathrm{act}}
        \mathfrak e_{\mathrm{act}}(L,T)
        \notag\\
        &\quad
        +
        2C_{\mathrm{mix},V}^{\mathrm{act}}
        \left(
        \frac{L}{T}\mathbb E\mathcal V_0
        +
        \mathfrak e_{\mathrm{act}}(L,T)^2\mathfrak V_{L,T}
        \right)
        \notag\\
        &\quad
        +
        2C_{\mathrm{win},0}^{\mathrm{act}}
        L^2
        \left(C_\theta^2\alpha^2+C_w^2\beta^2\right)
        +
        2C_{\mathrm{win},V}^{\mathrm{act}}
        C_\theta^2\alpha^2L^2\mathfrak V_{L,T}
        \notag\\
        &\quad
        +
        2L_{\nabla J}C_\theta\alpha L
        \left(
        B_{\tau,0}
        +
        B_{\tau,w}C_w\beta L
        \right)
        \notag\\
        &\quad
        +
        2L_{\nabla J}C_\theta\alpha L
        B_{\tau,V}\sqrt{\mathfrak V_{L,T}}
        +
        2C_{\mathrm{var}}^{\mathrm{act}}\alpha .
        \label{eq:finite-time-unreg-gradient-bound}
    \end{align}
    The fixed-step best-iterate claim follows because the minimum is at most
    the average.

    It remains only to justify the displayed scaling.  By definition of
    $\mathfrak V_{L,T}$,
    \[
        \mathfrak V_{L,T}
        =
        O\left(
        \frac{L}{T}
        +
        \frac1{\beta T}
        +
        \mathfrak e_{\mathrm{crit}}(L)
        +
        \frac{\alpha^2}{\beta^2}
        \right).
    \]
    Under $\alpha L\le1$ and $\beta L\le1$, the mixed-tracking,
    window-drift, delayed-correlation, and variance terms in
    \eqref{eq:finite-time-unreg-gradient-bound} are no larger than
    \[
        \frac1{\alpha T}
        +
        \frac1{\beta T}
        +
        \frac{L}{T}
        +
        \mathfrak e_{\mathrm{crit}}(L)
        +
        \mathfrak e_{\mathrm{act}}(L,T)
        +
        \frac{\alpha^2}{\beta^2}
        +
        \varepsilon_{\mathrm{FA}}^2
        +
        \rho^{2(k+1)}
    \]
    after increasing the hidden constant.  Now set $L=L_{\alpha,\beta}$ as in
    \eqref{eq:stepsize-dependent-window-def}.  This choice gives
    $L_{\alpha,\beta}=O(\log(1/\delta_{\alpha,\beta}))$,
    $\rho_{\mathrm{mix}}^{L_{\alpha,\beta}}\le\delta_{\alpha,\beta}$, and
    $\rho_{\mathrm{mix}}^{L_{\alpha,\beta}-1}
    \le\rho_{\mathrm{mix}}^{-1}\delta_{\alpha,\beta}$.  Since
    $\delta_{\alpha,\beta}\le e^{-1}$, the two geometric-mixing tails are
    bounded, up to constants, by
    \[
        \beta\log(1/\delta_{\alpha,\beta})
        +
        \alpha\log^2(1/\delta_{\alpha,\beta}).
    \]
    Also, $L_{\alpha,\beta}/T$ is bounded by a constant multiple of
    $1/(\alpha T)+1/(\beta T)$.  Substituting $L=L_{\alpha,\beta}$ proves
    \eqref{eq:finite-time-unreg-gradient-bound-stepsize-window}.

    \leavevmode\paragraph{Diminishing stepsizes.}
    Since
    $\mathfrak b<1$, the exponent condition
    \eqref{eq:overall-decaying-exponent-region} implies
    $\mathfrak a>\mathfrak b$.  Also,
    \begin{equation}
        A_{t_0,T}
        =
        \sum_{t=t_0}^{T-1}\alpha_t
        =
        \Theta(T^{1-\mathfrak a}).
        \label{eq:overall-decaying-A-order}
    \end{equation}
    The actor part of the following bounded-increment premise is the
    decaying-stepsize specialization of
    Lemma~\ref{lem:bounded-actor-increment}; the critic part is the
    corresponding post-burn-in premise required to couple
    Corollary~\ref{cor:moving_actor_time_varying_beta} with the actor
    stationarity result:
    \begin{equation}
        \|\theta_{t+1}-\theta_t\|_2\le C_\theta\alpha_t,
        \qquad
        \|w_i^{t+1}-w_i^t\|_2\le C_{w,i}\beta_t,
        \qquad
        t_0\le t<T,\quad i=1,\ldots,n.
        \label{eq:overall-decaying-increment-premise}
    \end{equation}
    For $t\ge t_0$, Corollary~\ref{cor:moving_actor_time_varying_beta}, whose
    assumptions are those of Lemma~\ref{lem:moving_actor_constant_beta} with
    the decaying-step local smallness conditions, gives
    \begin{equation}
        v_{t+1}
        \le
        (1-\lambda_\chi\beta_t)v_t
        +
        d_t^{\mathrm{dec}},
        \qquad
        d_t^{\mathrm{dec}}
        :=
        C_{\mathrm{sq}}\beta_t^2
        +
        C_{\mathrm{mov}}\beta_t\varepsilon_{P,t}^{\mathrm{mov}}
        +
        C_{\mathrm{tar}}C_\theta^2\frac{\alpha_t^2}{\beta_t}.
        \label{eq:overall-decaying-critic-recursion-proof}
    \end{equation}
    Since $\tau_t=O(\log(t+1))=o(t)$, the stepsizes are comparable inside the
    window:
    $\beta_\ell=O(\beta_t)$ and $\alpha_\ell=O(\alpha_t)$ for
    $\ell\in\{t-\tau_t,\ldots,t-1\}$.  The choice of $c_\tau$ in
    \eqref{eq:overall-decaying-window} also gives
    $\rho_{\mathrm{mix}}^{\tau_t}=O(\beta_t)$.  Hence the moving-target
    error defined in Corollary~\ref{cor:moving_actor_time_varying_beta}
    satisfies
    \begin{equation}
        \varepsilon_{P,t}^{\mathrm{mov}}
        =
        O\left(
        \beta_t\log(t+1)
        +
        \alpha_t\log^2(t+1)
        \right).
        \label{eq:overall-decaying-moving-error-order}
    \end{equation}
    Rearranging \eqref{eq:overall-decaying-critic-recursion-proof} yields
    $\lambda_\chi\beta_t v_t\le v_t-v_{t+1}+d_t^{\mathrm{dec}}$.  Let
    $r_t:=\alpha_t/\beta_t$.  Under
    \eqref{eq:overall-decaying-stepsizes},
    $r_t=(\alpha_0/\beta_0)(t+1)^{-(\mathfrak a-\mathfrak b)}$ is
    nonincreasing.  Therefore,
    \begin{align}
        \lambda_\chi
        \sum_{t=t_0}^{T-1}\alpha_t v_t
        &\le
        \sum_{t=t_0}^{T-1}r_t(v_t-v_{t+1})
        +
        \sum_{t=t_0}^{T-1}r_td_t^{\mathrm{dec}}
        \notag\\
        &\le
        r_{t_0}v_{t_0}
        +
        \sum_{t=t_0}^{T-1}
        \left(
        C_{\mathrm{sq}}\alpha_t\beta_t
        +
        C_{\mathrm{mov}}\alpha_t\varepsilon_{P,t}^{\mathrm{mov}}
        +
        C_{\mathrm{tar}}C_\theta^2
        \frac{\alpha_t^3}{\beta_t^2}
        \right).
        \label{eq:overall-decaying-weighted-critic-sum}
    \end{align}
    Under the reference exponent region
    \eqref{eq:overall-decaying-exponent-region}, the last sum is finite
    uniformly in $T$.  Indeed,
    \begin{align*}
        \alpha_t\beta_t
        &=
        O((t+1)^{-(\mathfrak a+\mathfrak b)}),\\
        \alpha_t\varepsilon_{P,t}^{\mathrm{mov}}
        &=
        O((t+1)^{-(\mathfrak a+\mathfrak b)}\log(t+1))
        +
        O((t+1)^{-2\mathfrak a}\log^2(t+1)),\\
        \frac{\alpha_t^3}{\beta_t^2}
        &=
        O((t+1)^{-(3\mathfrak a-2\mathfrak b)}).
    \end{align*}
    These sequences are summable because
    $\mathfrak b>1/2$,
    $\mathfrak a>\mathfrak b$,
    $\mathfrak a>(1+\mathfrak b)/2$, and $\mathfrak b<1$ imply
    $\mathfrak a+\mathfrak b>1$,
    $2\mathfrak a>1$, and $3\mathfrak a-2\mathfrak b>1$.

    The same window comparison gives
    \eqref{eq:overall-decaying-moving-error-order}, and therefore
    \begin{align*}
        d_t^{\mathrm{dec}}
        &=
        O\left(
        \beta_t^2\log(t+1)
        +
        \alpha_t\beta_t\log^2(t+1)
        +
        \frac{\alpha_t^2}{\beta_t}
        \right),\\
        \frac{d_t^{\mathrm{dec}}}{\beta_t}
        &=
        O\left(
        \beta_t\log(t+1)
        +
        \alpha_t\log^2(t+1)
        +
        \frac{\alpha_t^2}{\beta_t^2}
        \right)
        =
        o(1).
    \end{align*}
    Hence the critic recursion also yields
    $\sup_{t\ge t_0}v_t<\infty$.  To see this, choose $B<\infty$ and enlarge
    $t_0$ so that
    $d_t^{\mathrm{dec}}\le(\lambda_\chi/2)\beta_tB$ for all $t\ge t_0$.
    If $v_t>B$, then
    $v_{t+1}\le(1-\lambda_\chi\beta_t)v_t
    +(\lambda_\chi/2)\beta_tB\le v_t$; if $v_t\le B$, then
    $v_{t+1}\le B+\sup_{\ell\ge t_0}d_\ell^{\mathrm{dec}}$.  This gives a
    finite uniform bound after burn-in.

    Since $\mathfrak a>\mathfrak b$, $r_t=\alpha_t/\beta_t$ is nonincreasing,
    so the weighted
    critic inequality \eqref{eq:overall-decaying-weighted-critic-sum} applies.
    The summability estimates above and
    \eqref{eq:overall-decaying-A-order} give
    \begin{equation}
        \frac{1}{A_{t_0,T}}
        \sum_{t=t_0}^{T-1}\alpha_t\mathbb E\mathcal V_t
        =
        O\left(T^{-(1-\mathfrak a)}\right).
        \label{eq:overall-decaying-weighted-critic-average}
    \end{equation}

    It remains to bound the actor-side averages in
    Proposition~\ref{prop:actor-stationarity}.  By
    Lemma~\ref{lem:conditional-markovian-mixing-error} and
    \eqref{eq:actor-epsilon-mix-raw-definition}, with the common logarithmic
    window and the increment premise
    \eqref{eq:overall-decaying-increment-premise},
    \begin{equation}
        \varepsilon_{\mathrm{mix}}(t)
        =
        O\left(
        \beta_t
        +
        \alpha_t\log^2(t+1)
        \right),
        \label{eq:overall-decaying-actor-mixing-order}
    \end{equation}
    and hence
    \[
        \frac1{A_{t_0,T}}
        \sum_{t=t_0}^{T-1}
        \alpha_t\mathbb E[\varepsilon_{\mathrm{mix}}(t)]
        =
        O\left(T^{-(1-\mathfrak a)}\right).
    \]
    Since $v_t$ is uniformly bounded and
    $\varepsilon_{\mathrm{mix}}(t)\le1$, the stronger finite-sum bound following
    from \eqref{eq:overall-decaying-actor-mixing-order} also gives
    \[
        \frac1{A_{t_0,T}}
        \sum_{t=t_0}^{T-1}
        \alpha_t
        \mathbb E[
        \varepsilon_{\mathrm{mix}}(t)^2\mathcal V_{t_\tau}]
        =
        O\left(T^{-(1-\mathfrak a)}\right).
    \]
    By the definitions of $D_{\theta,t}$ and $D_{w,t}$ in
    \eqref{eq:actor-window-drift-defs}, the window-drift quantities obey
    \[
        D_{\theta,t}=O(\alpha_t\log(t+1)),
        \qquad
        D_{w,t}=O(\beta_t\log(t+1)),
    \]
    so
    \[
        \overline D_{\theta,t_0,T}^2
        +
        \overline D_{w,t_0,T}^2
        +
        \overline D_{\theta,V,t_0,T}^2
        =
        O\left(T^{-(1-\mathfrak a)}\right).
    \]
    Finally, the delayed-correlation remainder from
    Lemma~\ref{lem:actor-gradient-error}, namely
    \eqref{eq:actor-correlation-remainder}, the preceding drift bounds, and
    the uniform boundedness of $v_t$ imply
    \[
        \overline R_{\mathrm{corr},t_0,T}
        =
        O\left(T^{-(1-\mathfrak a)}\right),
        \qquad
        \overline S_{t_0,T}
        =
        \frac1{A_{t_0,T}}
        \sum_{t=t_0}^{T-1}\alpha_t^2
        =
        O\left(T^{-(1-\mathfrak a)}\right).
    \]
    Substituting these estimates and
    \eqref{eq:overall-decaying-weighted-critic-average} into
    Proposition~\ref{prop:actor-stationarity} gives
    \[
        \overline G_{t_0,T}^{\alpha}
        =
        O\left(
        T^{-(1-\mathfrak a)}
        +
        \varepsilon_{\mathrm{FA}}^2
        +
        \rho^{2(k+1)}
        \right).
    \]
    This proves \eqref{eq:finite-time-unreg-gradient-bound-decaying}.  The
    decaying-step best-iterate claim follows because a minimum is at most a
    positive weighted average.
\end{proof}

\begin{remark}
    \textcolor{revisionblue}{With constant stepsizes, the bound controls the
    average squared gradient norm up to a nonzero error floor. Smaller
    stepsizes reduce the stepsize-dependent part of this floor.} The diminishing
    schedules in \eqref{eq:overall-decaying-stepsizes} remove the
    stepsize-induced neighborhood: the remaining asymptotic floors in
    \eqref{eq:finite-time-unreg-gradient-bound-decaying} are the
    feature-approximation error and the exponential-decay localization error.
    A sharper Poisson-equation or resolvent argument may replace the
    logarithmic stepsize factors in the proof by mixing-time constants, but
    that would require rewriting the Markovian-bias arguments in the actor and
    critic proofs rather than only the overall combination step.
\end{remark}

\section{Convergence Proof with Regularization}\label{sec:regularized_convergence_results}

\paragraph{Algorithmic template.}
The fixed-state entropy-regularized method is the regularized actor-update
branch of Algorithm~\ref{alg:decentralized_ac_lfa}; this branch adds only the
analytic entropy-gradient term to the actor update and leaves the reward-only
TD critic unchanged.

\paragraph{Proof roadmap.}
The regularized proof reuses the critic analysis from Subsection~\ref{app:unregularized-critic}, because fixed-state entropy modifies only the actor objective and leaves the reward-only TD target unchanged.  Lemma~\ref{lem:fixed-state-regularized-actor-consequences} records the actor-side boundedness, Lipschitz, and localization consequences needed after adding entropy, and Lemma~\ref{lem:fixed-state-entropy-gradient-decomposition} identifies the exact local entropy-gradient blocks.  Lemma~\ref{lem:fixed-state-regularized-actor-gradient-error}, Lemma~\ref{lem:fixed-state-regularized-actor-second-moment}, and Lemma~\ref{lem:fixed-state-regularized-actor-drift} then repeat the unregularized actor logic for the regularized direction: control gradient bias, control variance, and prove one-step ascent for \(J_{\iota,\omega}\).  Theorem~\ref{thm:regularized-actor-convergence} telescopes this drift into finite-time stationarity, Theorem~\ref{thm:regularized-optional-pl-convergence} converts stationarity to a regularized value-gap bound under a PL condition, and Corollary~\ref{cor:regularized-to-unregularized} translates that bound back to the unregularized objective.
The strong-entropy route in Subsection~\ref{sec:strong-entropy-pl-route} is the conditional argument showing that fixed-state entropy can turn the product-policy landscape into one satisfying a PL inequality when its negative curvature dominates the positive curvature of the unregularized objective on the explored region.  Lemma~\ref{lem:strong-entropy-curvature-finite} bounds the direct-policy curvature of \(J\) on \(\Pi_0\) by \(\overline L_J\).  Lemma~\ref{lem:strong-entropy-direct-concavity} combines this curvature bound with the entropy Hessian lower bound from the positive state-weight condition to prove direct-policy strong concavity whenever \(\iota\omega_{\min}>\overline L_J\).  Lemma~\ref{lem:strong-entropy-direct-pl} converts that direct-policy strong concavity into a PL inequality for the tangent direct-policy gradient \(G_\Pi(\pi)\).  Lemma~\ref{lem:strong-entropy-softmax-pl} then transfers this direct-policy PL inequality to the softmax-logit gradient using the exploration lower bound \(\delta\), which is the PL condition required by Theorem~\ref{thm:regularized-optional-pl-convergence}.  Finally, Theorem~\ref{thm:strong-entropy-pl-consequence} plugs this verified PL constant into the finite-time recursion to obtain the stated strong-entropy global convergence consequence.

\subsection{Convergence of the Critic}
\label{app:regularized-critic}

The fixed-state entropy objective analyzed in this section does not change the
reward process or the Poisson equation that the critic estimates.  The
regularized objective is obtained by adding to the actor objective a
deterministic function of the current local policy tables.  Consequently the
critic continues to estimate the reward-only local action-value functions
$Q_i^\theta$ defined in the preliminaries, with the same TD target
\[
    r_i(s_i(t),a_i(t))-\mu_i^t
    +
    \psi_i(s_i^k(t+1),a_i^k(t+1))^\top w_i^t
    -
    \psi_i(s_i^k(t),a_i^k(t))^\top w_i^t .
\]
No new critic convergence proof is needed for this fixed-state regularization.
All critic tracking bounds from the unregularized proof, including the
definition of $\mathcal V_t$ in
\eqref{eq:actor-proof-critic-tracking-error}, apply verbatim.

\subsection{Convergence of the Actor}
\label{app:regularized-actor}

We now prove the actor convergence result for a fixed-state
entropy-regularized objective.  Fix a reference distribution
$\omega\in\Delta(\mathcal S_g)$ that is independent of the actor parameter, and
write its local marginals as
\begin{equation}
    \omega_i(s_i)
    :=
    \sum_{s_{-i}\in\mathcal S_{-i}}\omega(s_i,s_{-i}).
    \label{eq:fixed-state-marginal-def}
\end{equation}
For a local policy vector $p\in\Delta(\mathcal A_i)$, let
\[
    h_i(p):=-\sum_{a_i\in\mathcal A_i}p(a_i)\log p(a_i).
\]
The fixed-state entropy functional and the corresponding objective are
\begin{align}
    \mathcal H_\omega(\theta)
    &:=
    \sum_{i=1}^n
    \sum_{s_i\in\mathcal S_i}
    \omega_i(s_i)
    h_i(\pi_{\theta_i}(\cdot|s_i)),
    \label{eq:fixed-state-entropy-theta-def}\\
    J_{\iota,\omega}(\theta)
    &:=
    J(\theta)+\iota\mathcal H_\omega(\theta),
    \qquad
    \iota\ge0 .
    \label{eq:fixed-state-regularized-objective}
\end{align}
For each actor block define the exact fixed-state entropy-gradient vector
\begin{equation}
    \Gamma_{i,\omega}(\theta_i)(s_i,a_i)
    :=
    \omega_i(s_i)\pi_{\theta_i}(a_i|s_i)
    \left[
    -\log\pi_{\theta_i}(a_i|s_i)
    -
    h_i(\pi_{\theta_i}(\cdot|s_i))
    \right].
    \label{eq:fixed-state-entropy-gradient-block}
\end{equation}
This term is local to agent $i$ once $\omega_i$ is specified.  The critic still
estimates the unregularized reward critic.  For the
linear-function-approximation actor update, define
\begin{equation*}
    \widehat q_j^t
    :=
    \psi_j(s_j^k(t),a_j^k(t))^\top w_j^t,
    \qquad
    S_i^t
    :=
    \nabla_{\theta_i}
    \log\pi_{\theta_i^t}(a_i(t)|s_i(t)).
\end{equation*}
The fixed-state regularized actor increment is
\begin{equation}
    \theta_i^{t+1}
    =
    \theta_i^t+\alpha_t\widehat g_{\iota,\omega,i}^t,
    \qquad
    \widehat g_{\iota,\omega,i}^t
    :=
    S_i^t\sum_{j\in N_i^k}\widehat q_j^t
    +
    \iota\Gamma_{i,\omega}(\theta_i^t).
    \label{eq:fixed-state-regularized-actor-increment}
\end{equation}
Let
$\widehat g_{\iota,\omega}^t
:=(\widehat g_{\iota,\omega,1}^t,\ldots,
\widehat g_{\iota,\omega,n}^t)$.
The tabular critic version is obtained by replacing
$\widehat q_j^t$ by
$\widehat Q_j^t(s_j^k(t),a_j^k(t))$.

The proof uses the same delayed-window notation and unregularized critic
tracking quantities as the actor proof without regularization:
$X_t=(s_g(t),a_g(t))$, $\tau_t\le t$, $t_\tau=t-\tau_t$,
$D_{\theta,t}$ and $D_{w,t}$ are defined in
\eqref{eq:actor-window-drift-defs}, and $\mathcal V_t$ is defined in
\eqref{eq:actor-proof-critic-tracking-error}.

\begin{lemma}[Fixed-state regularized actor-side consequences]
    \label{lem:fixed-state-regularized-actor-consequences}
    Suppose Assumption~\ref{assump:actor-spatial-decay} holds on the
    actor-parameter region considered in the proof.  Under the standing
    finite-action product-softmax and bounded-reward setup, and using
    Assumption~\ref{assump:bounded_critic},
    Lemma~\ref{lem:lipschitzness}, \eqref{eq:actor-BW-def-new}, and
    \eqref{eq:actor-fa-error-new},
    the following regularized actor-side facts hold.
    \begin{enumerate}
        \item The unregularized localized actor direction formed from
        truncated critics has blockwise error
        \begin{equation}
            \delta_{\mathrm{loc}}(k)
            :=
            n\sqrt2\,c\rho^{k+1}
            =
            O(\rho^{k+1}).
            \label{eq:fixed-state-delta-loc-def}
        \end{equation}
        In particular, for every agent $i$,
        \begin{equation}
            \left\|
            \mathbb E_{\nu^\theta}
            \left[
            \nabla_{\theta_i}\log\pi_{\theta_i}(a_i|s_i)
            \sum_{j\in N_i^k}
            \widetilde Q_j^\theta(s_j^k,a_j^k)
            \right]
            -
            \nabla_{\theta_i}J(\theta)
            \right\|_2
            \le
            \delta_{\mathrm{loc}}(k).
            \label{eq:fixed-state-unreg-localization-error}
        \end{equation}

        \item The fixed-state entropy-gradient map is bounded and Lipschitz:
        \begin{align}
            K_{i,\omega}^{\Gamma}
            &:=
            \sup_{\theta_i}
            \left\|
            \Gamma_{i,\omega}(\theta_i)
            \right\|_2
            <\infty,
            \label{eq:fixed-state-Gamma-bound}\\
            L_{i,\omega}^{\Gamma}
            &:=
            \sup_{\theta_i\neq\theta_i'}
            \frac{
            \left\|
            \Gamma_{i,\omega}(\theta_i)
            -
            \Gamma_{i,\omega}(\theta_i')
            \right\|_2}
            {\|\theta_i-\theta_i'\|_2}
            <\infty .
            \label{eq:fixed-state-Gamma-lipschitz}
        \end{align}

        \item The fixed-state regularized objective is
        $L_{\nabla J_{\iota,\omega}}$-smooth and bounded above by
        $J_{\iota,\omega,\max}<\infty$ on the visited actor region:
        \begin{equation}
            J_{\iota,\omega}(\theta')
            \ge
            J_{\iota,\omega}(\theta)
            +
            \langle\nabla J_{\iota,\omega}(\theta),
            \theta'-\theta\rangle
            -
            \frac{L_{\nabla J_{\iota,\omega}}}{2}
            \|\theta'-\theta\|_2^2 .
            \label{eq:fixed-state-regularized-smoothness}
        \end{equation}
        With finite action spaces and $|r_i|\le r_{\max}$, one may take
        \[
            J_{\iota,\omega,\max}
            =
            nr_{\max}
            +
            \iota\sum_{i=1}^n\log|\mathcal A_i|.
        \]
    \end{enumerate}
\end{lemma}

\begin{proof}
    We first prove the localization statement.  Define
    \[
        \widetilde g_i(\theta)
        :=
        \mathbb E_{\nu^\theta}
        \left[
        \nabla_{\theta_i}\log\pi_{\theta_i}(a_i|s_i)
        \sum_{j\in N_i^k}
        \widetilde Q_j^\theta(s_j^k,a_j^k)
        \right].
    \]
    Lemma~\ref{lem:localized-actor-direction-bias-softmax}, applied to the
    spatial-decay premise in Assumption~\ref{assump:actor-spatial-decay}
    supplied by Theorem~\ref{thm:spatial-decay-sufficient-condition},
    gives
    \[
        \|\widetilde g_i(\theta)-\nabla_{\theta_i}J(\theta)\|_2
        \le
        nG_i c\rho^{k+1}.
    \]
    Under tabular softmax, Lemma~\ref{lem:product-policy-score-identities}
    gives $G_i=\sqrt2$, so
    \[
        \|\widetilde g_i(\theta)-\nabla_{\theta_i}J(\theta)\|_2
        \le
        n\sqrt2\,c\rho^{k+1}
        =
        \delta_{\mathrm{loc}}(k),
    \]
    which, together with the definition
    \eqref{eq:fixed-state-delta-loc-def}, proves
    \eqref{eq:fixed-state-unreg-localization-error}.

    We next prove the entropy-gradient bounds.  Fix an agent $i$ and local
    state $s_i$.  Let
    \[
        m_i:=|\mathcal A_i|,
        \qquad
        \omega_{i,\max}:=\max_{s_i\in\mathcal S_i}\omega_i(s_i),
        \qquad
        p_a:=\pi_{\theta_i}(a|s_i).
    \]
    The local entropy satisfies
    \[
        0\le h_i(p)\le \log m_i,
        \qquad
        0\le p_a(-\log p_a)\le e^{-1},
        \qquad a\in\mathcal A_i,
        \label{eq:fixed-state-entropy-elementary-bounds}
    \]
    with the convention $0\log0=0$.  Therefore each coordinate of
    $\Gamma_{i,\omega}$ obeys
    \begin{align}
        |\Gamma_{i,\omega}(\theta_i)(s_i,a)|
        &=
        \omega_i(s_i)p_a
        \left|
        -\log p_a-h_i(p)
        \right|
        \notag\\
        &\le
        \omega_i(s_i)
        \left[
        p_a(-\log p_a)+p_a h_i(p)
        \right]
        \notag\\
        &\le
        \omega_{i,\max}
        \left(e^{-1}+\log m_i\right).
        \label{eq:fixed-state-Gamma-coordinate-bound-proof}
    \end{align}
    Since there are $|\mathcal S_i|m_i$ coordinates in the $i$th actor block,
    \begin{equation}
        \|\Gamma_{i,\omega}(\theta_i)\|_2
        \le
        \sqrt{|\mathcal S_i|m_i}\,
        \omega_{i,\max}
        \left(e^{-1}+\log m_i\right).
        \label{eq:fixed-state-Gamma-explicit-bound-proof}
    \end{equation}
    Thus \eqref{eq:fixed-state-Gamma-bound} holds, for example, with the
    right-hand side of
    \eqref{eq:fixed-state-Gamma-explicit-bound-proof}.

    It remains to show Lipschitz continuity of $\Gamma_{i,\omega}$.  For a
    fixed state $s_i$, the softmax Jacobian is
    \begin{equation}
        \frac{\partial p_b}{\partial \theta_i(s_i,c)}
        =
        p_b(\mathbf 1\{b=c\}-p_c),
        \qquad
        b,c\in\mathcal A_i.
        \label{eq:fixed-state-softmax-jacobian-proof}
    \end{equation}
    Define
    \[
        z_a:=-\log p_a-h_i(p),
        \qquad
        \phi_a:=p_a z_a.
    \]
    By the same calculation used in
    Lemma~\ref{lem:fixed-state-entropy-gradient-decomposition},
    \begin{equation}
        \frac{\partial h_i(p)}
        {\partial \theta_i(s_i,c)}
        =
        p_c\left[-\log p_c-h_i(p)\right]
        =
        \phi_c.
        \label{eq:fixed-state-local-entropy-gradient-proof}
    \end{equation}
    Hence, for $a,c\in\mathcal A_i$,
    \begin{align}
        \frac{\partial}
        {\partial \theta_i(s_i,c)}
        \Gamma_{i,\omega}(\theta_i)(s_i,a)
        &=
        \omega_i(s_i)
        \frac{\partial}{\partial \theta_i(s_i,c)}
        \left(p_a z_a\right)
        \notag\\
        &=
        \omega_i(s_i)
        \left[
        p_a(\mathbf 1\{a=c\}-p_c)z_a
        +
        p_a
        \left(
        -(\mathbf 1\{a=c\}-p_c)-\phi_c
        \right)
        \right]
        \notag\\
        &=
        \omega_i(s_i)p_a
        \left[
        (\mathbf 1\{a=c\}-p_c)(z_a-1)-\phi_c
        \right].
        \label{eq:fixed-state-Gamma-hessian-coordinate}
    \end{align}
    The elementary bounds above imply
    \begin{align}
        p_a|z_a|
        &\le
        p_a(-\log p_a)+p_ah_i(p)
        \le e^{-1}+\log m_i,
        \notag\\
        |\phi_c|
        &\le
        e^{-1}+\log m_i.
        \label{eq:fixed-state-z-phi-bounds}
    \end{align}
    Since $|\mathbf 1\{a=c\}-p_c|\le1$ and $p_a\le1$,
    \begin{align}
        \left|
        \frac{\partial}
        {\partial \theta_i(s_i,c)}
        \Gamma_{i,\omega}(\theta_i)(s_i,a)
        \right|
        &\le
        \omega_{i,\max}
        \left[
        p_a|z_a|+p_a+p_a|\phi_c|
        \right]
        \notag\\
        &\le
        \omega_{i,\max}
        \left(
        2e^{-1}+2\log m_i+1
        \right)
        =:
        B_{i,\omega}^{\Gamma}.
        \label{eq:fixed-state-Gamma-hessian-entry-bound}
    \end{align}
    Derivatives across distinct local states are zero, because
    $\Gamma_{i,\omega}(\theta_i)(s_i,\cdot)$ depends only on the logits
    $\theta_i(s_i,\cdot)$.  Let $d_i^\theta:=|\mathcal S_i|m_i$.  The Hessian
    matrix $\nabla_{\theta_i}\Gamma_{i,\omega}(\theta_i)$ therefore has all
    entries bounded in absolute value by $B_{i,\omega}^{\Gamma}$, and hence
    \begin{equation}
        \left\|
        \nabla_{\theta_i}\Gamma_{i,\omega}(\theta_i)
        \right\|_{\mathrm{op}}
        \le
        d_i^\theta B_{i,\omega}^{\Gamma}.
        \label{eq:fixed-state-Gamma-hessian-op-bound}
    \end{equation}
    The mean-value theorem then gives, for any $\theta_i,\theta_i'$,
    \begin{align}
        \|\Gamma_{i,\omega}(\theta_i)-\Gamma_{i,\omega}(\theta_i')\|_2
        &\le
        \sup_{\xi_i}
        \left\|
        \nabla_{\theta_i}\Gamma_{i,\omega}(\xi_i)
        \right\|_{\mathrm{op}}
        \|\theta_i-\theta_i'\|_2
        \notag\\
        &\le
        d_i^\theta B_{i,\omega}^{\Gamma}
        \|\theta_i-\theta_i'\|_2.
        \label{eq:fixed-state-Gamma-lipschitz-proof}
    \end{align}
    Thus \eqref{eq:fixed-state-Gamma-lipschitz} holds, for example, with
    $L_{i,\omega}^{\Gamma}=d_i^\theta B_{i,\omega}^{\Gamma}$.

    Finally, we prove the regularized smoothness statement.  By
    Lemma~\ref{lem:smoothness-lipschitz-policy},
    \[
        \|\nabla J(\theta)-\nabla J(\theta')\|_2
        \le
        L_{\nabla J}\|\theta-\theta'\|_2,
        \qquad
        J(\theta)\le nr_{\max}.
    \]
    Since
    \[
        \nabla_{\theta_i}\mathcal H_\omega(\theta)
        =
        \Gamma_{i,\omega}(\theta_i),
    \]
    the blockwise Lipschitz bound gives
    \begin{align}
        \|\nabla\mathcal H_\omega(\theta)
        -
        \nabla\mathcal H_\omega(\theta')\|_2^2
        &=
        \sum_{i=1}^n
        \|\Gamma_{i,\omega}(\theta_i)
        -
        \Gamma_{i,\omega}(\theta_i')\|_2^2
        \notag\\
        &\le
        \left(\max_i L_{i,\omega}^{\Gamma}\right)^2
        \|\theta-\theta'\|_2^2 .
        \label{eq:fixed-state-entropy-gradient-lipschitz-full}
    \end{align}
    Thus one may take
    \begin{equation}
        L_{\nabla\mathcal H_\omega}
        :=
        \max_{1\le i\le n}L_{i,\omega}^{\Gamma},
        \qquad
        L_{\nabla J_{\iota,\omega}}
        :=
        L_{\nabla J}+\iota L_{\nabla\mathcal H_\omega}.
        \label{eq:fixed-state-regularized-smoothness-constant-proof}
    \end{equation}
    The standard descent/ascent lemma for functions with Lipschitz gradient
    then yields
    \[
        J_{\iota,\omega}(\theta')
        \ge
        J_{\iota,\omega}(\theta)
        +
        \langle\nabla J_{\iota,\omega}(\theta),\theta'-\theta\rangle
        -
        \frac{L_{\nabla J_{\iota,\omega}}}{2}
        \|\theta'-\theta\|_2^2.
    \]
    Also,
    \begin{align}
        J_{\iota,\omega}(\theta)
        &=
        J(\theta)
        +
        \iota
        \sum_{i=1}^n
        \sum_{s_i\in\mathcal S_i}
        \omega_i(s_i)
        h_i(\pi_{\theta_i}(\cdot|s_i))
        \notag\\
        &\le
        nr_{\max}
        +
        \iota
        \sum_{i=1}^n
        \sum_{s_i\in\mathcal S_i}
        \omega_i(s_i)\log|\mathcal A_i|
        \notag\\
        &=
        nr_{\max}
        +
        \iota
        \sum_{i=1}^n
        \log|\mathcal A_i|,
        \label{eq:fixed-state-Jmax-proof}
    \end{align}
    because $\sum_{s_i}\omega_i(s_i)=1$.  This proves the claimed upper bound
    on $J_{\iota,\omega,\max}$.
\end{proof}

\begin{remark}
    Entropy regularization does not by itself imply a product-policy PL
    inequality in networked MARL.  It can improve the geometry in some
    single-agent or tabular settings, but in the present product-policy
    multi-agent setting a PL or global-optimality conversion requires separate
    structural input, such as concentrability, a Nash-to-global comparison, or
    an entropy-dominance condition strong enough to rule out coordination
    traps.  Large entropy weight can therefore change constants and may reduce
    the regularized optimization error under additional assumptions, but it
    also increases the bias between $J_{\iota,\omega}$ and the original
    objective $J$.  The main theorem below proves finite-time stationarity of
    $J_{\iota,\omega}$ only.  Subsection~\ref{sec:strong-entropy-pl-route}
    gives a sufficient entropy-dominance route under which this stationarity
    result can be converted into a product-policy global-optimality result.
\end{remark}

\begin{lemma}[Fixed-state entropy gradient decomposition]
    \label{lem:fixed-state-entropy-gradient-decomposition}
    For every agent $i$,
    \begin{equation}
        \nabla_{\theta_i}J_{\iota,\omega}(\theta)
        =
        \nabla_{\theta_i}J(\theta)
        +
        \iota\Gamma_{i,\omega}(\theta_i).
        \label{eq:fixed-state-gradient-decomposition}
    \end{equation}
\end{lemma}

\begin{proof}
    The reward term is the unregularized average-reward objective, so its
    gradient is $\nabla_{\theta_i}J(\theta)$.  It remains to differentiate
    $\mathcal H_\omega$.  Fix $(i,s_i)$ and write
    $p_a=\pi_{\theta_i}(a|s_i)$.  For a softmax coordinate
    $\theta_i(s_i,a)$,
    \[
        \frac{\partial p_b}{\partial\theta_i(s_i,a)}
        =
        p_b(\mathbf 1\{a=b\}-p_a).
    \]
    Therefore
    \begin{align*}
        \frac{\partial}{\partial\theta_i(s_i,a)}
        h_i(p)
        &=
        \sum_{b\in\mathcal A_i}
        \frac{\partial p_b}{\partial\theta_i(s_i,a)}
        (-\log p_b-1)\\
        &=
        p_a
        \left[
        -\log p_a
        -
        \sum_{b\in\mathcal A_i}p_b(-\log p_b)
        \right],
    \end{align*}
    where the constant $-1$ vanishes because
    $\sum_b\partial p_b/\partial\theta_i(s_i,a)=0$.  Multiplying by
    $\omega_i(s_i)$ gives \eqref{eq:fixed-state-entropy-gradient-block}, and
    summing over local states gives \eqref{eq:fixed-state-gradient-decomposition}.
\end{proof}

\begin{lemma}[Fixed-state regularized actor gradient error bound]
    \label{lem:fixed-state-regularized-actor-gradient-error}
    Suppose Lemma~\ref{lem:fixed-state-regularized-actor-consequences}
    applies and Assumption~\ref{assump:uniform_mixing} holds.
    Define, for each actor block $i$,
    \begin{equation}\begin{aligned}
        A_{i,\iota,\omega}^{\mathrm{tr}}
        &:=
        G_i\bar\psi\sqrt{H_i},
        \qquad
        A_{i,\iota,\omega}^{\mathrm{FA}}
        :=
        G_iH_i,
        \qquad
        A_{i,\iota,\omega}^{\mathrm{loc}}
        :=
        1,
        \\
        A_{i,\iota,\omega}^{\mathrm{mix},0}
        &:=
        G_i\bar\psi B_i^W,
        \qquad
        A_{i,\iota,\omega}^{\mathrm{mix},V}
        :=
        G_i\bar\psi\sqrt{H_i},
        \\
        A_{i,\iota,\omega}^{\mathrm{win},0,\theta}
        &:=
        2L_{S,i}\bar\psi B_i^W,
        \qquad
        A_{i,\iota,\omega}^{\mathrm{win},V,\theta}
        :=
        2L_{S,i}\bar\psi\sqrt{H_i},
        \\
        A_{i,\iota,\omega}^{\mathrm{win},w}
        &:=
        2G_i\bar\psi\sqrt{H_i}.
        \label{eq:fixed-state-block-actor-error-constants}
    \end{aligned}\end{equation}
    Define the vector constants
    \begin{equation}\begin{aligned}
        C_{\iota,\omega,\mathrm{tr}}
        &:=
        \left(\sum_{i=1}^n
        (A_{i,\iota,\omega}^{\mathrm{tr}})^2\right)^{1/2},
        \qquad
        C_{\iota,\omega,\mathrm{FA}}
        :=
        \left(\sum_{i=1}^n
        (A_{i,\iota,\omega}^{\mathrm{FA}})^2\right)^{1/2},
        \\
        C_{\iota,\omega,\mathrm{loc}}
        &:=
        \left(\sum_{i=1}^n
        (A_{i,\iota,\omega}^{\mathrm{loc}})^2\right)^{1/2}
        =
        \sqrt n,
        \qquad
        C_{\iota,\omega,\mathrm{mix},0}
        :=
        \left(\sum_{i=1}^n
        (A_{i,\iota,\omega}^{\mathrm{mix},0})^2\right)^{1/2},
        \\
        C_{\iota,\omega,\mathrm{mix},V}
        &:=
        \left(\sum_{i=1}^n
        (A_{i,\iota,\omega}^{\mathrm{mix},V})^2\right)^{1/2},
        \qquad
        C_{\iota,\omega,\mathrm{win},0}^{\theta}
        :=
        \left(\sum_{i=1}^n
        (A_{i,\iota,\omega}^{\mathrm{win},0,\theta})^2\right)^{1/2},
        \\
        C_{\iota,\omega,\mathrm{win},V}^{\theta}
        &:=
        \left(\sum_{i=1}^n
        (A_{i,\iota,\omega}^{\mathrm{win},V,\theta})^2\right)^{1/2},
        \qquad
        C_{\iota,\omega,\mathrm{win}}^w
        :=
        \left(\sum_{i=1}^n
        (A_{i,\iota,\omega}^{\mathrm{win},w})^2\right)^{1/2}.
        \label{eq:fixed-state-vector-actor-error-constants}
    \end{aligned}\end{equation}
    Then, for every $t$,
    \begin{align}
        &\left\|
        \mathbb E[\widehat g_{\iota,\omega}^t\mid\mathcal H_{t_\tau}]
        -
        \nabla J_{\iota,\omega}(\theta_t)
        \right\|_2
        \notag\\
        &\qquad
        \leq
        C_{\iota,\omega,\mathrm{tr}}\sqrt{\mathcal V_t}
        +
        C_{\iota,\omega,\mathrm{FA}}\varepsilon_{\mathrm{FA}}
        +
        C_{\iota,\omega,\mathrm{loc}}\delta_{\mathrm{loc}}(k)
        +
        C_{\iota,\omega,\mathrm{mix},0}\varepsilon_{\mathrm{mix}}(t)
        \notag\\
        &\qquad\quad
        +
        C_{\iota,\omega,\mathrm{mix},V}
        \varepsilon_{\mathrm{mix}}(t)
        \sqrt{\mathcal V_{t_\tau}}
        +
        C_{\iota,\omega,\mathrm{win},0}^{\theta}D_{\theta,t}
        +
        C_{\iota,\omega,\mathrm{win},V}^{\theta}
        D_{\theta,t}\sqrt{\mathcal V_t}
        +
        C_{\iota,\omega,\mathrm{win}}^wD_{w,t}.
        \label{eq:fixed-state-regularized-actor-gradient-error}
    \end{align}
\end{lemma}

\begin{proof}
    Fix $t$ and write $\theta=\theta_t$.  For a stationary sample
    $(s_g,a_g)\sim\nu^\theta$, define
    \begin{equation*}
        S_i
        :=
        \nabla_{\theta_i}\log\pi_{\theta_i}(a_i|s_i),
        \qquad
        \psi_j
        :=
        \psi_j(s_j^k,a_j^k),
    \end{equation*}
    Let $\Gamma_i^\theta:=\Gamma_{i,\omega}(\theta_i)$.  The same exact
    vector $\iota\Gamma_i^\theta$ appears in the actor estimator and in
    $\nabla_{\theta_i}J_{\iota,\omega}(\theta)$, so the entropy term cancels in
    the bias analysis.  Only the reward-critic component needs sampling,
    feature-approximation, and localization control.
    Introduce the current stationary directions
    \begin{align}
        h_{i,0}
        &:=
        \mathbb E[\widehat g_{\iota,\omega,i}^t\mid\mathcal H_{t_\tau}],
        \notag\\
        h_{i,1}
        &:=
        \mathbb E_{\nu^\theta}
        \left[
        S_i\sum_{j\in N_i^k}\psi_j^\top w_j^t
        \right]
        +
        \iota\Gamma_i^\theta,
        \notag\\
        h_{i,2}
        &:=
        \mathbb E_{\nu^\theta}
        \left[
        S_i\sum_{j\in N_i^k}\psi_j^\top w_j^\star(\theta)
        \right]
        +
        \iota\Gamma_i^\theta,
        \notag\\
        h_{i,3}
        &:=
        \mathbb E_{\nu^\theta}
        \left[
        S_i
        \sum_{j\in N_i^k}
        \widetilde Q_j^{\theta}(s_j^k,a_j^k)
        \right]
        +
        \iota\Gamma_i^\theta.
        \label{eq:fixed-state-actor-error-intermediates}
    \end{align}
    By Lemma~\ref{lem:fixed-state-entropy-gradient-decomposition} and
    \eqref{eq:fixed-state-unreg-localization-error}, $h_{i,3}$ is within
    $\delta_{\mathrm{loc}}(k)$ of
    $\nabla_{\theta_i}J_{\iota,\omega}(\theta)$.

    Define the current and delayed frozen test functions
    \begin{align}
        f_{i,\iota,\omega}(x)
        &:=
        \nabla_{\theta_i}\log\pi_{\theta_i}(a_i|s_i)
        \sum_{j\in N_i^k}
        \psi_j(s_j^k,a_j^k)^\top w_j^t,
        \notag\\
        f_{i,\iota,\omega}^{\mathrm{fr}}(x)
        &:=
        \nabla_{\theta_i^{t_\tau}}
        \log\pi_{\theta_i^{t_\tau}}(a_i|s_i)
        \sum_{j\in N_i^k}
        \psi_j(s_j^k,a_j^k)^\top w_j^{t_\tau}.
        \label{eq:fixed-state-current-frozen-tests}
    \end{align}
    The current test function satisfies
    \begin{align}
        \|f_{i,\iota,\omega}\|_\infty
        &\leq
        G_i\bar\psi
        \left(
        B_i^W+\sqrt{H_i}\sqrt{\mathcal V_t}
        \right),
        \label{eq:fixed-state-current-test-bound}
    \end{align}
    and the delayed frozen test function satisfies
    \begin{equation}
        \|f_{i,\iota,\omega}^{\mathrm{fr}}\|_\infty
        \leq
        G_i\bar\psi
        \left(
        B_i^W+\sqrt{H_i}\sqrt{\mathcal V_{t_\tau}}
        \right).
        \label{eq:fixed-state-frozen-test-bound}
    \end{equation}
    Moreover, for every $x=(s_g,a_g)$,
    \begin{align}
        \|f_{i,\iota,\omega}(x)
        -
        f_{i,\iota,\omega}^{\mathrm{fr}}(x)\|_2
        &\leq
        L_{S,i}D_{\theta,t}\bar\psi
        \left(
        B_i^W+\sqrt{H_i}\sqrt{\mathcal V_t}
        \right)
        \notag\\
        &\quad+
        G_i\bar\psi
        \sum_{j\in N_i^k}
        \|w_j^t-w_j^{t_\tau}\|_2
        \notag\\
        &\leq
        \frac12A_{i,\iota,\omega}^{\mathrm{win},0,\theta}D_{\theta,t}
        +
        \frac12A_{i,\iota,\omega}^{\mathrm{win},V,\theta}
        D_{\theta,t}\sqrt{\mathcal V_t}
        +
        \frac12A_{i,\iota,\omega}^{\mathrm{win},w}D_{w,t}.
        \label{eq:fixed-state-current-frozen-window-bound}
    \end{align}
    Add and subtract the delayed frozen conditional and stationary
    expectations:
    \begin{align}
        h_{i,0}-h_{i,1}
        &=
        \mathbb E[
        (f_{i,\iota,\omega}-f_{i,\iota,\omega}^{\mathrm{fr}})(X_t)
        \mid\mathcal H_{t_\tau}]
        \notag\\
        &\quad+
        \left(
        \mathbb E[
        f_{i,\iota,\omega}^{\mathrm{fr}}(X_t)
        \mid\mathcal H_{t_\tau}]
        -
        \mathbb E_{\nu^\theta}[f_{i,\iota,\omega}^{\mathrm{fr}}(X)]
        \right)
        \notag\\
        &\quad+
        \mathbb E_{\nu^\theta}
        [(f_{i,\iota,\omega}^{\mathrm{fr}}
        -f_{i,\iota,\omega})(X)].
        \label{eq:fixed-state-h01-three-term}
    \end{align}
    The first and third terms are bounded by twice the supremum in
    \eqref{eq:fixed-state-current-frozen-window-bound}.  For the middle term,
    Lemma~\ref{lem:conditional-markovian-mixing-error} applies to the
    $\mathcal H_{t_\tau}$-measurable test function
    $f_{i,\iota,\omega}^{\mathrm{fr}}$ with the envelope in
    \eqref{eq:fixed-state-frozen-test-bound}.  Therefore
    \begin{align}
        \|h_{i,0}-h_{i,1}\|_2
        &\leq
        A_{i,\iota,\omega}^{\mathrm{mix},0}
        \varepsilon_{\mathrm{mix}}(t)
        +
        A_{i,\iota,\omega}^{\mathrm{mix},V}
        \varepsilon_{\mathrm{mix}}(t)
        \sqrt{\mathcal V_{t_\tau}}
        \notag\\
        &\quad+
        A_{i,\iota,\omega}^{\mathrm{win},0,\theta}D_{\theta,t}
        +
        A_{i,\iota,\omega}^{\mathrm{win},V,\theta}
        D_{\theta,t}\sqrt{\mathcal V_t}
        +
        A_{i,\iota,\omega}^{\mathrm{win},w}D_{w,t}.
        \label{eq:fixed-state-h01-bound}
    \end{align}

    The critic tracking term is
    \begin{align}
        \|h_{i,1}-h_{i,2}\|_2
        &\leq
        G_i\bar\psi
        \sum_{j\in N_i^k}
        \|w_j^t-w_j^\star(\theta)\|_2
        \notag\\
        &\leq
        G_i\bar\psi\sqrt{H_i}\sqrt{\mathcal V_t}.
        \label{eq:fixed-state-h12-bound}
    \end{align}
    The feature approximation term satisfies
    \begin{equation}
        \|h_{i,2}-h_{i,3}\|_2
        \leq
        G_iH_i\varepsilon_{\mathrm{FA}}.
        \label{eq:fixed-state-h23-bound}
    \end{equation}
    By \eqref{eq:fixed-state-unreg-localization-error} and
    Lemma~\ref{lem:fixed-state-entropy-gradient-decomposition},
    \begin{equation}
        \|h_{i,3}-\nabla_{\theta_i}J_{\iota,\omega}(\theta)\|_2
        \leq
        \delta_{\mathrm{loc}}(k).
        \label{eq:fixed-state-h3-gradient-bound}
    \end{equation}
    Combining
    \eqref{eq:fixed-state-h01-bound}--\eqref{eq:fixed-state-h3-gradient-bound}
    gives the blockwise estimate.  Taking the Euclidean norm over actor blocks
    and applying Minkowski's inequality yields
    \eqref{eq:fixed-state-regularized-actor-gradient-error}.
\end{proof}

\begin{lemma}[Fixed-state regularized actor estimator second moment]
    \label{lem:fixed-state-regularized-actor-second-moment}
    Suppose Assumption~\ref{assump:bounded_critic} and
    Lemma~\ref{lem:lipschitzness} hold, with $B_i^W$ defined in
    \eqref{eq:actor-BW-def-new}, and suppose the entropy-gradient bound
    \eqref{eq:fixed-state-Gamma-bound} holds.
    Define
    \begin{equation}
        \sigma_{\iota,\omega,a,0}^2
        :=
        3\sum_{i=1}^n
        \left(
        G_i^2\bar\psi^2(B_i^W)^2
        +
        \iota^2(K_{i,\omega}^{\Gamma})^2
        \right),
        \qquad
        \sigma_{\iota,\omega,a,V}^2
        :=
        3\bar\psi^2
        \sum_{i=1}^nG_i^2H_i .
        \label{eq:fixed-state-actor-second-moment-constants}
    \end{equation}
    Then
    \begin{equation}
        \mathbb E[
        \|\widehat g_{\iota,\omega}^t\|_2^2
        \mid\mathcal H_{t_\tau}]
        \leq
        \sigma_{\iota,\omega,a,0}^2
        +
        \sigma_{\iota,\omega,a,V}^2\mathcal V_t.
        \label{eq:fixed-state-actor-second-moment}
    \end{equation}
\end{lemma}

\begin{proof}
    From \eqref{eq:fixed-state-regularized-actor-increment} and
    \eqref{eq:fixed-state-current-test-bound},
    \begin{equation*}
        \|\widehat g_{\iota,\omega,i}^t\|_2
        \leq
        G_i\bar\psi B_i^W
        +
        G_i\bar\psi\sqrt{H_i}\sqrt{\mathcal V_t}
        +
        \iota K_{i,\omega}^{\Gamma}.
    \end{equation*}
    Using $(a+b+c)^2\leq3(a^2+b^2+c^2)$ and summing over $i$ proves
    \eqref{eq:fixed-state-actor-second-moment}.
\end{proof}

\begin{lemma}[Fixed-state regularized actor drift inequality]
    \label{lem:fixed-state-regularized-actor-drift}
    Suppose the hypotheses of
    Lemmas~\ref{lem:fixed-state-regularized-actor-gradient-error} and
    \ref{lem:fixed-state-regularized-actor-second-moment} hold and
    $\alpha_t\leq1$.  Define
    \begin{align}
        C_{\iota,\omega,V}^{\mathrm{act}}
        &:=
        4C_{\iota,\omega,\mathrm{tr}}^2
        +
        \frac{L_{\nabla J_{\iota,\omega}}}{2}
        \sigma_{\iota,\omega,a,V}^2,
        \notag\\
        C_{\iota,\omega,\mathrm{FA}}^{\mathrm{act}}
        &:=
        4C_{\iota,\omega,\mathrm{FA}}^2,
        \qquad
        C_{\iota,\omega,\mathrm{loc}}^{\mathrm{act}}
        :=
        4C_{\iota,\omega,\mathrm{loc}}^2,
        \notag\\
        C_{\iota,\omega,\mathrm{mix},0}^{\mathrm{act}}
        &:=
        4C_{\iota,\omega,\mathrm{mix},0}^2,
        \qquad
        C_{\iota,\omega,\mathrm{mix},V}^{\mathrm{act}}
        :=
        4C_{\iota,\omega,\mathrm{mix},V}^2,
        \notag\\
        C_{\iota,\omega,\mathrm{win},0}^{\mathrm{act}}
        &:=
        4\max
        \left\{
        (C_{\iota,\omega,\mathrm{win},0}^{\theta})^2,
        (C_{\iota,\omega,\mathrm{win}}^w)^2
        \right\},
        \notag\\
        C_{\iota,\omega,\mathrm{win},V}^{\mathrm{act}}
        &:=
        4(C_{\iota,\omega,\mathrm{win},V}^{\theta})^2,
        \qquad
        C_{\iota,\omega,\mathrm{var}}^{\mathrm{act}}
        :=
        \frac{L_{\nabla J_{\iota,\omega}}}{2}
        \sigma_{\iota,\omega,a,0}^2.
        \label{eq:fixed-state-actor-drift-constants}
    \end{align}
    Define the fixed-state regularized one-step error aggregate
    \begin{align}
        \mathcal R_{\iota,\omega,t}
        &:=
        C_{\iota,\omega,V}^{\mathrm{act}}\mathcal V_t
        +
        C_{\iota,\omega,\mathrm{FA}}^{\mathrm{act}}\varepsilon_{\mathrm{FA}}^2
        +
        C_{\iota,\omega,\mathrm{loc}}^{\mathrm{act}}\delta_{\mathrm{loc}}(k)^2
        +
        C_{\iota,\omega,\mathrm{mix},0}^{\mathrm{act}}
        \varepsilon_{\mathrm{mix}}(t)
        \notag\\
        &\quad+
        C_{\iota,\omega,\mathrm{mix},V}^{\mathrm{act}}
        \varepsilon_{\mathrm{mix}}(t)^2\mathcal V_{t_\tau}
        +
        C_{\iota,\omega,\mathrm{win},0}^{\mathrm{act}}
        \left(D_{\theta,t}^2+D_{w,t}^2\right)
        +
        C_{\iota,\omega,\mathrm{win},V}^{\mathrm{act}}
        D_{\theta,t}^2\mathcal V_t.
        \label{eq:fixed-state-one-step-error-aggregate}
    \end{align}
    Then
    \begin{align}
        \mathbb E[
        J_{\iota,\omega}(\theta_{t+1})\mid\mathcal H_{t_\tau}]
        &\geq
        J_{\iota,\omega}(\theta_t)
        +
        \frac{\alpha_t}{2}
        \left\|
        \nabla J_{\iota,\omega}(\theta_t)
        \right\|_2^2
        -
        \alpha_t\mathcal R_{\iota,\omega,t}
        -
        C_{\iota,\omega,\mathrm{var}}^{\mathrm{act}}\alpha_t^2 .
        \label{eq:fixed-state-regularized-actor-drift}
    \end{align}
\end{lemma}

\begin{proof}
    Write
    \begin{equation*}
        g_{\iota,\omega,t}:=\nabla J_{\iota,\omega}(\theta_t),
        \qquad
        b_{\iota,\omega,t}
        :=
        \mathbb E[\widehat g_{\iota,\omega}^t\mid\mathcal H_{t_\tau}]
        -
        g_{\iota,\omega,t},
        \qquad
        \xi_{\iota,\omega,t+1}
        :=
        \widehat g_{\iota,\omega}^t
        -
        \mathbb E[\widehat g_{\iota,\omega}^t\mid\mathcal H_{t_\tau}].
    \end{equation*}
    Then
    $\mathbb E[\xi_{\iota,\omega,t+1}\mid\mathcal H_{t_\tau}]=0$ and
    $\widehat g_{\iota,\omega}^t
    =
    g_{\iota,\omega,t}
    +
    b_{\iota,\omega,t}
    +
    \xi_{\iota,\omega,t+1}$.  By
    \eqref{eq:fixed-state-regularized-smoothness},
    \begin{align}
        \mathbb E[
        J_{\iota,\omega}(\theta_{t+1})-J_{\iota,\omega}(\theta_t)
        \mid\mathcal H_{t_\tau}]
        &\geq
        \alpha_t\|g_{\iota,\omega,t}\|_2^2
        +
        \alpha_t
        \langle g_{\iota,\omega,t},b_{\iota,\omega,t}\rangle
        -
        \frac{L_{\nabla J_{\iota,\omega}}}{2}\alpha_t^2
        \mathbb E[
        \|\widehat g_{\iota,\omega}^t\|_2^2
        \mid\mathcal H_{t_\tau}].
        \label{eq:fixed-state-drift-proof-start}
    \end{align}
    Young's inequality gives
    $\langle g_{\iota,\omega,t},b_{\iota,\omega,t}\rangle
    \geq
    -\frac12\|g_{\iota,\omega,t}\|_2^2
    -
    \frac12\|b_{\iota,\omega,t}\|_2^2$.
    Lemma~\ref{lem:fixed-state-regularized-actor-gradient-error} and
    $(x_1+\cdots+x_8)^2\leq8\sum_{\ell=1}^8x_\ell^2$ imply
    \begin{align}
        \|b_{\iota,\omega,t}\|_2^2
        \leq
        8C_{\iota,\omega,\mathrm{tr}}^2\mathcal V_t
        +
        8C_{\iota,\omega,\mathrm{FA}}^2\varepsilon_{\mathrm{FA}}^2
        +
        8C_{\iota,\omega,\mathrm{loc}}^2\delta_{\mathrm{loc}}(k)^2
        +
        8C_{\iota,\omega,\mathrm{mix},0}^2
        \varepsilon_{\mathrm{mix}}(t)^2
        \notag\\
        \quad+
        8C_{\iota,\omega,\mathrm{mix},V}^2
        \varepsilon_{\mathrm{mix}}(t)^2\mathcal V_{t_\tau}
        +
        8(C_{\iota,\omega,\mathrm{win},0}^{\theta})^2D_{\theta,t}^2
        \notag\\
        \quad+
        8(C_{\iota,\omega,\mathrm{win},V}^{\theta})^2
        D_{\theta,t}^2\mathcal V_t
        +
        8(C_{\iota,\omega,\mathrm{win}}^w)^2D_{w,t}^2.
        \label{eq:fixed-state-bias-square}
    \end{align}
    Since $0\le\varepsilon_{\mathrm{mix}}(t)\le1$, the scalar mixing term
    satisfies
    $\varepsilon_{\mathrm{mix}}(t)^2\le\varepsilon_{\mathrm{mix}}(t)$.  Using
    Lemma~\ref{lem:fixed-state-regularized-actor-second-moment} in
    \eqref{eq:fixed-state-drift-proof-start}, and using $\alpha_t\le1$ to
    absorb $\alpha_t^2\mathcal V_t$ into
    $\alpha_t\mathcal V_t$, gives
    \eqref{eq:fixed-state-regularized-actor-drift} with the constants in
    \eqref{eq:fixed-state-actor-drift-constants}.
\end{proof}

\begin{theorem}[Finite-time fixed-state regularized policy-gradient stationarity]
    \label{thm:regularized-actor-convergence}
    Suppose Assumptions~\ref{assump:ergodicity}-\ref{assump:actor-spatial-decay} hold, use Lemma~\ref{lem:lipschitzness}, \eqref{eq:actor-BW-def-new}, and
    \eqref{eq:actor-fa-error-new}, and suppose
    $\alpha_t\le1$ for all $t$ in the horizon.  For integers $0\le t_0<T$, define
    \begin{equation}
        A_{t_0,T}:=\sum_{t=t_0}^{T-1}\alpha_t,
        \qquad
        \overline G_{\iota,\omega,t_0,T}
        :=
        \frac{1}{A_{t_0,T}}
        \sum_{t=t_0}^{T-1}
        \alpha_t
        \mathbb E\|\nabla J_{\iota,\omega}(\theta_t)\|_2^2.
        \label{eq:fixed-state-average-gradient-def}
    \end{equation}
    Define the averaged error terms
    \begin{align}
        \overline V_{t_0,T}
        &:=
        \frac{1}{A_{t_0,T}}
        \sum_{t=t_0}^{T-1}\alpha_t\mathbb E\mathcal V_t,
        \notag\\
        \overline M_{t_0,T}
        &:=
        \frac{1}{A_{t_0,T}}
        \sum_{t=t_0}^{T-1}
        \alpha_t\varepsilon_{\mathrm{mix}}(t),
        \notag\\
        \overline M_{V,t_0,T}
        &:=
        \frac{1}{A_{t_0,T}}
        \sum_{t=t_0}^{T-1}
        \alpha_t
        \mathbb E[
        \varepsilon_{\mathrm{mix}}(t)^2\mathcal V_{t_\tau}],
        \notag\\
        \overline S_{t_0,T}
        &:=
        \frac{1}{A_{t_0,T}}
        \sum_{t=t_0}^{T-1}\alpha_t^2 .
        \label{eq:fixed-state-average-error-defs}
    \end{align}
    \begin{equation}
    \begin{aligned}
        \overline D_{\theta,t_0,T}^2
        &:=
        \frac{1}{A_{t_0,T}}
        \sum_{t=t_0}^{T-1}\alpha_tD_{\theta,t}^2,
        \qquad
        \overline D_{w,t_0,T}^2
        :=
        \frac{1}{A_{t_0,T}}
        \sum_{t=t_0}^{T-1}\alpha_tD_{w,t}^2,
        \\
        \overline D_{\theta,V,t_0,T}^2
        &:=
        \frac{1}{A_{t_0,T}}
        \sum_{t=t_0}^{T-1}
        \alpha_t
        \mathbb E[D_{\theta,t}^2\mathcal V_t].
        \label{eq:fixed-state-average-window-defs}
    \end{aligned}
    \end{equation}
    Then
    \begin{align}
        \overline G_{\iota,\omega,t_0,T}
        &\leq
        \frac{
        2\bigl(J_{\iota,\omega,\max}
        -
        \mathbb E J_{\iota,\omega}(\theta_{t_0})\bigr)}
        {A_{t_0,T}}
        +
        2C_{\iota,\omega,V}^{\mathrm{act}}\overline V_{t_0,T}
        +
        2C_{\iota,\omega,\mathrm{FA}}^{\mathrm{act}}
        \varepsilon_{\mathrm{FA}}^2
        \notag\\
        &\quad+
        2C_{\iota,\omega,\mathrm{loc}}^{\mathrm{act}}
        \delta_{\mathrm{loc}}(k)^2
        +
        2C_{\iota,\omega,\mathrm{mix},0}^{\mathrm{act}}
        \overline M_{t_0,T}
        +
        2C_{\iota,\omega,\mathrm{mix},V}^{\mathrm{act}}
        \overline M_{V,t_0,T}
        \notag\\
        &\quad+
        2C_{\iota,\omega,\mathrm{win},0}^{\mathrm{act}}
        \left(
        \overline D_{\theta,t_0,T}^2
        +
        \overline D_{w,t_0,T}^2
        \right)
        +
        2C_{\iota,\omega,\mathrm{win},V}^{\mathrm{act}}
        \overline D_{\theta,V,t_0,T}^2
        +
        2C_{\iota,\omega,\mathrm{var}}^{\mathrm{act}}\overline S_{t_0,T}.
        \label{eq:fixed-state-regularized-actor-finite-time}
    \end{align}
    Consequently,
    \begin{equation}
        \min_{t_0\le t\le T-1}
        \mathbb E\|\nabla J_{\iota,\omega}(\theta_t)\|_2^2
        \leq
        \overline G_{\iota,\omega,t_0,T},
        \label{eq:fixed-state-regularized-best-iterate}
    \end{equation}
    with $\overline G_{\iota,\omega,t_0,T}$ bounded by the right-hand side of
    \eqref{eq:fixed-state-regularized-actor-finite-time}.
\end{theorem}

\begin{proof}
    Take total expectation in
    Lemma~\ref{lem:fixed-state-regularized-actor-drift}, sum
    from $t=t_0$ to $T-1$, and rearrange.  Since
    $J_{\iota,\omega}(\theta_T)\le J_{\iota,\omega,\max}$,
    \begin{align*}
        \frac12
        \sum_{t=t_0}^{T-1}
        \alpha_t
        \mathbb E\|\nabla J_{\iota,\omega}(\theta_t)\|_2^2
        &\leq
        J_{\iota,\omega,\max}
        -
        \mathbb E J_{\iota,\omega}(\theta_{t_0})
        +
        \sum_{t=t_0}^{T-1}
        \alpha_t\mathbb E\mathcal R_{\iota,\omega,t}
        +
        C_{\iota,\omega,\mathrm{var}}^{\mathrm{act}}
        \sum_{t=t_0}^{T-1}\alpha_t^2.
    \end{align*}
    Expanding $\mathcal R_{\iota,\omega,t}$ with
    \eqref{eq:fixed-state-one-step-error-aggregate} and dividing by
    $A_{t_0,T}/2$ gives
    \eqref{eq:fixed-state-regularized-actor-finite-time}.  The
    best-iterate bound follows because the minimum is at most the weighted
    average.
\end{proof}

\begin{theorem}[Optional PL conversion for the fixed-state regularized objective]
    \label{thm:regularized-optional-pl-convergence}
    In addition to the assumptions of
    Theorem~\ref{thm:regularized-actor-convergence}, suppose the
    product-policy fixed-state entropy PL inequality
    \begin{equation}
        \|\nabla J_{\iota,\omega}(\theta)\|_2^2
        \geq
        \mu_{\iota,\omega}
        \left(
        J_{\iota,\omega}^\star
        -
        J_{\iota,\omega}(\theta)
        \right),
        \qquad
        \mu_{\iota,\omega}>0,
        \label{eq:fixed-state-product-pl-optional}
    \end{equation}
    holds on the considered actor-parameter region, where
    $J_{\iota,\omega}^\star
    :=
    \sup_{\theta\in\Theta_{\mathrm{prod}}}J_{\iota,\omega}(\theta)$.
    Subsection~\ref{sec:strong-entropy-pl-route} gives one set of sufficient
    conditions for \eqref{eq:fixed-state-product-pl-optional}.
    Suppose also $0<\alpha_t\le 2/\mu_{\iota,\omega}$.  Let
    $\Delta_{\iota,\omega,t}
    :=
    J_{\iota,\omega}^\star-J_{\iota,\omega}(\theta_t)$ and
    $m_t:=\mathbb E[\Delta_{\iota,\omega,t}]$.  Then
    \begin{equation}
        m_T
        \leq
        \mathcal P_{\iota,\omega,t_0,T}m_{t_0}
        +
        \sum_{\ell=t_0}^{T-1}
        \alpha_\ell
        \left(
        \prod_{r=\ell+1}^{T-1}
        \left(1-\frac{\mu_{\iota,\omega}\alpha_r}{2}\right)
        \right)
        \mathbb E[
        \mathcal R_{\iota,\omega,\ell}
        +
        C_{\iota,\omega,\mathrm{var}}^{\mathrm{act}}\alpha_\ell],
        \label{eq:fixed-state-optional-pl-bound}
    \end{equation}
    where
    \begin{equation*}
        \mathcal P_{\iota,\omega,t_0,T}
        :=
        \prod_{\ell=t_0}^{T-1}
        \left(1-\frac{\mu_{\iota,\omega}\alpha_\ell}{2}\right).
    \end{equation*}
    In particular, if
    $\mathbb E[
    \mathcal R_{\iota,\omega,t}
    +
    C_{\iota,\omega,\mathrm{var}}^{\mathrm{act}}\alpha_t]
    \leq\overline{\mathcal R}_{\iota,\omega}$ for all $t$, then
    \begin{equation}
        m_T
        \leq
        \mathcal P_{\iota,\omega,t_0,T}m_{t_0}
        +
        \frac{2\overline{\mathcal R}_{\iota,\omega}}
        {\mu_{\iota,\omega}}
        \left(1-\mathcal P_{\iota,\omega,t_0,T}\right).
        \label{eq:fixed-state-optional-pl-error-floor}
    \end{equation}
\end{theorem}

\begin{proof}
    Lemma~\ref{lem:fixed-state-regularized-actor-drift} and
    \eqref{eq:fixed-state-product-pl-optional} imply
    \begin{equation}
        \mathbb E[
        \Delta_{\iota,\omega,t+1}\mid\mathcal H_{t_\tau}]
        \leq
        \left(1-\frac{\mu_{\iota,\omega}\alpha_t}{2}\right)
        \Delta_{\iota,\omega,t}
        +
        \alpha_t\mathcal R_{\iota,\omega,t}
        +
        C_{\iota,\omega,\mathrm{var}}^{\mathrm{act}}\alpha_t^2.
        \label{eq:fixed-state-optional-pl-recursion}
    \end{equation}
    Taking total expectation and iterating the affine recursion gives
    \eqref{eq:fixed-state-optional-pl-bound}.  The error-floor bound follows
    from
    \begin{equation*}
        \sum_{\ell=t_0}^{T-1}
        \alpha_\ell
        \prod_{r=\ell+1}^{T-1}
        \left(1-\frac{\mu_{\iota,\omega}\alpha_r}{2}\right)
        \leq
        \frac{2}{\mu_{\iota,\omega}}
        \left(1-\mathcal P_{\iota,\omega,t_0,T}\right).
    \end{equation*}
\end{proof}

\begin{corollary}[Optional conversion to the unregularized objective]
    \label{cor:regularized-to-unregularized}
    Let
    $J^\star:=\sup_{\theta\in\Theta_{\mathrm{prod}}}J(\theta)$ and
    $H_{\omega,\max}:=\sum_i\log|\mathcal{A}_i|$.  Under the optional PL
    conversion theorem, the regularized optimality-gap bound implies
    \begin{equation*}
        \mathbb E[J^\star-J(\theta_T)]
        \leq
        m_T+\iota H_{\omega,\max}.
    \end{equation*}
\end{corollary}

\begin{proof}
    For every product policy,
    $0\leq\mathcal H_\omega(\theta)\leq H_{\omega,\max}$ and
    $J_{\iota,\omega}(\theta)
    =
    J(\theta)+\iota\mathcal H_\omega(\theta)$.  Hence
    $J^\star\leq J_{\iota,\omega}^\star$ and
    $-J(\theta_T)
    \leq
    -J_{\iota,\omega}(\theta_T)+\iota H_{\omega,\max}$.  Adding the two
    inequalities and taking expectations proves the claim.
\end{proof}

\subsection{A Strong-Entropy Route to the Product-Policy PL Inequality}
\label{sec:strong-entropy-pl-route}

The fixed-state regularized actor theorem above gives finite-time control of
$\|\nabla J_{\iota,\omega}(\theta)\|_2^2$.  To turn this stationarity statement into a
global-optimality statement, one needs a product-policy PL inequality of the
form \eqref{eq:fixed-state-product-pl-optional}.  This subsection gives a
self-contained sufficient route.  The result is deliberately conditional:
entropy regularization alone does not imply a PL inequality in networked MARL.

\paragraph{Direct product-policy variables.}
Let $\Pi$ denote the direct product-policy class
\begin{equation}
    \Pi
    :=
    \left\{
    \pi=\prod_{i=1}^n\pi_i:
    \pi_i(\cdot|s_i)\in\Delta(\mathcal A_i)
    \text{ for all }i,s_i
    \right\}.
    \label{eq:strong-entropy-product-class}
\end{equation}
In this subsection $J(\pi)$ and $J_{\iota,\omega}(\pi)$ denote the objectives
viewed as functions of the stacked local policy tables
$\{\pi_i(a_i|s_i)\}_{i,s_i,a_i}$.  Define
\begin{equation}
    \mathcal H_\omega(\pi)
    :=
    \sum_{i=1}^n
    \sum_{s_i\in\mathcal S_i}
    \omega_i(s_i)
    h_i(\pi_i(\cdot|s_i)),
    \qquad
    J_{\iota,\omega}(\pi)
    :=
    J(\pi)+\iota\mathcal H_\omega(\pi).
    \label{eq:strong-entropy-direct-objective}
\end{equation}
The feasible equality tangent space is
\begin{equation}
    T_\pi\Pi
    :=
    \left\{
    u:
    \sum_{a_i\in\mathcal A_i}u_i(s_i,a_i)=0
    \text{ for all }i,s_i
    \right\}.
    \label{eq:strong-entropy-tangent-space}
\end{equation}
Define the lower-bounded product-policy region
\[
    \Pi_0
    :=
    \left\{
    \pi\in\Pi:
    \pi_i(a_i|s_i)\ge\delta
    \text{ for all }i,s_i,a_i
    \right\}.
\]
Positive entropy weights are imposed in
Assumption~\ref{assump:strong-entropy-positive-weights}.
For every feasible direction $u\in T_\pi\Pi$,
\begin{align}
    u^\top\nabla_\pi^2\mathcal H_\omega(\pi)u
    &=
    -\sum_{i=1}^n
    \sum_{s_i\in\mathcal S_i}
    \omega_i(s_i)
    \sum_{a_i\in\mathcal A_i}
    \frac{u_i(s_i,a_i)^2}{\pi_i(a_i|s_i)}
    \notag\\
    &\leq
    -\omega_{\min}
    \sum_{i=1}^n
    \sum_{s_i\in\mathcal S_i}
    \sum_{a_i\in\mathcal A_i}
    u_i(s_i,a_i)^2
    =
    -\omega_{\min}\|u\|_2^2,
    \label{eq:fixed-state-entropy-strong-concavity}
\end{align}
where we used $\pi_i(a_i|s_i)\leq1$.  Hence the fixed-state entropy functional
is $\omega_{\min}$-strongly concave on the whole relative interior of the
product simplex.

\begin{lemma}[Finite direct-policy curvature on the explored region]
    \label{lem:strong-entropy-curvature-finite}
    Under Assumptions~\ref{assump:ergodicity},
    \ref{assump:uniform_mixing}, and
    \ref{assump:strong-entropy-exploration-region}, the direct-policy map
    $J$ is twice continuously differentiable on a neighborhood of $\Pi_0$.
    Let $\textcolor{revisionblue}{P_\pi}$ be the frozen state-action kernel, let
    $\nu_\pi$ be its stationary distribution, and define
    $\|G\|_{\infty,1}:=\sup_x\sum_y|G(x,y)|$ for signed kernels.  Set
    \begin{align}
        R&:=n\bar r,
        &
        \mathsf M&:=\frac{2m_{\mathrm{mix}}}{1-\rho_{\mathrm{mix}}},
        \notag\\
        \kappa_1
        &:=
        \sup_{\pi\in\Pi_0}
        \sup_{\substack{u\in T_\pi\Pi\\ \|u\|_2=1}}
        \|D\textcolor{revisionblue}{P_\pi}[u]\|_{\infty,1},
        &
        \kappa_2
        &:=
        \sup_{\pi\in\Pi_0}
        \sup_{\substack{u,v\in T_\pi\Pi\\ \|u\|_2=\|v\|_2=1}}
        \|D^2\textcolor{revisionblue}{P_\pi}[u,v]\|_{\infty,1}.
        \label{eq:strong-entropy-kappa-defs}
    \end{align}
    Then
    \begin{equation}
        \kappa_1\le
        \left(\sum_{i=1}^n|\mathcal A_i|\right)^{1/2},
        \qquad
        \kappa_2\le
        \sum_{i=1}^n|\mathcal A_i|.
        \label{eq:strong-entropy-kappa-bounds}
    \end{equation}
    Moreover, the curvature constant
    \begin{equation}
        L_J
        :=
        \sup_{\pi\in\Pi_0}
        \sup_{\substack{u\in T_\pi\Pi\\ \|u\|_2=1}}
        u^\top\nabla_\pi^2J(\pi)u
        \label{eq:strong-entropy-J-curvature}
    \end{equation}
    satisfies the explicit bound
    \begin{equation}
        L_J
        \le
        \overline L_J
        :=
        2R\mathsf M\kappa_2
        +
        4R\mathsf M^2\kappa_1^2
        \le
        2n\bar r
        \left(\sum_{i=1}^n|\mathcal A_i|\right)
        \mathsf M(1+2\mathsf M).
        \label{eq:strong-entropy-J-curvature-bound}
    \end{equation}
\end{lemma}

\begin{proof}
    The state and action spaces are finite, and the transition matrix and
    reward vector of the global Markov chain depend smoothly on the local
    policy tables.  Assumption~\ref{assump:ergodicity} gives a unique
    stationary distribution throughout the relative interior, and the
    definition of $\Pi_0$ above with the $\delta$ from
    Assumption~\ref{assump:strong-entropy-exploration-region} places $\Pi_0$
    inside a compact subset of that relative interior.  The stationary
    distribution and the average reward are therefore twice continuously
    differentiable on a neighborhood of $\Pi_0$.

    For $x=(s_g,a_g)$ and $y=(s_g',a_g')$,
    \[
        \textcolor{revisionblue}{P_\pi}(x,y)
        =
        P(s_g'|s_g,a_g)
        \prod_{i=1}^n\pi_i(a_i'|s_i').
    \]
    Therefore, for every feasible unit direction $u$,
    \begin{align*}
        \|D\textcolor{revisionblue}{P_\pi}[u]\|_{\infty,1}
        &\le
        \sup_{s_g'}
        \sum_{i=1}^n
        \sum_{a_i'\in\mathcal A_i}
        |u_i(s_i',a_i')|
        \\
        &\le
        \sup_{s_g'}
        \left(\sum_{i=1}^n|\mathcal A_i|\right)^{1/2}
        \left(
        \sum_{i=1}^n
        \|u_i(s_i',\cdot)\|_2^2
        \right)^{1/2}
        \le
        \left(\sum_{i=1}^n|\mathcal A_i|\right)^{1/2}.
    \end{align*}
    The same product-rule calculation gives, for feasible unit directions
    $u,v$,
    \[
        \|D^2\textcolor{revisionblue}{P_\pi}[u,v]\|_{\infty,1}
        \le
        \sup_{s_g'}
        \sum_{i\ne j}
        \|u_i(s_i',\cdot)\|_1
        \|v_j(s_j',\cdot)\|_1
        \le
        \sum_{i=1}^n|\mathcal A_i|,
    \]
    proving \eqref{eq:strong-entropy-kappa-bounds}.

    Let $\mathbf 1\nu_\pi$ denote the rank-one kernel whose rows equal
    $\nu_\pi$, and set
    \[
        Z_\pi:=
        \sum_{\ell=0}^{\infty}
        (\textcolor{revisionblue}{P_\pi}^\ell-\mathbf 1\nu_\pi).
    \]
    Assumption~\ref{assump:uniform_mixing} implies
    $\|Z_\pi f\|_\infty\le\mathsf M\|f\|_\infty$ uniformly over
    $\pi\in\Pi_0$.  If $r$ is the global reward vector, then
    $\|r-J(\pi)\mathbf 1\|_\infty\le2R$, and the Poisson solution
    \[
        Q_\pi:=Z_\pi(r-J(\pi)\mathbf 1)
    \]
    satisfies $\|Q_\pi\|_\infty\le2R\mathsf M$.

    The average-reward perturbation identity gives
    \[
        DJ(\pi)[u]=\nu_\pi D\textcolor{revisionblue}{P_\pi}[u]Q_\pi .
    \]
    Differentiating once more, and using that each row of $D\textcolor{revisionblue}{P_\pi}[u]$ and
    $D^2\textcolor{revisionblue}{P_\pi}[u,v]$ has zero sum, gives
    \begin{align*}
        D^2J(\pi)[u,v]
        &=
        \nu_\pi D^2\textcolor{revisionblue}{P_\pi}[u,v]Q_\pi
        +
        \nu_\pi D\textcolor{revisionblue}{P_\pi}[v]Z_\pi D\textcolor{revisionblue}{P_\pi}[u]Q_\pi
        \\
        &\quad+
        \nu_\pi D\textcolor{revisionblue}{P_\pi}[u]Z_\pi D\textcolor{revisionblue}{P_\pi}[v]Q_\pi .
    \end{align*}
    Hence, for $\|u\|_2=\|v\|_2=1$,
    \[
        |D^2J(\pi)[u,v]|
        \le
        2R\mathsf M\kappa_2
        +
        4R\mathsf M^2\kappa_1^2 .
    \]
    Taking $v=u$ and then the supremum over
    $\pi\in\Pi_0$ proves \eqref{eq:strong-entropy-J-curvature-bound}.
\end{proof}

\begin{lemma}[Strong entropy implies direct-policy strong concavity]
    \label{lem:strong-entropy-direct-concavity}
    Suppose Assumptions~\ref{assump:strong-entropy-exploration-region} and
    \ref{assump:strong-entropy-positive-weights} hold, and use the curvature
    bound from Lemma~\ref{lem:strong-entropy-curvature-finite}.  Assume
    \[
        \iota>\frac{\overline L_J}{\omega_{\min}} .
    \]
    Define
    \begin{equation}
        \mu_{\iota,\omega}^\Pi
        :=
        \iota\omega_{\min}-\overline L_J
        >
        0.
        \label{eq:strong-entropy-mu-pi}
    \end{equation}
    Then $J_{\iota,\omega}$ is $\mu_{\iota,\omega}^\Pi$-strongly concave on
    $\Pi_0$: for all
    $\pi,\pi'\in\Pi_0$,
    \begin{equation}
        J_{\iota,\omega}(\pi')
        \leq
        J_{\iota,\omega}(\pi)
        +
        \left\langle
        \nabla_\pi J_{\iota,\omega}(\pi),\pi'-\pi
        \right\rangle
        -
        \frac{\mu_{\iota,\omega}^\Pi}{2}
        \|\pi'-\pi\|_2^2 .
        \label{eq:strong-entropy-direct-strong-concavity}
    \end{equation}
\end{lemma}

\begin{proof}
    Fix $\pi,\pi'\in\Pi_0$ and set
    $v:=\pi'-\pi$.  Since $\Pi_0$ is convex,
    $\pi_\alpha:=(1-\alpha)\pi+\alpha\pi'\in\Pi_0$ for all
    $\alpha\in[0,1]$.  Also $v\in T_{\pi_\alpha}\Pi$ for every $\alpha$,
    because $\pi$ and $\pi'$ satisfy the same local simplex constraints.
    Define $\phi(\alpha):=J_{\iota,\omega}(\pi_\alpha)$.  By the chain rule,
    \eqref{eq:strong-entropy-J-curvature-bound}, and
    \eqref{eq:fixed-state-entropy-strong-concavity},
    \begin{align}
        \phi''(\alpha)
        &=
        v^\top\nabla_\pi^2J(\pi_\alpha)v
        +
        \iota v^\top\nabla_\pi^2\mathcal H_\omega(\pi_\alpha)v
        \notag\\
        &\leq
        \overline L_J\|v\|_2^2
        -
        \iota\omega_{\min}\|v\|_2^2
        =
        -\mu_{\iota,\omega}^\Pi\|v\|_2^2 .
        \label{eq:strong-entropy-line-second-derivative}
    \end{align}
    Taylor's theorem with integral remainder gives
    \begin{align*}
        J_{\iota,\omega}(\pi')
        &=
        \phi(1)
        =
        \phi(0)+\phi'(0)
        +
        \int_0^1(1-\alpha)\phi''(\alpha)\,d\alpha
        \notag\\
        &\leq
        J_{\iota,\omega}(\pi)
        +
        \left\langle\nabla_\pi J_{\iota,\omega}(\pi),v\right\rangle
        -
        \frac{\mu_{\iota,\omega}^\Pi}{2}\|v\|_2^2,
    \end{align*}
    which proves \eqref{eq:strong-entropy-direct-strong-concavity}.
\end{proof}

\begin{lemma}[Direct-policy PL inequality]
    \label{lem:strong-entropy-direct-pl}
    Under the assumptions of Lemma~\ref{lem:strong-entropy-direct-concavity},
    let
    \[
        J_{\iota,\omega,\Pi_0}^\star
        :=
        \max_{\pi\in\Pi_0}J_{\iota,\omega}(\pi),
        \qquad
        \pi_{\iota,\omega}^\star
        \in
        \arg\max_{\pi\in\Pi_0}J_{\iota,\omega}(\pi).
    \]
    Let
    \[
        G_\Pi(\pi)
        :=
        \operatorname{Proj}_{T_\pi\Pi}
        \nabla_\pi J_{\iota,\omega}(\pi)
    \]
    be the Euclidean projection of the direct-policy gradient onto the tangent
    space \eqref{eq:strong-entropy-tangent-space}.  Then, for every
    $\pi\in\Pi_0$,
    \begin{equation}
        J_{\iota,\omega,\Pi_0}^\star-J_{\iota,\omega}(\pi)
        \leq
        \frac{1}{2\mu_{\iota,\omega}^\Pi}
        \|G_\Pi(\pi)\|_2^2 .
        \label{eq:strong-entropy-direct-pl}
    \end{equation}
\end{lemma}

\begin{proof}
    Apply
    \eqref{eq:strong-entropy-direct-strong-concavity} with
    $\pi'=\pi_{\iota,\omega}^\star$:
    \begin{equation}
        J_{\iota,\omega,\Pi_0}^\star-J_{\iota,\omega}(\pi)
        \leq
        \left\langle
        \nabla_\pi J_{\iota,\omega}(\pi),
        \pi_{\iota,\omega}^\star-\pi
        \right\rangle
        -
        \frac{\mu_{\iota,\omega}^\Pi}{2}
        \|\pi_{\iota,\omega}^\star-\pi\|_2^2 .
        \label{eq:strong-entropy-direct-pl-start}
    \end{equation}
    Since $\pi_{\iota,\omega}^\star-\pi\in T_\pi\Pi$, the normal component of
    $\nabla_\pi J_{\iota,\omega}(\pi)$ does not contribute to the inner
    product:
    \[
        \left\langle
        \nabla_\pi J_{\iota,\omega}(\pi),
        \pi_{\iota,\omega}^\star-\pi
        \right\rangle
        =
        \left\langle
        G_\Pi(\pi),\pi_{\iota,\omega}^\star-\pi
        \right\rangle .
    \]
    Thus, using Cauchy's inequality and optimizing over
    $r:=\|\pi_{\iota,\omega}^\star-\pi\|_2\geq0$,
    \begin{align*}
        J_{\iota,\omega,\Pi_0}^\star-J_{\iota,\omega}(\pi)
        &\leq
        \|G_\Pi(\pi)\|_2r
        -
        \frac{\mu_{\iota,\omega}^\Pi}{2}r^2
        \notag\\
        &\leq
        \sup_{r\geq0}
        \left\{
        \|G_\Pi(\pi)\|_2r
        -
        \frac{\mu_{\iota,\omega}^\Pi}{2}r^2
        \right\}
        =
        \frac{1}{2\mu_{\iota,\omega}^\Pi}\|G_\Pi(\pi)\|_2^2 .
    \end{align*}
\end{proof}

\begin{lemma}[Softmax-logit PL inequality from direct-policy PL]
    \label{lem:strong-entropy-softmax-pl}
    Suppose the assumptions of Lemma~\ref{lem:strong-entropy-direct-pl} and
    Assumption~\ref{assump:strong-entropy-exploration-region} hold.
    Then, for every $\theta\in\Theta_0$,
    \begin{equation}
        \|\nabla_\theta J_{\iota,\omega}(\theta)\|_2^2
        \geq
        \mu_{\iota,\omega,\theta}
        \left(
        J_{\iota,\omega,\Pi_0}^\star
        -
        J_{\iota,\omega}(\theta)
        \right),
        \qquad
        \mu_{\iota,\omega,\theta}
        :=
        2\delta^2(\iota\omega_{\min}-\overline L_J).
        \label{eq:strong-entropy-softmax-pl}
    \end{equation}
\end{lemma}

\begin{proof}
    Fix a block $(i,s_i)$ and abbreviate
    $p(a):=\pi_{\theta_i}(a|s_i)$.  Define the direct-policy block gradient
    \begin{equation}
        G_i^\theta(s_i,a)
        :=
        \frac{\partial J_{\iota,\omega}(\pi)}
        {\partial \pi_i(a|s_i)}
        \bigg|_{\pi=\pi_\theta},
        \qquad
        \bar G_i^\theta(s_i)
        :=
        \sum_{b\in\mathcal A_i}
        \pi_{\theta_i}(b|s_i)G_i^\theta(s_i,b).
        \label{eq:strong-entropy-direct-block-gradient}
    \end{equation}
    The softmax Jacobian gives
    \begin{equation}
        \frac{\partial J_{\iota,\omega}(\theta)}
        {\partial\theta_i(s_i,a)}
        =
        \pi_{\theta_i}(a|s_i)
        \left(
        G_i^\theta(s_i,a)-\bar G_i^\theta(s_i)
        \right).
        \label{eq:strong-entropy-softmax-chain-rule}
    \end{equation}
    Equivalently, if
    $D_p:=\operatorname{diag}(p)-pp^\top$ and
    $G_i^\theta(s_i,\cdot)$ is viewed as a vector, then the logit-gradient block
    is
    \[
        D_pG_i^\theta(s_i,\cdot).
    \]
    Let
    \begin{equation}
        x_i^\theta(s_i,\cdot)
        :=
        \operatorname{Proj}_{\{\mathbf 1^\top u=0\}}
        G_i^\theta(s_i,\cdot)
        =
        G_i^\theta(s_i,\cdot)
        -
        \frac{\mathbf 1^\top G_i^\theta(s_i,\cdot)}
        {|\mathcal A_i|}\mathbf 1 .
        \label{eq:strong-entropy-block-tangent-projection}
    \end{equation}
    This is the Euclidean tangent projection of the direct-policy gradient
    block for the simplex constraint
    $\sum_{a_i}\pi_i(a_i|s_i)=1$.  The normal component is a constant multiple
    of $\mathbf 1$, and the softmax Jacobian kills it:
    \begin{equation}
        D_p\mathbf 1=0,
        \qquad
        D_pG_i^\theta(s_i,\cdot)
        =
        D_px_i^\theta(s_i,\cdot).
        \label{eq:strong-entropy-softmax-kills-normal}
    \end{equation}
    We now lower bound $D_p$ on the tangent space
    $\{\mathbf 1^\top x=0\}$.  For every vector $x$ satisfying
    $\mathbf 1^\top x=0$,
    \begin{align}
        x^\top D_px
        &=
        \sum_{a\in\mathcal A_i}p(a)x(a)^2
        -
        \left(\sum_{a\in\mathcal A_i}p(a)x(a)\right)^2
        \notag\\
        &=
        \min_{c\in\mathbb R}
        \sum_{a\in\mathcal A_i}p(a)(x(a)-c)^2
        \notag\\
        &\geq
        \delta
        \min_{c\in\mathbb R}
        \sum_{a\in\mathcal A_i}(x(a)-c)^2
        =
        \delta\|x\|_2^2 .
        \label{eq:strong-entropy-softmax-jacobian-spectrum}
    \end{align}
    The last equality uses $\mathbf 1^\top x=0$, because the minimizer of
    $\sum_a(x(a)-c)^2$ is the unweighted average
    $c=|\mathcal A_i|^{-1}\mathbf 1^\top x=0$.  Moreover, $D_p$ maps the
    tangent space into itself since $\mathbf 1^\top D_p=0$.  Hence the
    restriction of the symmetric matrix $D_p$ to the tangent space has all
    eigenvalues at least $\delta$: if
    $x=\sum_r y_r v_r$ is an orthonormal eigen-expansion in the tangent space,
    then
    \[
        \|D_px\|_2^2
        =
        \sum_r \lambda_r^2y_r^2
        \ge
        \delta^2\sum_r y_r^2
        =
        \delta^2\|x\|_2^2 .
    \]
    Applying this to
    $x=x_i^\theta(s_i,\cdot)$ and using
    \eqref{eq:strong-entropy-softmax-kills-normal} gives
    \begin{equation}
        \left\|
        D_pG_i^\theta(s_i,\cdot)
        \right\|_2^2
        \geq
        \delta^2
        \left\|
        \operatorname{Proj}_{\{\mathbf 1^\top u=0\}}
        G_i^\theta(s_i,\cdot)
        \right\|_2^2 .
        \label{eq:strong-entropy-block-logit-lower-bound}
    \end{equation}
    Finally, the direct-policy tangent space is a Cartesian product of the
    block tangent spaces:
    \[
        T_{\pi_\theta}\Pi
        =
        \prod_{i,s_i}
        \{u_i(s_i,\cdot):\mathbf 1^\top u_i(s_i,\cdot)=0\}.
    \]
    Therefore $G_\Pi(\pi_\theta)$ is exactly the collection of the block
    projections in
    \eqref{eq:strong-entropy-block-tangent-projection}.  Adding any blockwise
    constant to the direct-policy gradient changes neither this projection nor
    the softmax logit gradient, because both operations remove the normal
    $\mathbf 1$ component.
    Summing \eqref{eq:strong-entropy-block-logit-lower-bound} over all
    $(i,s_i)$ gives
    \begin{equation}
        \|\nabla_\theta J_{\iota,\omega}(\theta)\|_2^2
        \geq
        \delta^2
        \|G_\Pi(\pi_\theta)\|_2^2 .
        \label{eq:strong-entropy-logit-direct-gradient-bound}
    \end{equation}
    Combining this bound with
    \eqref{eq:strong-entropy-direct-pl} and
    $\mu_{\iota,\omega}^\Pi=\iota\omega_{\min}-\overline L_J$ proves
    \eqref{eq:strong-entropy-softmax-pl}.
\end{proof}

\begin{theorem}[Strong-entropy global convergence consequence]
    \label{thm:strong-entropy-pl-consequence}
    Suppose the assumptions of
    Theorem~\ref{thm:regularized-actor-convergence} and
    Lemma~\ref{lem:strong-entropy-softmax-pl} hold on the actor region visited
    by the iterates.  Let
    \[
        \Delta_{\iota,\omega,t}
        :=
        J_{\iota,\omega,\Pi_0}^\star
        -
        J_{\iota,\omega}(\theta_t),
        \qquad
        \mu_{\iota,\omega,\theta}
        =
        2\delta^2(\iota\omega_{\min}-\overline L_J),
    \]
    and write
    \begin{equation}
        \eta_{\iota,\omega}
        :=
        \frac{\mu_{\iota,\omega,\theta}}{2}
        =
        \delta^2(\iota\omega_{\min}-\overline L_J).
        \label{eq:strong-entropy-eta-def}
    \end{equation}
    and assume $0<\alpha_t\le 2/\mu_{\iota,\omega,\theta}$.  Then
    \begin{equation}
        \mathbb E[
        \Delta_{\iota,\omega,t+1}\mid\mathcal H_{t_\tau}]
        \leq
        \left(
        1-\frac{\alpha_t\mu_{\iota,\omega,\theta}}{2}
        \right)
        \Delta_{\iota,\omega,t}
        +
        \alpha_t\mathcal R_{\iota,\omega,t}
        +
        C_{\iota,\omega,\mathrm{var}}^{\mathrm{act}}\alpha_t^2 .
        \label{eq:strong-entropy-gap-recursion}
    \end{equation}
    Consequently, for integers $0\le t_0<T$,
    \begin{align}
        &\mathbb E[
        J_{\iota,\omega,\Pi_0}^\star
        -
        J_{\iota,\omega}(\theta_T)]
        \notag\\
        &\qquad\leq
        \mathcal P_{\iota,\omega,t_0,T}^{\mathrm{SE}}
        \mathbb E[
        J_{\iota,\omega,\Pi_0}^\star
        -
        J_{\iota,\omega}(\theta_{t_0})]
        \notag\\
        &\qquad\quad+
        \sum_{\ell=t_0}^{T-1}
        \alpha_\ell
        \left(
        \prod_{r=\ell+1}^{T-1}
        \left(1-\frac{\alpha_r\mu_{\iota,\omega,\theta}}{2}\right)
        \right)
        \mathbb E[
        \mathcal R_{\iota,\omega,\ell}
        +
        C_{\iota,\omega,\mathrm{var}}^{\mathrm{act}}\alpha_\ell],
        \label{eq:strong-entropy-unrolled-gap}
    \end{align}
    where
    \begin{equation}
        \mathcal P_{\iota,\omega,t_0,T}^{\mathrm{SE}}
        :=
        \prod_{t=t_0}^{T-1}
        \left(
        1-\alpha_t\eta_{\iota,\omega}
        \right).
        \label{eq:strong-entropy-gap-product}
    \end{equation}
    In particular, if
    \begin{equation}
        \mathbb E[
        \mathcal R_{\iota,\omega,t}
        +
        C_{\iota,\omega,\mathrm{var}}^{\mathrm{act}}\alpha_t]
        \leq
        \overline{\mathcal R}_{\iota,\omega}
        \qquad
        \text{for all }t,
        \label{eq:strong-entropy-uniform-error}
    \end{equation}
    then
    \begin{align}
        \mathbb E[
        J_{\iota,\omega,\Pi_0}^\star
        -
        J_{\iota,\omega}(\theta_T)]
        &\leq
        \mathcal P_{\iota,\omega,t_0,T}^{\mathrm{SE}}
        \mathbb E[
        J_{\iota,\omega,\Pi_0}^\star
        -
        J_{\iota,\omega}(\theta_{t_0})]
        \notag\\
        &\quad+
        \frac{\overline{\mathcal R}_{\iota,\omega}}
        {\delta^2(\iota\omega_{\min}-\overline L_J)}
        \left(
        1-\mathcal P_{\iota,\omega,t_0,T}^{\mathrm{SE}}
        \right).
        \label{eq:strong-entropy-explicit-gap-bound}
    \end{align}

    The preceding display is the exact finite-time statement.  The following
    two stepsize regimes expose the main dependences.
    \begin{enumerate}
        \item \textbf{Fixed stepsizes.}
        Take $t_0=0$ and suppose the actor and critic stepsizes are constant on
        the horizon, $\alpha_t=\alpha$ and $\beta_t=\beta$.  Choose a fixed
        mixing window $L\ge1$ by setting $\tau_t=\min\{t,L\}$, and use the same
        window in the moving-target critic recursion.  Assume the fixed-window
        increment bounds
        \[
            \|\theta_{t+1}-\theta_t\|_2\le C_\theta\alpha,
            \qquad
            \|w_i^{t+1}-w_i^t\|_2\le C_{w,i}\beta
        \]
        and the fixed-window critic estimate used in
        Theorem~\ref{thm:unreg-finite-time-stationarity}, with
        $\lambda_\chi$ defined in \eqref{eq:overall-lambda-chi-def}.  Let
        \begin{equation}
            \delta_{\alpha,\beta}:=\max\{\alpha,\beta\},
            \qquad
            L_{\alpha,\beta}
            :=
            \left\lceil
            \frac{\log(1/\delta_{\alpha,\beta})}
            {|\log\rho_{\mathrm{mix}}|}
            \right\rceil .
            \label{eq:strong-entropy-stepsize-window-def}
        \end{equation}
        If $0<\delta_{\alpha,\beta}\le e^{-1}$, $L=L_{\alpha,\beta}$, and
        $\alpha L_{\alpha,\beta}\le1$, $\beta L_{\alpha,\beta}\le1$, then
        \begin{align}
            \mathbb E[
            J_{\iota,\omega,\Pi_0}^\star
            -
            J_{\iota,\omega}(\theta_T)]
            &=
            O\Bigg(
            e^{-\eta_{\iota,\omega}\alpha T}
            \Delta_{\iota,\omega,0}
            \notag\\
            &\qquad\quad+
            \frac{1}{\eta_{\iota,\omega}}
            \bigg[
            e^{-\lambda_\chi\beta T}\mathbb E\mathcal V_0
            +
            \beta\log(1/\delta_{\alpha,\beta})
            \notag\\
            &\qquad\qquad\quad+
            \alpha\log^2(1/\delta_{\alpha,\beta})
            +
            \frac{\alpha^2}{\beta^2}
            +
            \varepsilon_{\mathrm{FA}}^2
            +
            \rho^{2(k+1)}
            +
            \alpha
            \bigg]
            \Bigg).
            \label{eq:strong-entropy-fixed-stepsize-gap-bigO}
        \end{align}
        Thus the regularized optimality gap contracts geometrically in the
        actor horizon up to a constant-stepsize neighborhood.

        \item \textbf{Diminishing stepsizes.}
        Fix exponents and constants satisfying
        \[
            \frac12<\mathfrak b<1,
            \qquad
            \frac{1+\mathfrak b}{2}<\mathfrak a<1,
            \qquad
            \alpha_0,\beta_0>0,
        \]
        and use the polynomial stepsizes
        \[
            \beta_t=\frac{\beta_0}{(t+1)^{\mathfrak b}},
            \qquad
            \alpha_t=\frac{\alpha_0}{(t+1)^{\mathfrak a}}.
        \]
        Use the logarithmic proof window
        \[
            \tau_t=\left\lceil c_\tau\log(t+1)\right\rceil,
            \qquad
            c_\tau>\frac{\mathfrak b}{-\log\rho_{\mathrm{mix}}}.
        \]
        These are the same decaying conditions as
        \eqref{eq:overall-decaying-exponent-region}--\eqref{eq:overall-decaying-window}.
        Fix a finite burn-in time $t_0$ after which
        $1\le\tau_t\le t$, the hypotheses of
        Corollary~\ref{cor:moving_actor_time_varying_beta}, and the regularized
        actor increment bounds in
        \eqref{eq:fixed-state-regularized-actor-increment} hold.  For $T>t_0$,
        let
        \begin{equation}
            A_{t_0,T}:=\sum_{t=t_0}^{T-1}\alpha_t
            =
            \Theta(T^{1-\mathfrak a}).
            \label{eq:strong-entropy-decaying-A-order}
        \end{equation}
        Then
        \begin{align}
            \mathbb E\Delta_{\iota,\omega,T}
            &=
            O\left(
            e^{-\eta_{\iota,\omega}A_{t_0,T}}
            \mathbb E\Delta_{\iota,\omega,t_0}
            \right.
            \notag\\
            &\qquad\left.
            +
            \frac{1}{\eta_{\iota,\omega}}
            \left[
            T^{-(1-\mathfrak a)}
            +
            \varepsilon_{\mathrm{FA}}^2
            +
            \rho^{2(k+1)}
            \right]
            \right).
            \label{eq:strong-entropy-decaying-stepsize-gap-bigO}
        \end{align}
        Equivalently, the optimization transient is
        $\exp(-\Omega_{\iota,\omega}(T^{1-\mathfrak a}))
        \mathbb E\Delta_{\iota,\omega,t_0}$.
        Moreover, for the concrete choice
        \[
            \mathfrak b=\frac12+\mathfrak c,
            \qquad
            \mathfrak a=\frac34+\mathfrak c,
            \qquad
            0<\mathfrak c<\frac14,
        \]
        the same schedules give $A_{t_0,T}=\Theta(T^{1/4-\mathfrak c})$ and the
        sharper bound
        \begin{align}
            \mathbb E\Delta_{\iota,\omega,T}
            &=
            O\left(
            e^{-\eta_{\iota,\omega}A_{t_0,T}}
            \mathbb E\Delta_{\iota,\omega,t_0}
            \right.
            \notag\\
            &\qquad\left.
            +
            \frac{1}{\eta_{\iota,\omega}}
            \left[
            T^{-1/2}\log T
            +
            \varepsilon_{\mathrm{FA}}^2
            +
            \rho^{2(k+1)}
            \right]
            \right).
            \label{eq:strong-entropy-decaying-specialized-gap-bigO}
        \end{align}
        The corresponding optimization transient is
        $\exp(-\Omega_{\iota,\omega}(T^{1/4-\mathfrak c}))
        \mathbb E\Delta_{\iota,\omega,t_0}$.
    \end{enumerate}
    The hidden constants in the two big-$O$ bounds are independent of
    $T,k,\varepsilon_{\mathrm{FA}}$, and in the fixed-stepsize case also of
    $\alpha,\beta$.  In the fixed-stepsize bound, the terms inside the brackets
    correspond to the critic transient, Markovian mixing and fixed-window drift,
    critic target motion, feature approximation, exponential spatial
    localization, and actor sampling variance, respectively.
\end{theorem}

\begin{proof}
    \leavevmode\paragraph{Exact PL recursion.}
    Lemma~\ref{lem:strong-entropy-softmax-pl} supplies
    \eqref{eq:fixed-state-product-pl-optional} with
    $\mu_{\iota,\omega}=\mu_{\iota,\omega,\theta}$ and
    $J_{\iota,\omega}^\star=J_{\iota,\omega,\Pi_0}^\star$.  Substituting this
    PL inequality into
    Lemma~\ref{lem:fixed-state-regularized-actor-drift} gives
    \eqref{eq:strong-entropy-gap-recursion}.  Taking total expectation and
    iterating the affine recursion gives
    \eqref{eq:strong-entropy-unrolled-gap}.  Finally,
    \[
        \sum_{\ell=t_0}^{T-1}
        \alpha_\ell
        \prod_{r=\ell+1}^{T-1}
        \left(1-\frac{\alpha_r\mu_{\iota,\omega,\theta}}{2}\right)
        \leq
        \frac{2}{\mu_{\iota,\omega,\theta}}
        \left(1-\mathcal P_{\iota,\omega,t_0,T}^{\mathrm{SE}}\right)
        =
        \frac{1}{\delta^2(\iota\omega_{\min}-\overline L_J)}
        \left(1-\mathcal P_{\iota,\omega,t_0,T}^{\mathrm{SE}}\right),
    \]
    and the uniform error condition
    \eqref{eq:strong-entropy-uniform-error} proves
    \eqref{eq:strong-entropy-explicit-gap-bound}.

    \leavevmode\paragraph{Fixed stepsizes.}
    With
    $\alpha_t=\alpha$,
    \[
        \mathcal P_{\iota,\omega,0,T}^{\mathrm{SE}}
        =
        (1-\eta_{\iota,\omega}\alpha)^T
        \le
        e^{-\eta_{\iota,\omega}\alpha T}.
    \]
    The fixed-window critic recursion quoted from
    Theorem~\ref{thm:unreg-finite-time-stationarity} gives, for the
    geometrically weighted residuals in
    \eqref{eq:strong-entropy-unrolled-gap},
    \[
        \sum_{\ell=0}^{T-1}
        \alpha
        (1-\eta_{\iota,\omega}\alpha)^{T-1-\ell}
        \mathbb E\mathcal V_\ell
        =
        O\left(
        \frac{
        e^{-\lambda_\chi\beta T}\mathbb E\mathcal V_0
        +
        \beta
        +
        \mathfrak e_{\mathrm{crit}}(L)
        +
        \alpha^2/\beta^2}
        {\eta_{\iota,\omega}}
        \right).
    \]
    The same fixed-window estimates bound the geometrically weighted
    contribution of the actor-mixing terms by
    $O((L/T+\rho_{\mathrm{mix}}^L+\alpha L^2)/\eta_{\iota,\omega})$, and the
    weighted contribution of the window-drift terms by
    $O(L^2(\alpha^2+\beta^2)/\eta_{\iota,\omega})$.  Substituting these
    estimates into the definition of
    $\mathcal R_{\iota,\omega,t}$ in
    \eqref{eq:fixed-state-one-step-error-aggregate}, using the window choice
    $L=L_{\alpha,\beta}$, and using
    $\alpha L_{\alpha,\beta}\le1$, $\beta L_{\alpha,\beta}\le1$, yields
    \[
        \mathfrak e_{\mathrm{crit}}(L_{\alpha,\beta})
        =
        O\left(
        \beta\log(1/\delta_{\alpha,\beta})
        +
        \alpha\log^2(1/\delta_{\alpha,\beta})
        \right),
    \]
    and the same order for the actor mixing/window terms.  The remaining
    pieces of $\mathcal R_{\iota,\omega,t}$ are exactly
    $\varepsilon_{\mathrm{FA}}^2$, $\rho^{2(k+1)}$, and the variance term
    $C_{\iota,\omega,\mathrm{var}}^{\mathrm{act}}\alpha$.  Combining these
    bounds with \eqref{eq:strong-entropy-unrolled-gap} proves
    \eqref{eq:strong-entropy-fixed-stepsize-gap-bigO}.

    \leavevmode\paragraph{Diminishing stepsizes.}
    The product term in \eqref{eq:strong-entropy-unrolled-gap} satisfies
    \begin{equation}
        \mathcal P_{\iota,\omega,t_0,T}^{\mathrm{SE}}
        \le
        \exp\left(-\eta_{\iota,\omega}A_{t_0,T}\right),
        \label{eq:strong-entropy-decaying-product-bound}
    \end{equation}
    and \eqref{eq:strong-entropy-decaying-A-order} follows from
    \eqref{eq:overall-decaying-stepsizes}.  We next bound the residual
    convolution in \eqref{eq:strong-entropy-unrolled-gap}.  For
    $t\ge t_0$, Corollary~\ref{cor:moving_actor_time_varying_beta}, together
    with the moving-error estimate
    \eqref{eq:overall-decaying-moving-error-order}, gives
    \[
        v_{t+1}
        \le
        (1-\lambda_\chi\beta_t)v_t
        +
        O\left(
        \beta_t^2
        +
        \beta_t\varepsilon_{P,t}^{\mathrm{mov}}
        +
        \frac{\alpha_t^2}{\beta_t}
        \right),
        \qquad
        v_t:=\mathbb E\mathcal V_t.
    \]
    The polynomial schedules and logarithmic window make the additive
    perturbation slowly varying relative to the contraction scale.  A standard
    scalar comparison therefore yields the pointwise critic-tracking estimate
    \begin{equation}
        \mathbb E\mathcal V_t
        =
        O\left(
        e^{-c(t^{1-\mathfrak b}-t_0^{1-\mathfrak b})}
        \mathbb E\mathcal V_{t_0}
        +
        \beta_t\log(t+1)
        +
        \alpha_t\log^2(t+1)
        +
        \frac{\alpha_t^2}{\beta_t^2}
        \right),
        \label{eq:strong-entropy-decaying-pointwise-critic}
    \end{equation}
    for some $c>0$ independent of $T,k,\varepsilon_{\mathrm{FA}}$.
    Since $\mathfrak a>\mathfrak b$, the
    $\alpha_t\log^2(t+1)$ term is absorbed by
    $\beta_t\log(t+1)$ for all sufficiently large $t$.

    The regularized one-step residual
    \eqref{eq:fixed-state-one-step-error-aggregate} uses the same critic
    tracking, actor-mixing, and window-drift quantities as the unregularized
    decaying-step proof.  Thus
    \eqref{eq:overall-decaying-actor-mixing-order},
    \eqref{eq:actor-window-drift-defs}, and
    \eqref{eq:strong-entropy-decaying-pointwise-critic} imply
    \begin{align}
        \mathbb E[
        \mathcal R_{\iota,\omega,t}
        +
        C_{\iota,\omega,\mathrm{var}}^{\mathrm{act}}\alpha_t]
        &=
        O\left(
        \varepsilon_{\mathrm{FA}}^2
        +
        \rho^{2(k+1)}
        +
        \beta_t\log(t+1)
        \right.
        \notag\\
        &\qquad\left.
        +
        \frac{\alpha_t^2}{\beta_t^2}
        +
        \alpha_t
        \right).
        \label{eq:strong-entropy-decaying-residual-order}
    \end{align}
    Here we used
    \eqref{eq:fixed-state-delta-loc-def} to replace
    $\delta_{\mathrm{loc}}(k)^2$ by $O(\rho^{2(k+1)})$; the exponentially
    decaying critic transient in
    \eqref{eq:strong-entropy-decaying-pointwise-critic} is smaller than
    $\alpha_t$ after increasing the burn-in constant.

    It remains to convolve the decaying residuals with the PL weights.  If
    $q_t$ is any one of
    $\beta_t\log(t+1)$, $\alpha_t^2/\beta_t^2$, or $\alpha_t$, then the exponent
    conditions in \eqref{eq:overall-decaying-exponent-region} imply
    $q_t=O((t+1)^{-s}\log^m(t+1))$ for some $s>1-\mathfrak a$ and
    $m\in\{0,1\}$.  Splitting the sum in
    \eqref{eq:strong-entropy-unrolled-gap} at
    $T-\lceil C T^{\mathfrak a}\log T\rceil$ and using
    $\sum_{r=\ell+1}^{T-1}\alpha_r=\Omega(T^{\mathfrak a}\log T\cdot
    T^{-\mathfrak a})$ on the early part gives the standard PL-convolution
    comparison
    \begin{equation}
        \sum_{\ell=t_0}^{T-1}
        \alpha_\ell
        \prod_{r=\ell+1}^{T-1}
        (1-\eta_{\iota,\omega}\alpha_r)
        q_\ell
        =
        O\left(
        \frac{T^{-(1-\mathfrak a)}}{\eta_{\iota,\omega}}
        \right).
        \label{eq:strong-entropy-decaying-pl-convolution}
    \end{equation}
    The constant residuals
    $\varepsilon_{\mathrm{FA}}^2+\rho^{2(k+1)}$ contribute at most
    $(\varepsilon_{\mathrm{FA}}^2+\rho^{2(k+1)})/\eta_{\iota,\omega}$ by the
    same summation argument used for
    \eqref{eq:strong-entropy-explicit-gap-bound}.  Combining
    \eqref{eq:strong-entropy-decaying-product-bound},
    \eqref{eq:strong-entropy-decaying-residual-order}, and
    \eqref{eq:strong-entropy-decaying-pl-convolution} in
    \eqref{eq:strong-entropy-unrolled-gap} proves
    \eqref{eq:strong-entropy-decaying-stepsize-gap-bigO}.

    It remains to prove the sharper concrete specialization.  Set
    $\mathfrak b=1/2+\mathfrak c$ and
    $\mathfrak a=3/4+\mathfrak c$ with
    $\mathfrak c\in(0,1/4)$.  Then
    \[
        \beta_t\log(t+1)
        =
        O((t+1)^{-1/2-\mathfrak c}\log(t+1)),
        \qquad
        \frac{\alpha_t^2}{\beta_t^2}
        =
        O((t+1)^{-1/2}),
    \]
    and $\alpha_t=O((t+1)^{-3/4-\mathfrak c})$.  Hence the decaying part of
    \eqref{eq:strong-entropy-decaying-residual-order} is
    \begin{equation}
        q_t
        :=
        \beta_t\log(t+1)
        +
        \frac{\alpha_t^2}{\beta_t^2}
        +
        \alpha_t
        =
        O((t+1)^{-1/2}\log(t+1)).
        \label{eq:strong-entropy-special-residual-order}
    \end{equation}
    We prove the corresponding PL-convolution estimate.  Let
    \[
        S_T(q)
        :=
        \sum_{\ell=t_0}^{T-1}
        \alpha_\ell
        \prod_{r=\ell+1}^{T-1}
        (1-\eta_{\iota,\omega}\alpha_r)
        q_\ell .
    \]
    Because $0<\eta_{\iota,\omega}\alpha_t\le1$,
    \begin{equation}
        \sum_{\ell=t_0}^{T-1}
        \alpha_\ell
        \prod_{r=\ell+1}^{T-1}
        (1-\eta_{\iota,\omega}\alpha_r)
        \le
        \frac{1}{\eta_{\iota,\omega}}.
        \label{eq:strong-entropy-pl-weight-mass}
    \end{equation}
    Fix
    $N_T:=\lceil 2T^{\mathfrak a}\log T/
    (\eta_{\iota,\omega}\alpha_0)\rceil$.  For all sufficiently large $T$,
    $N_T\le T/2$.  On the last $N_T$ indices,
    \eqref{eq:strong-entropy-special-residual-order} gives
    $q_\ell=O(T^{-1/2}\log T)$, and therefore
    \begin{equation}
        \sum_{\ell=T-N_T}^{T-1}
        \alpha_\ell
        \prod_{r=\ell+1}^{T-1}
        (1-\eta_{\iota,\omega}\alpha_r)
        q_\ell
        =
        O\left(
        \frac{T^{-1/2}\log T}{\eta_{\iota,\omega}}
        \right).
        \label{eq:strong-entropy-special-late-convolution}
    \end{equation}
    On the earlier indices $\ell<T-N_T$, the final $N_T$ actor steps satisfy
    \[
        \sum_{r=T-N_T}^{T-1}\alpha_r
        \ge
        \alpha_0N_TT^{-\mathfrak a}
        \ge
        \frac{2\log T}{\eta_{\iota,\omega}},
    \]
    so
    \[
        \prod_{r=\ell+1}^{T-1}
        (1-\eta_{\iota,\omega}\alpha_r)
        \le
        \exp\left(
        -\eta_{\iota,\omega}
        \sum_{r=T-N_T}^{T-1}\alpha_r
        \right)
        \le
        T^{-2}.
    \]
    Since
    $\sum_{\ell=t_0}^{\infty}\alpha_\ell q_\ell<\infty$
    under $\mathfrak a=3/4+\mathfrak c$, the early part is $O(T^{-2})$.
    Combining the early and late parts,
    \begin{equation}
        S_T(q)
        =
        O\left(
        \frac{T^{-1/2}\log T}{\eta_{\iota,\omega}}
        \right).
        \label{eq:strong-entropy-special-pl-convolution}
    \end{equation}
    The constant residuals again contribute at most
    $(\varepsilon_{\mathrm{FA}}^2+\rho^{2(k+1)})/\eta_{\iota,\omega}$ by
    \eqref{eq:strong-entropy-pl-weight-mass}.  Substituting
    \eqref{eq:strong-entropy-special-pl-convolution} into
    \eqref{eq:strong-entropy-unrolled-gap} proves
    \eqref{eq:strong-entropy-decaying-specialized-gap-bigO}.
\end{proof}

\begin{remark}
    The theorem proves convergence to the entropy-regularized optimum over the
    local product-policy region $\Pi_0$.  It does not assert optimality over
    unrestricted centralized policies.  The difference is essential in the
    networked setting: the decentralized actor only changes local policy
    tables, and the strong-entropy condition above is imposed on that
    product-policy landscape.
\end{remark}

\section{Auxiliary Lemmas}
\label{sec:auxiliary-lemmas}

\begin{lemma}[Product-policy score identities, Lemma~\ref{lem:product-policy-score-identities} restated]
    \label{auxproof:lem:product-policy-score-identities}
    Under the product policy
    $\pi_\theta(a_g|s_g)=\prod_{i=1}^n\pi_{\theta_i}(a_i|s_i)$, for every
    agent $i\in\mathcal{N}$,
    \begin{equation*}
        \nabla_{\theta_i}\log\pi_\theta(a_g|s_g)
        =
        \nabla_{\theta_i}\log\pi_{\theta_i}(a_i|s_i).
    \end{equation*}
    Moreover, under the softmax parameterization
    \[
        \pi_{\theta_i}(a_i|s_i)
        =
        \frac{\exp(\theta_i(s_i,a_i))}
        {\sum_{a_i'\in\mathcal{A}_i}\exp(\theta_i(s_i,a_i'))},
    \]
    for every $a\in\mathcal{A}_i$,
    \begin{equation*}
        \frac{\partial}{\partial \theta_i(s_i,a)}
        \log\pi_{\theta_i}(a_i|s_i)
        =
        \mathbb{I}\{a_i=a\}-\pi_{\theta_i}(a|s_i).
    \end{equation*}
    In particular, for finite action spaces, the local softmax score is
    uniformly bounded as
    \begin{equation}
        \left\|
        \nabla_{\theta_i}\log\pi_{\theta_i}(a_i|s_i)
        \right\|_2
        \leq
        \sqrt{2}
        \qquad
        \forall \theta_i,s_i,a_i.
        \label{auxproof:eq:softmax-score-sqrt-two-bound}
    \end{equation}
    Moreover, it has zero conditional mean:
    \begin{equation}
        \sum_{a_i\in\mathcal A_i}
        \pi_{\theta_i}(a_i|s_i)
        \nabla_{\theta_i}\log\pi_{\theta_i}(a_i|s_i)
        =
        0.
        \label{auxproof:eq:softmax-score-zero-mean}
    \end{equation}
\end{lemma}

\begin{proof}
    The product-policy identity follows from
    \[
        \log\pi_\theta(a_g|s_g)
        =
        \sum_{j=1}^n
        \log\pi_{\theta_j}(a_j|s_j),
    \]
    because only the $j=i$ term depends on $\theta_i$.

    For the softmax coordinate formula, fix $s_i$ and write
    $Z_i(s_i):=\sum_{u\in\mathcal A_i}\exp(\theta_i(s_i,u))$.  Then
    \[
        \log\pi_{\theta_i}(a_i|s_i)
        =
        \theta_i(s_i,a_i)-\log Z_i(s_i).
    \]
    Differentiating with respect to the logit $\theta_i(s_i,a)$ gives
    \[
        \frac{\partial}{\partial \theta_i(s_i,a)}
        \log\pi_{\theta_i}(a_i|s_i)
        =
        \mathbb I\{a_i=a\}
        -
        \frac{\exp(\theta_i(s_i,a))}{Z_i(s_i)}
        =
        \mathbb I\{a_i=a\}-\pi_{\theta_i}(a|s_i).
    \]
    Coordinates corresponding to states different from $s_i$ have derivative
    zero, so the displayed expression gives the whole tabular score vector.

    Let $p(\cdot):=\pi_{\theta_i}(\cdot|s_i)$ and let $e_{a_i}$ be the unit
    vector at action $a_i$.  The nonzero block of the score is $e_{a_i}-p$.
    Hence
    \begin{align*}
        \|e_{a_i}-p\|_2^2
        &=
        \|e_{a_i}\|_2^2
        -2\langle e_{a_i},p\rangle
        +\|p\|_2^2 \\
        &=
        1-2p(a_i)+\sum_{a\in\mathcal A_i}p(a)^2
        \leq
        1+\sum_{a\in\mathcal A_i}p(a)^2
        \leq
        2,
    \end{align*}
    which proves \eqref{auxproof:eq:softmax-score-sqrt-two-bound}.  Finally,
    \[
        \sum_{a_i}p(a_i)(e_{a_i}-p)
        =
        p-p\sum_{a_i}p(a_i)
        =
        0,
    \]
    proving \eqref{auxproof:eq:softmax-score-zero-mean}.
\end{proof}

\begin{lemma}[Softmax score and kernel Lipschitz constants, Lemma~\ref{lem:softmax-score-kernel-lipschitz} restated]
    \label{auxproof:lem:softmax-score-kernel-lipschitz}
    Suppose each local policy is the finite-action tabular softmax policy
    \begin{equation*}
        \pi_{\theta_i}(a_i|s_i)
        =
        \frac{\exp(\theta_i(s_i,a_i))}
        {\sum_{u\in\mathcal A_i}\exp(\theta_i(s_i,u))},
        \qquad
        \pi_\theta(a_g|s_g)=\prod_{i=1}^n\pi_{\theta_i}(a_i|s_i).
    \end{equation*}
    Define the local score
    $S_i(\theta_i;s_i,a_i):=\nabla_{\theta_i}\log\pi_{\theta_i}(a_i|s_i)$
    and the global score
    $S(\theta;s_g,a_g):=\nabla_\theta\log\pi_\theta(a_g|s_g)$.
    Then, for every $i,\theta_i,s_i,a_i$,
    \begin{equation}
        \|S_i(\theta_i;s_i,a_i)\|_2\le G_i,
        \qquad
        G_i:=\sqrt2,
        \label{auxproof:eq:actor-score-bound-new}
    \end{equation}
    and, for every $\theta,s_g,a_g$,
    \begin{equation*}
        \|S(\theta;s_g,a_g)\|_2\le G,
        \qquad
        G:=\sqrt{\sum_{i=1}^nG_i^2}=\sqrt{2n}.
    \end{equation*}
    Moreover, for every $i,s_i,a_i,\theta_i,\theta_i'$,
    \begin{equation}
        \left\|
        S_i(\theta_i;s_i,a_i)
        -
        S_i(\theta_i';s_i,a_i)
        \right\|_2
        \le
        L_{S,i}\|\theta_i-\theta_i'\|_2,
        \qquad
        L_{S,i}:=1 .
        \label{auxproof:eq:actor-score-lipschitz-new}
    \end{equation}
    Finally, define the frozen global state-action kernel
    \begin{equation*}
        \textcolor{revisionblue}{P_{\pi_\theta}}((s_g,a_g),(s_g',a_g'))
        :=
        P(s_g'|s_g,a_g)\pi_\theta(a_g'|s_g').
    \end{equation*}
    For every $\theta,\theta'$,
    \begin{equation}
        \sup_x
        \|\textcolor{revisionblue}{P_{\pi_\theta}}(x,\cdot)-\textcolor{revisionblue}{P_{\pi_{\theta'}}}(x,\cdot)\|_{\mathrm{TV}}
        \leq
        L_K\|\theta-\theta'\|_2,
        \qquad
        L_K:=G=\sqrt{2n}.
        \label{auxproof:eq:actor-kernel-lipschitz-new}
    \end{equation}
\end{lemma}

\begin{proof}
    Fix $i,s_i$ and write
    $p_{\theta_i}(\cdot):=\pi_{\theta_i}(\cdot|s_i)$.  By
    Lemma~\ref{auxproof:lem:product-policy-score-identities}, the only nonzero
    block of $S_i(\theta_i;s_i,a_i)$ is
    \begin{equation}
        e_{a_i}-p_{\theta_i},
        \label{auxproof:eq:softmax-local-score-vector}
    \end{equation}
    where $e_{a_i}$ is the unit vector on $\mathcal A_i$ at $a_i$.  Therefore
    \begin{align*}
        \|S_i(\theta_i;s_i,a_i)\|_2^2
        &=
        \|e_{a_i}-p_{\theta_i}\|_2^2
        \notag\\
        &=
        1-2p_{\theta_i}(a_i)
        +\sum_{a\in\mathcal A_i}p_{\theta_i}(a)^2
        \leq
        2,
    \end{align*}
    proving \eqref{auxproof:eq:actor-score-bound-new}.  The product-policy
    identity gives
    \begin{equation}
        S(\theta;s_g,a_g)
        =
        \left(
        S_1(\theta_1;s_1,a_1),
        \ldots,
        S_n(\theta_n;s_n,a_n)
        \right),
        \label{auxproof:eq:global-score-concatenation}
    \end{equation}
    and hence
    \begin{equation*}
        \|S(\theta;s_g,a_g)\|_2^2
        =
        \sum_{i=1}^n\|S_i(\theta_i;s_i,a_i)\|_2^2
        \leq
        \sum_{i=1}^nG_i^2
        =
        2n.
    \end{equation*}

    Fix $s_i,a_i$.  By the softmax score formula in
    Lemma~\ref{auxproof:lem:product-policy-score-identities},
    \begin{equation*}
        S_i(\theta_i;s_i,a_i)-S_i(\theta_i';s_i,a_i)
        =
        p_{\theta_i'}-p_{\theta_i}
    \end{equation*}
    on the block corresponding to $s_i$, and the difference is zero on all
    other state blocks.  The softmax Jacobian on the $s_i$ block is
    \begin{equation}
        \nabla_{\theta_i(s_i,\cdot)}p_{\theta_i}
        =
        \operatorname{diag}(p_{\theta_i})-p_{\theta_i}p_{\theta_i}^{\top}.
        \label{auxproof:eq:softmax-jacobian}
    \end{equation}
    For any vector $v\in\mathbb R^{|\mathcal A_i|}$,
    \begin{align*}
        v^\top
        \left(
        \operatorname{diag}(p_{\theta_i})-p_{\theta_i}p_{\theta_i}^{\top}
        \right)
        v
        &=
        \sum_{a}p_{\theta_i}(a)v(a)^2
        -
        \left(\sum_a p_{\theta_i}(a)v(a)\right)^2
        \notag\\
        &\leq
        \sum_{a}p_{\theta_i}(a)v(a)^2
        \leq
        \|v\|_2^2 .
    \end{align*}
    Hence
    $\|\operatorname{diag}(p_{\theta_i})-p_{\theta_i}p_{\theta_i}^{\top}\|_2
    \leq1$.  The mean-value theorem gives
    \begin{align*}
        \|p_{\theta_i'}-p_{\theta_i}\|_2
        &\leq
        \|\theta_i(s_i,\cdot)-\theta_i'(s_i,\cdot)\|_2
        \notag\\
        &\leq
        \|\theta_i-\theta_i'\|_2,
    \end{align*}
    which proves
    \eqref{auxproof:eq:actor-score-lipschitz-new}.

    For the kernel bound, for any current state-action pair
    $x=(s_g,a_g)$,
    \begin{equation}
        \textcolor{revisionblue}{P_{\pi_\theta}}(x,(s_g',a_g'))
        =
        P(s_g'|s_g,a_g)\pi_\theta(a_g'|s_g').
        \label{auxproof:eq:frozen-kernel-expanded}
    \end{equation}
    Since the environment transition $P(s_g'|s_g,a_g)$ does not depend on
    $\theta$, only the next-action distribution changes.  For every
    $(s_g',a_g')$,
    \begin{equation}
        \nabla_\theta\pi_\theta(a_g'|s_g')
        =
        \pi_\theta(a_g'|s_g')S(\theta;s_g',a_g').
        \label{auxproof:eq:product-policy-gradient-probability}
    \end{equation}
    Therefore
    \begin{equation}
        \sum_{a_g'}
        \|\nabla_\theta\pi_\theta(a_g'|s_g')\|_2
        =
        \sum_{a_g'}
        \pi_\theta(a_g'|s_g')
        \|S(\theta;s_g',a_g')\|_2
        \leq
        G .
        \label{auxproof:eq:action-distribution-gradient-l1-bound}
    \end{equation}
    Applying the mean-value theorem to the map
    $\theta\mapsto\pi_\theta(\cdot|s_g')$ and using
    \eqref{auxproof:eq:action-distribution-gradient-l1-bound} gives
    \begin{equation}
        \sum_{a_g'}
        |\pi_\theta(a_g'|s_g')-\pi_{\theta'}(a_g'|s_g')|
        \leq
        G\|\theta-\theta'\|_2.
        \label{auxproof:eq:product-action-l1-lipschitz}
    \end{equation}
    Hence, under the total-variation convention used in the paper,
    \begin{align*}
        \|\textcolor{revisionblue}{P_{\pi_\theta}}(x,\cdot)-\textcolor{revisionblue}{P_{\pi_{\theta'}}}(x,\cdot)\|_{\mathrm{TV}}
        &\leq
        \sum_{s_g'}P(s_g'|s_g,a_g)
        \|\pi_\theta(\cdot|s_g')-\pi_{\theta'}(\cdot|s_g')\|_{\mathrm{TV}}
        \notag\\
        &\leq
        \sum_{s_g'}P(s_g'|s_g,a_g)
        G\|\theta-\theta'\|_2
        =
        G\|\theta-\theta'\|_2.
    \end{align*}
    Taking the supremum over $x$ proves
    \eqref{auxproof:eq:actor-kernel-lipschitz-new}.
\end{proof}

\begin{lemma}[Conditional Markovian mixing error, Lemma~\ref{lem:conditional-markovian-mixing-error} restated]
    \label{auxproof:lem:conditional-markovian-mixing-error}
    Let $\Theta$ be the actor-parameter set considered in the actor proof.  For
    each $\theta\in\Theta$, define the frozen global state-action kernel
    \begin{equation*}
        \textcolor{revisionblue}{P_{\pi_\theta}}((s_g,a_g),(s_g',a_g'))
        =
        P(s_g'|s_g,a_g)\pi_\theta(a_g'|s_g').
    \end{equation*}
    Suppose there exist constants
    $m_{\mathrm{mix}}\geq1$, $\rho_{\mathrm{mix}}\in(0,1)$, and
    $L_K<\infty$ such that, for all $\theta,\theta'\in\Theta$ and all
    $\ell\ge0$,
    \begin{align}
        \sup_x\|\textcolor{revisionblue}{P_{\pi_\theta}}^\ell(x,\cdot)-\nu^\theta\|_{\mathrm{TV}}
        &\leq
        m_{\mathrm{mix}}\rho_{\mathrm{mix}}^\ell,
        \label{auxproof:eq:aux-uniform-geometric-mixing}\\
        \sup_x\|\textcolor{revisionblue}{P_{\pi_\theta}}(x,\cdot)-\textcolor{revisionblue}{P_{\pi_{\theta'}}}(x,\cdot)\|_{\mathrm{TV}}
        &\leq
        L_K\|\theta-\theta'\|,
        \label{auxproof:eq:aux-kernel-lipschitz-for-mixing}
    \end{align}
    Define
    \begin{equation}
        L_\nu
        :=
        \frac{2m_{\mathrm{mix}}L_K}{1-\rho_{\mathrm{mix}}}.
        \label{auxproof:eq:actor-stationary-lipschitz-constant}
    \end{equation}
    Then
    \begin{equation}
        \|\nu^\theta-\nu^{\theta'}\|_{\mathrm{TV}}
        \leq
        L_\nu\|\theta-\theta'\|.
        \label{auxproof:eq:stationary-lipschitz-for-mixing}
    \end{equation}

    Let $X_t:=(s_g(t),a_g(t))$ and let $\mathcal H_t$ be the history through
    sample $X_t$ and all updates up to time $t$.  Fix a time $t$, a window
    $\tau_t\le t$, and write $s:=t-\tau_t$.  Suppose that, conditional on
    $\mathcal H_s$, the adaptive trajectory over the window is generated by
    \begin{equation}
        \mathbb P(X_{\ell+1}\in A\mid\mathcal H_\ell)
        =
        \textcolor{revisionblue}{P_{\pi_{\vartheta_\ell}}}(X_\ell,A),
        \qquad
        \ell=s,\ldots,t-1,
        \label{auxproof:eq:adaptive-window-kernel-condition}
    \end{equation}
    for $\mathcal H_\ell$-measurable parameters
    $\vartheta_\ell\in\Theta$.  In the linear-FA algorithm with
    $\mathcal H_s$ interpreted after the time-$s$ updates, the effective choice
    is $\vartheta_\ell=\theta_{\ell+1}$; under a pre-update convention it is
    $\vartheta_\ell=\theta_\ell$.

    Then, for every $\mathcal H_s$-measurable test function
    $\phi_t:\mathcal S_g\times\mathcal A_g\to\mathbb R^d$ and every
    nonnegative $\mathcal H_s$-measurable random variable $B_t$ satisfying
    $\sup_x\|\phi_t(x)\|\le B_t$, the delayed conditional bias satisfies
    \begin{equation*}
        \left\|
        \mathbb{E}[\phi_t(X_t)\mid\mathcal{H}_{t-\tau_t}]
        -
        \mathbb{E}_{\nu^{\theta_t}}[\phi_t(X)]
        \right\|
        \leq
        B_t\varepsilon_{\mathrm{mix}}(t),
    \end{equation*}
    where
    \begin{equation}
        \bar\varepsilon_{\mathrm{mix}}(t)
        :=
        m_{\mathrm{mix}}\rho_{\mathrm{mix}}^{\tau_t}
        +
        L_K
        \sum_{\ell=t-\tau_t}^{t-1}
        \|\vartheta_\ell-\theta_{t-\tau_t}\|
        +
        L_\nu
        \sum_{\ell=t-\tau_t}^{t-1}
        \|\theta_{\ell+1}-\theta_\ell\|.
        \label{auxproof:eq:actor-epsilon-mix-raw-definition}
    \end{equation}
    With the paper's normalization of total variation, distances between
    probability laws are at most one; hence the same bound holds with the
    clipped quantity
    \begin{equation}
        \varepsilon_{\mathrm{mix}}(t)
        :=
        \min\{1,\bar\varepsilon_{\mathrm{mix}}(t)\}.
        \label{auxproof:eq:epsilon-mix-definition}
    \end{equation}
    If, in addition,
    $\|\theta_{\ell+1}-\theta_\ell\|\le C_\theta\alpha_\ell$,
    $\alpha_\ell\le C_\alpha(\ell+1)^{-a}$ for some $a\in(0,1]$, and
    $\vartheta_\ell\in\{\theta_\ell,\theta_{\ell+1}\}$ over the window, then
    the choice $\tau_t=\lceil c_\tau\log(t+2)\rceil$ gives
    \begin{equation*}
        \varepsilon_{\mathrm{mix}}(t)
        \leq
        m_{\mathrm{mix}}(t+2)^{-c_\tau|\log\rho_{\mathrm{mix}}|}
        +
        L_K C_\theta C_\alpha
        \tau_t^2(t-\tau_t+1)^{-a}
        +
        L_\nu C_\theta C_\alpha
        \tau_t(t-\tau_t+1)^{-a},
    \end{equation*}
    and hence
    \begin{equation*}
        \varepsilon_{\mathrm{mix}}(t)
        =
        O\left(t^{-c_\tau|\log\rho_{\mathrm{mix}}|}\right)
        +
        O\left((\log t)^2t^{-a}\right).
    \end{equation*}
\end{lemma}

\begin{proof}
    We first prove the stationary-law Lipschitz bound
    \eqref{auxproof:eq:stationary-lipschitz-for-mixing}.  Fix
    $\theta,\theta'\in\Theta$ and set
    \begin{equation*}
        \Delta K:=\textcolor{revisionblue}{P_{\pi_{\theta'}}}-\textcolor{revisionblue}{P_{\pi_\theta}},
        \qquad
        \delta:=\nu^{\theta'}\Delta K .
    \end{equation*}
    Since $\nu^{\theta'}\textcolor{revisionblue}{P_{\pi_{\theta'}}}=\nu^{\theta'}$, for every integer $T\ge1$,
    \begin{align}
        \nu^{\theta'}-\nu^{\theta'}\textcolor{revisionblue}{P_{\pi_\theta}}^T
        &=
        \sum_{m=0}^{T-1}
        \left(
        \nu^{\theta'}\textcolor{revisionblue}{P_{\pi_\theta}}^m
        -
        \nu^{\theta'}\textcolor{revisionblue}{P_{\pi_\theta}}^{m+1}
        \right)
        \notag\\
        &=
        \sum_{m=0}^{T-1}
        \nu^{\theta'}(\textcolor{revisionblue}{P_{\pi_{\theta'}}}-\textcolor{revisionblue}{P_{\pi_\theta}})\textcolor{revisionblue}{P_{\pi_\theta}}^m
        =
        \sum_{m=0}^{T-1}\delta \textcolor{revisionblue}{P_{\pi_\theta}}^m .
        \label{auxproof:eq:stationary-perturbation-telescope}
    \end{align}
    The signed measure $\delta$ has total mass zero, and the kernel-Lipschitz
    assumption gives
    \begin{equation}
        \|\delta\|_{\mathrm{TV}}
        \leq
        \sup_x\|\textcolor{revisionblue}{P_{\pi_{\theta'}}}(x,\cdot)-\textcolor{revisionblue}{P_{\pi_\theta}}(x,\cdot)\|_{\mathrm{TV}}
        \leq
        L_K\|\theta-\theta'\|.
        \label{auxproof:eq:delta-kernel-lipschitz}
    \end{equation}
    For any scalar test function $f$ with $\|f\|_\infty\le1$,
    \begin{equation}
        \delta \textcolor{revisionblue}{P_{\pi_\theta}}^m f
        =
        \delta\left(\textcolor{revisionblue}{P_{\pi_\theta}}^m f-\nu^\theta f\right),
        \label{auxproof:eq:zero-mass-centering}
    \end{equation}
    because
    \begin{equation*}
        \delta\mathbf 1
        =
        \nu^{\theta'}\textcolor{revisionblue}{P_{\pi_{\theta'}}}\mathbf 1
        -
        \nu^{\theta'}\textcolor{revisionblue}{P_{\pi_\theta}}\mathbf 1
        =
        \nu^{\theta'}\mathbf 1-\nu^{\theta'}\mathbf 1
        =
        0 .
    \end{equation*}
    Uniform geometric ergodicity implies
    \begin{equation}
        \|\textcolor{revisionblue}{P_{\pi_\theta}}^m f-\nu^\theta f\|_\infty
        \leq
        2m_{\mathrm{mix}}\rho_{\mathrm{mix}}^m .
        \label{auxproof:eq:mixing-test-function-bound}
    \end{equation}
    Combining
    \eqref{auxproof:eq:stationary-perturbation-telescope}--\eqref{auxproof:eq:mixing-test-function-bound}
    yields
    \begin{align*}
        \left|
        \left(\nu^{\theta'}-\nu^{\theta'}\textcolor{revisionblue}{P_{\pi_\theta}}^T\right)f
        \right|
        &\leq
        \sum_{m=0}^{T-1}
        |\delta \textcolor{revisionblue}{P_{\pi_\theta}}^m f|
        \notag\\
        &\leq
        2m_{\mathrm{mix}}L_K\|\theta-\theta'\|
        \sum_{m=0}^{T-1}\rho_{\mathrm{mix}}^m .
    \end{align*}
    Letting $T\to\infty$ and using
    $\nu^{\theta'}\textcolor{revisionblue}{P_{\pi_\theta}}^T\to\nu^\theta$ gives
    \begin{equation*}
        |(\nu^{\theta'}-\nu^\theta)f|
        \leq
        \frac{2m_{\mathrm{mix}}L_K}{1-\rho_{\mathrm{mix}}}
        \|\theta-\theta'\|.
    \end{equation*}
    Taking the supremum over $\|f\|_\infty\le1$ proves
    \eqref{auxproof:eq:stationary-lipschitz-for-mixing}.

    We next prove the delayed sampling bound.  Let
    $\mu_s(\cdot):=\mathbb P(X_s\in\cdot\mid\mathcal H_s)$ and define the
    conditional law of $X_t$ given $\mathcal H_s$ by
    \begin{equation}
        \lambda_t
        :=
        \mu_s\textcolor{revisionblue}{P_{\pi_{\vartheta_s}}}\textcolor{revisionblue}{P_{\pi_{\vartheta_{s+1}}}}\cdots \textcolor{revisionblue}{P_{\pi_{\vartheta_{t-1}}}} .
        \label{auxproof:eq:adaptive-window-law}
    \end{equation}
    This identity follows by iterating
    \eqref{auxproof:eq:adaptive-window-kernel-condition}.  A product
    telescoping identity gives
    \begin{align}
        &\mu_s\textcolor{revisionblue}{P_{\pi_{\vartheta_s}}}\cdots \textcolor{revisionblue}{P_{\pi_{\vartheta_{t-1}}}}
        -
        \mu_s\textcolor{revisionblue}{P_{\pi_{\theta_s}}}^{\tau_t}
        \notag\\
        &\quad=
        \sum_{\ell=s}^{t-1}
        \mu_s
        \textcolor{revisionblue}{P_{\pi_{\vartheta_s}}}\cdots \textcolor{revisionblue}{P_{\pi_{\vartheta_{\ell-1}}}}
        (\textcolor{revisionblue}{P_{\pi_{\vartheta_\ell}}}-\textcolor{revisionblue}{P_{\pi_{\theta_s}}})
        \textcolor{revisionblue}{P_{\pi_{\theta_s}}}^{t-1-\ell},
        \label{auxproof:eq:moving-kernel-product-telescope}
    \end{align}
    where an empty product is the identity kernel.  Since Markov kernels are
    contractions in total variation,
    \begin{align}
        \|\lambda_t-\mu_s\textcolor{revisionblue}{P_{\pi_{\theta_s}}}^{\tau_t}\|_{\mathrm{TV}}
        &\leq
        \sum_{\ell=s}^{t-1}
        \sup_x
        \|\textcolor{revisionblue}{P_{\pi_{\vartheta_\ell}}}(x,\cdot)-\textcolor{revisionblue}{P_{\pi_{\theta_s}}}(x,\cdot)\|_{\mathrm{TV}}
        \notag\\
        &\leq
        L_K
        \sum_{\ell=s}^{t-1}
        \|\vartheta_\ell-\theta_s\|.
        \label{auxproof:eq:aux-moving-kernel-telescoping}
    \end{align}
    The frozen-chain mixing assumption gives
    \begin{equation}
        \|\mu_s\textcolor{revisionblue}{P_{\pi_{\theta_s}}}^{\tau_t}-\nu^{\theta_s}\|_{\mathrm{TV}}
        \leq
        m_{\mathrm{mix}}\rho_{\mathrm{mix}}^{\tau_t}.
        \label{auxproof:eq:frozen-window-mixing}
    \end{equation}
    The stationary-law Lipschitz bound gives
    \begin{align}
        \|\nu^{\theta_s}-\nu^{\theta_t}\|_{\mathrm{TV}}
        &\leq
        L_\nu\|\theta_t-\theta_s\|
        \notag\\
        &\leq
        L_\nu
        \sum_{\ell=s}^{t-1}
        \|\theta_{\ell+1}-\theta_\ell\|.
        \label{auxproof:eq:stationary-drift-window}
    \end{align}
    Combining
    \eqref{auxproof:eq:aux-moving-kernel-telescoping},
    \eqref{auxproof:eq:frozen-window-mixing}, and
    \eqref{auxproof:eq:stationary-drift-window} gives
    \begin{equation}
        \|\lambda_t-\nu^{\theta_t}\|_{\mathrm{TV}}
        \leq
        \bar\varepsilon_{\mathrm{mix}}(t).
        \label{auxproof:eq:conditional-law-to-current-stationary}
    \end{equation}
    The clipped quantity $\varepsilon_{\mathrm{mix}}(t)$ also bounds the same
    distance.  Therefore, for any $\mathcal H_s$-measurable
    $\phi_t$ and nonnegative $\mathcal H_s$-measurable $B_t$ with
    $\sup_x\|\phi_t(x)\|\le B_t$,
    \begin{align*}
        &\left\|
        \mathbb E[\phi_t(X_t)\mid\mathcal H_s]
        -
        \mathbb E_{\nu^{\theta_t}}[\phi_t(X)]
        \right\|
        \notag\\
        &\qquad=
        \left\|
        \int \phi_t(x)\lambda_t(dx)
        -
        \int \phi_t(x)\nu^{\theta_t}(dx)
        \right\|
        \leq
        B_t\varepsilon_{\mathrm{mix}}(t),
    \end{align*}
    which proves the delayed conditional mixing inequality.

    It remains to verify the displayed logarithmic-window rate.  Put
    $s=t-\tau_t$.  If
    $\|\theta_{\ell+1}-\theta_\ell\|\le C_\theta C_\alpha(\ell+1)^{-a}$ and
    $\vartheta_\ell\in\{\theta_\ell,\theta_{\ell+1}\}$, then for
    $\ell=s,\ldots,t-1$,
    \begin{align*}
        \|\vartheta_\ell-\theta_s\|
        &\leq
        \sum_{r=s}^{\ell}
        \|\theta_{r+1}-\theta_r\|
        \notag\\
        &\leq
        C_\theta C_\alpha(\ell-s+1)(s+1)^{-a}.
    \end{align*}
    Hence
    \begin{align}
        \sum_{\ell=s}^{t-1}\|\vartheta_\ell-\theta_s\|
        &\leq
        C_\theta C_\alpha(s+1)^{-a}
        \sum_{q=1}^{\tau_t}q
        \notag\\
        &\leq
        C_\theta C_\alpha\tau_t^2(s+1)^{-a},
        \label{auxproof:eq:vartheta-window-rate}
    \end{align}
    and similarly
    \begin{equation}
        \sum_{\ell=s}^{t-1}\|\theta_{\ell+1}-\theta_\ell\|
        \leq
        C_\theta C_\alpha\tau_t(s+1)^{-a}.
        \label{auxproof:eq:theta-window-rate}
    \end{equation}
    Finally, with $\tau_t=\lceil c_\tau\log(t+2)\rceil$,
    \begin{equation}
        \rho_{\mathrm{mix}}^{\tau_t}
        \leq
        (t+2)^{-c_\tau|\log\rho_{\mathrm{mix}}|}.
        \label{auxproof:eq:geometric-window-rate}
    \end{equation}
    Substituting
    \eqref{auxproof:eq:vartheta-window-rate},
    \eqref{auxproof:eq:theta-window-rate}, and
    \eqref{auxproof:eq:geometric-window-rate} into
    \eqref{auxproof:eq:actor-epsilon-mix-raw-definition} gives the asserted
    finite-time bound and the stated asymptotic order.
\end{proof}

\begin{lemma}[Localized critic approximation; \citet{qu2020scalable}, Lemma~\ref{lem:localized-critic-approximation} restated]
    \label{auxproof:lem:localized-critic-approximation}
    Suppose the $(c,\rho)$ exponential decay property holds for the
    unregularized local $Q$ functions. For each agent $i$, define a truncated
    critic by
    \begin{align*}
        \widetilde Q_i^\theta(s_i^k,a_i^k)
        &:=
        \sum_{s_{-i}^k\in\mathcal{S}_{-i}^k}
        \sum_{a_{-i}^k\in\mathcal{A}_{-i}^k}
        \omega_i(s_{-i}^k,a_{-i}^k;s_i^k,a_i^k)
        Q_i^\theta(s_i^k,s_{-i}^k,a_i^k,a_{-i}^k),
    \end{align*}
    where the weights are nonnegative and satisfy
    \begin{equation*}
        \sum_{s_{-i}^k,a_{-i}^k}
        \omega_i(s_{-i}^k,a_{-i}^k;s_i^k,a_i^k)=1
        \qquad
        \forall (s_i^k,a_i^k).
    \end{equation*}
    Then:
    \begin{enumerate}
        \item The truncated critic uniformly approximates the full local critic:
    \begin{equation*}
        \sup_{s_g,a_g}
        \left|
        Q_i^\theta(s_g,a_g)
        -
        \widetilde Q_i^\theta(s_i^k,a_i^k)
        \right|
        \leq
        c\rho^{k+1}.
    \end{equation*}
        \item If
        $\|\nabla_{\theta_i}\log\pi_{\theta_i}(a_i|s_i)\|\leq G_i$, and
        \begin{equation*}
            \widetilde g_i(\theta)
            :=
            \mathbb{E}_{\nu^\theta}
            \left[
            \nabla_{\theta_i}\log\pi_{\theta_i}(a_i|s_i)
            \sum_{j\in N_i^k}
            \widetilde Q_j^\theta(s_j^k,a_j^k)
            \right],
        \end{equation*}
        then, under the sum-reward convention
        $J(\theta)=\sum_{j=1}^nJ_j(\theta)$ used in this paper,
        \begin{equation*}
            \left\|
            \widetilde g_i(\theta)-\nabla_{\theta_i}J(\theta)
            \right\|
            \leq
            nG_i c\rho^{k+1}.
        \end{equation*}
    \end{enumerate}
\end{lemma}

\begin{proof}
    Fix $i,\theta$ and
    $(s_g,a_g)=(s_i^k,s_{-i}^k,a_i^k,a_{-i}^k)$. Since the weights defining
    $\widetilde Q_i^\theta$ are nonnegative and sum to one,
    \begin{align*}
        &\left|
        Q_i^\theta(s_g,a_g)-\widetilde Q_i^\theta(s_i^k,a_i^k)
        \right|
        \notag\\
        &\quad\leq
        \sum_{\bar s_{-i}^k,\bar a_{-i}^k}
        \omega_i(\bar s_{-i}^k,\bar a_{-i}^k;s_i^k,a_i^k)
        \left|
        Q_i^\theta(s_i^k,s_{-i}^k,a_i^k,a_{-i}^k)
        -
        Q_i^\theta(s_i^k,\bar s_{-i}^k,a_i^k,\bar a_{-i}^k)
        \right|\\
        &\quad\leq c\rho^{k+1},
    \end{align*}
    where the last step is exactly the exponential decay property. This proves
    the first claim.

    For the second claim, write
    $S_i:=\nabla_{\theta_i}\log\pi_{\theta_i}(a_i|s_i)$. By the
    average-reward policy-gradient theorem and
    $Q^\theta=\sum_{j=1}^nQ_j^\theta$,
    \begin{equation*}
        \nabla_{\theta_i}J(\theta)
        =
        \mathbb{E}_{\nu^\theta}
        \left[
        S_i\sum_{j=1}^nQ_j^\theta(s_g,a_g)
        \right].
    \end{equation*}
    Therefore
    \begin{align*}
        \widetilde g_i(\theta)-\nabla_{\theta_i}J(\theta)
        &=
        \sum_{j\in N_i^k}
        \mathbb{E}_{\nu^\theta}
        \left[
        S_i\left(
        \widetilde Q_j^\theta(s_j^k,a_j^k)-Q_j^\theta(s_g,a_g)
        \right)
        \right]
        \notag\\
        &\quad
        -
        \sum_{j\notin N_i^k}
        \mathbb{E}_{\nu^\theta}
        \left[
        S_iQ_j^\theta(s_g,a_g)
        \right].
    \end{align*}
    For each included critic $j\in N_i^k$, the first part of the lemma gives
    \begin{equation*}
        \left\|
        \mathbb{E}_{\nu^\theta}
        \left[
        S_i\left(
        \widetilde Q_j^\theta(s_j^k,a_j^k)-Q_j^\theta(s_g,a_g)
        \right)
        \right]
        \right\|
        \leq
        G_i c\rho^{k+1}.
    \end{equation*}
    For an omitted critic $j\notin N_i^k$, the graph is undirected, so
    $i\notin N_j^k$. Hence $\widetilde Q_j^\theta(s_j^k,a_j^k)$ does not depend
    on $a_i$. Using the product policy and the zero-mean score identity,
    \begin{equation*}
        \mathbb{E}_{\nu^\theta}
        \left[
        S_i\widetilde Q_j^\theta(s_j^k,a_j^k)
        \right]
        =
        0.
    \end{equation*}
    Thus, again by the first part of the lemma,
    \begin{align*}
        \left\|
        \mathbb{E}_{\nu^\theta}[S_iQ_j^\theta(s_g,a_g)]
        \right\|
        &=
        \left\|
        \mathbb{E}_{\nu^\theta}
        \left[
        S_i
        \left(
        Q_j^\theta(s_g,a_g)-\widetilde Q_j^\theta(s_j^k,a_j^k)
        \right)
        \right]
        \right\|\\
        &\leq
        G_i c\rho^{k+1}.
    \end{align*}
    There are exactly $n$ local critics in the full sum. Summing the included
    and omitted contributions gives
    $\|\widetilde g_i(\theta)-\nabla_{\theta_i}J(\theta)\|
    \leq nG_i c\rho^{k+1}$.
\end{proof}

\begin{lemma}[Localized actor-direction bias under softmax, Lemma~\ref{lem:localized-actor-direction-bias-softmax} restated]
    \label{auxproof:lem:localized-actor-direction-bias-softmax}
    Suppose the $(c,\rho)$ exponential decay property holds for the
    unregularized local $Q$ functions, and define the truncated critics
    $\widetilde Q_j^\theta$ by the convex averaging construction in
    Lemma~\ref{auxproof:lem:localized-critic-approximation}.  For each agent $i$, define
    the localized stationary actor direction
    \begin{equation}
        \widetilde g_i(\theta)
        :=
        \mathbb E_{\nu^\theta}
        \left[
        \nabla_{\theta_i}\log\pi_{\theta_i}(a_i|s_i)
        \sum_{j\in N_i^k}
        \widetilde Q_j^\theta(s_j^k,a_j^k)
        \right].
        \label{auxproof:eq:aux-localized-actor-direction}
    \end{equation}
    Under the tabular softmax parameterization,
    \begin{equation}
        \left\|
        \widetilde g_i(\theta)-\nabla_{\theta_i}J(\theta)
        \right\|_2
        \leq
        n\sqrt{2}\,c\rho^{k+1}.
        \label{auxproof:eq:aux-localized-actor-direction-bias-softmax}
    \end{equation}
    More generally, if
    $\|\nabla_{\theta_i}\log\pi_{\theta_i}(a_i|s_i)\|_2\leq G_i$, then the
    right-hand side is $nG_i c\rho^{k+1}$.
\end{lemma}

\begin{proof}
    Write
    $S_i:=\nabla_{\theta_i}\log\pi_{\theta_i}(a_i|s_i)$.  By
    Lemma~\ref{lem:avg-reward-policy-gradient} and the convention
    $Q^\theta=\sum_{j=1}^nQ_j^\theta$,
    \begin{equation}
        \nabla_{\theta_i}J(\theta)
        =
        \mathbb E_{\nu^\theta}
        \left[
        S_i\sum_{j=1}^nQ_j^\theta(s_g,a_g)
        \right].
        \label{auxproof:eq:aux-localized-bias-exact-gradient}
    \end{equation}
    Adding and subtracting the truncated critics gives
    \begin{align}
        \widetilde g_i(\theta)-\nabla_{\theta_i}J(\theta)
        &=
        \sum_{j\in N_i^k}
        \mathbb E_{\nu^\theta}
        \left[
        S_i
        \left(
        \widetilde Q_j^\theta(s_j^k,a_j^k)-Q_j^\theta(s_g,a_g)
        \right)
        \right]
        \notag\\
        &\quad
        -
        \sum_{j\notin N_i^k}
        \mathbb E_{\nu^\theta}
        \left[
        S_iQ_j^\theta(s_g,a_g)
        \right].
        \label{auxproof:eq:aux-localized-bias-decomposition}
    \end{align}

    For each included critic $j\in N_i^k$,
    Lemma~\ref{auxproof:lem:localized-critic-approximation} gives
    \[
        \sup_{s_g,a_g}
        \left|
        \widetilde Q_j^\theta(s_j^k,a_j^k)-Q_j^\theta(s_g,a_g)
        \right|
        \leq
        c\rho^{k+1}.
    \]
    Therefore, if $\|S_i\|_2\leq G_i$,
    \begin{equation}
        \left\|
        \mathbb E_{\nu^\theta}
        \left[
        S_i
        \left(
        \widetilde Q_j^\theta(s_j^k,a_j^k)-Q_j^\theta(s_g,a_g)
        \right)
        \right]
        \right\|_2
        \leq
        G_i c\rho^{k+1}.
        \label{auxproof:eq:aux-included-localized-bias}
    \end{equation}

    Now let $j\notin N_i^k$.  Since the graph neighborhoods are symmetric,
    $i\notin N_j^k$, so
    $\widetilde Q_j^\theta(s_j^k,a_j^k)$ does not depend on $a_i$.  Conditional
    on all variables except $a_i$, the zero-mean score identity
    \eqref{auxproof:eq:softmax-score-zero-mean} gives
    \begin{equation}
        \mathbb E_{\nu^\theta}
        \left[
        S_i\widetilde Q_j^\theta(s_j^k,a_j^k)
        \right]
        =
        0.
        \label{auxproof:eq:aux-omitted-truncated-zero}
    \end{equation}
    Hence
    \begin{align}
        \left\|
        \mathbb E_{\nu^\theta}[S_iQ_j^\theta(s_g,a_g)]
        \right\|_2
        &=
        \left\|
        \mathbb E_{\nu^\theta}
        \left[
        S_i
        \left(
        Q_j^\theta(s_g,a_g)-\widetilde Q_j^\theta(s_j^k,a_j^k)
        \right)
        \right]
        \right\|_2 \\
        &\leq
        G_i c\rho^{k+1},
        \label{auxproof:eq:aux-omitted-localized-bias}
    \end{align}
    again by Lemma~\ref{auxproof:lem:localized-critic-approximation}.

    There are at most $n$ terms total across the included and omitted critic
    sums in \eqref{auxproof:eq:aux-localized-bias-decomposition}.  Combining
    \eqref{auxproof:eq:aux-included-localized-bias} and
    \eqref{auxproof:eq:aux-omitted-localized-bias} yields
    \[
        \|\widetilde g_i(\theta)-\nabla_{\theta_i}J(\theta)\|_2
        \leq
        nG_i c\rho^{k+1}.
    \]
    Under tabular softmax, Lemma~\ref{auxproof:lem:product-policy-score-identities}
    gives $G_i=\sqrt2$, which proves
    \eqref{auxproof:eq:aux-localized-actor-direction-bias-softmax}.
\end{proof}

\begin{lemma}[Smoothness of the unregularized objective under softmax, Lemma~\ref{lem:smoothness-lipschitz-policy} restated]
    \label{auxproof:lem:smoothness-lipschitz-policy}
    Suppose the policy is the finite-action product softmax policy and that
    $|r_i(s_i,a_i)|\le r_{\max}$ for every $i,s_i,a_i$.  Let
    \[
        R:=nr_{\max},
        \qquad
        G:=\sqrt{2n},
        \qquad
        H:=1 .
    \]
    Assume that the policy-induced global kernels are uniformly ergodic in the
    sense that
    \begin{equation*}
        K_{\mathrm{mix}}
        :=
        \sup_\theta
        \sum_{t=0}^{\infty}
        \sup_{s_g\in\mathcal{S}_g}
        \|(\textcolor{revisionblue}{P_{\pi_\theta}^{\mathrm{s}}})^t(s_g,\cdot)-d^\theta\|_1
        <\infty.
    \end{equation*}
    Define
    \begin{align*}
        L_d&:=K_{\mathrm{mix}}G,&
        L_J&:=R(L_d+G),&
        L_{\nu}^{\mathrm{sm}}&:=L_d+G,\\
        B_V&:=2RK_{\mathrm{mix}},&
        B_Q&:=2R+B_V,\\
        L_V&:=K_{\mathrm{mix}}\left(RG(K_{\mathrm{mix}}+2)+GB_V\right)
        +L_dB_V,&
        L_Q&:=L_J+L_V,
    \end{align*}
    and
    \begin{equation*}
        L_{\nabla J}
        :=
        L_dGB_Q+(G^2+H)B_Q+GL_Q .
    \end{equation*}
    Then the objective-smoothness and objective-upper-bound conclusions used in
    the actor proof hold with this $L_{\nabla J}$ and with
    \begin{equation}
        J_{\max}:=\sup_\theta J(\theta)\le nr_{\max}.
        \label{auxproof:eq:actor-Jmax-new}
    \end{equation}
    That is, for all $\theta,\theta'$,
    \begin{equation}
        \|\nabla J(\theta)-\nabla J(\theta')\|
        \leq
        L_{\nabla J}\|\theta-\theta'\|,
        \label{auxproof:eq:actor-J-smooth-new}
    \end{equation}
    and $J$ is bounded above by $J_{\max}$.
\end{lemma}

\begin{proof}
    We use the matrix norm
    $\|M\|_{\infty,1}:=\max_x\sum_y|M(x,y)|$.  Since
    $|r(s_g,a_g)|\le R$, every stationary average reward satisfies
    $J(\theta)\le R=nr_{\max}$, proving the asserted upper bound.

    By Lemma~\ref{auxproof:lem:product-policy-score-identities}, each local softmax
    score has norm at most $\sqrt{2}$.  Since the global product-policy score is
    the concatenation of the $n$ local scores, its norm is at most
    $G=\sqrt{2n}$.  The softmax coordinate formula in the same lemma also
    implies that the local score map is $1$-Lipschitz in the local logit vector:
    for a fixed local state-action pair,
    the only nonconstant part of the score is
    $-\pi_{\theta_i}(\cdot|s_i)$, and the softmax Jacobian
    $\operatorname{diag}(p)-pp^\top$ has operator norm at most $1$.  Summing the
    squared local bounds over agents gives the global score-Lipschitz bound
    with $H=1$.

    For product softmax,
    $\nabla_\theta\pi_\theta(a_g|s_g)
    =\pi_\theta(a_g|s_g)\nabla_\theta\log\pi_\theta(a_g|s_g)$, and hence
    \begin{equation*}
        \sum_{a_g}
        \|\nabla_\theta\pi_\theta(a_g|s_g)\|
        \leq G.
    \end{equation*}
    Applying the mean-value theorem to
    $\pi_\theta(\cdot|s_g)$ and to
    $\nabla_\theta\pi_\theta(\cdot|s_g)$ gives
    \begin{equation*}
        \|\pi_\theta(\cdot|s_g)-\pi_{\theta'}(\cdot|s_g)\|_1
        \leq G\|\theta-\theta'\|,
    \end{equation*}
    and
    \begin{equation*}
        \sum_{a_g}
        \|\nabla_\theta\pi_\theta(a_g|s_g)
        -\nabla_\theta\pi_{\theta'}(a_g|s_g)\|
        \leq (G^2+H)\|\theta-\theta'\|.
    \end{equation*}
    The second bound follows by writing
    $\nabla\pi=\pi\nabla\log\pi$ and bounding
    $\sum_a |\pi_\theta(a|s)-\pi_{\theta'}(a|s)|\|\nabla\log\pi_\theta(a|s)\|$
    by $G^2\|\theta-\theta'\|$ and
    $\sum_a\pi_{\theta'}(a|s)\|\nabla\log\pi_\theta(a|s)
    -\nabla\log\pi_{\theta'}(a|s)\|$ by $H\|\theta-\theta'\|$.
    Therefore
    \begin{equation*}
        \|\textcolor{revisionblue}{P_{\pi_\theta}^{\mathrm{s}}}-\textcolor{revisionblue}{P_{\pi_{\theta'}}^{\mathrm{s}}}\|_{\infty,1}
        \leq G\|\theta-\theta'\|.
    \end{equation*}

    Define the fundamental matrix
    \begin{equation*}
        Z_\theta:=\sum_{t=0}^{\infty}
        \left((\textcolor{revisionblue}{P_{\pi_\theta}^{\mathrm{s}}})^t-\mathbf{1}d^\theta{}^\top\right).
    \end{equation*}
    The definition of $K_{\mathrm{mix}}$ gives
    $\|Z_\theta\|_{\infty,1}\leq K_{\mathrm{mix}}$. The stationary perturbation
    identity
    \begin{equation*}
        d^{\theta'}-d^\theta
        =
        d^{\theta'}(\textcolor{revisionblue}{P_{\pi_{\theta'}}^{\mathrm{s}}}-\textcolor{revisionblue}{P_{\pi_\theta}^{\mathrm{s}}})Z_\theta
    \end{equation*}
    then implies
    \begin{equation*}
        \|d^\theta-d^{\theta'}\|_1
        \leq
        K_{\mathrm{mix}}\|\textcolor{revisionblue}{P_{\pi_\theta}^{\mathrm{s}}}-\textcolor{revisionblue}{P_{\pi_{\theta'}}^{\mathrm{s}}}\|_{\infty,1},
    \end{equation*}
    which proves the bound for $d^\theta$.

    Let
    $\bar r_\theta(s_g):=\sum_{a_g}\pi_\theta(a_g|s_g)r(s_g,a_g)$. Since
    $|\bar r_\theta(s_g)|\leq R$ and
    $|\bar r_\theta(s_g)-\bar r_{\theta'}(s_g)|
    \leq RG\|\theta-\theta'\|$,
    \begin{align*}
        |J(\theta)-J(\theta')|
        &\leq
        \sum_{s_g}|d^\theta(s_g)-d^{\theta'}(s_g)|\,|\bar r_\theta(s_g)|
        \notag\\
        &\quad+
        \sum_{s_g}d^{\theta'}(s_g)
        |\bar r_\theta(s_g)-\bar r_{\theta'}(s_g)|\\
        &\leq R(L_d+G)\|\theta-\theta'\|.
    \end{align*}
    Also,
    \begin{align*}
        \|\nu^\theta-\nu^{\theta'}\|_1
        &\leq
        \sum_{s_g}|d^\theta(s_g)-d^{\theta'}(s_g)|
        +
        \sup_{s_g}
        \|\pi_\theta(\cdot|s_g)-\pi_{\theta'}(\cdot|s_g)\|_1\\
        &\leq (L_d+G)\|\theta-\theta'\|.
    \end{align*}

    The centered reward
    $h_\theta:=\bar r_\theta-J(\theta)$ satisfies
    $\|h_\theta\|_\infty\leq 2R$ and
    $\|h_\theta-h_{\theta'}\|_\infty\leq RG(K_{\mathrm{mix}}+2)
    \|\theta-\theta'\|$. Since
    \begin{equation*}
        V^\theta=Z_\theta h_\theta,
    \end{equation*}
    we have $\|V^\theta\|_\infty\leq B_V$. Moreover, subtracting the Poisson
    equations
    $(I-\textcolor{revisionblue}{P_{\pi_\theta}^{\mathrm{s}}})V^\theta=h_\theta$ and
    $(I-\textcolor{revisionblue}{P_{\pi_{\theta'}}^{\mathrm{s}}})V^{\theta'}=h_{\theta'}$ gives
    \begin{equation*}
        (I-\textcolor{revisionblue}{P_{\pi_\theta}^{\mathrm{s}}})(V^\theta-V^{\theta'})
        =
        h_\theta-h_{\theta'}
        +(\textcolor{revisionblue}{P_{\pi_\theta}^{\mathrm{s}}}-\textcolor{revisionblue}{P_{\pi_{\theta'}}^{\mathrm{s}}})V^{\theta'}.
    \end{equation*}
    Let $b_{\theta,\theta'}:=h_\theta-h_{\theta'}
    +(\textcolor{revisionblue}{P_{\pi_\theta}^{\mathrm{s}}}-\textcolor{revisionblue}{P_{\pi_{\theta'}}^{\mathrm{s}}})V^{\theta'}$.  The vector
    $Z_\theta b_{\theta,\theta'}$ is centered under $d^\theta$, whereas
    $V^\theta-V^{\theta'}$ has mean
    \[
        d^\theta{}^\top(V^\theta-V^{\theta'})
        =
        -d^\theta{}^\top V^{\theta'}
        =
        (d^{\theta'}-d^\theta)^\top V^{\theta'},
    \]
    because the differential values are normalized by
    $d^\theta{}^\top V^\theta=0$ and
    $d^{\theta'}{}^\top V^{\theta'}=0$.  Hence
    \[
        V^\theta-V^{\theta'}
        =
        Z_\theta b_{\theta,\theta'}
        +
        \mathbf 1 (d^{\theta'}-d^\theta)^\top V^{\theta'} .
    \]
    Using the bounds above yields
    \begin{align*}
        \|V^\theta-V^{\theta'}\|_\infty
        &\leq
        K_{\mathrm{mix}}
        \left(
        \|h_\theta-h_{\theta'}\|_\infty
        +
        \|\textcolor{revisionblue}{P_{\pi_\theta}^{\mathrm{s}}}-\textcolor{revisionblue}{P_{\pi_{\theta'}}^{\mathrm{s}}}\|_{\infty,1}
        \|V^{\theta'}\|_\infty
        \right)
        +
        \|d^{\theta'}-d^\theta\|_1\|V^{\theta'}\|_\infty\\
        &\leq L_V\|\theta-\theta'\|.
    \end{align*}
    Since
    $Q^\theta(s_g,a_g)=r(s_g,a_g)-J(\theta)+\sum_{s_g'}P(s_g'|s_g,a_g)
    V^\theta(s_g')$, the $Q$-Lipschitz bound follows.

    By the policy-gradient lemma,
    \begin{equation*}
        \nabla J(\theta)
        =
        \sum_{s_g}d^\theta(s_g)\sum_{a_g}
        \nabla_\theta\pi_\theta(a_g|s_g)Q^\theta(s_g,a_g).
    \end{equation*}
    Also, $\|Q^\theta\|_\infty\leq B_Q$. Adding and subtracting the
    corresponding terms at $\theta'$ gives three terms:
    \begin{align*}
        \|\nabla J(\theta)-\nabla J(\theta')\|
        &\leq
        \|d^\theta-d^{\theta'}\|_1
        \sup_{s_g}\sum_{a_g}
        \|\nabla_\theta\pi_\theta(a_g|s_g)\|\,\|Q^\theta\|_\infty
        \notag\\
        &\quad+
        \sup_{s_g}\sum_{a_g}
        \|\nabla_\theta\pi_\theta(a_g|s_g)
        -\nabla_\theta\pi_{\theta'}(a_g|s_g)\|\,\|Q^\theta\|_\infty
        \notag\\
        &\quad+
        \sup_{s_g}\sum_{a_g}
        \|\nabla_\theta\pi_{\theta'}(a_g|s_g)\|
        |Q^\theta(s_g,a_g)-Q^{\theta'}(s_g,a_g)|\\
        &\leq
        \left(L_dGB_Q+(G^2+H)B_Q+GL_Q\right)
        \|\theta-\theta'\|.
    \end{align*}
\end{proof}

\begin{lemma}[Average-reward policy gradient]
    \label{lem:avg-reward-policy-gradient}
    Suppose that, for the product policy $\pi_\theta$, the induced global
    Markov chain admits a unique stationary distribution $d^\theta$ over
    $\mathcal{S}_g$ which is differentiable with respect to $\theta$. Let
    $\nu^\theta(s_g,a_g):=d^\theta(s_g)\pi_\theta(a_g|s_g)$ denote the
    stationary state-action distribution. Then, for every agent
    $i\in\mathcal{N}$,
    \begin{equation*}
        \nabla_{\theta_i}J(\theta)
        =
        \sum_{s_g\in\mathcal{S}_g}d^\theta(s_g)
        \sum_{a_g\in\mathcal{A}_g}
        \nabla_{\theta_i}\pi_\theta(a_g|s_g)Q^\theta(s_g,a_g).
    \end{equation*}
    Equivalently,
    \begin{equation*}
        \nabla_{\theta_i}J(\theta)
        =
        \mathbb{E}_{(s_g,a_g)\sim\nu^\theta}
        \left[
        \nabla_{\theta_i}\log\pi_{\theta_i}(a_i|s_i)
        Q^\theta(s_g,a_g)
        \right].
    \end{equation*}
    For the softmax parameterization, define the one-agent averaged advantage
    \begin{equation*}
        \bar A_i^\theta(s_g,a_i)
        :=
        \sum_{a_{-i}\in\mathcal{A}_{-i}}
        \pi_{\theta_{-i}}(a_{-i}|s_{-i})
        Q^\theta(s_g,a_i,a_{-i})
        -
        V^\theta(s_g),
    \end{equation*}
    where $s_{-i}$ and $a_{-i}$ collect the states and actions of all agents
    except $i$,
    $\mathcal{S}_{-i}:=\prod_{j\neq i}\mathcal{S}_j$,
    $\mathcal{A}_{-i}:=\prod_{j\neq i}\mathcal{A}_j$, and
    $\pi_{\theta_{-i}}(a_{-i}|s_{-i})
    :=\prod_{j\neq i}\pi_{\theta_j}(a_j|s_j)$. Then, for every
    $(\bar s_i,\bar a_i)\in\mathcal{S}_i\times\mathcal{A}_i$,
    \begin{equation*}
        \frac{\partial J(\theta)}
        {\partial\theta_i(\bar s_i,\bar a_i)}
        =
        \sum_{s_{-i}\in\mathcal{S}_{-i}}
        d^\theta(\bar s_i,s_{-i})
        \pi_{\theta_i}(\bar a_i|\bar s_i)
        \bar A_i^\theta((\bar s_i,s_{-i}),\bar a_i).
    \end{equation*}
\end{lemma}

\begin{proof}
    Let
    \begin{equation*}
        \textcolor{revisionblue}{P_{\pi_\theta}^{\mathrm{s}}}(s_g'|s_g)
        :=
        \sum_{a_g\in\mathcal{A}_g}
        \pi_\theta(a_g|s_g)P(s_g'|s_g,a_g)
    \end{equation*}
    be the transition kernel of the global Markov chain under $\pi_\theta$, and
    let
    \begin{equation*}
        \bar r_\theta(s_g)
        :=
        \sum_{a_g\in\mathcal{A}_g}\pi_\theta(a_g|s_g)r(s_g,a_g).
    \end{equation*}
    The average reward is
    $J(\theta)=\sum_{s_g}d^\theta(s_g)\bar r_\theta(s_g)$, and the differential
    value function satisfies the Poisson equation
    \begin{equation*}
        J(\theta)+V^\theta(s_g)
        =
        \bar r_\theta(s_g)
        +
        \sum_{s_g'\in\mathcal{S}_g}
        \textcolor{revisionblue}{P_{\pi_\theta}^{\mathrm{s}}}(s_g'|s_g)V^\theta(s_g').
    \end{equation*}
    Differentiating $J(\theta)=d^\theta{}^\top \bar r_\theta$ with respect to
    any component of $\theta_i$ gives
    \begin{equation*}
        \nabla_{\theta_i}J(\theta)
        =
        (\nabla_{\theta_i}d^\theta)^\top \bar r_\theta
        +
        d^\theta{}^\top\nabla_{\theta_i}\bar r_\theta.
    \end{equation*}
    Since $d^\theta{}^\top \textcolor{revisionblue}{P_{\pi_\theta}^{\mathrm{s}}}=d^\theta{}^\top$ and
    $d^\theta{}^\top\mathbf{1}=1$, differentiating the stationarity equation
    yields
    \begin{equation*}
        (\nabla_{\theta_i}d^\theta)^\top(I-\textcolor{revisionblue}{P_{\pi_\theta}^{\mathrm{s}}})
        =
        d^\theta{}^\top\nabla_{\theta_i}\textcolor{revisionblue}{P_{\pi_\theta}^{\mathrm{s}}},
        \qquad
        (\nabla_{\theta_i}d^\theta)^\top\mathbf{1}=0.
    \end{equation*}
    Using the Poisson equation to replace
    $\bar r_\theta=J(\theta)\mathbf{1}+V^\theta-\textcolor{revisionblue}{P_{\pi_\theta}^{\mathrm{s}}} V^\theta$, we obtain
    \begin{align*}
        (\nabla_{\theta_i}d^\theta)^\top \bar r_\theta
        &=
        (\nabla_{\theta_i}d^\theta)^\top
        \left(J(\theta)\mathbf{1}+V^\theta-\textcolor{revisionblue}{P_{\pi_\theta}^{\mathrm{s}}} V^\theta\right)\\
        &=
        d^\theta{}^\top(\nabla_{\theta_i}\textcolor{revisionblue}{P_{\pi_\theta}^{\mathrm{s}}})V^\theta.
    \end{align*}
    Therefore,
    \begin{align*}
        \nabla_{\theta_i}J(\theta)
        &=
        d^\theta{}^\top\nabla_{\theta_i}\bar r_\theta
        +
        d^\theta{}^\top(\nabla_{\theta_i}\textcolor{revisionblue}{P_{\pi_\theta}^{\mathrm{s}}})V^\theta\\
        &=
        \sum_{s_g\in\mathcal{S}_g}d^\theta(s_g)
        \sum_{a_g\in\mathcal{A}_g}
        \nabla_{\theta_i}\pi_\theta(a_g|s_g)
        \left(
        r(s_g,a_g)
        +
        \sum_{s_g'\in\mathcal{S}_g}
        P(s_g'|s_g,a_g)V^\theta(s_g')
        \right).
    \end{align*}
    Since
    $\sum_{a_g}\nabla_{\theta_i}\pi_\theta(a_g|s_g)=0$, we may subtract
    $J(\theta)$ inside the parentheses. By the definition of $Q^\theta$,
    \begin{equation*}
        \nabla_{\theta_i}J(\theta)
        =
        \sum_{s_g\in\mathcal{S}_g}d^\theta(s_g)
        \sum_{a_g\in\mathcal{A}_g}
        \nabla_{\theta_i}\pi_\theta(a_g|s_g)Q^\theta(s_g,a_g).
    \end{equation*}
    Using
    $\nabla_{\theta_i}\pi_\theta(a_g|s_g)
    =
    \pi_\theta(a_g|s_g)\nabla_{\theta_i}\log\pi_\theta(a_g|s_g)$ and the
    product-policy score identity gives
    \begin{equation*}
        \nabla_{\theta_i}J(\theta)
        =
        \mathbb{E}_{(s_g,a_g)\sim\nu^\theta}
        \left[
        \nabla_{\theta_i}\log\pi_{\theta_i}(a_i|s_i)
        Q^\theta(s_g,a_g)
        \right].
    \end{equation*}

    It remains to derive the componentwise softmax expression. Fix
    $(\bar s_i,\bar a_i)\in\mathcal{S}_i\times\mathcal{A}_i$, and write
    $\bar s_g=(\bar s_i,s_{-i})$. From the product form of the policy and the
    softmax derivative,
    \begin{align*}
        &\frac{\partial J(\theta)}
        {\partial\theta_i(\bar s_i,\bar a_i)}
        =
        \sum_{s_{-i}\in\mathcal{S}_{-i}}
        d^\theta(\bar s_g)
        \sum_{a_{-i}\in\mathcal{A}_{-i}}
        \pi_{\theta_{-i}}(a_{-i}|s_{-i})
        \sum_{a_i\in\mathcal{A}_i}
        \frac{\partial\pi_{\theta_i}(a_i|\bar s_i)}
        {\partial\theta_i(\bar s_i,\bar a_i)}
        Q^\theta(\bar s_g,a_i,a_{-i})\\
        &=
        \sum_{s_{-i}\in\mathcal{S}_{-i}}
        d^\theta(\bar s_g)
        \sum_{a_{-i}\in\mathcal{A}_{-i}}
        \pi_{\theta_{-i}}(a_{-i}|s_{-i})
        \notag\\
        &\qquad\cdot
        \sum_{a_i\in\mathcal{A}_i}
        \pi_{\theta_i}(a_i|\bar s_i)
        \left(\mathbb{I}\{a_i=\bar a_i\}
        -\pi_{\theta_i}(\bar a_i|\bar s_i)\right)
        Q^\theta(\bar s_g,a_i,a_{-i})\\
        &=
        \sum_{s_{-i}\in\mathcal{S}_{-i}}
        d^\theta(\bar s_g)
        \pi_{\theta_i}(\bar a_i|\bar s_i)
        \left[
        \sum_{a_{-i}\in\mathcal{A}_{-i}}
        \pi_{\theta_{-i}}(a_{-i}|s_{-i})
        Q^\theta(\bar s_g,\bar a_i,a_{-i})
        -
        V^\theta(\bar s_g)
        \right]\\
        &=
        \sum_{s_{-i}\in\mathcal{S}_{-i}}
        d^\theta(\bar s_g)
        \pi_{\theta_i}(\bar a_i|\bar s_i)
        \bar A_i^\theta(\bar s_g,\bar a_i),
    \end{align*}
    where the third equality uses
    $V^\theta(s_g)=\sum_{a_g}\pi_\theta(a_g|s_g)Q^\theta(s_g,a_g)$. This
    completes the proof.
\end{proof}

\section{Numerical Illustration of the Localization--Compression Tradeoff}
\label{app:numerical-localization-compression}
\setcounter{figure}{0}
\setcounter{table}{0}
\renewcommand{\thefigure}{\thesection.\arabic{figure}}
\renewcommand{\thetable}{\thesection.\arabic{table}}
\providecommand*{\theHfigure}{}
\providecommand*{\theHtable}{}
\renewcommand*{\theHfigure}{\thesection.\arabic{figure}}
\renewcommand*{\theHtable}{\thesection.\arabic{table}}

This section provides a numerical illustration of the interaction between the
critic neighborhood radius $k$ and the dimension $D=d_i$ of a linear
representation.  The experiment is designed to visualize the structural
tradeoff in the bounds that a larger neighborhood can reduce omitted-information
error while making a fixed-dimensional critic harder to fit.

\begin{figure}[H]
\centering
\begin{tikzpicture}[
    agent/.style={circle, draw=networkblue, fill=white,
        minimum size=4.8mm, inner sep=0pt, line width=0.7pt},
    link/.style={draw=networkblue!65, line width=0.8pt},
    every node/.style={font=\small, text=networkblue}]
    \path[use as bounding box] (0,0) rectangle (12.4,3.3);
    \node[font=\small\bfseries] at (3.1,3.05) {(a) Path: 8 agents};
    \node[font=\small\bfseries] at (9.33,3.05) {(b) Grid: $4\times4$ agents};
    \foreach \j in {0,...,6} {
        \draw[link] ({0.55+0.73*\j},1.43) -- ({0.55+0.73*(\j+1)},1.43);
    }
    \foreach \j in {0,...,7} {
        \node[agent] at ({0.55+0.73*\j},1.43) {};
    }
    \foreach \j in {2,4} {
        \node[agent, fill=networkblue!20] at ({0.55+0.73*\j},1.43) {};
    }
    \node[agent, fill=networkblue, text=white] at (2.74,1.43) {$i$};
    \foreach \r in {0,...,3} {
        \draw[link] (8.25,{0.35+0.72*\r}) -- (10.41,{0.35+0.72*\r});
        \draw[link] ({8.25+0.72*\r},0.35) -- ({8.25+0.72*\r},2.51);
    }
    \foreach \r in {0,...,3} {
        \foreach \c in {0,...,3} {
            \node[agent] at ({8.25+0.72*\c},{0.35+0.72*\r}) {};
        }
    }
    \foreach \c/\r in {0/1,2/1,1/0,1/2} {
        \node[agent, fill=networkblue!20]
            at ({8.25+0.72*\c},{0.35+0.72*\r}) {};
    }
    \node[agent, fill=networkblue, text=white] at (8.97,1.07) {$i$};
\end{tikzpicture}
\caption{\textcolor{revisionblue}{Examples of the experimental networks.
Edges connect adjacent agents, with no wrap-around connections. The dark node
marks agent $i$ and light nodes mark its immediate neighbors; together they
form the one-hop neighborhood $N_i^1$.}}
\label{fig:experiment-network-topologies}
\end{figure}
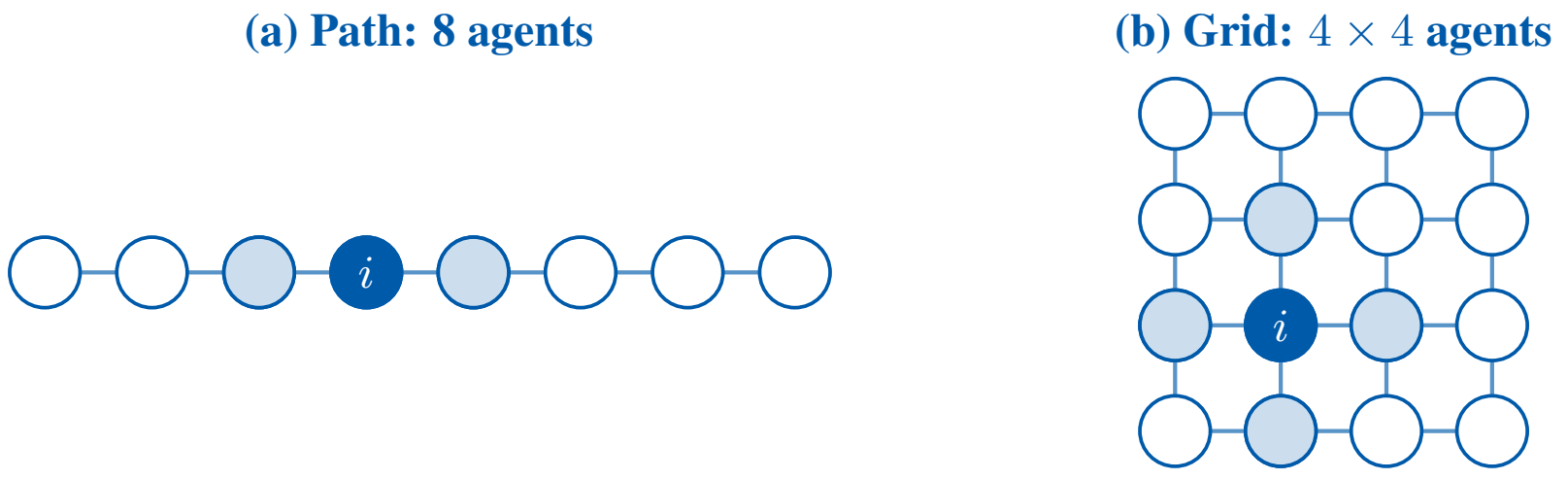

\paragraph{Environment and protocol.}
We use a finite-state networked maintenance environment with local state
$s_i=(h_i,m_i)\in\{0,1\}^2$, action $a_i\in\{0,1\}$, and local reward
$r_i(s_i,a_i)=h_i-0.20a_i$.  Here $h_i$ is a health indicator and $m_i$ is the
action taken by agent $i$ in the previous round.  With $\bar h_i$ and
$\bar m_i$ denoting the averages of these coordinates over the neighbors of
agent $i$, the transition is
\begin{align*}
    \Pr(h_i'=1\mid s,a_i)
    =\operatorname{clip}\!\left(
        0.05+0.03h_i+0.01a_i+0.05\bar h_i+0.75\bar m_i,0.05,0.95
    \right),
    \qquad m_i'=a_i.
\end{align*}
The equation assigns small positive effects to the agent's current health and
intervention, a modest effect to its neighbors' current health, and a much
stronger effect to their previous-round interventions; clipping bounds the
health probability between $0.05$ and $0.95$, while $m_i'=a_i$ stores the
current action for the next round.
Thus, the current action of a neighbor affects agent $i$ only after becoming a
state coordinate at the next step, so the transition retains the local
factorization used in the main text.  We consider paths with
$8,16,32,$ and $64$ agents and grids of sizes $3\times3$, $4\times4$,
$6\times6$, and $8\times8$.
\textcolor{revisionblue}{Figure~\ref{fig:experiment-network-topologies}
illustrates both network families.} The actor observes its own state, while the
critic uses radii $k\in\{0,1,2,3\}$.  We compare additive and pairwise
deterministic hashed embeddings with $D\in\{4,8,16,32,64,128,256,512\}$ and a
tabular baseline.  Each main run uses $3{,}000$ training steps
and a fresh $300$-step greedy evaluation.  Curves pool four graph sizes and
three seeds; error bars are descriptive normal-approximation 95\% confidence
intervals.  Hyperparameters were tuned separately for each graph family,
embedding, $D$, and $k$ on a representative path or grid.

\begingroup
\color{revisionblue}
\paragraph{Tuning and selected settings.}
Tuning uses a 16-node path and a $4\times4$ grid, seeds $0,1$,
$1{,}500$ training steps, and $180$ greedy-evaluation steps.
The four joint configurations $H_j=(\alpha,\beta,\eta,C_w,C_\theta)$ are
\[
\begin{aligned}
H_1&=(0.03,0.05,0,50,8), & H_2&=(0.06,0.08,0,70,10),\\
H_3&=(0.10,0.10,0.001,70,10), & H_4&=(0.16,0.10,0.003,100,12).
\end{aligned}
\]
Here $\alpha$ and $\beta$ are constant actor and critic stepsizes, $\eta$
weights the sampled-state entropy bonus, and $C_w,C_\theta$ bound critic
and actor parameters as described below.  For each graph family, feature
map, $D$, and $k$, we select the individual successful tuning run with the
largest greedy reward minus $0.0002$ times the Euclidean norm of its critic
parameters, including all table entries for the tabular baseline.
Table~\ref{tab:experiment-selected-hypers} lists the selected settings.
Final runs use seeds $0,1,2$ at every size, so tuning and final seeds
partially overlap.  Greedy evaluation starts from a fresh environment with
seed $s+4000$ for training seed $s$, averages all $300$ steps without
burn-in, and breaks action ties in favor of action $0$.

\begin{table}[H]
\centering
\color{revisionblue}
\caption{Selected configurations. Each tuple lists the indices of $H_j$
for $k=0,1,2,3$, respectively, shared across all four sizes and three final
seeds. The $D=4$ entries use the corrected runs.}
\label{tab:experiment-selected-hypers}
\begin{tabular}{llcc}
\toprule
Features & $D$ & Path ($k=0,1,2,3$) & Grid ($k=0,1,2,3$) \\
\midrule
Additive & 4 & $(1,1,4,1)$ & $(2,1,1,1)$ \\
Additive & 8 & $(4,2,1,1)$ & $(1,1,4,1)$ \\
Additive & 16 & $(4,1,4,1)$ & $(4,3,4,4)$ \\
Additive & 32 & $(1,1,4,2)$ & $(1,2,3,4)$ \\
Additive & 64 & $(1,4,4,4)$ & $(3,4,3,3)$ \\
Additive & 128 & $(1,1,2,4)$ & $(4,2,2,3)$ \\
Additive & 256 & $(1,3,1,4)$ & $(1,3,3,4)$ \\
Additive & 512 & $(1,3,3,4)$ & $(1,2,3,4)$ \\
\midrule
Pairwise & 4 & $(1,3,1,3)$ & $(1,1,1,3)$ \\
Pairwise & 8 & $(3,2,2,4)$ & $(1,4,1,1)$ \\
Pairwise & 16 & $(1,4,2,2)$ & $(1,4,1,2)$ \\
Pairwise & 32 & $(4,4,4,4)$ & $(1,2,1,1)$ \\
Pairwise & 64 & $(1,4,4,2)$ & $(1,4,2,3)$ \\
Pairwise & 128 & $(1,3,3,4)$ & $(1,4,3,4)$ \\
Pairwise & 256 & $(1,4,4,4)$ & $(1,4,3,3)$ \\
Pairwise & 512 & $(1,3,3,4)$ & $(1,3,3,3)$ \\
\midrule
Tabular & -- & $(4,4,4,1)$ & $(4,1,1,1)$ \\
\bottomrule
\end{tabular}
\end{table}

\paragraph{Feature construction and updates.}
Neighborhood observations concatenate $(h_j,m_j)$ in increasing node-index
order. Critic keys append the focal action and the number of active immediate
neighbors, capped at four; this count is retained even at $k=0$.
For a token sequence $q$, the first four feature coordinates are initialized
as $1$, $\operatorname{mean}(q)/4$, $\max(q)/4$, and
$\min(1,|q|/32)$. Deterministic signed hashes add individual-token
contributions; pairwise features also hash adjacent tokens in this sequence.
Critic features additionally hash the action, count, and their interaction.
Hashes use the remaining coordinates for $D>4$ and share the three non-bias
coordinates for $D=4$; the resulting vector is normalized to unit Euclidean
norm. The exact integer mixer and token weights are specified in
\texttt{src/sampled\_actor\_critic.py} in the accompanying source package.

Initial health and memory coordinates are independent
Bernoulli$(1/2)$ draws. Linear actor entries start as independent
$0.01\,\mathcal N(0,1)$ draws; critics and average-reward estimates start
at zero. Each transition updates critics and then actors sequentially by
agent index, with average-reward stepsize $\beta$. The actor signal combines
the neighborhood mean TD error with the mean critic value minus its
counterfactual baseline. Linear critic weights are projected onto the
Euclidean ball of radius $C_w$, and actor entries are clipped to
$[-C_\theta,C_\theta]$. The tabular baseline instead uses per-agent critic
tables and local plus shared actor-logit tables, splitting each actor update
equally between the latter two. Its logits use the same Gaussian
initialization, unseen critic entries start at zero, and critic entries are
clipped individually to $[-C_w,C_w]$.

\paragraph{Environment selection and aggregation.}
The environment was selected to illustrate the tradeoff. Calibration screened
five parameter settings on the representative graphs, using seed $0$,
$800$ training steps, and $120$ evaluation steps. For additive features at
$D=8$ and pairwise features at $D=8,16$, it retained the best-scoring
configuration at each $k$ and maximized, across graph families and feature
settings, the average of
$\max(R_1,R_2)-\max(R_0,R_3)$, where $R_k$ is the corresponding greedy
reward; mean peak reward broke ties. Thus, the observed curve shape describes
this selected example and need not hold across environments.

Reported intervals are $\bar R\pm1.96s_R/\sqrt{m}$, where $s_R$ is the
sample standard deviation and $m=12$ for pooled curves or $m=3$ for
size-specific results. They summarize variation across the included runs.
An initial feature-indexing bug caused the original $D=4$ runs to fail;
these records are retained but excluded, and all reported $D=4$ results
use successful runs after the repair. Dimensions $128,256,512$ were
completed in a later extension with the same tuning and evaluation protocol.

\paragraph{Reproduction records and computation.}
The accompanying \texttt{anc/experiment\_reproducibility} directory contains
the implementation, calibration and tuning records, final-run data, selected
settings, and execution instructions. It also includes the script for
Figure~\ref{fig:main-experiment-tradeoff}. Recorded elapsed counters sum to
approximately $6.33$ hours across retained calibration, tuning, final, and
auxiliary runs. This is partial timing: setup, final policy evaluations,
and failed-run costs are excluded. Runs used NumPy on a local macOS CPU;
the historical CPU model, RAM capacity, and peak memory were not logged.
\par
\endgroup
\newpage

\begin{figure}[H]
    \centering
    \includegraphics[width=\textwidth]{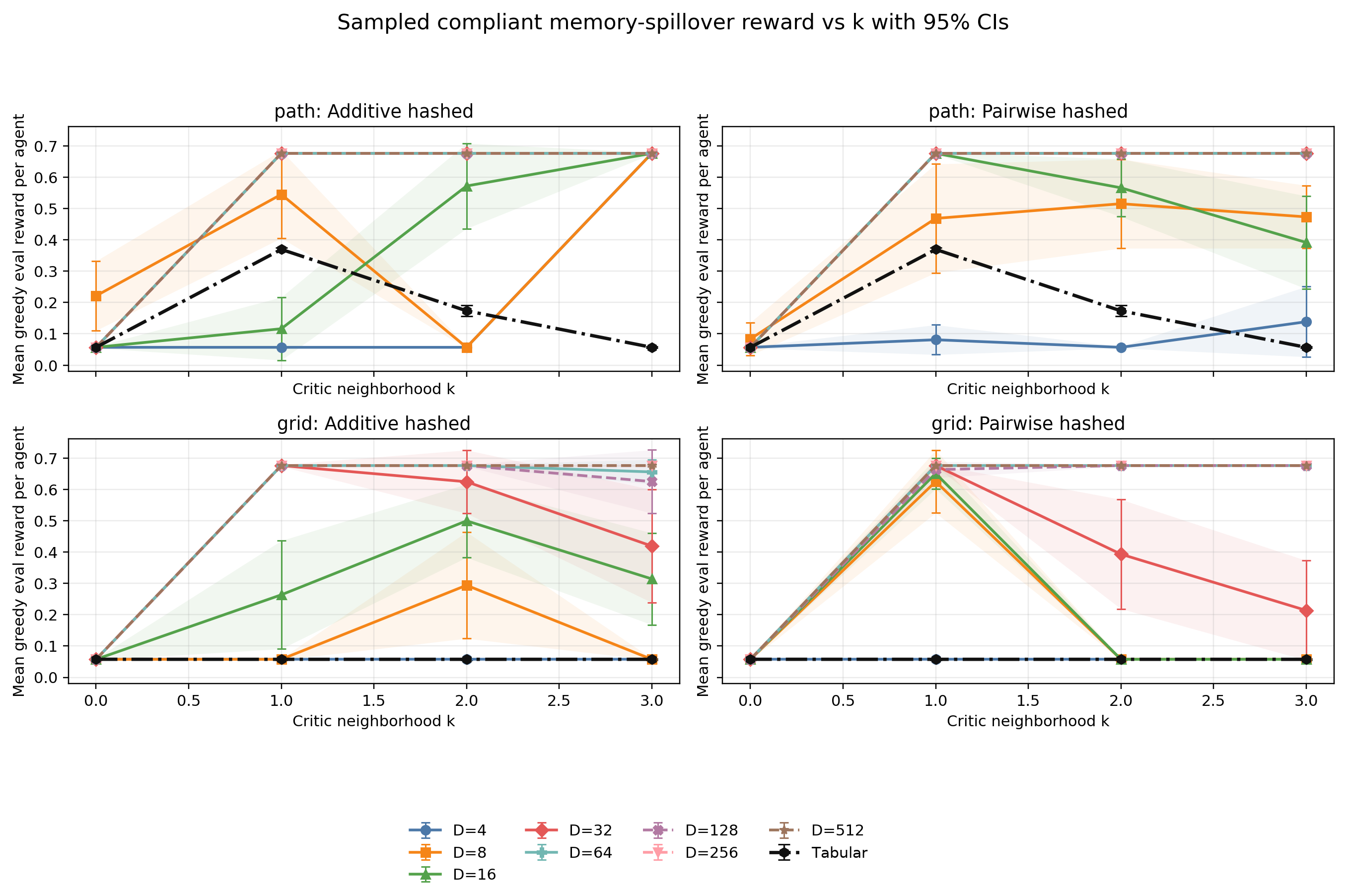}
    \caption{Mean greedy-evaluation reward per agent as the critic radius $k$
    and representation dimension $D$ vary.  Each panel pools the four tested
    sizes in one graph family and the three main-run seeds.  The plot uses the
    corrected summaries that exclude failed pre-repair $D=4$ runs.}
    \label{fig:experiment-localization-compression}
\end{figure}

\begin{table}[H]
    \centering
    \caption{Representative entries from Figure~\ref{fig:experiment-localization-compression}.
    Values are mean reward per agent $\pm$ the descriptive 95\% confidence interval.}
    \label{tab:experiment-localization-compression}
    \begingroup
    \small
    \setlength{\tabcolsep}{3.5pt}
    \begin{tabular}{llccccc}
        \toprule
        Graph & Representation & $D$ & $k=0$ & $k=1$ & $k=2$ & $k=3$ \\
        \midrule
        Path & Pairwise hashed & 16
            & $0.057\!\pm\!0.003$ & $0.676\!\pm\!0.002$
            & $0.566\!\pm\!0.092$ & $0.392\!\pm\!0.148$ \\
        Grid & Additive hashed & 16
            & $0.058\!\pm\!0.003$ & $0.263\!\pm\!0.173$
            & $0.499\!\pm\!0.117$ & $0.314\!\pm\!0.146$ \\
        Grid & Pairwise hashed & 8
            & $0.058\!\pm\!0.003$ & $0.625\!\pm\!0.100$
            & $0.058\!\pm\!0.003$ & $0.058\!\pm\!0.003$ \\
        Grid & Pairwise hashed & 64
            & $0.058\!\pm\!0.003$ & $0.676\!\pm\!0.003$
            & $0.676\!\pm\!0.003$ & $0.676\!\pm\!0.003$ \\
        Path & Tabular & --
            & $0.057\!\pm\!0.003$ & $0.369\!\pm\!0.006$
            & $0.173\!\pm\!0.017$ & $0.057\!\pm\!0.003$ \\
        Grid & Tabular & --
            & $0.058\!\pm\!0.003$ & $0.058\!\pm\!0.003$
            & $0.058\!\pm\!0.003$ & $0.058\!\pm\!0.003$ \\
        \bottomrule
    \end{tabular}
    \endgroup
\end{table}

\begin{table}[H]
    \centering
    \caption{Grid results separated by network size for the pairwise-hashed
    critic used in Figure~\ref{fig:main-experiment-tradeoff}.  Values are mean
    greedy reward per agent $\pm$ the descriptive 95\% confidence interval
    across the three seeds for that size; unlike the main figure, no graph
    sizes are pooled.}
    \label{tab:grid-size-specific-results}
    \begingroup
    \scriptsize
    \setlength{\tabcolsep}{2.5pt}
    \begin{tabular}{ccccccc}
        \toprule
        Grid & Agents & $d_i$ & $k=0$ & $k=1$ & $k=2$ & $k=3$ \\
        \midrule
        $3\times3$ & $9$ & 4 & $0.0556\!\pm\!0.0097$ & $0.0556\!\pm\!0.0097$ & $0.0556\!\pm\!0.0097$ & $0.0556\!\pm\!0.0097$ \\
        $3\times3$ & $9$ & 8 & $0.0556\!\pm\!0.0097$ & $0.4759\!\pm\!0.4044$ & $0.0556\!\pm\!0.0097$ & $0.0556\!\pm\!0.0097$ \\
        $3\times3$ & $9$ & 16 & $0.0556\!\pm\!0.0097$ & $0.6800\!\pm\!0.0098$ & $0.0556\!\pm\!0.0097$ & $0.0556\!\pm\!0.0097$ \\
        $3\times3$ & $9$ & 32 & $0.0556\!\pm\!0.0097$ & $0.6800\!\pm\!0.0098$ & $0.5754\!\pm\!0.2095$ & $0.6800\!\pm\!0.0098$ \\
        $3\times3$ & $9$ & 64 & $0.0556\!\pm\!0.0097$ & $0.6800\!\pm\!0.0098$ & $0.6800\!\pm\!0.0098$ & $0.6800\!\pm\!0.0098$ \\
        \addlinespace[2pt]
        $4\times4$ & $16$ & 4 & $0.0581\!\pm\!0.0056$ & $0.0581\!\pm\!0.0056$ & $0.0581\!\pm\!0.0056$ & $0.0581\!\pm\!0.0056$ \\
        $4\times4$ & $16$ & 8 & $0.0581\!\pm\!0.0056$ & $0.6753\!\pm\!0.0052$ & $0.0581\!\pm\!0.0056$ & $0.0581\!\pm\!0.0056$ \\
        $4\times4$ & $16$ & 16 & $0.0581\!\pm\!0.0056$ & $0.5736\!\pm\!0.1942$ & $0.0581\!\pm\!0.0056$ & $0.0581\!\pm\!0.0056$ \\
        $4\times4$ & $16$ & 32 & $0.0581\!\pm\!0.0056$ & $0.6753\!\pm\!0.0052$ & $0.6753\!\pm\!0.0052$ & $0.0581\!\pm\!0.0056$ \\
        $4\times4$ & $16$ & 64 & $0.0581\!\pm\!0.0056$ & $0.6753\!\pm\!0.0052$ & $0.6753\!\pm\!0.0052$ & $0.6753\!\pm\!0.0052$ \\
        \addlinespace[2pt]
        $6\times6$ & $36$ & 4 & $0.0588\!\pm\!0.0011$ & $0.0588\!\pm\!0.0011$ & $0.0588\!\pm\!0.0011$ & $0.0588\!\pm\!0.0011$ \\
        $6\times6$ & $36$ & 8 & $0.0593\!\pm\!0.0021$ & $0.6733\!\pm\!0.0042$ & $0.0588\!\pm\!0.0011$ & $0.0588\!\pm\!0.0011$ \\
        $6\times6$ & $36$ & 16 & $0.0588\!\pm\!0.0011$ & $0.6733\!\pm\!0.0042$ & $0.0588\!\pm\!0.0011$ & $0.0588\!\pm\!0.0011$ \\
        $6\times6$ & $36$ & 32 & $0.0588\!\pm\!0.0011$ & $0.6733\!\pm\!0.0042$ & $0.0588\!\pm\!0.0011$ & $0.0588\!\pm\!0.0011$ \\
        $6\times6$ & $36$ & 64 & $0.0588\!\pm\!0.0011$ & $0.6733\!\pm\!0.0042$ & $0.6733\!\pm\!0.0042$ & $0.6733\!\pm\!0.0042$ \\
        \addlinespace[2pt]
        $8\times8$ & $64$ & 4 & $0.0578\!\pm\!0.0002$ & $0.0578\!\pm\!0.0002$ & $0.0578\!\pm\!0.0002$ & $0.0578\!\pm\!0.0002$ \\
        $8\times8$ & $64$ & 8 & $0.0578\!\pm\!0.0002$ & $0.6751\!\pm\!0.0034$ & $0.0578\!\pm\!0.0002$ & $0.0578\!\pm\!0.0002$ \\
        $8\times8$ & $64$ & 16 & $0.0578\!\pm\!0.0002$ & $0.6751\!\pm\!0.0034$ & $0.0578\!\pm\!0.0002$ & $0.0578\!\pm\!0.0002$ \\
        $8\times8$ & $64$ & 32 & $0.0578\!\pm\!0.0002$ & $0.6751\!\pm\!0.0034$ & $0.2631\!\pm\!0.4025$ & $0.0578\!\pm\!0.0002$ \\
        $8\times8$ & $64$ & 64 & $0.0578\!\pm\!0.0002$ & $0.6751\!\pm\!0.0034$ & $0.6751\!\pm\!0.0034$ & $0.6751\!\pm\!0.0034$ \\
        \bottomrule
    \end{tabular}
    \endgroup
\end{table}

\begin{figure}[H]
    \centering
    \includegraphics[width=\textwidth]{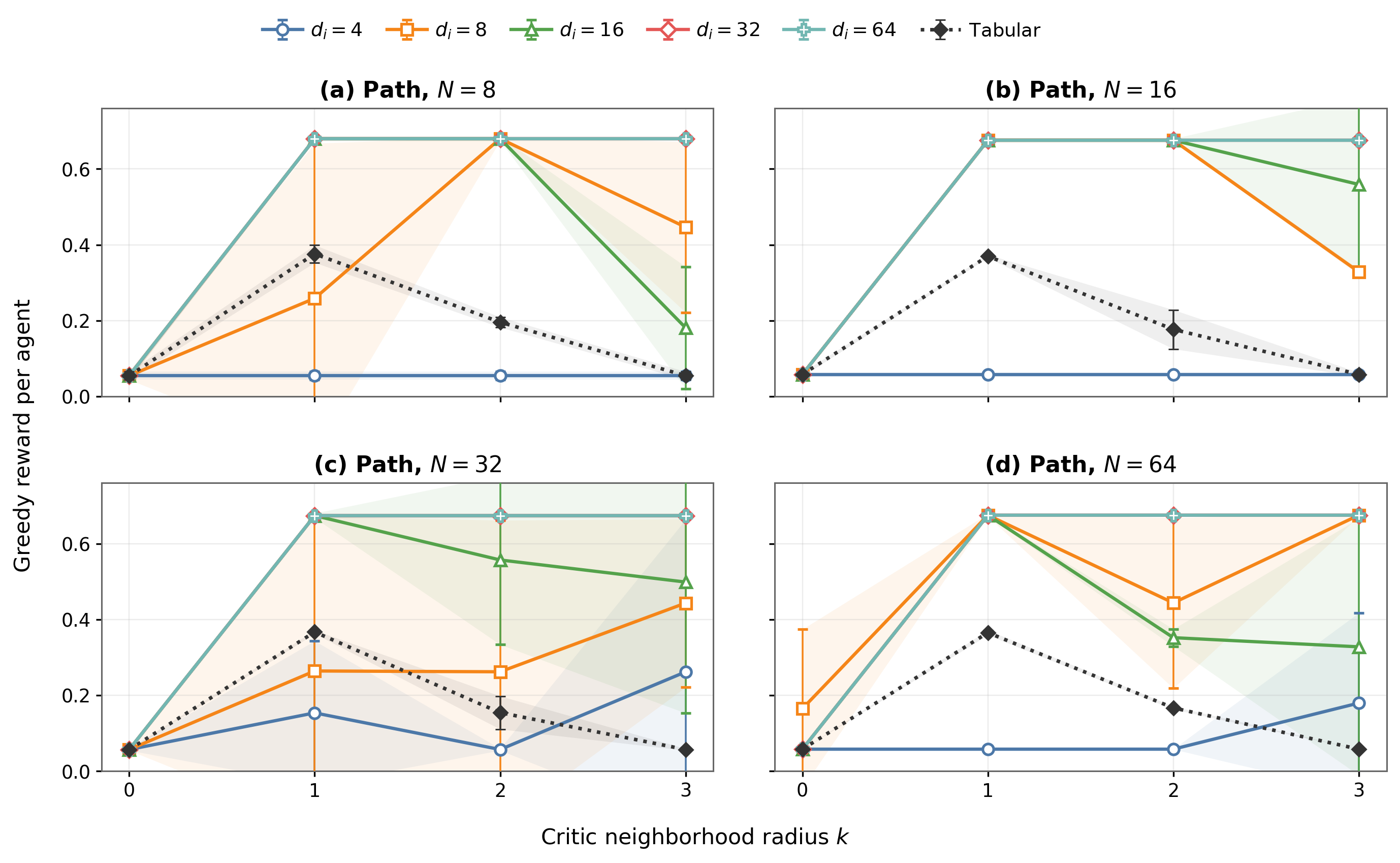}
    \caption{Path results separated by network size for the pairwise-hashed
    critic.  Each panel averages only the three seeds for the indicated path
    size, so no network sizes are pooled.  Curves show
    $d_i\in\{4,8,16,32,64\}$ and the tabular critic across
    $k\in\{0,1,2,3\}$; error bars are descriptive normal-approximation 95\%
    confidence intervals across the three seeds.}
    \label{fig:path-size-specific-results}
\end{figure}

\begin{figure}[H]
    \centering
    \includegraphics[width=\textwidth]{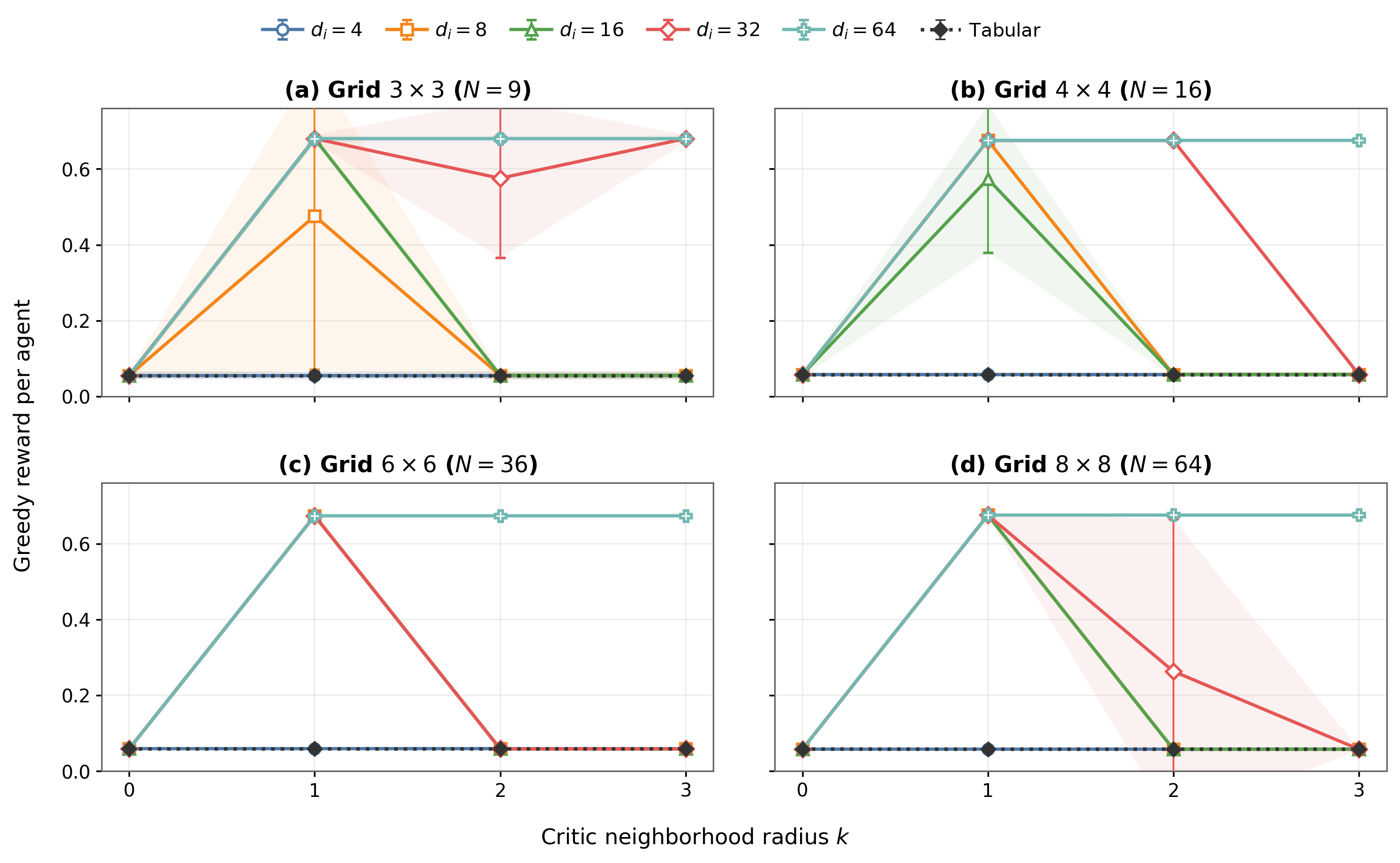}
    \caption{Grid results separated by network size for the pairwise-hashed
    critic.  Each panel averages only the three seeds for the indicated grid
    size, so no network sizes are pooled.  Curves show
    $d_i\in\{4,8,16,32,64\}$ and the tabular critic across
    $k\in\{0,1,2,3\}$; error bars are descriptive normal-approximation 95\%
    confidence intervals across the three seeds.  The corresponding numerical
    values appear in Table~\ref{tab:grid-size-specific-results}.}
    \label{fig:grid-size-specific-results}
\end{figure}

\paragraph{Results.}
The pooled curves in Figure~\ref{fig:experiment-localization-compression} and
representative entries in Table~\ref{tab:experiment-localization-compression}
show that at small and intermediate $D$, reward can first improve and then
deteriorate as $k$ grows.  For example, on grids with additive features and
$D=16$, reward increases from $0.058\pm0.003$ at $k=0$ to $0.499\pm0.117$ at
$k=2$, then falls to $0.314\pm0.146$ at $k=3$.  The grid pairwise result at
$D=8$ and the path pairwise result at $D=16$ show analogous reversals.  The
size-specific path and grid radius plots in
Figures~\ref{fig:path-size-specific-results} and
\ref{fig:grid-size-specific-results}, together with the grid values in
Table~\ref{tab:grid-size-specific-results}, show how these reversals vary
across network sizes.  These patterns are consistent with an
information--compression tradeoff: a positive radius makes neighbor-state
information available, while a growing neighborhood must be represented with
the same number of features.  The precise best radius depends on the embedding
and feature dimension, and the wider intervals at smaller $D$ indicate
sensitivity to training and hash collisions.

\begin{figure}[H]
    \centering
    \includegraphics[width=\textwidth]{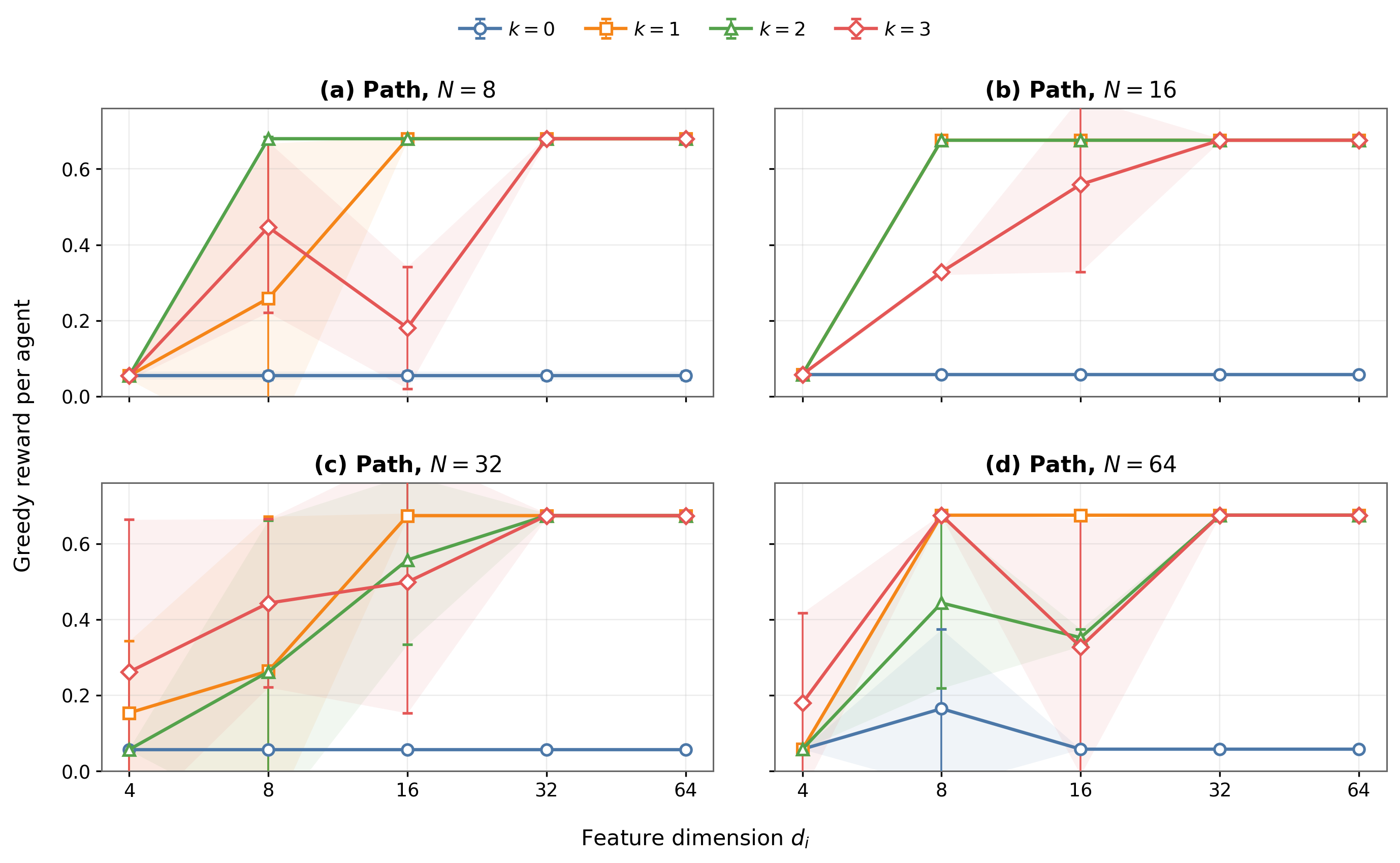}
    \caption{Path reward versus feature dimension, separated by network size,
    for the pairwise-hashed critic.  Each panel averages only the three seeds
    for the indicated path size, so no network sizes are pooled.  Curves show
    $k\in\{0,1,2,3\}$; error bars are descriptive normal-approximation 95\%
    confidence intervals across the three seeds.}
    \label{fig:path-size-specific-dimension-results}
\end{figure}

\begin{figure}[H]
    \centering
    \includegraphics[width=\textwidth]{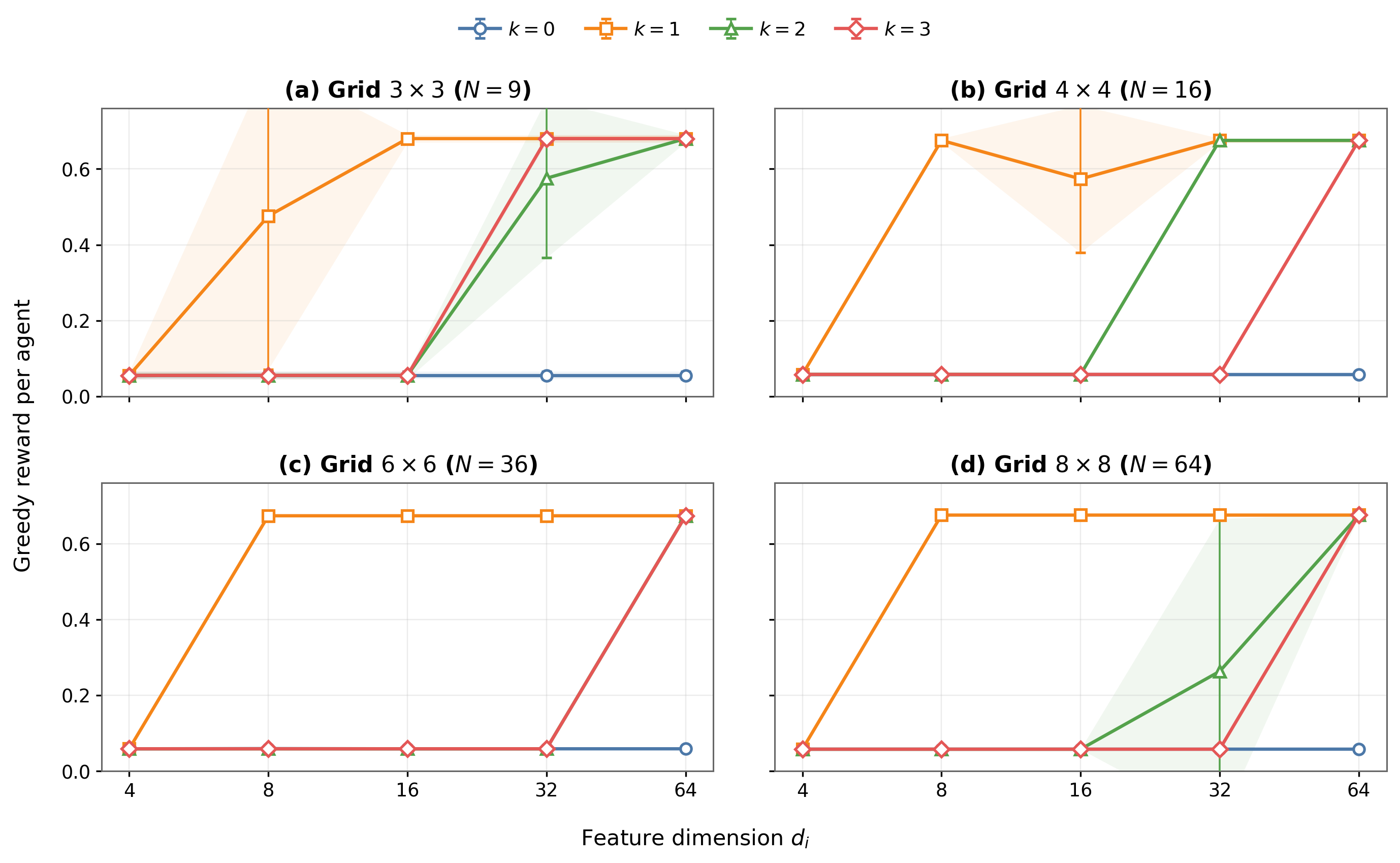}
    \caption{Grid reward versus feature dimension, separated by network size,
    for the pairwise-hashed critic.  Each panel averages only the three seeds
    for the indicated grid size, so no network sizes are pooled.  Curves show
    $k\in\{0,1,2,3\}$; error bars are descriptive normal-approximation 95\%
    confidence intervals across the three seeds.  The corresponding numerical
    values appear in Table~\ref{tab:grid-size-specific-results}.}
    \label{fig:grid-size-specific-dimension-results}
\end{figure}

The dimension sweeps in Figures~\ref{fig:path-size-specific-dimension-results}
and~\ref{fig:grid-size-specific-dimension-results} show that increasing $D$
largely removes this bottleneck in the tested instances.  Every positive path
radius reaches approximately $0.676$ by $D=32$, and every positive radius for
the grid pairwise embedding reaches $0.676\pm0.003$ at $D=64$.  Under the same
finite trajectory budget, the tabular critic remains near $0.058$ for every
grid radius and reaches only $0.369\pm0.006$ on paths, consistent with sparse
visitation of a rapidly growing local table.  These comparisons illustrate why
a compact linear critic can be useful even though it introduces a
feature-approximation floor.

\paragraph{Relation to the theoretical framework.}
These experiments follow the same high-level localization--representation idea
as Algorithm~\ref{alg:decentralized_ac_lfa}: actors use local observations,
critics incorporate information from a $k$-hop neighborhood, rewards and
transitions are locally structured, and finite-dimensional features control
critic representation capacity.  For scalable implementation across the
tested graph families, \textcolor{revisionblue}{linear} actor and critic parameterizations are shared across
agents, neighborhood critic quantities are averaged using a counterfactual
baseline, and the actor feature map is varied together with $D$.  A few
protocol details differ slightly: evaluation uses greedy actions rather than
sampling from the stochastic softmax policy, and the tuning and main runs use
partially overlapping seeds.  The deterministic memory update preserves the
local factorization and delayed neighbor interactions; its unit
self-interaction places this instance at the boundary of, rather than strictly
inside, the coefficient condition used as a sufficient route to spatial
decay.  These choices preserve the mechanism of interest, and the results
provide a reproducible finite-state illustration of the practical
localization--representation tradeoff highlighted by the analysis.

\end{document}